%% file: main.tex
\documentclass[11pt,openany]{book}
\usepackage{egbook}

\title{\Huge\bfseries Elimination Geometry\\[1.2ex]
\Large From Local Optima to Global Realizability\\[2ex]
\large Structural Realizability, Certification, and Repair in Statistical Learning and AI}
\author{Mian Huang \and Xueqin Wang}
\date{Research monograph
}

\begin{document}
\frontmatter
\maketitle
\input{frontmatter/copyright}
\input{frontmatter/preface}
\input{frontmatter/notation}
\clearpage
\pdfbookmark[0]{Contents}{contents}
\tableofcontents
\listoffigures
\listoftables

\mainmatter
\egpart{The Structural Realizability Problem}{Modern learning systems reuse
representations, parameterizations, memories, and output interfaces across
many local problems.  The first question is therefore not only whether local
optima exist, but whether one declared deployment system can realize them
simultaneously.}
\input{chapters/ch01_deployability}
\input{chapters/ch02_certified_systems}
\input{chapters/ch03_risk_decomposition}

\input{chapters/ch04_laboratories}
{\small\input{part_notes/part01_history}}

\partbridge{From structural realizability to native loss scales}{Part I
separated local solvability, global realizability, and finite-sample
certifiability, and introduced a four-component risk decomposition.  Part II
now derives the loss scale in which structural nonrealizability should be
measured: the excess objective generated by the original elimination itself.}

\egpart{Native Defects and Elimination Calculus}{Elimination does more than
simplify an objective.  It generates an objective-native notion of excess loss
and determines which transformations preserve the statistical target.}
\input{chapters/ch05_conjugate_lifts}

\input{chapters/ch06_towers_pgxvc}

\input{chapters/ch07_exactification}

\input{chapters/ch08_graph_crps}

{\small\input{part_notes/part02_history}}

\partbridge{From native loss to valid structural objects}{A defect can enter
an architecture theorem only after two questions have been answered.  Do
locally reported increments integrate to one global objective?  If an
auxiliary lift is not directly native, does it arise from an intrinsic
factorization rather than an arbitrary target-calling coordinate?  Part III
closes these validity gates before obstruction is quantified.}

\egpart{Validity of Defects and Representations}{An architecture obstruction
is meaningful only when its local defect system is integrable and its carrier
is intrinsic to the declared target.}
\input{chapters/ch09_integrability}

\input{chapters/ch10_lift_complexity}

{\small\input{part_notes/part03_history}}

\partbridge{From valid defects to structural nonrealizability}{Parts II--III
have fixed the native loss scale and certified the defect system and carrier
through which it is represented.  Part IV changes the quantifiers: can one
shared, compressed, smooth, or coordinated deployment realize the entire
oracle family?  Regular variation, parameter sharing, and branch inconsistency around loops
(monodromy) produce different lower bounds and call for different repairs.}

\egpart{Structural Obstructions under Shared Deployment}{A shared deployment
contract turns local optimization into a global compatibility problem.
Regularity, coordination, and singularity impose different structural limits
and call for mechanism-matched repairs.}
\input{chapters/ch11_second_elimination}
\input{chapters/ch12_regular_transfer}

\input{chapters/ch13_coordination}

\input{chapters/ch14_singular}

\input{chapters/ch15_repairs}

{\small\input{part_notes/part04_history}}

\partbridge{From obstruction to resources, tasks, and composition}{A positive
architecture obstruction is not yet an operational conclusion.  Part V
separates carrier information loss from decoder limitations, asks which
distinctions legal downstream tasks can observe, and states when certified
defect decompositions remain valid across conditioning, representations, and
temperature limits.}

\egpart{Resources, Operational Semantics, and Composition}{Resources
determine which oracle distinctions can be retained; task and context classes
determine which surviving distinctions are operationally visible.}
\input{chapters/ch16_resource_rd}

\input{chapters/ch17_operational_semantics}

\input{chapters/ch18_composition}

{\small\input{part_notes/part05_history}}

\partbridge{From population structure to finite-sample certification}{Parts
I--V described what is true when the population objective, oracle family, and
deployment grammar are known.  Finite data do not select one population
world.  They generate a covered set of population states that retain inferential
standing under the declared error contract; Chapter~\ref{ch:statistical-eg}
formalizes this object as a \emph{confidence world}.  Part
VI projects that set through declared queries, authorizes a resolved
certificate only when all compatible worlds agree, and asks how much
information is required to separate the worlds that would reverse the
conclusion.}

\egpart{Statistical Certification}{Data produce a confidence world, not one
privileged population story.  Its identified image is the complete
finite-data answer to a declared query; a resolved certificate is authorized
only when that image lies on one side of the operational boundary.}
\input{chapters/ch19_statistical_eg}

\input{chapters/ch20_certificate_statistics}

{\small\input{part_notes/part06_history}}

\partbridge{From finite-sample certification to structural intervention}{Part
VI established confidence sets, three-way decisions, resolution limits, and
the evidence that must be retained.  A certificate is not yet a deployable
repair.  Part VII requires one common deployment witness, turns a typed structural failure into a controlled repair workflow,
develops a partial-transport implementation for multibranch oracles, and ends
with independent validation.  Chapter~\ref{ch:oali-workflow} introduces the
general intervention protocol and its terminology.

\par\smallskip
{\centering\bfseries Five reductions and what survives\par}
\smallskip
{\small
\setlength{\tabcolsep}{3pt}%
\begin{tabularx}{\linewidth}{@{}>{\raggedright\arraybackslash}p{0.18\linewidth}>{\raggedright\arraybackslash}p{0.21\linewidth}>{\raggedright\arraybackslash}X>{\raggedright\arraybackslash}p{0.18\linewidth}@{}}
\toprule
Operation & Removes & Must preserve & It is not\\
\midrule
Vertical fiber quotient & optimized-away auxiliary state & native defect: original-objective height above the oracle fiber & an arbitrary parameter distance\\
Source-side oracle quotient & target-invisible labels, gauge, and padding in the raw carrier & oracle signatures and target-visible distinctions & raw parameter count\\
Sink-side contextual quotient & kernel differences invisible to every legal context and task & canonical operational behavior & native carrier sufficiency\\
Statistical projection & coordinates suppressed by the declared audit map & its entire identified image over all data-compatible populations & a structural quotient or a point estimate\\
Recursive quotient & only distinctions invisible to color and every labeled future update & future-closed certificate and action distinctions & compression by current posterior or current action alone\\
\bottomrule
\end{tabularx}}
}

\egpart{Obstruction-Aware Learning and Structural Repair}{A structural
certificate becomes actionable only when it localizes the failed interface,
authorizes a compatible intervention, and predicts an improvement that
survives independent validation and matched controls.}
\input{chapters/ch21_certified_learning}

\input{chapters/ch22_oali_workflow}

\input{chapters/ch23_transport_cycles}

\input{chapters/ch24_applied_laboratories}
{\small\input{part_notes/part07_history}}

\partbridge{From worked systems to stress tests and limits}{The applications
show how structural realizability depends on the local oracle, observation
unit, representation, and deployment contract.  They also expose where the
theory still depends on finite-state structure, stable transports, or
independent validation.  Part VIII subjects the framework to a
noncommutative stress test, records its claim boundaries, and closes with a
falsifiable research program.}

\egpart{Stress Tests, Boundaries, and Research Program}{A structural theory
must survive foreign geometries, expose its boundaries, and make its claims
falsifiable. Its final question is what knowledge a learning system can
coherently realize.}
\input{chapters/ch25_quantum}

\input{chapters/ch26_boundaries}
\input{chapters/ch27_agenda}
{\small\input{part_notes/part08_history}}

\appendix
\part*{Appendices: Technical Background, Terminology, and Crosswalks}
\addcontentsline{toc}{part}{Appendices: Technical Background, Terminology, and Crosswalks}
\input{appendices/appA_convex}
\input{appendices/appB_graphs}
\input{appendices/appC_topology}
\input{appendices/appD_statistics}
\input{appendices/appE_dependency}
\input{appendices/appF_glossary}
\input{appendices/appG_source_map}

\backmatter
\bibliographystyle{plainnat}
\bibliography{references}
\input{thematic_index_crossrefs}
\begingroup
\addtolength{\textheight}{12pt}
\printindex
\endgroup
\end{document}

%% file: frontmatter/copyright.tex
\cleardoublepage
\thispagestyle{empty}
\vspace*{0.25\textheight}
\begin{center}
{\small
\textbf{Elimination Geometry: From Local Optima to Global Realizability}\\[0.8em]
Structural Realizability, Certification, and Repair in Statistical Learning and AI\\[2em]
Copyright \textcopyright\ 2026 Mian Huang and Xueqin Wang\\[0.5em]
Draft for research circulation. All theorem statements, proofs, examples,\
and computational protocols remain subject to author review before formal publication.\\[2em]
Typeset in \LaTeX. Computational outputs are reproduced only where their\
status and validation scope are explicitly stated in the text.
}
\end{center}
\vfill
\noindent\textit{Citation note.} Until a publisher version is available, cite this work by title, authors, version number, and year.
\cleardoublepage

%% file: frontmatter/preface.tex
\chapter*{Preface}
\addcontentsline{toc}{chapter}{Preface}
\markboth{Preface}{Preface}

Modern statistical learning and artificial intelligence increasingly rely on
a common mechanism being reused across many local problems.  Amortized
inference replaces repeated optimization by one inference network; multitask
systems share representations or parameters; sequential decision systems
compress growing histories into finite states; scientific learning pipelines
seek common labels or comparable representations across samples, batches, or
conditions.  Reuse creates efficiency, but it also creates a structural
question that is distinct from ordinary finite-sample error and numerical
optimization.

Suppose each input $x$ has a locally optimal object $a_x^\star$.  Pointwise
solvability asks only
\[
  \forall x\;\exists a_x^\star.
\]
Deployment asks for one admissible rule
\[
  \exists A\in\mathfrak A\;\forall x,
  \qquad A(x)\approx a_x^\star,
\]
where $\mathfrak A$ may enforce shared parameters, continuity, finite memory,
a fixed output type, communication limits, or a computation budget.  The
first statement does not imply the second.  When the implication fails, more
data, longer training, or greater width inside the same deployment contract
need not remove the resulting loss; the required intervention may instead be
a different representation, sharing pattern, memory state, output semantics,
or resource allocation.

The phenomenon is not new.  Approximation theory, amortized inference,
multitask learning, representation learning, control, and topology all study
nearby failures.  The book therefore makes a narrower claim.  Once a local
statistical or optimization problem has fixed its oracle objects, can the loss
caused specifically by a shared deployment contract be separated from local
model approximation, finite-sample generalization, and implementation error?
Can that loss be expressed in the scale generated by the original objective
rather than in a convenient post hoc distance?

Three levels must be kept distinct.  \emph{Local solvability} asks whether the
pointwise oracle is well defined and what it costs to deviate from it.
\emph{Global realizability} asks whether one rule satisfying the declared
deployment contract can realize the collection of local optima
simultaneously.  \emph{Finite-sample certifiability} asks whether the available
data are sufficient to distinguish realizability, nonrealizability, and an
unresolved case with stated error control.  None of these levels subsumes the
others: a population obstruction need not be identifiable from finite data,
and statistical unresolvedness does not imply that the population structure
is absent.

The technical starting point is elimination.  If an auxiliary object $a$ is
optimized out of a local objective,
\[
  J(x)=\inf_a H_x(a),
\]
then
\[
  \Def_x(a)=H_x(a)-J(x)\ge0
\]
is the \emph{native defect}: the exact excess objective incurred by deploying
$a$ rather than a local optimum.  This quantity is inherited from the
likelihood, divergence, proper score, Bellman objective, free energy, or other
declared criterion; it is not an arbitrary distance chosen after the fact.
For a deployment class $\mathfrak A$ and population law $P$,
\[
  \inf_{A\in\mathfrak A}\mathbb E_P D_X\{A(X)\}
\]
is the architecture obstruction: the smallest native population risk left by
the deployment contract after the local oracle has been fixed.

The book assembles an audit workflow rather than a new universal
approximation or representation theory.  Its mathematical ingredients come
from several established traditions, including convex duality, approximation
and representation theory, information theory, control, topology, partial
identification, and statistical learning.  The proposed contribution is to
type their assumptions and conclusions on a common native-loss interface and
to expose the additional assumptions needed when moving from one interface to
the next.  The ordering is methodological, not a theorem saying that a native
defect implies an architecture obstruction, or that an obstruction implies a
successful repair.

The resulting workflow is deliberately falsifiable.  It derives a native
loss, audits a declared deployment contract, constructs an appropriate
population or statistical certificate, and proposes a mechanism-matched
intervention.  A stronger architecture-choice claim additionally requires
evidence of saturation within the fixed contract, independent validation,
and parameter-, compute-, and selection-matched negative controls.  The
empirical studies in this edition illustrate only parts of that workflow; none
completes a general certificate-saturation-repair validation chain.

This Version closes one architecture-specific mathematical chain for
full-column-rank nonnegative sparse inference: strict active-set neighborhoods
yield a computable one-pass native-loss floor, and the declared proximal
repair has an explicit sufficient crossing depth.  The sealed diabetes case
also closes a separate two-step, patient-average, equal-information and
equal-compute decision.  These two advances are deliberately not conflated:
the theorem-alignment calculation is post-confirmation, its priority remains
independently unresolved, and no general architecture-choice theory or
external replication is claimed.

This scope imposes important limits.  Architecture obstruction is not a new
name for generic approximation error.  A positive worst-case topological cost
does not automatically imply a positive average risk.  A deterministic
population obstruction is not automatically identifiable from finite data.
A native defect need not be visible to the final scientific endpoint.  And a
structural repair does not by itself guarantee held-out improvement.  Each
transition requires its own theorem or empirical validation.

The recurring examples—variational inference and EM, graph-indexed
distribution learning, shared policies and finite memory, roots and
eigenvectors, resource-limited representations, and multi-sample population
alignment—serve one purpose: to test whether the same sequence of questions
can be answered across different mathematical objects.  What is locally
optimal?  What must be reused globally?  Which deployment constraint creates
the obstruction?  What is the minimal structural change that removes it?  Can
finite data support that diagnosis?

The intended contribution is therefore a typed, native-loss, audit-oriented
synthesis complementary to existing learning theory.  It organizes results
about when local optima exist, when a declared shared contract can realize
them, how particular lower bounds are expressed in objective units, and what
additional statistical and experimental evidence would be needed before an
architecture intervention is justified.  It does not claim that these
questions constitute a wholly new approximation or representation theory.

\begin{flushright}
Mian Huang and Xueqin Wang\\
August 2026 
\end{flushright}

%% file: frontmatter/notation.tex
\chapter*{Notation and Standing Conventions}
\addcontentsline{toc}{chapter}{Notation and Standing Conventions}
\markboth{Notation and Standing Conventions}{Notation and Standing Conventions}

\begingroup
\small
\setlength{\tabcolsep}{4pt}
\renewcommand{\arraystretch}{1.04}
\begin{longtable}{@{}p{0.27\textwidth}p{0.69\textwidth}@{}}
\toprule
Symbol & Meaning\\
\midrule
\endfirsthead
\toprule
Symbol & Meaning\\
\midrule
\endhead
\midrule
\endfoot
\bottomrule
\endlastfoot
\multicolumn{2}{@{}l}{\bfseries Core elimination and architecture}\\*[2pt]
$\X$ & instance, covariate, history, or problem-index space\\
$P$ & population law on $\X$ or on observed biological/statistical units\\
$\A_x$ & auxiliary fiber at instance $x$\\
$H_x(a)$ & lifted objective before elimination\\
$J(x)$ & eliminated objective $\inf_{a\in\A_x}H_x(a)$\\
$\mathcal O(x)$ & oracle set $\argmin_{a\in\A_x}H_x(a)$\\
$\Def_x(a)$ & native defect $H_x(a)-J(x)$\\
$\infconv$ & typed infimal (min-plus) composition of compatible stage costs\\
$\Arch$ & declared class of deployable fields or architectures\\
$A$ & one deployed field $x\mapsto A(x)$\\
$\Obs_P(\Arch)$ & population architecture obstruction $\inf_{A\in\Arch}\E_P\Def_X\{A(X)\}$\\
$\Obs_\infty(\Arch)$ & uniform obstruction $\inf_{A\in\Arch}\sup_x\Def_x\{A(x)\}$\\
$\implgap(A;\Arch)$ & excess defect of $A$ above the architecture infimum\\
$Z$ & carrier or representation\\
$g$ & decoder or readout\\
$S_U,\ a_U^\dagger$ & conditional dual oracle signature $\E(S\mid U)$ and its canonical Bregman decoder $\nabla\Phi^\star(S_U)$\\
$\mathcal C_\Phi(U),\ \operatorname{vc}_\Phi(U)$ & carrier information loss and visible signature cardinality\\
$\mathfrak F_{J,\mathcal K}(m,r,\delta)$ & quotient-reduced lift frontier at cone, coherence, and fidelity budgets\\
$\mathfrak G_{m,r},\ \Psi_{m,r}(Z)$ & resource grammar and decoder nonsaturation for carrier $Z$\\
$\mathcal D_{\rm nat}(m,r)$ & optimal native architecture distortion under the resource grammar\\
$W_i,\ \rho_i$ & rectangular Bellman value and its local coordination residual at stage $i$\\
$\Task$ & declared downstream task contract\\
$K$ & min-plus or operational kernel\\
$\cert$ & typed finite-data certificate\\
$\modelgap$ & local model or variational approximation component\\
$\archgap$ & global architecture obstruction component\\
$\gengap$ & generalization component\\
$\optgap$ & optimization or implementation component\\
\addlinespace[4pt]
\multicolumn{2}{@{}l}{\bfseries Certified learning and intervention}\\*[2pt]
$\mathcal W,\ C_n,\ (C_t)_{t\ge0}$
  & Population-world universe, fixed-record confidence world, and anytime
    confidence-world sequence.  Their declared coverage events contain the
    true population world at the protected records or times.\\
$J_C(q),\ \delta_C^q,\ m_C(q)$
  & Identified image, standard typed certificate, and certificate margin for
    query $q$ over confidence world $C$.\\
$q:\Theta\to\mathsf Q$
  & Declared certificate truth or architecture-action color;
    $\operatorname{Alt}_q(\theta)=\{\lambda:q(\lambda)\ne q(\theta)\}$.\\
$\mathcal I_X^{q,*}(\theta),\ T_X^{q,*}(\theta)$
  & Raw-data max--min KL rate and its inverse characteristic time.\\
$\mathcal I_Z^{q,*}(\theta),\ T_Z^{q,*}(\theta)$
  & Retained-evidence counterparts; data processing gives
    $\mathcal I_Z^{q,*}\le\mathcal I_X^{q,*}$ and
    $T_Z^{q,*}\ge T_X^{q,*}$.\\
$H_{\mathrm{dep}},\ d_{\mathrm{dep}}$
  & Deployment carrier and action decoder; exactness requires
    $a^\star=d_{\mathrm{dep}}\circ H_{\mathrm{dep}}$.\\
$T_{\mathrm{ev}}$
  & World-independent channel from raw observations or transcripts to retained
    evidence $Z$.\\
$\phi_{\mathrm{rec}}$
  & Recursive carrier preserving the declared color and every
    observation-labeled successor law.\\
$\zeta=(c_{\mathrm{dep}},c_{\mathrm{cert}},c_{\mathrm{ctx}})$
  & Joint deployment, certificate, and contextual-equivalence color.\\
$N_h,\ N_\infty,\ N_{\mathrm{joint}}$
  & Class counts of the $h$-step, stable, and joint-color stable recursive
    quotients.\\
$\mathfrak F_R^\eta(\lambda),\ A_\tau$
  & Worldwise feasible set and adapted common witness; certification requires
    $A_\tau\in\bigcap_{\lambda\in C_\tau}\mathfrak F_R^\eta(\lambda)$.\\
\end{longtable}
\endgroup

\section*{Orientation of asymmetric divergences}
Throughout the book,
\[
  D_\Phi(a\|b)
  =\Phi(a)-\Phi(b)-\langle\nabla\Phi(b),a-b\rangle,
\]
so the first argument is the evaluation point and the second is the tangent or
reference point.  Likewise,
\[
  \KL(Q\|P)=\int\log\!\left(\frac{\dd Q}{\dd P}\right)\dd Q
\]
when $Q\ll P$, and $\KL(Q\|P)=+\infty$ otherwise.  The first argument is the
integration law and the second the reference; conditioning and pushforward
preserve this order.  Native defects therefore place the trial or deployed
object first and the oracle second.

\section*{Standing distinctions}
The following distinctions are maintained throughout.
\begin{itemize}
  \item A zero infimum does not imply that an exact minimizer or global section is attained.
  \item Uniform, average, and task-weighted risks are different contracts and are never interchanged without an explicit theorem.
  \item Local oracle error, architecture obstruction, generalization error, and implementation error are recorded separately unless an exact identity combines them.
  \item Point-valued, set-valued, quotient-valued, projector-valued, and distribution-valued outputs are different semantic contracts.
  \item Deterministic obstruction, statistical impossibility, posterior credibility, and computational intractability are different claims.
  \item The independent statistical unit is the declared sampling unit, not automatically an individual cell, token, edge, or repeated measurement.
\end{itemize}

\section*{Infima and measurability}
Unless stated otherwise, all infima are over nonempty declared classes.  Measurability, lower semicontinuity, compactness, coercivity, or closure assumptions needed for attainment are stated locally rather than imposed globally.  When only approximation is required, the book works with infima and $\varepsilon$-optimal selections.

%% file: chapters/ch01_deployability.tex
\chapter{Structural Realizability under Shared Deployment}
\label{ch:deployability}

\section{Reuse changes the learning problem}
\index{deployability}
\index{oracle!field}
\index{global section}
\index{shared deployment}

Many modern learning systems solve a family of local problems and then reuse a
single mechanism across them.  Let
\[
  a_x^\star\in\argmin_{a\in\A_x}H_x(a)
\]
denote a local posterior, conditional policy, spectral object, population
identity, or inner optimization state.  A pointwise procedure asks only that
each local problem be solvable.  Deployment asks for one field
\[
  A:\X\longrightarrow\bigsqcup_{x\in\X}\A_x,
  \qquad A(x)\in\A_x,
\]
subject to a contract such as continuity, Lipschitz regularity, finite memory,
shared parameters, a fixed output type, bounded communication, or a prescribed
computation graph.

The quantifiers are different:
\[
  \forall x\;\exists a_x^\star
  \qquad\not\Rightarrow\qquad
  \exists A\in\Arch\;\forall x,\ A(x)=a_x^\star.
\]
The right-hand statement is a global realizability problem.  Its failure is
structural when it persists after the local oracles have been fixed and after
optimization within the declared deployment class has been completed.

A two-point example isolates the issue.  Let $X\in\{0,1\}$ be equiprobable and
let the native loss be $\Def_x(a)=(a-x)^2$.  Each local optimum is exact:
$a^\star(x)=x$.  If deployment is restricted to constant rules
$A(x)\equiv c$, then
\[
  \inf_c \mathbb E(c-X)^2
  =
  \inf_c\frac12\{c^2+(c-1)^2\}
  =
  \frac14.
\]
No reparameterization of the same constant-output contract removes this
$1/4$.  The obstruction disappears only when the deployment rule is allowed
to depend on $x$.

The same pattern appears in less trivial forms.  Amortized inference asks one
encoder to reproduce many instancewise variational optima.  Sequential
decision systems reuse parameters or a compressed memory state across many
histories.  Spectral procedures may need one continuous representative of an
eigenspace whose oriented eigenvector changes sign around a loop.  The
mechanisms differ, but the quantifier mismatch is the same: local solvability
does not determine global realizability under reuse.

\section{Three levels: solvability, realizability, and certifiability}
\index{local solvability}
\index{global realizability}
\index{finite-sample certifiability}

The book separates three questions that are often mixed together.
\begin{enumerate}
\item \emph{Local solvability.}  For each input, is the local oracle well
defined, and what excess objective is paid by deviating from it?
\item \emph{Global realizability.}  Does one rule satisfying the declared
representation, sharing, memory, regularity, and resource constraints realize
the local oracle family simultaneously?
\item \emph{Finite-sample certifiability.}  When the local oracles and
population structure must themselves be estimated, do the available data
support a statistically valid conclusion of realizability,
nonrealizability, or unresolvedness?
\end{enumerate}

These levels do not substitute for one another.  A genuine population
obstruction need not be identifiable at a realistic sample size.  Conversely,
an unresolved statistical certificate does not imply that the obstruction is
absent.  If the local model itself is misspecified, then a perfectly
realizable deployment can still be scientifically inadequate.

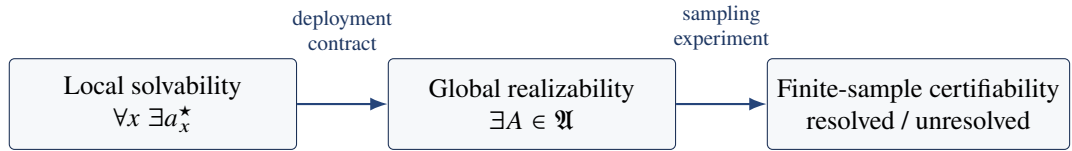
\begin{figure}[htbp]
\centering
\begin{tikzpicture}[node distance=12mm,>=Latex]
  \tikzset{
    box/.style={draw=egblue!70!black,rounded corners=2pt,fill=egblue!4,
      minimum width=38mm,minimum height=12mm,align=center,font=\small},
    bridge/.style={align=center,font=\scriptsize,text=egblue!80!black,
      inner sep=0pt}
  }
  \node[box] (local) {Local solvability\\$\forall x\;\exists a_x^\star$};
  \node[box,right=of local] (global) {Global realizability\\$\exists A\in\Arch$};
  \node[box,right=of global] (stat) {Finite-sample certifiability\\resolved / unresolved};
  \draw[->,thick,egblue] (local) --
    node[midway,above=7mm,bridge]{deployment\\contract} (global);
  \draw[->,thick,egblue] (global) --
    node[midway,above=7mm,bridge]{sampling\\experiment} (stat);
\end{tikzpicture}
\caption{Three distinct questions. Local optimization, global deployment, and finite-data certification require different assumptions and different remedies.}
\label{fig:three-levels}
\end{figure}

This separation determines the intervention.  Model misspecification calls
for a richer local model; finite-sample ambiguity calls for more informative
data or a different experiment; implementation error calls for a better
solver.  Structural nonrealizability calls instead for a change in the
deployment contract---for example a quotient representation, an atlas,
additional memory, a different sharing pattern, or a larger resource budget.

\paragraph{Position relative to neighboring theory.}
The book does not claim priority for the observation that sharing can create
suboptimality.  The amortization gap has been separated from variational-family
approximation since early work on VAEs \citep{CremerLiDuvenaud2018}, and recent
results characterize when amortized variational inference can attain its
instancewise optimum \citep{MargossianBlei2024}.  Multitask representation
learning studies benefits and limits of shared features
\citep{LiEtAl2025MultitaskRep}, while finite-memory RL quantifies errors caused
by compressed histories \citep{EberhardEtAl2025}.  Elimination geometry makes
a more specific claim: once the local oracle and its native excess objective
are fixed, it seeks objective-denominated lower bounds, finite-data
certificates, and mechanism-specific repairs for the remaining deployment
constraint.

The architecture-specific candidate in Chapter~\ref{ch:statistical-eg} is
correspondingly narrower.  \citet{ONeillGumranKlindt2025} already establish a
global amortisation gap for a one-layer linear--nonlinear sparse autoencoder
and study inference-time optimization.  This book does not reclaim that
conclusion.  It asks whether two strict KKT neighborhoods in one fitted
nonnegative sparse problem expose incompatible oracle Jacobian rows, converts
that conflict into a quantitative native-loss floor for the declared
one-pass class, and computes a sufficient proximal repair depth.  Its exact
priority remains unresolved and is recorded as such in Appendix~\ref{app:source-map}.

\section{Native risk and architecture obstruction}
\index{defect!native}
\index{obstruction!architecture}
\index{native loss scale}

A structural lower bound is useful only if it is measured in a loss scale
connected to the original problem.  If an auxiliary object $a$ is eliminated
from
\[
  J(x)=\inf_{a\in\A_x}H_x(a),
\]
then the native defect is
\[
  \Def_x(a)=H_x(a)-J(x)\ge0.
\]
It is an exact excess objective, not a post hoc parameter distance.  The
population risk of a deployment $A$ is
\[
  \mathcal L_P(A)=\mathbb E_P\Def_X\{A(X)\},
\]
and the architecture obstruction is
\[
  \Obs_P(\Arch)
  =
  \inf_{A\in\Arch}\mathcal L_P(A).
\]
When $\Obs_P(\Arch)>0$, the declared deployment class has a strictly positive
population floor relative to the already fixed local oracle family.

This is the book's organizing interface with generic expressivity and
approximation theory.  In settings with a unique oracle and a norm-equivalent
defect, the resulting obstruction may reduce to an ordinary approximation
error.  The useful additional discipline is then not a new mathematical
mechanism, but the requirement that the local target, loss scale, deployment
contract, and population quantifier remain explicit.

\begin{boundarybox}
Architecture obstruction is not a synonym for generic approximation error.
It has an independent interpretation only after the local oracle, native
defect, representation semantics, deployment class, statistical unit, and
implementation error have been declared separately.
\end{boundarybox}

\section{Mission and falsifiable consequences}
\index{structural realizability}
\index{falsification!structural explanation}

The purpose of this book is to provide a common audit language for identifying,
quantifying, and testing claims of structural nonrealizability under shared
deployment.  The chapters that follow ask five questions in a recommended
workflow order:
\begin{enumerate}
\item Which native loss scale is generated by the original elimination?
\item Are the local defect descriptions and lifted representations
structurally valid?
\item What regular, coordination, singular, or resource obstruction is forced
by the deployment contract?
\item Can finite data certify the relevant structural conclusion, and can the
declared downstream task observe it?
\item Which minimal structural intervention is predicted to remove the
obstruction, and does that intervention improve independent held-out risk?
\end{enumerate}

A structural explanation should be falsifiable.  A performance plateau alone is
not evidence of architecture obstruction: misspecification, optimization
failure, data scarcity, regularization, and metric ceilings can produce the
same curve.  A stronger empirical signature has several parts.  First, a
structural certificate predicts a nonzero floor under a fixed deployment
contract.  Second, capacity or compute increases that preserve that contract
do not remove the diagnosed gap.  Third, a mechanism-matched change of
representation, sharing, memory, or output semantics reduces the native
defect.  Finally, the gain survives independent validation and matched
controls.  Failure of any step weakens the structural interpretation.

The book therefore does not propose a replacement for approximation,
generalization, optimization, information theory, or representation learning.
It develops a complementary structural question: once local optima are
defined, what can a shared learning system coherently realize, what native
risk is unavoidable when it cannot, and what evidence justifies changing the
deployment contract?

\section*{Exercises}

\begin{exercise}
Construct a finite instance space on which every local optimum exists but a
constant deployment class has positive architecture obstruction.  Then enlarge
the deployment class minimally so that the obstruction vanishes.
\end{exercise}

\begin{exercise}
For
\[
v(\theta)=
\begin{pmatrix}\cos(\theta/2)\\ \sin(\theta/2)\end{pmatrix},
\qquad
P(\theta)=v(\theta)v(\theta)^\top,
\]
show that $P(0)=P(2\pi)$ while $v(2\pi)=-v(0)$.  Explain why the projector
representation changes the realizability problem without discarding the
eigenspace.
\end{exercise}

\begin{exercise}
Choose a modern learning system and write three separate questions:
local solvability, global realizability, and finite-sample certifiability.
Propose one negative control that could falsify a structural explanation.
\end{exercise}

%% file: chapters/ch02_certified_systems.tex
\chapter{Certified Elimination Systems}
\label{ch:certified-systems}
\section{The primitive interface}
\index{elimination!certified}
\index{fiber!auxiliary}
\index{oracle!fiber}
\index{defect!native}

The theory begins with a family of inner optimization problems.  Let $\X$ be an instance space.  Let $\pi:\mathcal E\to\X$ be a fibered auxiliary space with fiber $\A_x=\pi^{-1}(x)$.  A lifted objective is a function
\[
  H:\mathcal E\to\R\cup\{+\infty\},
  \qquad H_x(a)=H(x,a).
\]

\begin{definition}[Certified elimination]
\label{def:certified-elimination}
A \emph{certified elimination} is a triple $(H,J,\Def)$ satisfying
\[
  J(x)=\inf_{a\in\A_x}H_x(a)\in\R,
  \qquad
  \Def_x(a)=H_x(a)-J(x)\in[0,+\infty].
\]
Thus every auxiliary fiber is nonempty and its lifted objective is proper
and bounded below by a finite value; attainment is not required.  These
conditions make the residual well defined even when a trial state has
infinite lifted cost.
The oracle fiber is
\[
  \mathcal O(x)=\{a\in\A_x:\Def_x(a)=0\}.
\]
\end{definition}

The adjective \emph{certified} refers to the exact objective identity.  A distance from $a$ to a chosen optimizer may still be useful, but it is not a certificate until a proved exchange inequality links that distance to $H_x(a)-J(x)$.

\begin{example}[Quadratic profiling]
Let $H_x(a)=\frac12\|a-m(x)\|^2+c(x)$.  Then $J(x)=c(x)$, $\mathcal O(x)=\{m(x)\}$, and
\[
  \Def_x(a)=\frac12\|a-m(x)\|^2.
\]
Here Euclidean distance and native defect coincide up to scale.
\end{example}

\begin{example}[Nonidentifiable mixture]
Suppose $H_x(a)$ is invariant under a permutation group $G$.  The oracle fiber is an orbit rather than a point.  Parameter distance between two representatives may be positive while the native defect is zero.  The correct geometry lives on a quotient or an unordered output space.
\end{example}

\section{A complete learning contract}
\index{learning system!certified}
\index{architecture!deployment contract}
\index{task contract}
\index{statistical experiment}

A learning system is not determined by $H$ alone.  We use the following package.

\begin{definition}[Certified learning system]
A certified learning system is
\[
  \mathfrak E=
  (\X,P,\mathcal E,H,J,\Def,\Arch,\mathcal R,\Task,\mathcal P),
\]
where:
\begin{itemize}
  \item $(\X,P)$ is the population probability space;
  \item $\mathcal E\to\X$ is the auxiliary fibration;
  \item $(H,J,\Def)$ is a certified elimination;
  \item $\Arch$ is a class of $P$-measurable sections $A$ of
  $\mathcal E\to\X$ for which $x\mapsto\Def_x\{A(x)\}$ is measurable;
  \item $\mathcal R$ is the representation and resource grammar;
  \item $\Task$ is the legal downstream task/context contract;
  \item $\mathcal P$ is the statistical experiment generating observed data.
\end{itemize}
\end{definition}

The components play different roles: the objective determines the defect; deployment determines the architecture class; the task contract determines operational visibility; and the statistical experiment determines what can be certified.

\begin{boundarybox}
Changing the architecture or representation while keeping the objective fixed changes the obstruction but not the native defect.  Changing the lifted objective can change the statistical target itself.  These are distinct interventions.
\end{boundarybox}

\section{Pointwise, population, and uniform risk functionals}
\label{sec:architecture-risk-functionals}
\index{obstruction!population}
\index{obstruction!uniform}
\index{implementation gap}

For a deployed field $A\in\Arch$, define pointwise and population defects
\[
  \Def_A(x)=\Def_x\{A(x)\},
  \qquad
  \mathcal L_P(A)=\E_P\Def_A(X).
\]
The expectation is allowed to be extended-valued.  For the finite population decomposition below, assume that $\Arch$ is nonempty and contains at least one field of
finite population defect, and take $A$ itself to have finite population
defect.  Then the architecture obstruction is finite:
\[
  \Obs_P(\Arch)=\inf_{A\in\Arch}\mathcal L_P(A).
\]
Every deployed field satisfies the exact architecture decomposition
\[
  \mathcal L_P(A)
  =
  \Obs_P(\Arch)
  +
  \implgap_P(A;\Arch),
\]
where
\[
  \implgap_P(A;\Arch)
  =
  \mathcal L_P(A)-\Obs_P(\Arch)\ge0.
\]

The identity is simple, but conceptually decisive.  It separates an irreducible class-level floor from failure to reach the class optimum.

Uniform deployment uses
\[
  \mathcal L_\infty(A)=\sup_{x\in\X}\Def_A(x),
  \qquad
  \Obs_\infty(\Arch)=\inf_{A\in\Arch}\mathcal L_\infty(A).
\]
Uniform and average risks can differ sharply.  A continuous field may be forced to incur a fixed worst-case tax at a topological seam while concentrating that seam on a set of arbitrarily small $P$-mass.

\section{Regular and singular oracle geometry}
\index{oracle geometry!regular}
\index{oracle geometry!finite cover}
\index{oracle geometry!singular}
\index{oracle!set-valued}

The oracle incidence set
\[
  \mathcal O
  =
  \{(x,a):a\in\mathcal O(x)\}
  \subseteq \mathcal E
\]
can have several geometries.

\paragraph{Unique regular oracle.}
Each fiber contains one oracle $a^\star(x)$ and the map $x\mapsto a^\star(x)$ is regular.  Obstruction then comes from limited variation, resources, or decoder capacity.

\paragraph{Separated finite cover.}
Each fiber contains finitely many separated branches.  Local branches form a covering space; a global labeled section may fail because of monodromy.

\paragraph{Singular oracle family.}
Branches collide, disappear, or change multiplicity.  The incidence map is stratified rather than a covering.  Strong convexity and uniform branch separation fail, and the defect itself must price the singularity.

\paragraph{Set- or distribution-valued oracle.}
The natural oracle object may be an orbit, a projector, an unordered set, or a law.  A point-valued representation can introduce an artificial obstruction.

\section{Defect fibrations and fine-to-coarse maps}
\index{defect!fibration}
\index{defect!fine-to-coarse decomposition}
\index{fiber realization tax}

Suppose $\rho:\mathcal E\to\mathcal B$ maps a fine auxiliary object to a coarse one.  A useful elimination hierarchy has an exact decomposition
\[
  \Def_x^{\mathcal E}(e)
  =
  \Def_x^{\mathcal B}\{\rho(e)\}
  +
  \delta_x(e),
  \qquad \delta_x(e)\ge0.
\]
The coarse defect measures the price of choosing the wrong coarse state; the vertical term measures the price of realizing that coarse state in the fine fiber.

For a restricted fine architecture $\Arch_{\mathcal E}$, define the fiber realization tax
\[
  \Psi_x(b)
  =
  \inf\{\delta_x(e):e\in\Arch_{\mathcal E}(x),\ \rho(e)=b\}.
\]
Then the fine obstruction takes the infimal form
\[
  \Obs^{\mathcal E}
  =
  \inf_b\bigl\{\Def^{\mathcal B}(b)+\Psi(b)\bigr\}.
\]
This is the basic obstruction tower.  It is a Bellman or min-plus recursion over representation levels.

\begin{decompositionbox}
The tower separates two questions: which coarse state should be selected, and how much must the architecture pay to realize its canonical fine fiber?  An objective-faithful coarse representation need not be obstruction-faithful.
\end{decompositionbox}

\section{Common loss scales, attainment, and approximation}
\index{declared-loss scale rule}

A central discipline of elimination geometry is that different nonnegative quantities are not added merely because they appear in the same problem.

\begin{principlebox}
Two terms may be summed only if they are parts of an exact identity, an infimal decomposition, or a theorem that converts them to a common declared loss scale.
\end{principlebox}

For example, KL defect, topological degree, parameter distance, runtime, and task regret have different units and semantics.  A theorem may exchange one for another under strong convexity, data processing, an exposure gate, or a complexity model.  Without that theorem, the quantities should be reported separately.

\paragraph{Attainment and approximation.}
\index{attainment}
\index{approximate selection}

A zero architecture obstruction has several interpretations.
\begin{itemize}
  \item If the infimum is attained and equals zero, the architecture contains an exact oracle section.
  \item If the infimum is zero but unattained, the architecture approximates the oracle arbitrarily well in the declared risk.
  \item If the average obstruction is zero but the uniform obstruction is positive, the architecture can squeeze failure into a vanishing seam but cannot remove it everywhere.
\end{itemize}

Compactness, lower semicontinuity, coercivity, or finite-dimensional closure are needed for attainment.  The notation $\min$ must not replace $\inf$ without such a theorem.

\section*{Exercises}

\begin{exercise}
For a finite instance space and finite auxiliary fibers, prove that the population architecture obstruction is attained for every nonempty architecture class.
\end{exercise}

\begin{exercise}
Construct an example with $\Obs_P(\Arch)=0$ but $\Obs_\infty(\Arch)>0$.  Hint: use a continuous selector on a circle with a seam of shrinking measure.
\end{exercise}

\begin{exercise}
Let $\rho$ forget a label permutation in a mixture model.  Describe a coarse oracle object for which the coarse defect is zero while a labeled fine architecture pays a positive realization tax.
\end{exercise}

%% file: chapters/ch03_risk_decomposition.tex
\chapter{A Four-Component Decomposition of Population Risk}
\label{ch:four-components}
\index{risk decomposition!four-component}
\index{four-component risk decomposition!architecture obstruction component}

The same excess risk can have different causes, and those causes call for
different interventions.  The local statistical model may be inadequate; a
shared deployment contract may fail to realize the local oracle family;
finite data may not identify a good deployment; or the numerical procedure
may not have reached the best member of the declared class.  The decomposition
below separates these mechanisms by their reference objects before any
probability bound is applied.

\section{Reference risks and the four components}

Let $R^\star$ be the scientifically relevant population optimum.  Let
$R_{\rm oracle}$ be the population risk attained by the local oracle family
generated by the declared local model or variational approximation.  Let
$\Arch$ be the deployment class, and let $\widehat A$ be the learned
deployment.  Define
\[
  \modelgap=R_{\rm oracle}-R^\star,
  \qquad
  \archgap=\inf_{A\in\Arch}R(A)-R_{\rm oracle},
\]
and
\[
  \optgap_n
  =
  \widehat R_n(\widehat A)
  -
  \inf_{A\in\Arch}\widehat R_n(A).
\]
For a reference deployment $A^\dagger\in\Arch$, define
\[
  \widehat\rho_n(A^\dagger)
  =
  \widehat R_n(A^\dagger)
  -
  \inf_{A\in\Arch}\widehat R_n(A)
\]
and
\[
  \rho(A^\dagger)
  =
  R(A^\dagger)
  -
  \inf_{A\in\Arch}R(A).
\]

The four principal components have different meanings.

\begin{itemize}
\item $\modelgap$ is the \emph{model approximation component}: it compares the
best local oracle family with the scientific target.
\item $\archgap$ is the \emph{architecture obstruction component}: after the
local oracle has been fixed, it measures the smallest population loss forced
by the deployment contract.
\item The population--empirical deviations at $\widehat A$ and a comparator
form the \emph{generalization component}.
\item $\optgap_n$ is the \emph{implementation component}: it measures
empirical suboptimality within the declared deployment class.
\end{itemize}

All components that are added quantitatively must refer to a common risk
scale, or be connected by an explicit exchange theorem.  A KL defect, a
parameter distance, a topological degree, a runtime, and a scientific utility
difference are not automatically commensurable.

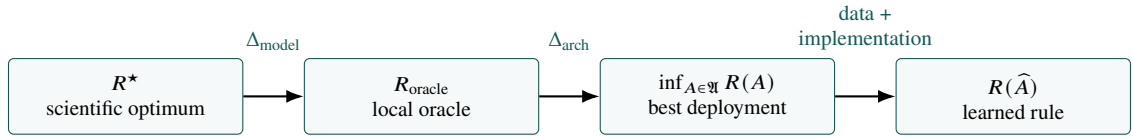
\begin{figure}[htbp]
\centering
\begin{tikzpicture}[>=Latex,node distance=8mm]
  \tikzset{
    rbox/.style={draw=egteal!75!black,fill=egteal!4,rounded corners=2pt,
      minimum height=10mm,minimum width=31mm,align=center,font=\scriptsize},
    gaplabel/.style={align=center,font=\scriptsize,text=egteal!80!black,
      inner sep=0pt}
  }
  \node[rbox] (star) {$R^\star$\\scientific optimum};
  \node[rbox,right=of star] (oracle) {$R_{\rm oracle}$\\local oracle};
  \node[rbox,right=of oracle] (arch) {$\inf_{A\in\Arch}R(A)$\\best deployment};
  \node[rbox,right=of arch] (learned) {$R(\widehat A)$\\learned rule};
  \draw[->,thick] (star)--
    node[midway,above=6mm,gaplabel]{$\Delta_{\rm model}$}(oracle);
  \draw[->,thick] (oracle)--
    node[midway,above=6mm,gaplabel]{$\Delta_{\rm arch}$}(arch);
  \draw[->,thick] (arch)--
    node[midway,above=6mm,gaplabel]{data +\\implementation}(learned);
\end{tikzpicture}
\caption{Reference risks in the four-component decomposition. The first two gaps are population-level structural terms; generalization and implementation control the distance from the class optimum to the learned rule.}
\label{fig:risk-references}
\end{figure}

\begin{conditionbox}
The population and empirical class infima and every risk evaluation appearing
in the identity are finite, and $A^\dagger$ belongs to $\Arch$; hence no
undefined extended-real subtraction occurs.  No
stochastic assumption or minimizer attainment is needed for the exact
identity below.  The generalization bracket and the comparator correction may
have either sign; only the architecture and model components are structural
population floors under the stated nesting of targets.
\end{conditionbox}

\begin{proposition}[Exact four-component risk identity]
\label{prop:four-component-identity}
For any reference $A^\dagger\in\Arch$,
\begin{align*}
R(\widehat A)-R^\star
={}&\modelgap+\archgap\\
&+\Bigl[R(\widehat A)-\widehat R_n(\widehat A)
      +\widehat R_n(A^\dagger)-R(A^\dagger)\Bigr]\\
&+\optgap_n
+\rho(A^\dagger)-\widehat\rho_n(A^\dagger).
\end{align*}
If $A^\dagger$ is a population minimizer, then
$\rho(A^\dagger)=0$; if it is an empirical minimizer, then
$\widehat\rho_n(A^\dagger)=0$.
\end{proposition}

\begin{interpretationbox}
The proposition is an algebraic identity, not a probabilistic bound.  Its
role is to fix the reference objects.  In particular, the architecture
obstruction is measured relative to an already declared local oracle family,
so it cannot be silently merged with local model misspecification.
\end{interpretationbox}

\begin{proofroadmap}
Insert, in order, $R_{\rm oracle}$, $\inf_{A\in\Arch}R(A)$,
$R(A^\dagger)$, $\widehat R_n(A^\dagger)$, the empirical class infimum, and
$\widehat R_n(\widehat A)$.  The chapter appendix gives the full telescoping
calculation.
\end{proofroadmap}

When the population risk has the native form
\[
  R(A)
  =
  R_{\rm oracle}
  +
  \mathbb E\Def_X\{A(X)\},
\]
the second component becomes exactly
\[
  \archgap
  =
  \inf_{A\in\Arch}\mathbb E\Def_X\{A(X)\}
  =
  \Obs_P(\Arch).
\]
This is the principal interface between elimination geometry and statistical
learning: generic approximation relative to $R^\star$ is refined into local
model inadequacy and structural nonrealizability under shared deployment.

In amortized variational inference, the distinction is especially simple.
Fix the decoder and local variational family and take
$H_x=-\operatorname{ELBO}_x$.  Then $\Def_x$ is the per-datum amortization
gap of \citet{CremerLiDuvenaud2018}, while $\archgap$ is the smallest
population-averaged amortization gap attainable by the declared encoder
class---not the gap of a particular fitted encoder.  For a deployed encoder
$A$, the population-averaged amortization gap is therefore the sum of this
class-level floor and a within-class residual,
$\archgap+\implgap_P(A;\Arch)$.  \citet{MargossianBlei2024} characterize when
the first term can vanish under the sharing rule they study: one deterministic
inference function must produce all local variational parameters.

\section{A high-probability learning bound}

Assume that, with probability at least $1-\delta$,
\[
  \sup_{A\in\Arch}
  |R(A)-\widehat R_n(A)|
  \le
  \varepsilon_n(\delta),
\]
and that the training algorithm satisfies
\[
  \widehat R_n(\widehat A)
  \le
  \inf_{A\in\Arch}\widehat R_n(A)
  +
  \eta_n.
\]

\begin{conditionbox}
The deployment class is fixed on the event being analyzed, the
population--empirical deviation is uniform over that class, and the training
procedure is an $\eta_n$-approximate empirical minimizer.  In addition,
$R^\star\le R_{\rm oracle}\le\inf_{A\in\Arch}R(A)$ so that the model and
architecture components are nonnegative.
\end{conditionbox}

\begin{theorem}[Four-component certified learning bound]
\label{thm:four-component-certified}
On the uniform generalization event,
\[
  \modelgap+\archgap
  \le
  R(\widehat A)-R^\star
  \le
  \modelgap+\archgap
  +
  2\varepsilon_n(\delta)+\eta_n.
\]
\end{theorem}

\begin{interpretationbox}
The theorem places the learned excess risk around a structural population
floor.  More data can reduce $\varepsilon_n(\delta)$ and better optimization
can reduce $\eta_n$; neither operation changes a positive
$\modelgap+\archgap$ unless the local model or deployment contract itself is
changed.  If the architecture is selected from the same data, the fixed-class
uniform event must be replaced by sample splitting, a finite-library bound,
stability, or a data-dependent random-set argument; PAC--Bayesian bounds for
random hypothesis sets give one such route \citep{DupuisEtAl2024}.
\end{interpretationbox}

\begin{proofroadmap}
The lower bound uses only $\widehat A\in\Arch$.  For the upper bound, compare
$\widehat A$ with an empirical class minimizer or a minimizing sequence and
apply the uniform deviation twice.  The appendix gives the argument without
assuming attainment.
\end{proofroadmap}

\section{Structural meaning and mechanism coupling}

The decomposition is exact, but its mechanisms are not independent.
Enlarging a local variational family can reduce $\modelgap$ while producing a
multibranch oracle family that increases $\archgap$.  A data-driven atlas
makes the deployment class itself random.  Test-time computation can reduce
implementation error while also enlarging the effective deployment class.
A nonconvex optimizer may reach only a subset of a nominal architecture class.

The correct use of the decomposition is therefore diagnostic, not merely
algebraic.  The order of analysis is:
\[
  \text{local oracle}
  \longrightarrow
  \text{architecture floor}
  \longrightarrow
  \text{finite-sample control}
  \longrightarrow
  \text{implementation residual}.
\]
The intervention then follows the dominant mechanism.

\begin{center}
\begin{tabularx}{\textwidth}{@{}p{0.22\textwidth}p{0.30\textwidth}X@{}}
\toprule
Component & Typical cause & First intervention to examine\\
\midrule
Model approximation & local family omits relevant structure &
change the local model or scientific target\\
Architecture obstruction & representation, sharing, memory, regularity, or
resource restriction & change the deployment contract\\
Generalization & limited independent data or data-dependent selection &
collect information or control statistical complexity\\
Implementation & incomplete training or numerical solution &
improve the solver or computation budget\\
\bottomrule
\end{tabularx}
\end{center}

A positive architecture obstruction can itself admit more refined exact or
infimal decompositions.  Carrier information loss and decoder nonsaturation
are separated in Chapter~\ref{ch:resource-rd}; rectangularization and
coordination are separated in Chapter~\ref{ch:coordination}; fine-to-coarse
representations generate fiber realization costs in
Chapter~\ref{ch:certified-systems}.  These decompositions are used only under
their declared assumptions.  They are not combined into a universal
multi-term formula without an exact identity or a proved exchange bound.

\begin{boundarybox}
Different loss scales cannot be added by analogy.  Topological degree, native
KL defect, parameter error, communication, runtime, and downstream utility
remain distinct quantities until a theorem converts one into another.
\end{boundarybox}

\section{What an empirical study should separate}

A convincing structural study should report evidence for the four components
rather than infer mechanism from one performance gap.

\begin{table}[H]
\centering
\caption{Evidence associated with the four components of risk.}
\begin{tabularx}{\textwidth}{p{0.22\textwidth}X}
\toprule
Component & Diagnostic evidence\\
\midrule
Model approximation & stronger local oracle family, oracle refit, simulation
truth, or local-family comparison\\
Architecture obstruction & transport or monodromy certificate, coordination
decomposition, resource lower bound, or zero-obstruction witness\\
Generalization & independent-unit split, confidence sequence, stability, or
data-dependent complexity bound\\
Implementation & primal--dual gap, KKT residual, multiple restarts, or
certified inner-solver tolerance\\
\bottomrule
\end{tabularx}
\end{table}

A performance plateau is therefore only a symptom.  It supports a structural
interpretation only after local model inadequacy, finite-sample uncertainty,
and implementation error have been separately controlled.
The prospectively frozen MNIST audit in
Laboratory~\ref{sec:lab-mnist-amortization} applies this separation to a
finite one-pass inference grammar: it holds the local variational contract
fixed, uses held-out selection and evaluation, audits the imagewise reference,
and tests data, width, and training-effort alternatives before changing the
deployment computation.

\section*{Exercises}

\begin{exercise}
Verify the exact four-component identity term by term.  Simplify it when
$A^\dagger$ is a population minimizer and when it is an empirical minimizer.
\end{exercise}

\begin{exercise}
Suppose $\Arch_1\subseteq\Arch_2$.  Prove that the architecture obstruction is
monotone nonincreasing.  Explain why the generalization component need not be.
\end{exercise}

\begin{exercise}
Construct a problem in which $\modelgap=0$, $\archgap>0$, and both the
generalization and implementation terms vanish asymptotically.
\end{exercise}

\input{chapter_appendices/ch03_proofs}

%% file: chapter_appendices/ch03_proofs.tex
\chapterproofappendix

\subsection*{Proof of the exact four-component risk identity}
\proofdependency{Only the definitions of the four components and finiteness of every displayed population/empirical evaluation and class infimum are used.  No stochastic assumption and no attainment of the architecture infimum are needed.}
\begin{proof}
Write
\[
R(\widehat A)-R^\star
=\{R_{\rm oracle}-R^\star\}
 +\{\inf_{A\in\Arch}R(A)-R_{\rm oracle}\}
 +\{R(\widehat A)-\inf_{A\in\Arch}R(A)\}.
\]
For an arbitrary reference $A^\dagger\in\Arch$, decompose the last bracket by adding and subtracting $R(A^\dagger)$, $\widehat R_n(A^\dagger)$, the empirical class infimum, and $\widehat R_n(\widehat A)$:
\begin{align*}
R(\widehat A)-\inf_{A\in\Arch}R(A)
={}&R(\widehat A)-\widehat R_n(\widehat A)\\
&+\left\{\widehat R_n(\widehat A)
       -\inf_{A\in\Arch}\widehat R_n(A)\right\}\\
&-\left\{\widehat R_n(A^\dagger)
       -\inf_{A\in\Arch}\widehat R_n(A)\right\}\\
&+\widehat R_n(A^\dagger)-R(A^\dagger)\\
&+R(A^\dagger)-\inf_{A\in\Arch}R(A).
\end{align*}
The first two population brackets are respectively $\modelgap$ and $\archgap$; the two empirical-infimum brackets are $\optgap_n$ and $\widehat\rho_n(A^\dagger)$; and the final population term is $\rho(A^\dagger)$.  Substitution proves the identity.  The two residuals vanish under population and empirical attainment, respectively.
\end{proof}

\subsection*{Proof of the four-component certified learning bound}
\proofdependency{The lower bound uses only membership $\widehat A\in\Arch$.  The upper bound uses the uniform generalization event and the empirical optimization residual.  Population attainment is not required.}
\begin{proof}
Since $\widehat A\in\Arch$,
\[
R(\widehat A)\ge \inf_{A\in\Arch}R(A)
=R^\star+\modelgap+\archgap,
\]
which proves the lower bound.

For the upper bound, let $(A_k)_{k\ge1}\subseteq\Arch$ be a minimizing sequence such that
\[
R(A_k)\downarrow\inf_{A\in\Arch}R(A).
\]
On the event
\(
\sup_{A\in\Arch}|R(A)-\widehat R_n(A)|\le\varepsilon_n(\delta)
\),
we have, for every $k$,
\begin{align*}
R(\widehat A)
&\le \widehat R_n(\widehat A)+\varepsilon_n(\delta)\\
&\le \inf_{A\in\Arch}\widehat R_n(A)+\eta_n+\varepsilon_n(\delta)\\
&\le \widehat R_n(A_k)+\eta_n+\varepsilon_n(\delta)\\
&\le R(A_k)+2\varepsilon_n(\delta)+\eta_n.
\end{align*}
Letting $k\to\infty$ gives
\[
R(\widehat A)
\le \inf_{A\in\Arch}R(A)+2\varepsilon_n(\delta)+\eta_n.
\]
Subtracting $R^\star$ and using
\(
\inf_{A\in\Arch}R(A)-R^\star=\modelgap+\archgap
\)
proves the upper bound.
\end{proof}

%% file: chapters/ch04_laboratories.tex
\chapter{Four Running Laboratories}
\label{ch:laboratories}
\section{Four recurring structural problems}
\index{running laboratories}

Four recurring laboratories keep the abstract machinery tied to concrete
failure modes.  Each isolates one obstruction and its matched repair;
Chapter~\ref{ch:applied-laboratories} later recombines them inside more complex
scientific interfaces.

\begin{enumerate}
  \item Oracle-preserving coarse-graining illustrates carrier information loss, safe quotienting, back-mapping, and representation refinement.
  \item Graph-indexed distribution learning illustrates signed high-order smoothing, probability validity, nonlinear projection, and sharp statistical rates.
  \item Sequential conditional decision making illustrates rectangularity, shared-parameter coordination, memory compression, and operational context.
  \item One-pass prediction illustrates architecture obstruction, local refinement, and an exact computation--defect frontier indexed by deployment budget.
\end{enumerate}

The same terminology will mean the same thing in all four settings.

\section{Laboratory I: oracle-preserving coarse-graining}
\index{coarse-graining!oracle-preserving}
\index{carrier!coarsening tax}

\emph{What may a representation safely forget?}
Let a fine state $\widetilde X$ have oracle output $a_{\widetilde X}^\star$,
and let $Z=q(\widetilde X)$ be the carrier exposed to a decoder.  Under
squared native loss, the best unrestricted decoder returns
$\E(a_{\widetilde X}^\star\mid Z)$ and pays the carrier-information component
\[
  \mathcal C(Z)
  =\frac12\E\left\|
    a_{\widetilde X}^\star-\E(a_{\widetilde X}^\star\mid Z)
  \right\|^2.
\]
If $W=h(Z)$ is a further coarsening, the tower property gives the exact tax
\[
  \mathcal C(W)-\mathcal C(Z)
  =\frac12\E\left\|
    \E(a_{\widetilde X}^\star\mid Z)
    -\E(a_{\widetilde X}^\star\mid W)
  \right\|^2.
\]
Thus forgetting is free exactly when $W$ preserves the conditional oracle
signature; raw state count and hidden dimension are not the relevant capacity.

Molecular coarse-graining supplies a concrete fiber model.  A reaction
coordinate $z=q(\widetilde x)$ descends a fine potential $U$ to the potential
of mean force
\[
  \mathsf W_\beta(z)
  =-\frac1\beta\log\left[
    v(z)\int_{q^{-1}(z)}e^{-\beta U(\widetilde x)}
    \,\nu_z(d\widetilde x)
  \right].
\]
Here $v(z)\nu_z$ records the declared disintegration of the fine reference
measure.

If $\widetilde\Pi_\beta$ is the fine Gibbs law,
$\Pi_\beta^{\mathsf W}$ its coarse marginal, and $\eta_{\beta,z}$ its oracle
fiber kernel, then any trial fine law $Q$, with coarse marginal $P$ and
conditional back-mapping $Q_z$, obeys the exact conditional-KL decomposition
\[
  \frac1\beta\KL(Q\|\widetilde\Pi_\beta)
  =\frac1\beta\KL(P\|\Pi_\beta^{\mathsf W})
   +\frac1\beta\E_P\KL(Q_z\|\eta_{\beta,z}).
\]
When $P$ is induced by a learned coarse potential, enriching that potential
can reduce only the first component.  The second is controlled by the
conditional back-map; if the scientific oracle requires distinctions not
measurable from $z$, no coarse predictor can restore them.  The matched repair
is to refine the reaction coordinate or retain a faithful conditional
back-mapping, according to which component is dominant.

Not every quotient is destructive.  Replacing a subspace frame $U$ by the
projector $UU^\top$ removes only the $U\sim UQ$ basis gauge and preserves the
subspace oracle.  The criterion is always oracle survival, not whether the
representation became smaller.

\paragraph{Structural prediction.}
Scaling a predictor behind a fixed coarse carrier saturates at $\mathcal C(Z)$
when distinct conditional oracle means have been merged.  Witness-directed
carrier refinement can reduce that floor.  For fine-law reconstruction, a
faithful stochastic back-mapping instead removes the separate conditional-fiber
term; dummy states and irrelevant gauge coordinates reduce neither.

\section{Laboratory II: graph-indexed distribution learning}
\index{graph-indexed learning!distributional prediction}
\index{continuous ranked probability score}

Let $t=1,\ldots,T$ index vertices of a graph with Laplacian $L$.  At vertex $t$, observations $Y_{t1},\ldots,Y_{tn_t}$ define an empirical CDF $\widehat F_t$.  A graph filter $S$ produces a raw field
\[
  \widetilde F=S\widehat F.
\]
For higher-order qualification, the filter may have signed off-diagonal weights.  Then $\widetilde F_t$ need not be a CDF.

Fix a compact outcome interval $I=[a,b]$ and work in the product space
$L^2(I)^T$.  The valid set
\[
  \mathcal C=\{F: F_t \text{ is a CDF for every }t\}
\]
is closed and convex in that space.  Projection
\[
  F=\Pi_{\mathcal C}(\widetilde F)
\]
restores validity.  Because the truth $F^\star$ belongs to $\mathcal C$,
\[
  \|F-F^\star\|^2
  \le
  \|\widetilde F-F^\star\|^2.
\]
The projection can also contract graph roughness under the declared product geometry.
On an unbounded outcome line, CDFs themselves need not lie in $L^2$; the
corresponding statement must instead be formulated for an affine difference
class (for example, functions whose difference from a fixed reference CDF is
square integrable).

The laboratory separates three objects that are often conflated:
\begin{enumerate}
  \item the raw ambient variational solution;
  \item the nonlinear validity-repaired field;
  \item the constrained variational optimizer over the CDF set.
\end{enumerate}
The first and second are not generally the same as the third.

\paragraph{Structural prediction.}
For smoothness beyond the graph-universal Markov range, linear high-order validity fails.  Nonlinear repair can retain the statistical rate while restoring the semantic contract.

\section{Laboratory III: shared conditional decisions}
\index{conditional policy!shared deployment}
\index{rectangularity}

Consider a finite history tree.  At history $h$, a local oracle kernel is
$q_h^\star(\cdot\mid h)$.  The formal rectangular-hull construction is given
in Chapter~\ref{ch:coordination}; here we use only its motivating contrast.  A
fully rectangular architecture can choose every conditional kernel
independently, whereas a shared architecture may require
\[
  q_h(\cdot\mid h)=q_\theta(\cdot\mid \phi(h))
\]
for one parameter $\theta$ and perhaps one compressed memory state $\phi(h)$.

The rectangular hull $\operatorname{Rect}(\Arch)$, defined formally in Chapter~\ref{ch:coordination}, frees the locally available kernels and allows independent pasting.  The rectangular value obeys an exact Bellman recursion.  The gap
\[
  \coordtax(\Arch)
  =
  \Obs(\Arch)-\Obs\{\operatorname{Rect}(\Arch)\}
\]
prices the cross-history coupling induced by sharing.

A coarse memory representation can be objective-faithful for one task while failing to realize the canonical fine conditional fibers.  The resulting fiber tax depends on occupancy and on the declared oracle law.

\paragraph{Structural prediction.}
Increasing network width without refining memory may not remove the coordination tax.  Rectangularization, memory refinement, or a task-specific quotient can.

\section{Laboratory IV: one-pass prediction and local refinement}
\sectionmark{Laboratory IV: one-pass refinement}
\label{sec:lab-one-pass-refinement}
\index{one-pass prediction}
\index{test-time refinement!quadratic laboratory}
\index{amortized inference!resource frontier}

\emph{When does test-time computation change the deployed class rather than
merely improve training?}
At instance $x$, consider the scalar local problem
\[
  H_x(a)=\frac12\{a-m(x)\}^2,
  \qquad
  J(x)=\inf_a H_x(a)=0,
  \qquad
  \Def_x(a)=H_x(a)-J(x).
\]
The local oracle is $a_x^\star=m(x)$.  Let $P$ be the deployment law and let
$\Arch_0$ be a declared class of one-pass predictors.  Its architecture
obstruction is
\[
  \Obs_P(\Arch_0)
  =\inf_{A\in\Arch_0}\E_P\Def_X\{A(X)\}.
\]
This infimum is architecture-relative by definition.  A measured positive
residual estimates it only after reachability error inside $\Arch_0$ has been
controlled.

Starting from $a_0=A(x)$, now permit exact local-gradient steps with a fixed
$0<\eta<1$:
\[
  a_{j+1}=a_j-\eta\{a_j-m(x)\}.
\]
The recursion is explicit,
\[
  a_j=m(x)+(1-\eta)^j\{A(x)-m(x)\},
  \qquad
  \Def_x(a_j)=(1-\eta)^{2j}\Def_x\{A(x)\}.
\]
Write $R_jA$ for the resulting predictor and define the budgeted deployment
class by
\[
  \Arch_{\le k}
  =\{R_jA:A\in\Arch_0,\ 0\le j\le k\},
\]
where $j$ is integer.  These classes are nested, and the pointwise identity
gives the exact class-level resource--defect frontier
\begin{equation}
  \boxed{
  \Obs_P(\Arch_{\le k})
  =(1-\eta)^{2k}\Obs_P(\Arch_0).}
  \label{eq:one-pass-refinement-frontier}
\end{equation}

\paragraph{An exact three-point ledger.}
Let $P$ be uniform on $\{-1,0,1\}$, set $m(x)=x^2$, and take the one-pass
grammar
\[
  \Arch_0=\Arch_{\mathrm{aff}}
  =\{x\mapsto\alpha+\beta x:\alpha,\beta\in\mathbb R\}.
\]
Its exact population risk is
\[
  \E_P\Def_X\{\alpha+\beta X\}
  =\frac16\left[
    (\alpha-\beta-1)^2+\alpha^2+(\alpha+\beta-1)^2
  \right].
\]
The unique minimizer is $(\alpha^\star,\beta^\star)=(2/3,0)$, so
\[
  \Obs_P(\Arch_{\mathrm{aff}})=\frac19\approx0.111111.
\]
With $\eta=1/2$, the class-level identity gives the complete numerical
ledger in Table~\ref{tab:three-point-refinement}.  These are exact population
values under the declared three-point law, not Monte Carlo estimates.

\begin{table}[H]
\centering
\caption{Exact one-pass and local-refinement ledger for the three-point
quadratic laboratory.}
\label{tab:three-point-refinement}
\small
\begin{tabularx}{\textwidth}{@{}Xccc@{}}
\toprule
Deployment contract & Step budget $k$ & Residual factor & Exact minimum (decimal)\\
\midrule
Affine one pass & $0$ & $1$ & $1/9\;(0.111111)$\\
Affine plus local refinement & $1$ & $1/4$ & $1/36\;(0.027778)$\\
Affine plus local refinement & $2$ & $1/16$ & $1/144\;(0.006944)$\\
Affine plus local refinement & $4$ & $1/256$ & $1/2304\;(0.000434)$\\
Quadratic-feature one pass & $0$ & --- & $0\;(0.000000)$\\
\bottomrule
\end{tabularx}
\end{table}

The last row enlarges the one-pass grammar to
$\{\alpha+\beta x+\gamma x^2\}$ and realizes the oracle with
$(\alpha,\beta,\gamma)=(0,0,1)$.  It therefore removes the same baseline
obstruction through a representation change rather than test-time compute.
Operationally the refined system runs a local optimizer; structurally it has
moved from $\Arch_0$ to the computation-indexed contract $\Arch_{\le k}$,
not merely trained the same class more accurately.

\paragraph{Interface accounting.}
Changing predictor width or features changes $\Arch_0$; adding training data
changes estimation; improving the one-pass optimizer changes reachability
inside $\Arch_0$; and increasing $k$ changes the deployment computation
budget.  Calling all four changes ``more model capacity'' loses the
intervention map.

\paragraph{Structural prediction and boundary.}
If a well-optimized one-pass class has $\Obs_P(\Arch_0)>0$, local refinement
traces the monotone frontier in
\eqref{eq:one-pass-refinement-frontier}.  The quadratic calculation transfers
an existing baseline floor; it does not prove that the floor is positive.
For fixed-variance Gaussian variational inference, the negative-ELBO defect
is this same laboratory up to a constant scale factor.  Nonquadratic problems
replace the equality by a curvature-controlled bound or an empirical audit;
Chapter~\ref{ch:applied-laboratories} later applies that audit logic to
learned one-pass inference on MNIST.

\Needspace{26\baselineskip}
\section{A common table}
\index{running laboratories!cross-laboratory comparison}

The four laboratories can now be compared row by row.  The table aligns
their oracle, native defect, deployment contract, obstruction, repair, and
statistical evidence before the final cross-laboratory lesson.

\begin{table}[H]
\centering
\caption{The four laboratories under one structural vocabulary.}
\small
\begin{tabularx}{\textwidth}{@{}
  >{\raggedright\arraybackslash}p{0.15\textwidth}
  >{\raggedright\arraybackslash}X
  >{\raggedright\arraybackslash}X
  >{\raggedright\arraybackslash}X@{}}
\toprule
Laboratory & Oracle and native defect & Contract and obstruction & Repair and evidence\\
\midrule
Coarse-graining
  & conditional oracle/fiber law; carrier loss/fiber KL
  & carrier plus decoder/back-map; lost oracle distinctions
  & refine carrier/faithful back-map; marginal and transfer risk\\
Graph distributions
  & vertexwise CDF; integrated squared-CDF loss
  & graph filter plus validity interface; validity barrier
  & nonlinear projection; CRPS/minimax rate\\
Sequential decisions
  & historywise kernel; conditional KL or Bellman defect
  & shared network/memory; coordination/memory tax
  & rectangular hull/memory refinement; held-out policy value\\
One-pass refinement
  & instancewise local optimum; excess local objective
  & one-pass class plus step budget; architecture floor
  & representation/local refinement; exact frontier and held-out audit\\
\bottomrule
\end{tabularx}
\end{table}

\paragraph{Why one example is not enough.}

A theory based only on coarse-graining could be dismissed as information
compression.  A theory based only on graph CDFs could be dismissed as
shape-constrained smoothing.  A theory based only on sequential kernels could
be dismissed as robust control or memory approximation.  A theory based only
on one-pass refinement could be dismissed as amortized inference or local
optimization.  The common lesson is deliberately limited: the same decomposition discipline
applies in all four laboratories, while the obstruction mechanism and the
repair remain model specific.

\section*{Exercises}

\begin{exercise}
Under squared native loss, prove the displayed coarsening identity for
$W=h(Z)$.  Give one coarsening with zero tax and one that merges distinct
conditional oracle means and therefore has positive tax.
\end{exercise}

\begin{exercise}
Show that the coordinatewise projection of an arbitrary function onto the set of CDFs is nonexpansive in $L^2$.  Which step uses convexity of the CDF set?
\end{exercise}

\begin{exercise}
Construct a two-history shared-kernel example in which both local oracle kernels belong to the architecture locally, but no single shared parameter realizes them simultaneously.
\end{exercise}

\begin{exercise}
For the three-point one-pass laboratory, derive the optimizer $(2/3,0)$ and
the obstruction $1/9$.  Prove the class-level frontier for general
$0<\eta<1$, verify the $k=1,2,4$ rows of
Table~\ref{tab:three-point-refinement}, and explain why adding the quadratic
feature changes $\Arch_0$ whereas increasing $k$ changes the deployment
computation budget.
\end{exercise}

%% file: part_notes/part01_history.tex
\parthistoricalnotes{}
\index{four-component risk decomposition!historical comparison}

\subsection*{Where the four accounts sit in learning theory}

Classical statistical learning separates approximation and estimation, and
computational analyses add optimization error
\citep{BartlettMendelson2002,BousquetElisseeff2002,BottouBousquet2008}.
Amortized inference then makes a further distinction between the best member
of a local variational family and the output delivered by a shared inference
mechanism \citep{CremerLiDuvenaud2018,Amos2023,MargossianBlei2024}.  Part I
does not claim priority for these decompositions or for the telescoping
identity in Proposition~\ref{prop:four-component-identity}.

The narrower book contribution is to insert a contract-relative reference
between local approximation and learned deployment:
\[
R^\star
\longrightarrow R_{\rm oracle}
\longrightarrow \inf_{A\in\Arch}R(A)
\longrightarrow R(\widehat A).
\]
The first arrow measures whether the declared local family is scientifically
adequate.  The second asks whether one declared global architecture can
realize the already-fixed local oracle family.  The last arrow is governed by
data and implementation.  This split can coincide with an amortization gap
in a variational-inference example, but it is not defined by amortization: it
also applies to shared policies, finite memories, continuous selectors,
compressed carriers, and other deployment contracts.

\begin{center}
\small
\begin{tabularx}{0.96\textwidth}{@{}p{0.23\textwidth}p{0.31\textwidth}X@{}}
\toprule
Account & Historical neighbor & Part I distinction \\
\midrule
Local model gap & approximation or variational-family gap & compares the
local oracle with the scientific target before global sharing is imposed \\
Architecture gap & amortization, approximation under a restricted function
class & fixes one deployment contract and prices failure to realize the
whole oracle field in population-risk units \\
Generalization & estimation and uniform convergence & controls population
versus empirical behavior for the declared, possibly selected, class \\
Implementation & optimization or reachability error & compares the fitted
rule with the empirical class optimum and is not folded into structural
nonrealizability \\
\bottomrule
\end{tabularx}
\end{center}

\subsection*{Ownership of the Part I results}

The exact identity is a book-level accounting synthesis.  The certified
bound is the standard two-deviation empirical-risk argument specialized to
that accounting chain.  Their value is diagnostic separation: changing the
local family, the shared architecture, the data, or the optimizer changes a
different reference comparison.  The detailed row-by-row provenance is in
Appendix~\ref{app:principal-result-audit}.

%% file: chapters/ch05_conjugate_lifts.tex
\chapter{Conjugate Lifts and Native Loss Scales}
\label{ch:conjugate-lifts}
\index{defect!objective-generated}
\index{native loss scale}

Suppose an inner state is introduced because it makes an objective easier to
optimize.  After elimination, a trial inner state should be evaluated in the
same units as the objective.  A Euclidean penalty may be algorithmically
convenient, but it is not automatically the insertion price.

The canonical residual is
\[
  \Def_x(a)=H_x(a)-\inf_b H_x(b).
\]
In conjugate-complete lifts this residual is a directed Bregman divergence.  This observation fixes the orientation of KL and other asymmetric divergences.

\section{Legendre conjugacy}
\index{Legendre conjugacy}
\index{Fenchel conjugacy}
\index{Bregman geometry!divergence}

Let $\Phi:\mathcal A\to\R\cup\{+\infty\}$ be a proper, closed, strictly convex Legendre function.  Its conjugate is
\[
  \Phi^\star(u)=\sup_{a}\{\langle u,a\rangle-\Phi(a)\}.
\]
For $u$ in the interior of the dual domain,
\[
  a^\star(u)=\nabla\Phi^\star(u)
\]
solves
\[
  \inf_a\{\Phi(a)-\langle u,a\rangle\}
  =-\Phi^\star(u).
\]
The Bregman divergence generated by $\Phi$ is \citep{Bregman1967}
\[
  D_\Phi(a\|b)
  =
  \Phi(a)-\Phi(b)-\langle\nabla\Phi(b),a-b\rangle.
\]

\begin{conditionbox}
The generator $\Phi$ is differentiable and strictly convex on the relevant domain, and the lift is conjugate-complete: every auxiliary occurrence is paired with the full conjugate term $\Phi^*$.  The pointwise oracle $a^\star$ lies in the common Legendre domain.  These assumptions fix both the nonnegativity and the orientation of the Bregman divergence.
\end{conditionbox}

\begin{theorem}[Conjugate defect identity]
\label{thm:conjugate-defect}
Let
\[
  H_u(a)=\Phi(a)-\langle u,a\rangle+c(u).
\]
Then the eliminated objective is
\[
  J(u)=c(u)-\Phi^\star(u),
\]
and
\[
  H_u(a)-J(u)
  =
  D_\Phi\{a\|a^\star(u)\}.
\]
\end{theorem}

\begin{interpretationbox}
The theorem says that the insertion price is not a chosen regularizer: it is exactly the directed Bregman divergence generated by the eliminated objective.  Reversing the arguments generally changes the loss.  In the entropy case this distinction fixes the correct direction of KL used by the algorithm.
\end{interpretationbox}

\begin{proofroadmap}
Use the Fenchel--Young equality at the oracle and subtract the lifted objective evaluated at a trial auxiliary state.  Rearranging the conjugate terms produces the three-point definition of $D_\Phi(a\|a^\star)$.  The chapter appendix supplies the full calculation.
\end{proofroadmap}

The order of the arguments is not cosmetic.  The trial state appears in the first argument and the oracle in the second.  Reversing the divergence generally changes both geometry and performance guarantees.

\section{Examples of native Bregman defects}

\subsection{Relative entropy}
\index{relative entropy}
\index{Bregman geometry!relative entropy}

For probability vectors $q$ and $p$ on the same finite set, take
\[
  \Phi(q)=\sum_z q_z\log q_z.
\]
Then
\[
  D_\Phi(q\|p)=\KL(q\|p).
\]
Classical \(I\)-divergence geometry supplies the projection background for
this probability-simplex example \citep{Csiszar1975}.  A variational lift
based on entropy nevertheless generates one particular KL direction.  The
reverse direction is not certified unless it arises from a different
elimination.

\subsection{LogDet divergence}
\index{LogDet divergence}

For positive-definite matrices $X,Y\in\mathbb S_{++}^{m}$, take
\[
  \Phi(X)=-\log\det X.
\]
The corresponding Bregman divergence is
\[
  D_\Phi(X\|Y)
  =
  \operatorname{tr}(Y^{-1}X)-\log\det(Y^{-1}X)-m.
\]
This is the Burg or LogDet divergence \citep{KulisSustikDhillon2009}.  Its natural centroid and consensus endpoint differ from Euclidean matrix averaging.

\subsection{Quantum relative entropy}
\index{quantum theory!relative entropy}

On the faithful density-matrix domain, the entropy functional
\[
  \Phi(\rho)=\operatorname{tr}(\rho\log\rho)
\]
induces Umegaki relative entropy \citep{Umegaki1962}
\[
  D(\rho\|\sigma)
  =
  \operatorname{tr}\rho(\log\rho-\log\sigma).
\]

For general density matrices, this formula is understood through its
lower-semicontinuous extension: $D(\rho\|\sigma)<\infty$ only when
$\operatorname{supp}(\rho)\subseteq\operatorname{supp}(\sigma)$, and it is
$+\infty$ otherwise.  On the faithful domain both logarithms and the Bregman
gradient are ordinary finite-dimensional operators.
The noncommutative case preserves the defect orientation but changes the
validity of base change---the fine-to-coarse passage formalized in
Chapter~\ref{ch:composition}---and measurement.

\section{Further certified eliminations and boundaries}
\paragraph{EM as a certified elimination.}
\index{expectation--maximization}
\index{elimination!certified}

Let $p_\theta(x,z)$ be a latent-variable model.  For a trial conditional law $q(z)$ define
\[
  H_x(\theta,q)
  =
  -\E_q\log p_\theta(x,Z)
  -\mathsf H(q),
\]
where $\mathsf H(q)=-\E_q\log q(Z)$.  The Gibbs variational identity gives
\[
  -\log p_\theta(x)
  =
  \inf_q H_x(\theta,q).
\]
Moreover,
\[
  H_x(\theta,q)+\log p_\theta(x)
  =
  \KL\{q\|p_\theta(\cdot\mid x)\}.
\]
Thus inserting a shared or approximated posterior pays the forward variational KL from the trial law to the exact posterior.  This is the variational identity underlying classical EM and its free-energy interpretation \citep{DempsterLairdRubin1977,NealHinton1998}.

\begin{decompositionbox}
In EM and variational inference, the native insertion certificate is not ``posterior distance'' in the abstract.  It is the KL orientation generated by the Gibbs--Fenchel lift.
\end{decompositionbox}

\paragraph{Left and right Bregman pooling.}
\index{Bregman geometry!pooling}

For interior points $a_1,\ldots,a_T$ with weights $w_t\ge0$ satisfying
$\sum_t w_t=1$, two pooling problems differ:
\[
  \min_a\sum_t w_tD_\Phi(a_t\|a),
  \qquad
  \min_a\sum_t w_tD_\Phi(a\|a_t).
\]
When the displayed means remain in the relevant Legendre domain, the first
produces the primal arithmetic average
\[
  a=\sum_t w_ta_t,
\]
whereas the second produces the dual-coordinate average
\[
  \nabla\Phi(a)=\sum_t w_t\nabla\Phi(a_t).
\]
For KL on probability vectors, these are respectively arithmetic mixture
pooling and normalized geometric pooling; the distinction is standard in
Bregman centroid geometry \citep{BanerjeeEtAl2005}.

This distinction matters when a shared auxiliary field is designed by averaging.  The correct pooling direction depends on the certified defect orientation.

\paragraph{Saddle elimination.}
\index{elimination!saddle}

Certified residuals also arise from convex--concave elimination.  Suppose
\[
  J(x)=\inf_a\sup_b L_x(a,b)
\]
under strong duality.  Let $(a_x^\star,b_x^\star)$ be a saddle point.  A trial pair can be audited by primal and dual insertion gaps,
\[
  L_x(a,b_x^\star)-J(x)\ge0,
  \qquad
  J(x)-L_x(a_x^\star,b)\ge0.
\]
Their sum is a primal--dual certificate.  The same orientation discipline applies: the two signs are fixed by the min--max order.

\paragraph{Exchange inequalities.}
\index{defect!exchange inequality}

The native defect can be exchanged for a more geometric error when the objective has curvature.  If
\[
  \Def_x(a)\ge \frac\mu2 d^2\{a,\mathcal O(x)\},
\]
then
\[
  d\{a,\mathcal O(x)\}
  \le
  \sqrt{2\Def_x(a)/\mu}.
\]
Conversely, work in Euclidean or Hilbert geometry, with
$d(a,b)=\lVert a-b\rVert$.  Suppose a nearest oracle
\[
  b\in\argmin_{c\in\mathcal O(x)}\lVert a-c\rVert
\]
exists, $\Def_x$ is differentiable on a neighborhood containing the whole
segment $[b,a]$, $\nabla\Def_x(b)=0$, and $\nabla\Def_x$ is $L$-Lipschitz on
that neighborhood.  The descent lemma and $\Def_x(b)=0$ then give
\[
  \Def_x(a)
  \le
  \Def_x(b)+\langle\nabla\Def_x(b),a-b\rangle
  +\frac L2\lVert a-b\rVert^2
  =\frac L2 d^2\{a,\mathcal O(x)\}.
\]
These exchanges are local analytic theorems, not definitions of the defect.

\paragraph{When the defect is set-valued.}
\index{defect!set-valued}
\index{oracle!set-valued}

If $\mathcal O(x)$ has multiple elements, the residual still satisfies
\[
  \Def_x(a)=0\iff a\in\mathcal O(x).
\]
A distance-to-oracle bound may use
\[
  d\{a,\mathcal O(x)\}=\inf_{b\in\mathcal O(x)}d(a,b),
\]
but a particular labeled representative should not be selected unless the output contract requires it.  This is the first place where representation semantics enters.

\section*{Exercises}

\begin{exercise}
Derive the Bregman defect for the Poisson log-partition function and identify its statistical interpretation.
\end{exercise}

\begin{exercise}
For KL divergence, compute the left and right centroids of two Bernoulli distributions.  Compare their limiting behavior near the boundary of the simplex.
\end{exercise}

\begin{exercise}
Show that strong convexity gives a defect-to-distance exchange, but not a distance-to-task-regret exchange without an additional task theorem.
\end{exercise}

\input{chapter_appendices/ch05_proofs}

%% file: chapter_appendices/ch05_proofs.tex
\chapterproofappendix

\subsection*{Proof of the conjugate defect identity}
\proofdependency{Fenchel conjugacy and the Legendre identity $\nabla\Phi(a^\star(u))=u$ are the only ingredients.  Strict convexity is used for uniqueness of the oracle, not for the algebraic equality itself.}
\begin{proof}
By definition of the convex conjugate,
\[
\Phi^\star(u)=\langle u,a^\star(u)\rangle-\Phi\{a^\star(u)\},
\]
where $a^\star(u)=\nabla\Phi^\star(u)$ and
$u=\nabla\Phi\{a^\star(u)\}$.  Therefore
\begin{align*}
J(u)
&=\inf_a\{\Phi(a)-\langle u,a\rangle+c(u)\}\\
&=c(u)-\Phi^\star(u).
\end{align*}
For an arbitrary trial state $a$,
\begin{align*}
H_u(a)-J(u)
&=\Phi(a)-\langle u,a\rangle+\Phi^\star(u)\\
&=\Phi(a)-\Phi(a^\star)
  -\langle u,a-a^\star\rangle\\
&=\Phi(a)-\Phi(a^\star)
  -\left\langle\nabla\Phi(a^\star),a-a^\star\right\rangle\\
&=D_\Phi(a\|a^\star).
\end{align*}
Nonnegativity and equality only at $a=a^\star$ follow from strict convexity of $\Phi$.
\end{proof}

%% file: chapters/ch06_towers_pgxvc.tex
\chapter{Elimination Towers and the P/G/X/V/C Calculus}
\label{ch:towers-pgxvc}

The P/G/X/V/C calculus and an elimination tower describe orthogonal
directions.  The five letters classify what happens within one elimination
level after its local oracle has been exposed, whereas a tower records how
states and defects pass between levels.  The mode says what was changed; the
tower says where its cost propagates.

\section{Five distinct operations on an elimination system}
\index{P/G/X/V/C calculus}

The calculus begins from one noncommutation:
\[
  \operatorname{Eliminate}\circ\operatorname{Couple}
  \ne
  \operatorname{Couple}\circ\operatorname{Eliminate}.
\]
The decisive question is not which numerical smoother is used, but where
the coupling enters the declared contract.

Suppose $J(\Theta)=\inf_A H_0(\Theta,A)$ and $A^\star(\Theta)$ is the
pointwise oracle.  Across an index set $\mathcal T$ of locations, tasks, or
replicas, the oracles form a field $A^\star=(A_t^\star)_{t\in\mathcal T}$.
A practitioner may wish to smooth, pool, share, or couple that field.

The common contract fixes the original lift $H_0$, its eliminated target
$J$, and the oracle field $A^\star$.  It may also declare an external field
operator $S$, a coherence penalty $R$, or marginal laws $(\mu_t)$ whose
couplings form $\Gamma(\mu_t:t\in\mathcal T)$.  A P/G/X/V/C label records
which of these objects is acted on and what remains invariant.  It does not
name a numerical algorithm: the same smoother can implement different
operations under different contracts.

\subsection{P: pointwise elimination}
\index{elimination!pointwise}

The baseline is
\[
  A_t=A_t^\star(\Theta).
\]
No cross-index coherence is imposed.  The target remains exactly $J$ and every local defect is zero.
P is the reference contract against which the other four operations are
measured.

\subsection{G: plug-in globalization}
\index{globalization!plug-in}

An external operator $S$ transforms the oracle field:
\[
  \widetilde A=S A^\star.
\]
The field is inserted back into the original lift.  The target value becomes
\[
  H_0(\Theta,\widetilde A)
  =
  J(\Theta)+\Def_\Theta(\widetilde A).
\]
The insertion defect is exact.  G does not optimize a new joint objective; it deploys an externally constructed field.
Thus G changes the deployed field while holding $H_0$ and $J$ fixed.

\subsection{X: exactification}
\index{exactification}

If a frozen G step at anchor $\bar\Theta$ gives
$Q_G(\Theta)=J(\Theta)+d(\Theta)$, X forms
\[
  Q_X^{(k)}(\Theta\mid\bar\Theta)
  =Q_G(\Theta)-T_kd(\Theta;\bar\Theta),
  \qquad
  j_{\bar\Theta}^kQ_X^{(k)}=j_{\bar\Theta}^kJ.
\]
X changes the frozen surrogate, not the field or the joint law.
Chapter~\ref{ch:exactification} develops the construction and its converse.

\subsection{V: variational coupling}
\index{coupling!variational}

V places the field and a coherence penalty inside a new objective:
\[
  J_{\lambda}(\Theta)
  =
  \inf_A\{H_0(\Theta,A)+\lambda R(A)\}.
\]
The statistical or optimization target changes.  Block monotonicity can be exact, but the estimator need not target the original $J$.
The returned estimand is $J_\lambda$, not merely $J$ evaluated at an external
field.

\subsection{C: fixed-marginal coupling}
\index{coupling!fixed-marginal}

C keeps every coordinate procedure and marginal law fixed, choosing only
\[
  \gamma\in\Gamma(\mu_t:t\in\mathcal T).
\]
It is relevant to common random numbers, antithetic couplings, ensemble
dependence, and variance reduction.  Additive coordinate expectations cannot
improve under a fixed-marginal coupling; only genuinely joint functionals can.

\begin{table}[H]
\centering
\caption{The five operational contracts.}
\small
\begin{tabularx}{\textwidth}{@{}
  >{\raggedright\arraybackslash}p{0.06\textwidth}
  >{\raggedright\arraybackslash}p{0.16\textwidth}
  >{\raggedright\arraybackslash}X
  >{\raggedright\arraybackslash}X
  >{\raggedright\arraybackslash}X@{}}
\toprule
Mode & Acts on & Held fixed & Returned object & Certified consequence\\
\midrule
P & local oracle choice & original lift and target & $A^\star$ & zero local defect\\
G & deployed field & $H_0$ and $J$ & $SA^\star$ & exact insertion defect\\
X & frozen surrogate & original target $J$ & $Q_G-T_kd$ & target jet restored\\
V & joint objective & declared feasible class & optimizer of $H_0+\lambda R$ & target generally changes\\
C & joint law & every marginal $\mu_t$ & $\gamma\in\Gamma(\mu_t)$ & additive means fixed\\
\bottomrule
\end{tabularx}
\end{table}

\section{A compact classification test}
\index{P/G/X/V/C calculus!classification test}

Classify a procedure from the provenance of its returned object, not from the
last numerical array it produces.
\begin{enumerate}
  \item If it stops after independent pointwise elimination, label it P.
  \item If it transforms the oracle field outside the original objective and
  then evaluates that field in $H_0$, label it G.
  \item If it subtracts a defect jet to restore target contact at an anchor,
  label that correction X.
  \item If it moves coherence inside an objective and reoptimizes jointly,
  label it V.
  \item If it fixes all marginals and changes only their joint coupling,
  label it C.
\end{enumerate}
A workflow may answer more than one question affirmatively.  Record the
sequence---for example, $\mathrm{G}\to\mathrm{X}$---because the last step
does not erase the contract of the earlier one.

\begin{boundarybox}
Numerical equality is weaker than operational equality.  G and V can return
the same field in a special quadratic problem while targeting different
objects; X repairs a target jet without turning G into V; and C cannot remove
an insertion defect or alter an additive coordinate expectation.  A change
of architecture, output semantics, or resource grammar is a separate contract
change and must not be hidden inside one of the five letters.
\end{boundarybox}

\section{Horizontal operations and multistage towers}

If a stage uses one or more horizontal modes, its certified defect
contribution enters the vertical decomposition; the stage costs then compose
by infimal convolution.

\paragraph{A multistage elimination tower.}
\index{elimination!tower}
\index{infimal convolution}

Let a fine state $e$ map to an intermediate state $b$ and then to a coarse state $c$:
\[
  \mathcal E\xrightarrow{\pi}\mathcal B\xrightarrow{\sigma}\mathcal C.
\]
Suppose the defect splits at each stage,
\[
  \Def^{\mathcal E}(e)
  =
  \Def^{\mathcal B}(\pi e)+\delta_{\pi}(e),
\]
\[
  \Def^{\mathcal B}(b)
  =
  \Def^{\mathcal C}(\sigma b)+\delta_{\sigma}(b).
\]
Then
\[
  \Def^{\mathcal E}(e)
  =
  \Def^{\mathcal C}(\sigma\pi e)
  +
  \delta_{\sigma}(\pi e)
  +
  \delta_{\pi}(e).
\]

After minimizing over architecture-constrained fibers, the realization costs compose by infimal convolution:
\[
  \Psi_{\sigma\pi}(c)
  =
  \inf_{b:\sigma b=c}
  \{\delta_\sigma(b)+\Psi_\pi(b)\}.
\]
This is the usual Bellman/min-plus composition law
\citep{Bellman1957,BaccelliEtAl1992}.

\begin{conditionbox}
The successive coarse and fine states must be compatible, and infeasible transitions are encoded by the value $+\infty$.  All intermediate infima are taken over the same declared fibers.  No minimizer need be unique; the statement concerns values of infimal composition.
\end{conditionbox}

\begin{proposition}[Tower associativity]
\label{prop:tower-associativity}
For compatible consecutive state spaces, define the typed stage-cost
composition by
\[
  (c_1\infconv c_2)(x_2,x_0)
  :=
  \inf_{x_1\ {\rm compatible}}
  \{c_1(x_1,x_0)+c_2(x_2,x_1)\}.
\]
Whenever the relevant feasible sets are nonempty, the following identity
holds pointwise for every compatible endpoint pair $(x_3,x_0)$:
\[
  \{(c_1\infconv c_2)\infconv c_3\}(x_3,x_0)
  =
  \{c_1\infconv(c_2\infconv c_3)\}(x_3,x_0).
\]
\end{proposition}

\begin{interpretationbox}
A tower can be bracketed in any order because elimination is min-plus composition.  This justifies dynamic-programming and multistage obstruction decompositions.  It does not imply that two numerical algorithms have the same runtime, conditioning, or finite-precision behavior.
\end{interpretationbox}

\begin{proofroadmap}
Write both bracketings as a nested infimum over all intermediate states, then exchange the order of the finite or extended-value infima.  The chapter appendix gives the explicit Bellman/min-plus calculation.
\end{proofroadmap}

\paragraph{Why the tower matters.}

The tower converts a vague statement such as ``the representation loses information'' into a decomposition with identifiable stages.

\begin{itemize}
  \item A coarse representation can choose the wrong coarse object.
  \item Even the correct coarse object may be expensive to realize in the fine architecture.
  \item A deployed algorithm may then incur additional implementation error inside the chosen fine class.
\end{itemize}

The same structure appears in COT under conditional-law base change, in resource-constrained architectures under carrier--decoder decomposition, and in certificate statistics under evidence compression.

\section{Worked decompositions and coupling boundaries}
\paragraph{Bregman Pythagorean identities.}
\index{Bregman geometry!Pythagorean identity}

Suppose $\mathcal C\subseteq\mathcal A$ is convex and the Bregman projection
in the first argument exists:
\[
  a_{\mathcal C}
  \in
  \argmin_{c\in\mathcal C}D_\Phi(c\|a^\star).
\]
Thus the projected point varies in the first argument while $a^\star$ remains
the second argument.  Under the standard differentiability and convex
projection conditions \citep{BauschkeBorwein1997Legendre}, every comparison
point $a\in\mathcal C$ satisfies
\[
  D_\Phi(a\|a^\star)
  \ge
  D_\Phi(a\|a_{\mathcal C})
  +
  D_\Phi(a_{\mathcal C}\|a^\star).
\]
This gives a three-level interpretation:
\begin{enumerate}
  \item irreducible constraint defect;
  \item within-constraint implementation defect;
  \item possible nonorthogonality remainder if exact Pythagoras fails.
\end{enumerate}

\paragraph{A worked Gaussian field.}
\index{Gaussian field}

Let $a_t^\star$ be local Gaussian means and let
\[
  H_t(a_t)=\frac12(a_t-a_t^\star)^2.
\]
On a graph with Laplacian $L$, V solves
\[
  \min_a
  \frac12\|a-a^\star\|^2
  +
  \frac\lambda2 a^\top La.
\]
The solution is
\[
  a^{\rm V}=(I+\lambda L)^{-1}a^\star.
\]
If G uses the same filter externally, the deployed field coincides numerically with $a^{\rm V}$ in this quadratic unconstrained example.  Nevertheless the interpretation differs: G evaluates the original lift at an external field; V defines the field as the optimizer of a changed objective.  Under constraints or nonlinear repair, the numerical equality generally disappears.

\paragraph{C coupling and its boundary.}
\index{coupling!fixed-marginal}
\index{claim boundary!fixed-marginal coupling}

Let $Y_1,\ldots,Y_T$ have fixed marginals.  Whenever every
$\ell_t(Y_t)$ is integrable, any additive target satisfies
\[
  \E\sum_t \ell_t(Y_t)=\sum_t\E\ell_t(Y_t),
\]
changing the joint coupling cannot change the expectation.  A benefit requires a joint functional, such as variance of an average, maximum loss, simultaneous coverage, or a path-dependent objective.

\begin{boundarybox}
C is not a hidden way to improve every coordinatewise estimator.  Fixed marginals protect all additive coordinate expectations.
\end{boundarybox}

\section*{Exercises}

\begin{exercise}
For the quadratic graph example, compute the exact G insertion defect and V objective improvement.  Explain why they are numerically related but conceptually distinct.
\end{exercise}

\begin{exercise}
Prove tower associativity for finite state spaces using min-plus matrix multiplication.
\end{exercise}

\begin{exercise}
Give a joint functional for which fixed-marginal antithetic coupling improves performance, and an additive functional for which it cannot.
\end{exercise}

\input{chapter_appendices/ch06_proofs}

%% file: chapter_appendices/ch06_proofs.tex
\chapterproofappendix

\subsection*{Proof of tower associativity}
\proofdependency{Only the compatibility of intermediate states and the convention that the infimum of an empty feasible set is $+\infty$ are used.}
\begin{proof}
Let $x_0,x_1,x_2,x_3$ denote compatible states at four consecutive levels, and let the three stage costs be
$c_1(x_1,x_0)$, $c_2(x_2,x_1)$, and $c_3(x_3,x_2)$.  For fixed endpoints $(x_3,x_0)$,
\begin{align*}
\{(c_1\infconv c_2)\infconv c_3\}(x_3,x_0)
&=\inf_{x_2}
  \left[
    \inf_{x_1}\{c_1(x_1,x_0)+c_2(x_2,x_1)\}
    +c_3(x_3,x_2)
  \right]\\
&=\inf_{x_1,x_2}
  \{c_1(x_1,x_0)+c_2(x_2,x_1)+c_3(x_3,x_2)\}.
\end{align*}
Similarly,
\begin{align*}
\{c_1\infconv(c_2\infconv c_3)\}(x_3,x_0)
&=\inf_{x_1}
  \left[
    c_1(x_1,x_0)
    +\inf_{x_2}\{c_2(x_2,x_1)+c_3(x_3,x_2)\}
  \right]\\
&=\inf_{x_1,x_2}
  \{c_1(x_1,x_0)+c_2(x_2,x_1)+c_3(x_3,x_2)\}.
\end{align*}
The two expressions are identical.  The same argument applies to extended-valued realization costs and proves the stated tower law.
\end{proof}

%% file: chapters/ch07_exactification.tex
\chapter{Exactification and Optimization Certificates}
\label{ch:exactification}
\index{touching identity}
\index{globalization!loss of touching}

Fix an anchor $\bar\Theta$ and freeze a feasible, possibly nonlocal,
auxiliary field $\widetilde A$.  The field may have been constructed by a
globalization step at the anchor, but it is held fixed as the outer parameter
varies.  Define
\[
  Q_G(\Theta)=H_0(\Theta,\widetilde A)
  =J(\Theta)+d(\Theta),
\]
where
\[
  d(\Theta)=\Def_\Theta(\widetilde A)
  =H_0(\Theta,\widetilde A)-J(\Theta)\ge0.
\]
Even if the frozen lift is smooth in $\Theta$, $Q_G$ need not touch $J$ at
the current iterate.  Minimizing $Q_G$ can therefore optimize the defect
rather than the target.

Exactification repairs this problem by subtracting the local defect jet.
In the P/G/X/V/C calculus of Chapter~\ref{ch:towers-pgxvc}, the frozen
construction is G and the jet correction is X; no V objective or C coupling
is introduced.

In this chapter, an optimization certificate combines target-jet contact with
explicit inner-oracle, defect-score, and step-acceptance error budgets;
exactification supplies the target-contact component.
\index{optimization certificate}

\section{Jet exactification}
\index{exactification!jet}
\index{defect jet}

Let $T_kd(\Theta;\bar\Theta)$ be the order-$k$ Taylor polynomial of the
frozen defect $d$ at $\bar\Theta$.  Define
\[
  Q_X^{(k)}(\Theta\mid\bar\Theta)
  =
  J(\Theta)+d(\Theta)-T_kd(\Theta;\bar\Theta).
\]
Then the $k$-jet of $Q_X^{(k)}$ agrees with that of $J$ at the anchor.

\begin{conditionbox}
The order $k$ is a nonnegative integer, and $J$ and the frozen defect $d$
are $C^k$ on a neighborhood of $\bar\Theta$ in one common local coordinate
chart.  The polynomial $T_kd$ is their ordinary order-$k$ Taylor
polynomial.  Taylor expansion is in $\Theta$ with $\widetilde A$ fixed; no
global convexity or descent is claimed.  A function is called $k$-flat at
$\bar\Theta$ when all of its derivatives through order $k$ vanish there.
\end{conditionbox}

\begin{theorem}[Jet exactification]
\label{thm:jet-exactification}
If both $J$ and $d$ are $C^k$ on a neighborhood of $\bar\Theta$, then
\[
  j_{\bar\Theta}^k Q_X^{(k)}
  =
  j_{\bar\Theta}^k J.
\]
In particular, for $k=1$,
\[
  Q_X(\bar\Theta\mid\bar\Theta)=J(\bar\Theta),
  \qquad
  \nabla Q_X(\bar\Theta\mid\bar\Theta)=\nabla J(\bar\Theta).
\]
\end{theorem}

\begin{interpretationbox}
Subtracting the low-order defect jet makes the surrogate agree with the original target in value and derivatives through order $k$, while retaining higher-order auxiliary geometry.  Exactification therefore repairs target tangency; it does not by itself guarantee a useful step size or global convergence.
\end{interpretationbox}

\begin{proofroadmap}
Express the frozen surrogate as $J+d$, subtract the order-$k$ Taylor
polynomial of $d$, and compare derivatives at the anchor.  Every derivative
of $d-T_kd$ through order $k$ vanishes.  The chapter appendix records the
complete jet calculation.
\end{proofroadmap}

The result is elementary once $d$ is known.  The substantive point is that
$d$ is the exact native insertion defect, not an arbitrary correction.  The
construction and converse below are the exactification normal form of the
companion framework; their relation to first-order surrogate functions is
discussed there as well \citep{FrameworkPartII,Mairal2013FirstOrder}.

\paragraph{Candidate-wise descent identity.}
\index{descent identity!candidate-wise}

For first-order exactification,
\[
  Q_X(\Theta\mid\bar\Theta)
  =
  J(\Theta)+d(\Theta)-d(\bar\Theta)
  -\langle\nabla d(\bar\Theta),\Theta-\bar\Theta\rangle.
\]
If a candidate $\Theta^+$ satisfies
\[
  Q_X(\Theta^+\mid\bar\Theta)
  \le
  Q_X(\bar\Theta\mid\bar\Theta)-\gamma\|\Theta^+-\bar\Theta\|^2,
\]
then
\begin{align*}
J(\Theta^+)-J(\bar\Theta)
={}&Q_X(\Theta^+\mid\bar\Theta)-Q_X(\bar\Theta\mid\bar\Theta)\\
&-\Bigl[d(\Theta^+)-d(\bar\Theta)
-\langle\nabla d(\bar\Theta),\Theta^+-\bar\Theta\rangle\Bigr].
\end{align*}
If the defect gradient is locally $L_d$-Lipschitz, the bracket is bounded below by $-L_d\|\Theta^+-\bar\Theta\|^2/2$.  Hence
\[
  J(\Theta^+)-J(\bar\Theta)
  \le
  -(\gamma-L_d/2)\|\Theta^+-\bar\Theta\|^2.
\]
A trust region or backtracking line search enforces $\gamma>L_d/2$.

\section{The converse normal form}
\index{exactification!converse normal form}

The forward construction subtracts the Taylor jet of the frozen defect.  The
converse asks whether any scalar correction can preserve the target
$k$-jet without doing the same.  The hypothesis box fixes the local comparison
under which the theorem gives the answer.

\begin{conditionbox}
A candidate scalar correction $c-S$ is compared with the actual frozen defect $d$ in a neighborhood of the anchor, and all objects possess the required $k$-jets.  The conclusion is modulo a $k$-flat remainder; it does not identify a unique global surrogate.
\end{conditionbox}

\begin{theorem}[Converse exactification normal form]
\label{thm:converse-exactification}
Suppose $J,d,c,S$ are $C^k$ on a neighborhood of $\bar\Theta$ and a
surrogate can be written as
\[
  Q(\Theta)=J(\Theta)+d(\Theta)-c(\Theta)+S(\Theta).
\]
Then
\[
  j_{\bar\Theta}^kQ=j_{\bar\Theta}^kJ
\]
if and only if
\[
  j_{\bar\Theta}^k\{c-S\}
  =
  j_{\bar\Theta}^k d.
\]
Thus any target-$k$-jet-preserving correction subtracts the defect jet, modulo a $k$-flat term.
Equivalently, the canonical representative is
\[
  Q_{X,\mathrm{can}}^{(k)}(\Theta\mid\bar\Theta)
  =H_0(\Theta,\widetilde A)-T_kd(\Theta;\bar\Theta),
\]
and every other target-$k$-jet-preserving surrogate of the displayed form is
$Q_{X,\mathrm{can}}^{(k)}+R$, where $R$ is $k$-flat at $\bar\Theta$.
Conversely, every such $k$-flat perturbation preserves the target $k$-jet.
\end{theorem}

\begin{interpretationbox}
The defect-jet correction is not merely one convenient construction.  Any scalar surrogate that preserves the target jet through order $k$ must subtract the same defect jet, up to a term whose derivatives through order $k$ vanish.  This is a local normal form, not a statement about global optimizer equivalence.
\end{interpretationbox}

\begin{proofroadmap}
Write the candidate surrogate as $Q=J+d-c+S$ and take its $k$-jet at the anchor.  Equality with the target jet is equivalent to equality of the $k$-jets of $c-S$ and $d$.  The chapter appendix makes both implications explicit.
\end{proofroadmap}

The converse gives exactification a structural status: first-order target preservation forces a defect-score correction, not just any tangent surrogate.

\section{Approximation, transmission, and certification error}
\paragraph{Approximate inner oracles.}
\index{oracle!approximate}
\index{certification error}

In practice, the auxiliary oracle and its defect gradient are computed
approximately.  Suppose $H_0(\Theta,\cdot)$ is differentiable and
$\mu$-strongly convex, its minimizer $a^\star$ is interior and satisfies
$\nabla_aH_0(\Theta,a^\star)=0$, and the outer score
$\nabla_\Theta H_0(\Theta,a)$ is $L$-Lipschitz in $a$.  Then an inner
gradient residual
\[
  \|\nabla_aH_0(\Theta,\widetilde a)\|\le \epsilon
\]
implies
\[
  \|\widetilde a-a^\star\|\le \epsilon/\mu,
\]
\[
  0\le H_0(\Theta,\widetilde a)-J(\Theta)
  \le \epsilon^2/(2\mu),
\]
and, under the usual envelope/Danskin regularity giving
$\nabla J(\Theta)=\nabla_\Theta H_0(\Theta,a^\star)$,
\[
  \|\nabla_\Theta H_0(\Theta,\widetilde a)-\nabla J(\Theta)\|
  \le L\epsilon/\mu.
\]
These bounds turn numerical
stopping tolerances into objective and gradient certificates.

\paragraph{Summable certification error.}
\index{certification error!summable}

Let $Q_k$ be an approximate exactified surrogate at anchor $\Theta^k$ and write $r_k=Q_k-J$.  Suppose an accepted candidate satisfies
\[
  Q_k(\Theta^{k+1})\le Q_k(\Theta^k)-\Delta_k,
  \qquad
  r_k(\Theta^{k+1})-r_k(\Theta^k)
  \ge-\eta_k\Delta_k-\varepsilon_k,
\]
where $\Delta_k\ge0$, $0\le\eta_k<1$, and $\varepsilon_k\ge0$.  If $J$ is bounded below, $\sup_k\eta_k<1$, $\sum_k\varepsilon_k<\infty$, and
\[
  \Delta_k\ge c\|\Theta^{k+1}-\Theta^k\|^2
  \quad\text{for some }c>0,
\]
then $\sum_k\Delta_k<\infty$ and
\[
  \sum_k\|\Theta^{k+1}-\Theta^k\|^2<\infty.
\]
For the stationary-point conclusion, additionally assume
\[
  \|\nabla r_k(\Theta^k)\|\le\delta_k\to0,
  \qquad
  \|\nabla Q_k(\Theta^{k+1})\|\le\zeta_k\to0,
\]
and that $\nabla Q_k$ is $K$-Lipschitz on the segment from $\Theta^k$ to $\Theta^{k+1}$, uniformly in $k$.  Then
\[
  \|\nabla J(\Theta^k)\|
  \le K\|\Theta^{k+1}-\Theta^k\|+\zeta_k+\delta_k
  \longrightarrow0.
\]
If $\nabla J$ is continuous, every cluster point is stationary for $J$.  Compactness or level boundedness is needed only to guarantee that cluster points exist.

\begin{boundarybox}
Stationarity is a target-level conclusion only because exactification restores the target score.  A generic majorization or smoothed surrogate may converge to a stationary point of a different objective.
\end{boundarybox}

\paragraph{Conjugate transmission.}
\index{conjugate transmission}

For the conjugate lift
\[
  H_0(\Theta,A)=c(\Theta)+\Phi(A)-\langle A,\eta(\Theta)\rangle,
\]
let $Q_{X,A}^{(k)}$ and $Q_{X,B}^{(k)}$ be the canonical exactifications
obtained by freezing $A$ and $B$.  Direct cancellation gives the transmission
identity
\[
  Q_{X,A}^{(k)}-Q_{X,B}^{(k)}
  =-\langle A-B,\eta-T_k\eta\rangle.
\]
Thus, for $k=1$, their Hessians at the anchor differ by
$-\langle A-B,D^2\eta(\bar\Theta)\rangle$.  If the natural-parameter map
$\eta$ is affine, the Taylor tail vanishes and no nonlinear curvature is
transmitted.  This is the conjugate-transmission calculation in the
companion framework \citep{FrameworkPartII}.

\section{Practical algorithm}
\index{exactification!algorithm}

The preceding error bounds become an implementable outer loop only when the
acceptance test keeps the inner-oracle, defect-score, and outer-solver budgets
separate.  The algorithm records that interface.

\begin{algorithm}[H]
\caption{First-order exactification with certified backtracking}
\begin{algorithmic}[1]
\State Input anchor $\Theta^k$ and predeclared inner-oracle, defect-score, and outer-solver tolerances.
\State Compute and freeze an approximate globalized field $\widetilde A^k$; certify the value and score $(\widetilde d_k,\widetilde g_k)$ of $d_k(\Theta)=H_0(\Theta,\widetilde A^k)-J(\Theta)$ at $\Theta^k$.
\State Form $\widetilde Q_k(\Theta)=H_0(\Theta,\widetilde A^k)-\widetilde d_k-\langle\widetilde g_k,\Theta-\Theta^k\rangle$.
\State Propose $\Theta^{k+1}$ by approximately minimizing $\widetilde Q_k$ in a trust region.
\State Accept when the certified surrogate decrease exceeds the defect-remainder and inner-solver budgets.
\State Otherwise shrink the trust region and repeat.
\end{algorithmic}
\end{algorithm}

The algorithm separates three tolerances: inner-oracle error, defect-score error, and outer minimization error.  Collapsing them into one generic tolerance obscures the proof.

\section*{Exercises}

\begin{exercise}
For $d(\theta)=\theta^4$, write the first- and second-order exactified surrogates at an anchor $\bar\theta$.  Verify jet matching directly.
\end{exercise}

\begin{exercise}
Under $\mu$-strong convexity, prove the value bound $H(\widetilde a)-H(a^\star)\le \|\nabla H(\widetilde a)\|^2/(2\mu)$.
\end{exercise}

\begin{exercise}
Construct a surrogate that is value-touching but not score-touching.  Show that its stationary points need not be stationary for the target.
\end{exercise}

\input{chapter_appendices/ch07_proofs}

%% file: chapter_appendices/ch07_proofs.tex
\chapterproofappendix

\subsection*{Proof of jet exactification}
\proofdependency{The frozen defect $d$ and target $J$ are $C^k$ on a neighborhood of the anchor.  The proof then uses only the defining derivative identities of the order-$k$ Taylor polynomial; no convexity, majorization, or optimization assumption is required.}
\begin{proof}
Let $R_k(\Theta)=d(\Theta)-T_kd(\Theta;\bar\Theta)$.  By the defining property of the Taylor polynomial, for every multi-index $\alpha$ with $|\alpha|\le k$,
\[
\partial^\alpha R_k(\bar\Theta)=0.
\]
Since
\[
Q_X^{(k)}(\Theta\mid\bar\Theta)=J(\Theta)+R_k(\Theta),
\]
all derivatives of $Q_X^{(k)}-J$ through order $k$ vanish at $\bar\Theta$.  Hence
\(
j_{\bar\Theta}^kQ_X^{(k)}=j_{\bar\Theta}^kJ
\).
For $k=1$, the zeroth- and first-order identities are precisely
\[
Q_X(\bar\Theta\mid\bar\Theta)=J(\bar\Theta),
\qquad
\nabla Q_X(\bar\Theta\mid\bar\Theta)=\nabla J(\bar\Theta).
\]
\end{proof}

\subsection*{Proof of the converse exactification normal form}
\proofdependency{All functions are $C^k$ on a neighborhood of the anchor.  The conclusion is local: it identifies the correction jet and a canonical representative, uniquely only modulo a $k$-flat remainder.}
\begin{proof}
From
\[
Q-J=d-c+S
\]
we obtain
\[
j_{\bar\Theta}^kQ=j_{\bar\Theta}^kJ
\quad\Longleftrightarrow\quad
j_{\bar\Theta}^k(d-c+S)=0.
\]
Linearity of the jet operator gives
\[
j_{\bar\Theta}^k(d-c+S)=0
\quad\Longleftrightarrow\quad
j_{\bar\Theta}^k(c-S)=j_{\bar\Theta}^kd.
\]
This proves both directions.  Because $T_kd$ has the same $k$-jet as $d$,
the difference
\[
  Q-\{H_0(\Theta,\widetilde A)-T_kd(\Theta;\bar\Theta)\}
  =-c+S+T_kd
\]
is $k$-flat whenever the target-jet condition holds.  Conversely, adding any
$k$-flat remainder to the canonical representative leaves the target
$k$-jet unchanged.  This proves the claimed local uniqueness modulo
$k$-flat terms.
\end{proof}

%% file: chapters/ch08_graph_crps.tex
\chapter{Probability Validity, Graph Smoothing, and CRPS}
\label{ch:graph-crps}
\index{graph-indexed learning!distributional prediction}
\index{graph smoothing}

The native statistical loss is the continuous ranked probability score
(CRPS), whose expected excess equals squared $L^2(I)$ distance between the
forecast and true CDFs in the compact-support setup declared below.

Let $G=(V,E)$ be a graph with $T=|V|$ and Laplacian $L$.  Vertex $t$ has distribution function $F_t^\star$ and observations $Y_{t1},\ldots,Y_{tn_t}$.  Throughout this chapter's projection and risk statements, outcomes are supported on a declared compact interval $I=[a,b]$.  Put
\[
  \mathbb H=L^2(I),
  \qquad
  \mathcal C_{\rm cdf}
  =
  \{F\in\mathbb H:0\le F\le1\text{ and $F$ is nondecreasing a.e.}\}.
\]
Choose right-continuous representatives with $F(b)=1$, extended by zero to the left of $a$ and by one to the right of $b$.  Then $\mathcal C_{\rm cdf}$ is the nonempty closed convex set of CDF restrictions in $\mathbb H$.  The empirical CDF restriction is
\[
  \widehat F_t(y)
  =
  \frac1{n_t}\sum_{i=1}^{n_t}\ind\{Y_{ti}\le y\}.
\]
Stack the CDFs as a vector-valued function $\widehat F(y)\in\R^T$.

For a self-adjoint graph filter $S$, define the raw estimator
\[
  \widetilde F(y)=S\widehat F(y).
\]
A higher-order Richardson filter can cancel low-order bias but may use signed weights.  The raw coordinates may then fail monotonicity or leave $[0,1]$.

\section{The CDF set and projection repair}
\index{probability validity}
\index{CDF set}
\index{repair!projection}

Let $\mathcal C=\mathcal C_{\rm cdf}^T\subset\mathbb H^T$, and let $\boldsymbol\Pi_{\rm cdf}$ be its coordinatewise metric projection.  Define
\[
  F=\boldsymbol\Pi_{\rm cdf}(\widetilde F)=\Pi_{\mathcal C}(\widetilde F).
\]

\begin{conditionbox}
The metric projection is taken in $\mathbb H^T=L^2(I)^T$, the population target belongs to $\mathcal C$, and the graph-roughness conclusion uses the same nonexpansive coordinate projection $\Pi_{\rm cdf}$ at every vertex.
\end{conditionbox}

On an unbounded outcome space, one must instead formulate the argument in an affine $L^2$-difference class under suitable moment conditions; no $L^2(\mathbb R)$ membership of each individual CDF is asserted here.

\begin{proposition}[Validity repair]
\label{prop:cdf-validity-repair}
For every true CDF field $F^\star\in\mathcal C$,
\[
  \|F-F^\star\|^2
  \le
  \|\widetilde F-F^\star\|^2.
\]
For the coordinatewise product projection above, graph roughness also satisfies
\[
  \langle F,LF\rangle
  \le
  \langle\widetilde F,L\widetilde F\rangle.
\]
\end{proposition}

\begin{interpretationbox}
Projection can repair an invalid signed high-order proposal without increasing squared CDF error, hence without increasing CRPS risk.  Under the graph product geometry it also cannot increase edge roughness.  The projected G estimator is not automatically the constrained V optimizer in a different metric.
\end{interpretationbox}

\begin{proofroadmap}
Apply the Hilbert projection variational inequality with the true CDF field as a feasible comparison.  The Pythagorean inequality yields risk contraction; coordinatewise nonexpansiveness yields edgewise roughness contraction.  Full details appear in the chapter appendix.
\end{proofroadmap}

In the P/G/X/V/C calculus of Chapter~\ref{ch:towers-pgxvc}, the repaired
field is a G construction: an ambient proposal followed by a validity
interface.  It is not generally equal to the V solution obtained by
minimizing a penalized objective directly over $\mathcal C$.

\section{CRPS risk and its observable decomposition}
\paragraph{CRPS as squared-CDF risk.}
\index{continuous ranked probability score}

For a forecast CDF $F$ and observation $Y$, the usual real-line CRPS integral reduces, under the support convention above, to
\[
  \CRPS(F,Y)
  =
  \int_I\{F(y)-\ind(Y\le y)\}^2\dd y.
\]
Its expected excess risk is
\[
  \E\CRPS(F,Y)-\E\CRPS(F^\star,Y)
  =
  \int_I\{F(y)-F^\star(y)\}^2\dd y.
\]
The continuous-distribution scoring-rule lineage and the strict-propriety
framework are classical \citep{MathesonWinkler1976,GneitingRaftery2007}.
What is specific here is the probability-valid graph repair and its matching
rate analysis.  The displayed identity implies that the Hilbert projection
repair cannot increase excess CRPS.

\paragraph{Exact bias--variance decomposition.}
\index{bias--variance decomposition}

Assume that observations are i.i.d. within each vertex and that the samples
are independent across vertices.  Let $B=(S-I)F^\star$ be the smoothing bias.  Then
\[
  \widetilde F-F^\star
  =
  B+S(\widehat F-F^\star).
\]
The cross term vanishes after expectation, giving
\[
  \E\|\widetilde F-F^\star\|^2
  =
  \|B\|^2
  +
  \sum_{t=1}^T(S^2)_{tt}\frac{\kappa_t}{n_t},
\]
where
\[
  \kappa_t
  =
  \int_I F_t^\star(y)\{1-F_t^\star(y)\}\dd y.
\]
The variance complexity is therefore controlled by
\[
  \operatorname{tr}(S^2),
\]
not merely by $\operatorname{tr}(S)$.

\begin{decompositionbox}
The graph effective dimension for squared-CDF risk is $\operatorname{tr}(S^2)$ because the observation noise is filtered twice in the second moment.
\end{decompositionbox}

\paragraph{An observable raw-risk estimate.}
\index{risk!observable estimate}

For a fixed data-independent filter, pairwise empirical-CDF identities yield an unbiased estimate of the raw risk after correcting the diagonal self-influence.  The correction involves $2S_{tt}-1$.  The nonlinear projection then provides the pathwise Pythagorean bound
\[
  \|F-F^\star\|^2+\|\widetilde F-F\|^2
  \le \|\widetilde F-F^\star\|^2.
\]
Thus $\|\widetilde F-F\|^2$ is an observable guaranteed deduction from the
raw squared error.  The raw estimator can therefore be tuned using an exact
risk decomposition while the final projection restores validity.

\section{The probability-validity barrier}
\index{probability validity!barrier}

Projection repairs a raw forecast after it has left the CDF set.  A sharper
question is when the ambient linear filter preserves probability validity
without repair.  The source box fixes the provenance and scope of that
classification and of the path-rate result used in the next section.

\begin{sourcebox}[title={Source result; proof not reproduced here}]
The graph-universal Markov-validity classification is established in \EGII,
whereas the sharp path minimax theorem is established in \EGI\ and audited in
\EGII\ \citep{FrameworkPartII,FrameworkPartI}.  The book records their
statements and operative scope but does not count them among its principal
formal results.  The chapter appendix proves only
Proposition~\ref{prop:cdf-validity-repair}.
\end{sourcebox}

Within that scope, suppose the ambient V filter is
\[
  S=(I+\lambda\psi(L))^{-1}.
\]
Here $\psi:[0,\infty)\to[0,\infty)$ and $\psi(0)=0$; the latter condition is the spectral mass-preservation gate.  A positive-semidefinite contraction need not be entrywise nonnegative or mass preserving.  Graph-universal Markov validity requires the generator $\psi(L)$ to be a graph Laplacian for every weighted graph.

Every Bernstein function with $\psi(0)=0$ satisfies this Laplacian gate,
possibly producing a dense graph; this uses the classical link among
Bernstein functions, matrix functions, and graph generators
\citep{SchillingSongVondracek2012,MicchelliWilloughby1979,GregorioMugnolo2020}.
If exact kernel preservation on every connected graph is also required,
assume in addition that $\psi(x)>0$ for $x>0$.  In particular,
\[
  (I+\lambda L^\nu)^{-1}
\]
is Markov for every finite weighted graph and every $\lambda>0$ exactly when
\[
  0<\nu\le1.
\]
For $\nu>1$, a weighted three-node path produces a negative entry for a range of $\lambda$.  Likewise, among real polynomials $\psi$ with $\psi(0)=0$, the only graph-universally valid spectral penalties are nonnegative multiples of $L$.

This barrier applies to linear probability-preserving mechanisms.  It does not rule out nonlinear projection repair.

\section{Sharp path rates}
\index{graph smoothing!path rates}

On a growing path, suppose the graph-source smoothness is $s>0$ and a Richardson filter has integer qualification $m\ge s$.  With calibrated bandwidth, the repaired forecast attains average excess CRPS
\[
  N^{-1}
  +
  R^{2/(4s+1)}N^{-4s/(4s+1)},
  \qquad N=nT,
\]
in the stated regime.  An embedded Bernoulli experiment gives a matching lower bound.

The rate has two pieces.  The $N^{-1}$ term reflects the graph nullspace or global average.  The nonparametric term balances spectral bias and effective dimension.

\begin{boundarybox}
The source path-rate theorem assumes the declared growing-path source class,
compact outcome support, independent vertex samples, and calibrated
qualification.  It is not a universal rate for arbitrary graphs, dependence,
or unbounded outcomes.
\end{boundarybox}

\paragraph{The broader design principle.}

The example suggests a general workflow for constrained statistical objects:
\[
\begin{aligned}
\text{high-order ambient proposal}
&\longrightarrow \text{diagnose semantic invalidity}\\
&\longrightarrow \text{nonlinear validity repair}\\
&\longrightarrow \text{risk and rate certificate}.
\end{aligned}
\]
The workflow is relevant to covariance matrices, densities, quantiles, stochastic matrices, projectors, and other shape-constrained outputs, but each case requires its own projection and loss geometry.

\section*{Exercises}

\begin{exercise}
Derive the expected excess CRPS identity from the definition of CRPS.
\end{exercise}

\begin{exercise}
For a two-vertex graph, compute $\operatorname{tr}(S^2)$ for $S=(I+\lambda L)^{-1}$.  Interpret the limiting values as $\lambda\downarrow0$ and $\lambda\uparrow\infty$.
\end{exercise}

\begin{exercise}
Construct a signed linear combination of two valid CDFs that is not a CDF.  Show that projection repairs the violation without increasing squared-CDF error to any valid target.
\end{exercise}

\input{chapter_appendices/ch08_proofs}

%% file: chapter_appendices/ch08_proofs.tex
\chapterproofappendix

\subsection*{Proof of validity repair}
\proofdependency{The first assertion uses the Hilbert-space projection theorem.  The graph-roughness assertion additionally uses coordinatewise nonexpansiveness of the projection and the edge representation of the graph Dirichlet form.}
\begin{proof}
Let $P=\Pi_{\mathcal C}$ be metric projection onto the nonempty closed convex set $\mathcal C$.  The projection variational inequality states that
\[
\langle u-Pu,v-Pu\rangle\le0
\qquad(v\in\mathcal C).
\]
Taking $u=\widetilde F$ and $v=F^\star$ and expanding the square gives
\begin{align*}
\|\widetilde F-F^\star\|^2
&=\|\widetilde F-F\|^2+\|F-F^\star\|^2
 +2\langle\widetilde F-F,F-F^\star\rangle\\
&\ge \|F-F^\star\|^2,
\end{align*}
which proves the risk contraction.

For the graph claim, write the weighted Dirichlet form as
\[
\langle G,LG\rangle
=\frac12\sum_{u,v}w_{uv}\|G_u-G_v\|_{\mathbb H}^2.
\]
The metric projection $P_0=\Pi_{\mathcal C_{\rm cdf}}$ is nonexpansive, so
for every edge
\[
\|P_0(\widetilde F_u)-P_0(\widetilde F_v)\|_{\mathbb H}
\le
\|\widetilde F_u-\widetilde F_v\|_{\mathbb H}.
\]
Squaring, multiplying by $w_{uv}/2$, and summing over edges yields
\[
\langle F,LF\rangle\le\langle\widetilde F,L\widetilde F\rangle.
\]
\end{proof}

%% file: part_notes/part02_history.tex
\parthistoricalnotes{}
\index{native loss scale!historical comparison}

\subsection*{Classical engines}

Part II uses four mature mathematical mechanisms.  Fenchel--Young equality
produces Bregman divergences \citep{Bregman1967,Rockafellar1970}; infimal
composition is Bellman/min-plus algebra \citep{Bellman1957,BaccelliEtAl1992};
Taylor subtraction produces a touching jet and is adjacent to first-order
surrogate methodology \citep{Mairal2013FirstOrder}; and metric projection
onto a convex CDF set gives the validity-repair contraction.  CRPS as
squared-CDF risk and its propriety are likewise established
\citep{MathesonWinkler1976,GneitingRaftery2007}.

\begin{center}
\small
\begin{tabularx}{0.96\textwidth}{@{}p{0.22\textwidth}p{0.31\textwidth}X@{}}
\toprule
Book object & Classical core & Elimination-calculus use \\
\midrule
Conjugate defect & Fenchel--Bregman identity & the eliminated objective fixes
the loss units and the direction of an asymmetric divergence \\
Obstruction tower & infimal convolution and dynamic programming & intermediate
states are typed and every stage cost is read as an objective-generated tax \\
Exactification & Taylor touching and surrogate correction & the subtracted
jet is the actual frozen insertion defect; the converse is stated modulo a
flat remainder within that class \\
CDF repair & Hilbert projection and nonexpansiveness & validity is restored
without increasing squared-CDF/CRPS risk or the declared graph roughness \\
\bottomrule
\end{tabularx}
\end{center}

\subsection*{What is and is not claimed}

The conjugate identity, tower associativity, jet-matching calculation, and
projection inequalities are not claimed as new classical mathematics.  The
program-specific layer is the insistence that a defect be generated by the
same elimination as the target, that transformations preserve its units and
orientation, and that target-jet preservation be audited against that exact
defect.  The converse exactification normal form is recorded as a source-
program claimed increment, not as an exhaustive priority assertion.

%% file: chapters/ch09_integrability.tex
\chapter{Integrability, Hodge Repair, and Global Defect Consistency}
\label{ch:integrability}
\chaptermark{Integrability and Defect Consistency}
\section{When local defect reports may be inconsistent}
\index{integrability}
\index{defect!local consistency}

A complex learning pipeline may eliminate several coordinates in different orders.  Local modules can report nonnegative, touching residuals while failing to arise from one global objective.  The first structural audit is therefore integrability.

Consider a finite elimination complex whose vertices represent partial eliminations and whose edges represent eliminating one additional coordinate.  Let $e$ assign a reported increment to every oriented edge.

If $e$ is the gradient of a potential $V$,
\[
  e=d_0V,
\]
then every path sum between two vertices is the same.  Conversely, on a simply connected complex, zero circulation around every elementary face implies path independence.
This discrete potential/cycle condition is the same integrability mechanism
that underlies exact potential games \citep{MondererShapley1996}; here the
edge quantities are elimination defects rather than unilateral payoffs.

\section{Square curvature}
\index{curvature!square}

For the elementary square shown below, $e_1,e_4$ form the lower-then-right path and $e_2,e_3$ form the left-then-upper path.  Define
\[
  \Omega=e_1+e_4-e_2-e_3.
\]
A nonzero $\Omega$ measures elimination-order dependence.

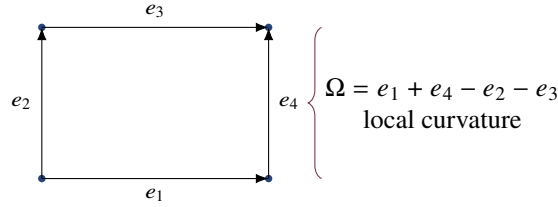
\begin{figure}[htbp]
\centering
\begin{tikzpicture}[scale=1.0,>=Latex]
  \coordinate (a) at (0,0); \coordinate (b) at (3,0);
  \coordinate (c) at (0,2); \coordinate (d) at (3,2);
  \fill[egblue] (a) circle (1.4pt); \fill[egblue] (b) circle (1.4pt);
  \fill[egblue] (c) circle (1.4pt); \fill[egblue] (d) circle (1.4pt);
  \draw[->] (a)--node[below,font=\scriptsize]{$e_1$}(b);
  \draw[->] (a)--node[left,font=\scriptsize]{$e_2$}(c);
  \draw[->] (c)--node[above,font=\scriptsize]{$e_3$}(d);
  \draw[->] (b)--node[right,font=\scriptsize]{$e_4$}(d);
  \node[align=center,font=\small] at (5.3,1) {$\Omega=e_1+e_4-e_2-e_3$\\local curvature};
  \draw[decorate,decoration={brace,amplitude=5pt},egwine] (3.7,0)--(3.7,2);
\end{tikzpicture}
\caption{An elementary elimination square. Nonzero circulation records order dependence. On a general complex, zero face curvature must still be supplemented by a period audit.}
\label{fig:integrability-square}
\end{figure}

\begin{conditionbox}
The elimination index is a finite product of nonempty local choice sets.  For every $S\subseteq[K]$, $j\notin S$, and fixed partial state $a_S$, the edge report satisfies $e_{j\mid S}(a_S,a_j)\ge0$ for every $a_j$ and $\inf_{a_j}e_{j\mid S}(a_S,a_j)=0$.  If a touching witness is declared, one must be supplied in each fiber and attain zero there.  Every elementary square has zero circulation.  These are compatibility conditions on a complete local defect system, not regularity assumptions on a statistical estimator.
\end{conditionbox}

\begin{theorem}[Flat elimination criterion]
\label{thm:flat-elimination}
On a product-of-stars elimination complex, let a real-valued local increment $e_{j\mid S}(a_S,a_j)$ be given for every $S\subseteq[K]$ and $j\notin S$.  There are potentials $V_S$ satisfying
\[
  e_{j\mid S}=V_{S\cup\{j\}}-V_S,
  \qquad
  V_S(a_S)=\inf_{a_j}V_{S\cup\{j\}}(a_S,a_j),
\]
if and only if, for every $S$, $j$, and fixed $a_S$, the edge reports are nonnegative with $\inf_{a_j}e_{j\mid S}(a_S,a_j)=0$, and every elementary square has zero circulation.
\end{theorem}

\begin{interpretationbox}
Local nonnegative gaps come from one global elimination tower exactly when their discrete curvature vanishes and fiberwise touching identifies every partial minimum.  Nonzero square curvature is therefore a falsification certificate.  On a general complex, harmonic periods must be added to the audit.
\end{interpretationbox}

\begin{proofroadmap}
Integrate the edge reports along a path from the base vertex.  Square-flatness makes the path sum invariant under adjacent swaps, hence well defined.  Fiberwise touching identifies the resulting vertex potential with the required partial minima.  The converse follows by telescoping a genuine potential around every square.  The chapter appendix gives both directions.
\end{proofroadmap}

The nonnegativity and fiberwise touching conditions make the potential a certified defect potential rather than an arbitrary scalar potential.

\section{Approximate flatness and Hodge repair}
\paragraph{Approximate flatness.}
\index{integrability!approximate flatness}

Two elimination orders differ by adjacent swaps.  Every swap crosses one elementary square.  Hence the discrepancy between path sums is bounded by the sum of absolute square curvatures along a swap sequence.  A Kendall-distance bound follows when a uniform curvature bound is available.

This provides a quantitative diagnostic: small local curvature implies limited order ambiguity, but does not guarantee exact integrability.

\par\medskip
\noindent\textbf{Hodge decomposition.}\par\smallskip
\index{Hodge decomposition}
\index{integrability!Hodge repair}

\begin{sourcebox}[title={Source result; proof not reproduced here}]
The full weighted Hodge decomposition, nearest-flat and nearest-exact
projections, period tax, and nearest-certified repair summarized below are
established in \EGII.  The chapter appendix proves the flat elimination
criterion, not this broader repair theory \citep{FrameworkPartII}.
\end{sourcebox}

Let $C^0,C^1,C^2$ be weighted cochain spaces with coboundary maps
\[
  d_0:C^0\to C^1,
  \qquad
  d_1:C^1\to C^2.
\]
An observed edge-defect field $e\in C^1$ decomposes orthogonally as
\[
  e=d_0V+d_1^\star\psi+h,
\]
where
\begin{itemize}
  \item $d_0V$ is exact;
  \item $d_1^\star\psi$ carries local curvature;
  \item $h$ is harmonic and carries global periods.
\end{itemize}

Combinatorial Hodge decompositions have been used to separate gradient,
cyclic, and harmonic components of graph data, notably in statistical ranking
\citep{JiangEtAl2011}.  The present use is narrower and auditable: the cochain
is an elimination defect field, and exactness is the condition for a global
potential with the declared units and orientation.

The exact and flat projections must be distinguished.  The nearest exact
defect field is
\[
  e^{\rm ex}=P_{\operatorname{im}d_0}e=d_0V,
\]
whereas the nearest flat defect field is
\[
  e^{\rm fl}=P_{\ker d_1}e=d_0V+h.
\]
On a contractible elimination complex the harmonic component vanishes and
the two projections coincide.  On a general complex, local face tests cannot
detect $h$.

\begin{decompositionbox}
A complete repair may pay three taxes: curvature removal, global period removal, and projection onto the cone of certified nonnegative/touching defect fields.
\end{decompositionbox}

\section{Repair, falsification, and audit boundaries}
\paragraph{Nearest certified repair.}
\index{repair!nearest certified}

Let $C_{\rm cert}\subseteq\operatorname{im}d_0$ be a declared nonempty
closed convex set of exact defect fields satisfying witness constraints.  Starting
from the exact projection $e^{\rm ex}$, the certified repair is
\[
  e_{\rm cert}=P_{C_{\rm cert}}e^{\rm ex}.
\]
Orthogonality gives
\[
  \|e-e_{\rm cert}\|_W^2
  =
  \|d_1^\star\psi\|_W^2
  +
  \|h\|_W^2
  +
  \|e^{\rm ex}-e_{\rm cert}\|_W^2.
\]
The three terms are curvature, period, and certification taxes.

\paragraph{Local falsification.}
\index{falsification!local integrability}

If a square has circulation $\Omega$, then at least one of its four edge reports must be wrong by at least $|\Omega|/4$ under the sup norm.  This turns local curvature into a falsification certificate for the collection of local claims.

\paragraph{When integrability must be audited.}
\index{integrability!audit ordering}

Architecture obstruction assumes that the local defect field belongs to one
declared objective.  This contract is automatic when the defect is derived
directly as $H_x(a)-J(x)$ from a single global objective.  It is not automatic
when local modules report increments that are later assembled into a defect field.
In the latter case, optimizing before validation can produce a precise answer
to an incoherent question.

This chapter closes the integrability gate before the architecture
obstruction results of Part IV are invoked.  The logical validation order is
\[
\begin{aligned}
  \text{global derivation}
  &\Longrightarrow \text{interpret obstruction},\\
  \text{assembled defect field}
  &\Longrightarrow \text{audit/repair}
  \Longrightarrow \text{interpret obstruction}.
\end{aligned}
\]
A reader may postpone the full Hodge repair construction, but any theorem
that combines separately reported local increments must either invoke this
audit or derive its defect directly from one global objective.

\subsection{Local generative vector fields under density and transport contracts}
\index{score field!integrability}
\index{generative model!density versus transport contract}

Score-based generation supplies a useful contract-sensitive example.  If a
reported field $s:\mathcal X\to\R^d$ is declared to be the score of one
positive density,
\[
  s(x)=\nabla\log p(x),
\]
then, on a simply connected smooth domain, a continuously differentiable
field must have symmetric Jacobian:
\[
  \partial_i s_j=\partial_j s_i.
\]
On a domain with holes, vanishing local curl must additionally be supplemented
by zero periods around noncontractible loops.  An antisymmetric Jacobian entry
or a nonzero loop integral is therefore a finite falsification certificate for
the declared global density or energy interpretation.  Projecting the field
onto an exact component, or parameterizing it directly as a scalar-energy
gradient, repairs the gradient-consistency part of that contract; the removed
coexact and harmonic components are the corresponding integrability taxes.
A density claim still requires the exponential potential to be integrable and
normalized.  This viewpoint is consistent with empirical audits of
conservativeness in learned score fields
\citep{ChaoEtAl2023ConservativeScore}.

The conclusion changes when the output contract changes.  A vector field used
only to define a transport ODE or sampler need not be conservative.  Nonzero
curl alone then does not certify poor samples, invalid transport, or excess
native generative loss.  The audit must instead use the declared terminal law
or transport objective.  Thus the same numerical field can fail a
global-score contract while remaining admissible under a sampling contract;
integrability is a typed obligation, not a model-name diagnosis.

\begin{boundarybox}
The gradient criterion, period obstruction, and Hodge projection are classical
Poincar\'e/Hodge facts.  This example introduces no new theorem about diffusion
or score-based models; it instantiates the chapter's audit sequence and marks
the precise boundary at which a density certificate ceases to be a sampling
certificate.
\end{boundarybox}

\paragraph{Operational ordering curvature.}
\index{curvature!operational ordering}

Algebraic elimination curvature asks whether two orders produce the same internal defect field.  Operational curvature, developed in Chapter~\ref{ch:operational-semantics}, asks whether a legal task and context can observe the difference.  An internal discrepancy may be operationally invisible; conversely, a context can expose a difference hidden at the base input.

\section*{Exercises}

\begin{exercise}
For a square elimination complex, solve the least-squares projection of arbitrary edge values onto exact increments.
\end{exercise}

\begin{exercise}
Construct a cochain on a cycle graph with zero local curvature but nonzero period.  Explain why local face tests cannot detect it.
\end{exercise}

\begin{exercise}
Prove the $|\Omega|/4$ local falsification bound and show that the constant is sharp.
\end{exercise}

\input{chapter_appendices/ch09_proofs}

%% file: chapter_appendices/ch09_proofs.tex
\chapterproofappendix

\subsection*{Proof of the flat certified-elimination criterion}
\proofdependency{The state space is a finite product of nonempty coordinate sets.  Nonnegativity and the condition $\inf_{a_j}e_{j\mid S}(a_S,a_j)=0$ in every fixed partial-state fiber certify conditional minimization; square flatness certifies path independence.}
\begin{proof}
Assume first that potentials $V_S$ exist with
\[
e_{j\mid S}=V_{S\cup\{j\}}-V_S,
\qquad
V_S(a_S)=\inf_{a_j}V_{S\cup\{j\}}(a_S,a_j).
\]
For every fixed $a_S$, the second identity immediately yields
\[
e_{j\mid S}\ge0,
\qquad
\inf_{a_j}e_{j\mid S}=0.
\]
For distinct $i,j\notin S$, both path sums around the elementary square telescope to
$V_{S\cup\{i,j\}}-V_S$, proving square flatness.

Conversely, fix the additive constant $V_\varnothing$.  For a nonempty subset $S$, choose an ordering $\pi=(\pi_1,\ldots,\pi_{|S|})$, put
$S_k^\pi=\{\pi_1,\ldots,\pi_k\}$, and define
\[
V_S(a_S)
=
V_\varnothing+
\sum_{k=1}^{|S|}
 e_{\pi_k\mid S_{k-1}^\pi}(a_{S_k^\pi}).
\]
Any two orderings of $S$ are connected by adjacent transpositions.  An adjacent transposition replaces one two-edge path around an elementary square by the other; square flatness says the two sums agree.  Hence $V_S$ is independent of the chosen ordering.

Append $j$ to an ordering of $S$.  The definition then gives
\[
V_{S\cup\{j\}}(a_S,a_j)-V_S(a_S)=e_{j\mid S}(a_S,a_j).
\]
For the fixed partial state $a_S$, nonnegativity and fiberwise touching give
\[
\inf_{a_j}V_{S\cup\{j\}}(a_S,a_j)
=V_S(a_S)+\inf_{a_j}e_{j\mid S}(a_S,a_j)
=V_S(a_S).
\]
Iterating this identity over the complement of $S$ yields
\[
V_S(a_S)=\inf_{a_{S^c}}V_{[K]}(a).
\]
Finally, once $V_\varnothing$ is fixed, every $V_S$ is the path integral of the edge field and is therefore unique.
\end{proof}

%% file: chapters/ch10_lift_complexity.tex
\chapter{Lift Admissibility, Slack Factorization, and Complexity}
\label{ch:lift-complexity}

In this chapter, \emph{lift admissibility} means target-faithfulness of a
nonnative lifted representation after slack, minimal-face, quotient, and
gauge reduction.  This is a representation-level gate.  A lift used to support
a deployment claim must also pass a second interface: quotient-faithful
extraction into a state whose oracle-relevant distinctions are measured in the
native defect.  The audit therefore proceeds from target slack, through
minimal-face and gauge reduction, to deployment extraction; finite auxiliary
dimension alone certifies none of these steps.

\begin{figure}[htbp]
\centering
\begin{tikzpicture}[>=Latex,node distance=6mm]
  \tikzset{ibox/.style={draw=egblue!70!black,fill=egblue!4,rounded corners=2pt,
    minimum width=28mm,minimum height=9mm,align=center,font=\scriptsize}}
  \node[ibox] (target) {target epigraph\\and slack};
  \node[ibox,right=of target] (lift) {admissible\\cone factorization};
  \node[ibox,right=of lift] (quot) {minimal-face and\\gauge quotient};
  \node[ibox,right=of quot] (dep) {target-visible\\deployment carrier};
  \draw[->,thick] (target)--(lift);
  \draw[->,thick] (lift)--(quot);
  \draw[->,thick] (quot)--(dep);
\end{tikzpicture}
\caption{The admissibility chain.  Exact elimination alone is insufficient: a lift must be tied to target slack, stripped of slack-invisible gauge, and then connected to a deployment state whose distinctions are measured by native defect.}
\label{fig:lift-admissibility-chain}
\end{figure}
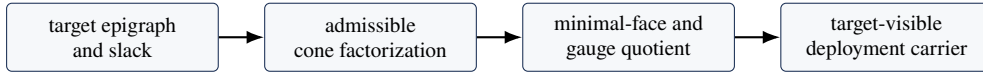

\section{Why exact lifts can be vacuous}
\index{lift!target-calling}
\index{lift!admissibility}

Given any target objective $J(x)$ and any chosen field $g(x)$,
\[
  H_g(x,a)=J(x)+\frac12\|a-g(x)\|^2
\]
is an exact lift.  Its oracle is $a^\star(x)=g(x)$.  By choosing $g$ we can manufacture arbitrary coherence, local metrics, or apparent geometric structure that was not present in the target.

This first observation is analytical: it does not assert that an arbitrary
$g$ has a finite representation in a declared lift grammar.  The finite-size
statement below therefore restricts $g$ to an affine field, or more generally
to a field whose graph and penalty composition are explicitly representable
by that grammar.

A constant $g$ makes the oracle globally coherent.  An affine $g(x)=Mx+b$
manufactures frozen-oracle metric $M^\top M$.  With suitable primitives,
arbitrary positive-semidefinite Hessians can be appended at fixed
dimension-dependent overhead.

\begin{conditionbox}
An exact lift $L$ of size $m$ is already available, with objective
$H_L(x,z)$.  The appended field is affine, $g(x)=Mx+b$; the same construction
applies to a nonaffine field only when its graph and the composed penalty are
explicitly representable in the grammar.  A closed convex penalty
$\psi:\R^q\to[0,\infty]$ vanishes only at zero, its epigraph has fixed
representation cost $c_\psi(q)$, and the cone dictionary is stable under the
required product.
\end{conditionbox}

\begin{theorem}[Finite-complexity target-calling no-go]
\label{thm:target-calling-nogo}
Under the preceding conditions, $L$ has an exact augmentation
\[
  H_{L[g,\psi]}(x,z,a,s)
  =H_L(x,z)+s+\iota_{\{s\ge\psi(a-g(x))\}}
\]
of size at most $m+c_\psi(q)$, whose appended exact oracle is
$a^\star(x)=g(x)$.  The overhead is independent of the coefficients $M,b$.
Consequently a constant $g$ appends a dummy coherent coordinate; a
polyhedral $\ell_1$ primitive appends any affine field at fixed
dimension-dependent overhead; and a quadratic primitive produces frozen-oracle
Hessian $M^\top M$, hence any prescribed positive-semidefinite local metric.
\end{theorem}

\begin{interpretationbox}
A finite extension-size claim is meaningless if the grammar permits target-calling or dummy coordinates: one can manufacture an exact, apparently coherent lift without explaining the target.  The theorem is a no-go for an insufficient notion of complexity, not for all finite lifts.
\end{interpretationbox}

\begin{proofroadmap}
Append the representable epigraph of $\psi(a-g(x))$, eliminate $(a,s)$, and
verify that its unique minimum is $(g(x),0)$.  Product stability gives the
fixed overhead, while the affine and quadratic specializations yield the
dummy and prescribed-metric conclusions.  The chapter appendix gives the
complete construction.
\end{proofroadmap}

\section{Epigraph geometry and intrinsic lift structure}
\index{epigraph geometry}
\index{slack factorization}
\index{conic lift}

Thus finite auxiliary dimension or extension size alone does not define lift
admissibility.  The target-calling no-go and the quotient-faithful lift
interface in this chapter come from the Lift Complexity companion
\citep{LiftComplexity}.  For a convex objective on a compact domain, consider
a truncated epigraph body
\[
  K_J=\{(x,t):x\in X,\ J(x)\le t\le M\}.
\]
A proper conic lift of the objective corresponds to an extended formulation of $K_J$.  The intrinsic target object is the slack operator
\[
  S_J(u,v)=1-\langle u,v\rangle
\]
after the appropriate normalization of extreme points and supporting functionals.

Classical extension-complexity theory relates lift size to factorization of the
slack operator: nonnegative rank governs polyhedral lifts, and cone
factorizations extend the principle to semidefinite and other convex lifts
\citep{Yannakakis1991,GouveiaParriloThomas2013,FawziEtAl2015}.  Semidefinite
hierarchies and extension-complexity lower bounds provide complementary
parts of this literature
\citep{Lasserre2001,Parrilo2003,FioriniEtAl2015,FioriniRothvossTiwary2012}.

\begin{conditionbox}
The target epigraph is a normalized compact convex body, the cone lift is proper and finite dimensional, and redundant directions invisible to the slack operator are quotiented out.  Minimal-face reduction is used to remove artificial cone dimensions.
\end{conditionbox}

\begin{theorem}[Conic lift--factorization gate]
\label{thm:lift-factorization}
Every proper $\mathcal K$-lift of the truncated epigraph induces a
$\mathcal K$-factorization of its slack operator, and every
$\mathcal K$-factorization induces a $\mathcal K$-lift.  A nonproper lift is
handled after restriction to the minimal face containing its affine slice.
Polyhedral and semidefinite lifts correspond to nonnegative and
positive-semidefinite factorizations.
\end{theorem}

This is the lift--factorization theorem of
\citet{GouveiaParriloThomas2013}, extending the polyhedral theorem of
\citet{Yannakakis1991}.

\begin{interpretationbox}
Admissible conic lifts are exactly factorizations of the target slack operator, up to the declared quotient.  This makes lift geometry target-generated and prevents dummy directions from creating fake coherence.  The theorem classifies a representation interface; it does not equate extension size with unrestricted runtime.
\end{interpretationbox}

\begin{proofroadmap}
From a proper conic lift, pair primal cone representatives with dual exposing functionals to factor the slack.  Conversely, use a cone factorization to reconstruct a lifted feasible set and projection.  Minimal-face reduction and the slack-kernel quotient establish target faithfulness.  The chapter appendix gives both constructions.
\end{proofroadmap}

The theorem does not by itself define a useful auxiliary coherence field.  It supplies the gate through which lift design must pass.

\section[Gauge reduction and lift frontier]{Gauge reduction and the fidelity--coherence--complexity frontier}
\label{sec:reduced-slack-carrier}
\index{lift!gauge reduction}
\index{lift!slack carrier}
\index{representation!quotient reduction}

Factorizations are nonunique.  Cone automorphisms, duplicated factors, and
slack-invisible directions can change raw coordinates without changing the
represented objective.  A reduced slack carrier removes directions not seen
by the target slack operator.  A coherence rule is admissible only if it is
invariant under cone gauge and faithful under quotient refinement.

\begin{example}[Factor splitting artifact]
Duplicate each nonnegative factor $k$ times and divide its weight among copies.  A naive Euclidean factor-coherence penalty can decrease by a factor $1/k$ while the represented convex set is unchanged.  On the quotient-reduced carrier the apparent improvement disappears.
\end{example}

\paragraph{Fidelity--coherence--complexity frontier.}
\index{lift!fidelity--coherence--complexity frontier}
\index{extension complexity}

For cone family $\mathcal K$, size budget $m$, coherence budget $r$, and one-sided objective-fidelity error $\delta$, define
\[
  \mathfrak F_{J,\mathcal K}(m,r,\delta)
\]
as the least true objective defect achievable by a quotient-reduced $\mathcal K$-lift satisfying the budgets.

The exact-lift frontier is the slice $\delta=0$.  Approximate lifts introduce a model or formulation bias that must be separated from coupling obstruction.

A deployed lift admits the decomposition
\[
\begin{aligned}
\text{total defect}
={}&\text{complexity-limited obstruction}\\
&+\text{formulation-selection tax}\\
&+\text{within-formulation implementation tax},
\end{aligned}
\]
under the declared decomposition convention.

\section{Target-visible reduction and extraction}
\label{sec:lift-target-visible}
\index{lift!target-visible reduction}
\index{carrier!target-visible quotient}
\index{visible capacity}

Slack reduction answers which lift directions belong to the target.  It does
not yet answer which distinctions in a retained deployment state matter to
native defect.  In the common-fiber Legendre--Bregman regime there is a
second, canonical reduction.  Let $a^\star$ be the unique oracle, put
\[
  S=\nabla\Phi(a^\star),
  \qquad
  S_U=\E(S\mid U),
  \qquad
  a_U^\dagger=\nabla\Phi^\star(S_U)
\]
for a standard-Borel deployment statistic $U$, and define
\[
  \mathcal C_\Phi(U)
  =\E D_\Phi(a_U^\dagger\|a^\star)
  =\E D_{\Phi^\star}(S\|S_U).
\]
For every factorized deployment $\widehat a=q(U)$, conditional Bregman
projection gives the exact defect field
\begin{equation}
  \E D_\Phi(\widehat a\|a^\star)
  =\mathcal C_\Phi(U)
   +\E D_\Phi(\widehat a\|a_U^\dagger).
  \label{eq:lift-visible-decomposition}
\end{equation}
Thus $\mathcal C_\Phi(U)$ is the native obstruction caused by the carrier,
while the second term is a decoder tax.  The identity is the classical
conditional Bregman predictor decomposition
\citep{BanerjeeGuoWang2005,Pfau2025,AdlamEtAl2022}; its use as a canonical quotient
of an admissible lift is the additional step here.

Count only distinct conditional dual signatures:
\[
  \operatorname{vc}_\Phi(U)
  =\bigl|\operatorname{essran}S_U\bigr|,
  \qquad
  \mathcal C_\Phi(k)
  =\inf_{\operatorname{vc}_\Phi(U)\le k}\mathcal C_\Phi(U).
\]
Raw labels inducing the same $S_U$ therefore consume one visible state, not
several.  For finitely many separated oracle types, this is the ordinary
Bregman quantization frontier and is positive below the number of distinct
types.

\begin{conditionbox}
$\Phi$ is Legendre on a common open convex fiber; $S$ is integrable; all
conditional means lie in $\operatorname{ri}(\operatorname{dom}\Phi^\star)$.
The coarsening identity below is asserted only when
$\mathcal C_\Phi(U)$ and $\mathcal C_\Phi(V)$ are both finite, so that its
left-hand difference is defined.
For the transfer clause below, every exact deployment counted by the lift
frontier has a measurable, quotient-faithful extraction
$\widehat a_{L,z}=q_{L,z}(U_{L,z})$, a uniform native exchange constant
$c>0$, and the stated finite visible-capacity bound.
\end{conditionbox}

\begin{theorem}[Visible quotient and lift-to-carrier transfer]
\label{thm:lift-visible-transfer}
Let $R_U=S_U$.  Then $R_U$ is a quotient of $U$,
\[
  \E(S\mid R_U)=R_U,
  \qquad
  \mathcal C_\Phi(R_U)=\mathcal C_\Phi(U).
\]
If $V$ is a coarsening of $U$ and both
$\mathcal C_\Phi(U),\mathcal C_\Phi(V)<\infty$, then
\begin{equation}
\begin{aligned}
  \mathcal C_\Phi(V)-\mathcal C_\Phi(U)
  &=\E D_{\Phi^\star}(S_U\|S_V)\\
  &=\E D_\Phi(a_V^\dagger\|a_U^\dagger)\ge0,
\end{aligned}
\label{eq:lift-coarsening-tax}
\end{equation}
with equality exactly when $S_U=S_V$ almost surely.  Hence
$\sigma(R_U)$ is, modulo null sets, the coarsest quotient of $U$ preserving
optimal native defect.

Suppose, in addition, that every quotient-reduced exact lift of size at most
$m$ and every deployment of coherence at most $r$ admits the declared
extraction and satisfies
\[
  \text{true defect}
  \ge c\,\E D_\Phi(\widehat a_{L,z}\|a^\star),
  \qquad
  \operatorname{vc}_\Phi(U_{L,z})
  \le\kappa_{\mathcal K}(m,r).
\]
Then the exact lift frontier obeys
\begin{equation}
  \mathfrak F_{J,\mathcal K}(m,r,0)
  \ge
  c\,\mathcal C_\Phi\!\left(\kappa_{\mathcal K}(m,r)\right).
  \label{eq:lift-extraction-transfer}
\end{equation}
\end{theorem}

\begin{interpretationbox}
There are two different quotients.  The reduced slack carrier removes
representation gauge; $R_U$ then removes deployment labels invisible to the
native defect.  A quotient-faithful extraction theorem is the bridge between
them.  Cone size $m$ and visible capacity $k$ remain different units:
there is no universal $\kappa_{\mathcal K}$ determined by cone order alone.
\end{interpretationbox}

\begin{proofroadmap}
Conditional expectation makes $R_U$ sufficient for its own dual mean.  The
Bregman predictor decomposition applied to a coarsening gives the exact tax
\eqref{eq:lift-coarsening-tax}.  Applying \eqref{eq:lift-visible-decomposition} to
each extracted deployment, then using the native exchange and taking the
frontier infimum, gives \eqref{eq:lift-extraction-transfer}.  The chapter
appendix supplies the complete argument.
\end{proofroadmap}

The upgraded lift chain can therefore be read as
\[
  (m,r)
  \xrightarrow{\text{quotient-faithful extraction}}
  \kappa_{\mathcal K}(m,r)
  \xrightarrow{\text{visible Bregman frontier}}
  c\,\mathcal C_\Phi\!\left(\kappa_{\mathcal K}(m,r)\right).
\]
Chapter~\ref{ch:resource-rd} starts after such a deployment carrier has been
declared or extracted and separates its information loss from decoder
nonsaturation.

\section{From lift size to native defect}
\paragraph{Scaling obstruction.}
\index{obstruction!scaling}

Positive rescaling of the objective preserves extension size while rescaling the native defect.  Therefore extension size cannot be a universal defect scale.  Complexity and objective loss live in different units until an exchange theorem is supplied.

\paragraph{Barrier-to-defect exchange.}
\index{barrier method!defect exchange}
\index{computational complexity!conditioning}

For a fixed encoded, well-conditioned conic lift with a self-concordant barrier, a primal--dual gap provides an objective defect certificate.  Iteration complexity depends on:
\begin{itemize}
  \item barrier parameter;
  \item coefficient encoding length;
  \item conditioning;
  \item cone-oracle cost;
  \item per-iteration linear algebra;
  \item requested numerical accuracy.
\end{itemize}

\begin{boundarybox}
An extension-complexity lower bound counts the size of an exact extended
formulation.  It is not a bit-complexity or runtime lower bound: coefficient
encoding length and numerical conditioning may grow even when extension size
does not.  A computational conclusion requires a declared oracle model,
barrier, conditioning regime, and linear-algebra cost.
\end{boundarybox}

\paragraph{Relationship to architecture obstruction.}
\index{extension complexity!versus runtime}
\index{obstruction!representation admissibility}

When an obstruction argument uses a nonnative lift, lift complexity answers a
logically earlier question than EOT or COT: do the auxiliary states arise from
an intrinsic factorization of the target rather than an arbitrary appended
coordinate?  Native auxiliary states require no separate lift audit, and a
lift whose target-faithfulness follows by construction has already discharged
the gate.

This chapter closes the target-faithfulness gate before nonnative lifts are
used in the architecture-obstruction results of Part IV.  The logical order is
\[
\begin{aligned}
\text{target epigraph and slack}
&\longrightarrow \text{reduced admissible carrier}\\
&\longrightarrow \text{coherence grammar}\\
&\longrightarrow \text{architecture obstruction}.
\end{aligned}
\]

\section*{Exercises}

\begin{exercise}
Verify that the target-calling lift is exact for every field $g$.  Explain why this makes unrestricted lift geometry nonfalsifiable.
\end{exercise}

\begin{exercise}
Construct the duplicated-factor artifact for a simplex slack matrix and compute the naive coherence before and after duplication.
\end{exercise}

\begin{exercise}
List the additional assumptions needed to convert a semidefinite extension into a numerical runtime statement.
\end{exercise}

\input{chapter_appendices/ch10_proofs}

%% file: chapter_appendices/ch10_proofs.tex
\chapterproofappendix

\subsection*{Proof of the finite-complexity target-calling no-go}
\proofdependency{The cone dictionary must represent the epigraph of the declared penalty primitive with fixed overhead and must be stable under products.  The oracle map is affine so that its graph can be imposed by affine formulation constraints.  A nonaffine field is covered only if the declared grammar separately represents its graph and the composed epigraph with the claimed overhead.}
\begin{proof}
Let the original exact lift be represented by auxiliary variable $z$ and objective $H_L(x,z)$, so that
\[
\inf_z H_L(x,z)=J(x).
\]
Append variables $a\in\R^q$ and $s\in\R$, together with the representable epigraph constraint
\[
s\ge\psi\{a-g(x)\}.
\]
The enlarged lifted objective is
\[
H_{L[g,\psi]}(x,z,a,s)
=H_L(x,z)+s+\iota_{\{s\ge\psi(a-g(x))\}}.
\]
Because $\psi\ge0$ and vanishes only at zero,
\[
\inf_{a,s}\{s:s\ge\psi(a-g(x))\}=0,
\]
and the infimum is attained uniquely at $a=g(x)$ and $s=0$.  Eliminating first $(a,s)$ and then $z$ therefore returns $J(x)$, so the enlarged formulation is exact.  Product stability of the cone dictionary adds at most the fixed representation cost $c_\psi(q)$, proving the size bound.

If $g$ is constant, the appended exact oracle is constant.  If $\psi=\|\cdot\|_1$, any affine field is appended with polyhedral overhead depending only on $q$.  If
$\psi(v)=\frac12\|v\|_2^2$ and the auxiliary state is frozen at $a=g(x_0)$, then
\[
\frac12\|g(x_0)-g(x)\|^2
=
\frac12\|M(x-x_0)\|^2
=
\frac12(x-x_0)^\top M^\top M(x-x_0).
\]
Its Hessian is $M^\top M$.  Every positive semidefinite matrix has such a factorization, establishing the final claim.
\end{proof}

\subsection*{Proof of the conic lift--factorization gate}
\proofdependency{The body is full dimensional, compact, and affinely normalized with the origin in its interior.  Properness ensures conic dual certificates on the ambient cone; a nonproper lift is first restricted to its minimal face.}
\begin{proof}
We give the standard lift--factorization construction.  Suppose first that
\[
C=\pi(K\cap L)
\]
is a proper $K$-lift, where $L$ is an affine subspace meeting $\operatorname{int}K$.  For every extreme point $x\in C$, choose a lifted representative $A(x)\in K\cap L$ with $\pi A(x)=x$.  For every $u\in\operatorname{ext}(C^\circ)$, the inequality
\[
\langle u,x\rangle\le1
\]
is valid on the projection.  Conic strong duality for the proper slice supplies a dual vector $B(u)\in K^\ast$ whose pairing with any lifted feasible point equals the slack of that inequality.  In particular,
\[
1-\langle u,x\rangle=\langle A(x),B(u)\rangle.
\]
Thus the intrinsic slack operator has a $K$-factorization.

Conversely, suppose maps
\[
A:\operatorname{ext}(C)\to K,
\qquad
B:\operatorname{ext}(C^\circ)\to K^\ast
\]
satisfy
\[
1-\langle u,x\rangle=\langle A(x),B(u)\rangle.
\]
Form the affine set of pairs $(x,z)$ satisfying
\[
\langle z,B(u)\rangle=1-\langle u,x\rangle
\quad\text{for every }u\in\operatorname{ext}(C^\circ),
\qquad z\in K.
\]
Every extreme point $x$ of $C$ has feasible witness $z=A(x)$, and convexity supplies witnesses for all of $C$.  Conversely, if $(x,z)$ satisfies the displayed system, then
\[
1-\langle u,x\rangle=\langle z,B(u)\rangle\ge0
\]
for every extreme point of $C^\circ$, hence for every $u\in C^\circ$.  By the bipolar theorem, $x\in C$.  Therefore the projection of this affine slice of $K$ is exactly $C$.

If the original lift is not proper, intersect $K$ with the minimal face containing the affine slice.  The slice is proper relative to that face, and the same argument applies there.
\end{proof}

\subsection*{Proof of the visible quotient and lift-to-carrier transfer}
\proofdependency{The conditional Bregman decomposition uses the common-fiber
Legendre assumptions stated in the chapter.  Doob--Dynkin is applied to the
standard-Borel statistic $U$.  The frontier conclusion additionally uses the
declared quotient-faithful extraction, native exchange, and capacity bound;
none is inferred from cone order.}
\begin{proof}
The random variable $R_U=S_U$ is a measurable function of $U$.  Since it is
also $R_U$-measurable, the tower property gives
\[
  \E(S\mid R_U)
  =\E\{\E(S\mid U)\mid R_U\}=R_U.
\]
Substitution in the dual expression for $\mathcal C_\Phi$ shows that
$\mathcal C_\Phi(R_U)=\mathcal C_\Phi(U)$.

If $\sigma(V)\subseteq\sigma(U)$, then
$S_V=\E(S_U\mid V)$.  The conditional Bregman predictor identity, now in
dual coordinates, yields
\[
  \E D_{\Phi^\star}(S\|S_V)
  =\E D_{\Phi^\star}(S\|S_U)
   +\E D_{\Phi^\star}(S_U\|S_V).
\]
Legendre duality converts the last term to
$\E D_\Phi(a_V^\dagger\|a_U^\dagger)$.  Strict convexity makes it zero
exactly when $S_U=S_V$ almost surely.  Any defect-preserving coarsening must
therefore retain $S_U$, proving the coarsest-quotient claim modulo null sets.

Finally fix a deployment counted by
$\mathfrak F_{J,\mathcal K}(m,r,0)$.  Its factorized extraction and
\eqref{eq:lift-visible-decomposition} imply
\[
  \text{true defect}
  \ge c\,\E D_\Phi(\widehat a_{L,z}\|a^\star)
  \ge c\,\mathcal C_\Phi(U_{L,z})
  \ge c\,\mathcal C_\Phi\!\left(\kappa_{\mathcal K}(m,r)\right).
\]
The last inequality is the definition of the visible-capacity frontier.
Taking the infimum over all exact deployments proves
\eqref{eq:lift-extraction-transfer}.
\end{proof}

%% file: part_notes/part03_history.tex
\parthistoricalnotes{}
\index{lift!historical comparison}
\index{integrability!historical comparison}

\subsection*{Two different validity questions}

The Part separates an integrability question from a representation question.
Cycle-flat edge data are gradients of a potential under familiar discrete
integrability conditions, as in exact potential games and combinatorial
Hodge theory \citep{MondererShapley1996,JiangEtAl2011}.  Separately, a convex
lift is intrinsic only when it factors the target slack operator: the
polyhedral equivalence is due to \citet{Yannakakis1991}, and the general
closed-cone equivalence to \citet{GouveiaParriloThomas2013}.

These are not interchangeable audits.  A perfectly integrable local defect
field can be attached to a target-calling auxiliary coordinate; a valid slack
factorization can still carry duplicated or gauge directions whose raw
coherence has no target meaning.

\begin{center}
\small
\begin{tabularx}{0.96\textwidth}{@{}p{0.21\textwidth}p{0.34\textwidth}X@{}}
\toprule
Gate & Established antecedent & Part III increment or specialization \\
\midrule
Flatness & potential/cycle tests & adds nonnegativity and fiberwise touching
so the potential is a partial-minimum elimination tower \\
Lift admissibility & slack-factorization/extension complexity & imports the
classical iff theorem as a gate and removes minimal-face and gauge artifacts \\
Target-calling no-go & exact augmentation by representable penalties &
elementary diagnostic specialization showing why size alone is not
admissibility \\
Visible quotient & conditional Bregman prediction
\citep{BanerjeeGuoWang2005} & retains only conditional dual
signatures and then requires an explicit resource-to-carrier extraction
theorem \\
\bottomrule
\end{tabularx}
\end{center}

The conic lift--factorization theorem is therefore explicitly classical.
The finite target-calling no-go and the quotient/extraction bridge are the
Part's program-specific claims, with the qualified priority statuses stated
in Appendix~\ref{app:principal-result-audit}.

%% file: chapters/ch11_second_elimination.tex
\chapter{Architecture Obstruction as a Second Elimination}
\label{ch:second-elimination}
\section{From a local defect to a class-level floor}
\index{elimination!second}
\index{obstruction!architecture}
\index{obstruction!class-level floor}

Once the native defect is fixed, architecture design becomes another
optimization problem.  Throughout Part IV, that defect is assumed either to
be derived directly from one declared global objective or to come with an
integrable local defect system.  Any nonnative lifted carrier is assumed to satisfy
the target-faithfulness gate.  Chapters~\ref{ch:integrability} and
\ref{ch:lift-complexity} audit these conditions retrospectively when they are
not guaranteed by construction.

For a population law $P$ and architecture class $\Arch$,
\[
  \Obs_P(\Arch)
  =
  \inf_{A\in\Arch}
  \E_P\Def_X\{A(X)\}.
\]
This is a second elimination: the first eliminates the local auxiliary state at each $x$; the second eliminates over globally deployable fields.

Every deployment $A\in\Arch$ has the exact defect field
\[
  \E_P\Def_X\{A(X)\}
  =
  \Obs_P(\Arch)
  +
  \implgap_P(A;\Arch).
\]
The architecture obstruction is a property of the triple
\[
  (\Def,P,\Arch),
\]
not of the architecture alone.  Changing the loss, population weighting, or output contract can change the tax.

\section{Saturation}
\index{saturation}
\index{obstruction!zero floor}

The first class-level question is whether the second elimination leaves any
irreducible defect.  The definition distinguishes a zero infimum from its
attainment by an exact oracle section.

\begin{definition}[Saturation]
An architecture class $\Arch$ is \emph{saturated} for an oracle family if its obstruction is zero under the declared contract.  It is \emph{exactly saturated} if it contains an exact oracle section.
\end{definition}

Saturation is the correct zero-tax condition.  Universal approximation in a generic norm is neither necessary nor sufficient without an exchange theorem connecting that norm to the native defect.

\begin{example}[Quotient saturation]
A vector-valued architecture may be nonsaturated for an eigenline oracle because of sign ambiguity.  A projector-valued architecture can be exactly saturated while using no more intrinsic information.
\end{example}

\section{Architecture comparison and witnesses}
\index{architecture!comparison}

If $\Arch_1\subseteq\Arch_2$, then
\[
  \Obs_P(\Arch_2)\le \Obs_P(\Arch_1).
\]
But a larger class can have worse generalization or optimization behavior.  The structural comparison isolates only the irreducible floor.

For a resource-indexed family $\{\Arch_r:r\ge0\}$ with $\Arch_r\subseteq\Arch_{r'}$ for $r\le r'$, the frontier
\[
  r\longmapsto \Obs_P(\Arch_r)
\]
is nonincreasing.  Plateaus identify ranges in which additional resources do not change realizability.

\paragraph{Lower and upper architecture witnesses.}
\index{obstruction!lower witness}
\index{obstruction!upper witness}

The same native defect supports both directions of the theory: a concrete
deployment gives a constructive upper witness, while a flow, topological,
resource, or information argument gives a converse lower certificate;
together, the two should bracket the obstruction.

\paragraph{Lower witness.}
A dual flow, topological argument, resource counting theorem, or information contraction gives
\[
  \underline{\Obs}\le \Obs_P(\Arch).
\]

\paragraph{Upper witness.}
An explicit deployment $A_{\rm wit}\in\Arch$ gives
\[
  \Obs_P(\Arch)
  \le
  \E_P\Def_X\{A_{\rm wit}(X)\}.
\]

A sharp theory seeks matching witnesses.  Even when they do not match, the gap is itself an auditable unresolved region.

\section{Architecture obstruction versus approximation and optimization}
\index{obstruction!versus approximation error}

Classical approximation theory studies
\[
  \inf_{f\in\mathcal F}\|f-f^\star\|.
\]
Architecture obstruction differs in four ways.

\begin{enumerate}
  \item The target may be a fiber, orbit, or set rather than one function.
  \item The error loss scale is generated by an eliminated objective.
  \item The architecture contract may include sharing, memory, topology, and output semantics.
  \item The obstruction may disappear under a quotient or atlas without enlarging ordinary function capacity.
\end{enumerate}

The two theories are complementary.  When the oracle is a unique function and the defect is equivalent to a norm, architecture obstruction reduces to a familiar approximation problem.

\paragraph{Architecture obstruction versus optimization.}
\index{obstruction!versus optimization error}

Suppose a training algorithm returns $\widehat A$.  A large achieved defect may arise because
\[
  \implgap_P(\widehat A;\Arch)>0
\]
or because
\[
  \Obs_P(\Arch)>0.
\]
Training loss alone does not separate the two.  One needs a class-level lower certificate and an upper witness.

A particularly strong empirical design compares:
\begin{enumerate}
  \item more optimization within the baseline class;
  \item more parameters within the same contract;
  \item a targeted contract change predicted by the obstruction;
  \item random or capacity-matched repairs.
\end{enumerate}

\section{Output, risk, and theorem contracts}
\index{architecture!output contract}
\index{architecture!risk contract}
\index{architecture!regularity contract}

The same oracle incidence can have different taxes.

\paragraph{Output contract.}
A point, unordered set, orbit, projector, probability law, or charted section are different outputs.

\paragraph{Aggregation contract.}
Uniform, average, tail, or task-weighted risk can see different parts of the obstruction.

\paragraph{Regularity contract.}
Continuous, measurable, Lipschitz, finite-memory, and neural-parameterized deployments form different classes.

\begin{principlebox}
An architecture tax is never stated without naming the output object, regularity class, population aggregation, and native defect.
\end{principlebox}

\paragraph{A minimal theorem template.}
\index{obstruction!theorem template}

A complete architecture theorem should contain four clauses.
\begin{enumerate}
  \item \textbf{Oracle clause:} identify the local oracle set.
  \item \textbf{Obstruction clause:} prove a lower bound over the declared architecture.
  \item \textbf{Witness clause:} exhibit an architecture or deployment attaining or approaching the bound.
  \item \textbf{Repair clause:} identify a contract change that changes the bound, with its own proof.
\end{enumerate}

The regular, coordination, singular, and resource theories in the next chapters instantiate this template.

\section*{Exercises}

\begin{exercise}
Prove monotonicity of the obstruction under architecture inclusion and under weakening of the regularity contract.
\end{exercise}

\begin{exercise}
Construct two output contracts on the same oracle family for which one has zero obstruction and the other has positive obstruction.
\end{exercise}

\begin{exercise}
Explain why a class-level lower bound is needed to distinguish architecture saturation from an optimizer that simply failed.
\end{exercise}

%% file: chapters/ch12_regular_transfer.tex
\chapter{Regular Obstruction Transfer and Flow Duality}
\label{ch:regular-transfer}
\section{The regular regime and quarter transport bound}
\index{obstruction transfer!regular regime}
\index{oracle geometry!regular}

The regular-transfer, flow-duality, and population-thickening program developed
in this chapter is presented in full in the companion manuscript
\emph{Elimination Obstruction Transfer} \citep{EOT}; here it is stated in the
monograph's unified notation.

Assume a unique oracle $a^\star(x)$ and a defect with local growth
\[
  \Def_x(a)
  \ge
  \frac\mu2 d^2\{a,a^\star(x)\}.
\]
Let the architecture class consist of $L$-Lipschitz maps $A:\X\to\A$.  If the oracle field varies faster than $L$, the architecture must deviate somewhere.

For two points $x,y$,
\[
  d\{A(x),A(y)\}\le Ld_\X(x,y).
\]
By the triangle inequality,
\[
  d\{a^\star(x),a^\star(y)\}
  \le
  e_x+Ld_\X(x,y)+e_y,
\]
where $e_x=d\{A(x),a^\star(x)\}$.  Hence
\[
  e_x+e_y
  \ge
  \bigl[d\{a^\star(x),a^\star(y)\}-Ld_\X(x,y)\bigr]_+.
\]

\index{obstruction transfer!quarter transport bound}
\index{Lipschitz architecture}

Using $u^2+v^2\ge (u+v)^2/2$, strong growth gives
\[
  \Def_x\{A(x)\}+\Def_y\{A(y)\}
  \ge
  \frac\mu4
  \bigl[d\{a^\star(x),a^\star(y)\}-Ld_\X(x,y)\bigr]_+^2.
\]
Averaging over selected pairs yields an architecture lower bound.

\begin{conditionbox}
The input law $P$ and oracle map $a^\star$ are measurable, the oracle is
single-valued in a metric space, and every deployed architecture is
$L$-Lipschitz.  The displayed expectations must be finite (or interpreted as
extended nonnegative values).  The native-defect conclusion additionally
requires the pointwise exchange inequality
$\Def_x(a)\ge(\mu/2)d^2\{a,a^\star(x)\}$ with $\mu>0$.
\end{conditionbox}

\begin{theorem}[Oracle-variation transport bound]
\label{thm:oracle-transport}
Let $\Pi(P,P)$ denote the set of couplings of $P$ with itself.  Then
\[
  \Obs_P(\Arch_L)
  \ge
  \frac\mu8
  \sup_{\pi\in\Pi(P,P)}
  \E_{(X,Y)\sim\pi}
  \bigl[d\{a^\star(X),a^\star(Y)\}-Ld_\X(X,Y)\bigr]_+^2.
\]
Equivalently, every fixed self-coupling $\pi$ supplies a valid, possibly
weaker, lower certificate by omitting the supremum.
\end{theorem}

\begin{interpretationbox}
The theorem converts variation of the oracle field that exceeds what an $L$-Lipschitz deployment can follow into a quantitative architecture lower bound.  It may be zero when the oracle variation lies within the architecture budget; it is a certificate, not a universal claim that every smooth architecture fails.
\end{interpretationbox}

\begin{proofroadmap}
For each coupled pair, use the triangle inequality to lower-bound the sum of the two pointwise deployment errors by the excess oracle displacement.  Square, average, and sum over edges; then use the exchange inequality to translate metric error into native objective defect.  The complete constants are verified in the chapter appendix.
\end{proofroadmap}

The coefficient $\mu/8$ is fixed by the convention that $\pi$ is a
probability coupling with both marginals equal to $P$.  Alternative
unnormalized edge sums require their own normalization; this does not alter the
population theorem.

\section{Finite calibration, dual flow, and population thickening}
\paragraph{Finite calibration graphs.}
\index{calibration graph}

In data analysis the instance space is represented by a graph $G=(V,E)$.  At vertex $i$, an estimated local oracle is $\widetilde a_i$ with uncertainty radius $\epsilon_i$.  Define robust edge demand
\[
  b_{ij}
  =
  \bigl[d(\widetilde a_i,\widetilde a_j)-\epsilon_i-\epsilon_j-Ld_{ij}\bigr]_+.
\]
Every admissible field has vertex errors $e_i\ge0$ satisfying
\[
  e_i+e_j\ge b_{ij}
  \qquad ((i,j)\in E).
\]
A quadratic lower certificate is obtained by minimizing
\[
  \min_{e_i\ge0}
  \sum_i w_i e_i^2
  \quad\text{subject to}\quad
  e_i+e_j\ge b_{ij}.
\]
Strong defect growth converts the solution to native objective units.

\paragraph{Dual flow.}
\index{flow duality}
\index{transport!dual flow}

The convex program has a flow-like dual.  Introduce nonnegative edge multipliers $\lambda_{ij}$.  The Lagrangian is
\[
  \sum_i w_ie_i^2
  +
  \sum_{(i,j)}\lambda_{ij}(b_{ij}-e_i-e_j).
\]
Minimization over $e_i$ gives
\[
  e_i=\frac{1}{2w_i}\sum_{j:(i,j)\in E}\lambda_{ij}
\]
when the nonnegativity constraint is inactive; the complete formula uses the positive part.  The dual objective becomes
\[
  \sum_{(i,j)}\lambda_{ij}b_{ij}
  -
  \sum_i\frac{1}{4w_i}
  \left(\sum_{j}\lambda_{ij}\right)^2.
\]
Maximizing over $\lambda\ge0$ yields the exact convex lower certificate in the quadratic graph model.

This calculation is a direct Lagrange dual of the displayed finite program,
not a claim to originate network-flow duality or order-restricted regression.
Its closest classical lineage includes the least-squares isotonic formulation
and its cumulative-sum/dual characterizations
\citep{BarlowBrunk1972,Rockafellar1984}.  The contribution here is the
oracle-mismatch interpretation and conversion of the optimum into native
objective units.

\begin{decompositionbox}
The dual variables select incompatible oracle differences and charge their total demand against congestion at the vertices.  The certificate is objective-denominated after multiplication by the defect-growth constant.
\end{decompositionbox}

\paragraph{Population thickening.}
\index{population thickening}

A pointwise mismatch becomes a positive population tax only if the population places mass near the mismatch.  Suppose $A$ and $a^\star$ are regular and at $x_0$ the distance is at least $\delta$.  If both fields are Lipschitz, the distance remains at least $\delta/2$ on a ball of radius proportional to $\delta$.  A lower mass condition
\[
  P\{B(x,r)\}\ge c r^d
\]
then gives a population lower bound of order
\[
  \mu c\,\delta^{d+2}.
\]

This step is essential.  A deterministic point obstruction on a nonatomic population can otherwise have zero average cost.

\section{Set-valued oracles and certificate boundaries}
\index{oracle!set-valued}

For a finite oracle cover, replace point distance by
\[
  d\{A(x),\mathcal O(x)\}.
\]
Local branch transport is needed to compare oracle elements across nearby points.  If the cover is separated and transport-consistent, a regular lower bound can be applied branchwise.  When branches collide, the assumptions fail and the singular theory of Chapter~\ref{ch:singular} is required.

\paragraph{What the certificate does not prove.}
\index{claim boundary!regular transfer}

A positive graph certificate proves that the declared baseline architecture cannot follow the estimated oracle field within its uncertainty radii.  It does not by itself prove:
\begin{itemize}
  \item that the local oracle model is scientifically correct;
  \item that an atlas will generalize;
  \item that the defect is visible to the final task;
  \item that the graph faithfully represents the population geometry.
\end{itemize}
Each claim requires a later gate.

\section*{Exercises}

\begin{exercise}
Derive the two-point quarter bound carefully and track constants under equal vertex weights.
\end{exercise}

\begin{exercise}
Compute the primal and dual certificates for a three-vertex path with demands $b_{12}=b_{23}=1$ and equal weights.
\end{exercise}

\begin{exercise}
Show how a lower mass condition converts a pointwise defect lower bound into a population bound.  Identify where regularity of the deployed field is used.
\end{exercise}

\input{chapter_appendices/ch12_proofs}

%% file: chapter_appendices/ch12_proofs.tex
\chapterproofappendix

\subsection*{Proof of the oracle-variation transport bound}
\proofdependency{The metric part uses only the triangle inequality and the $L$-Lipschitz architecture contract.  The native-risk conclusion additionally uses the quadratic exchange inequality $\Def_x(a)\ge(\mu/2)d_{\A}^2(a,a_x^\star)$.}
\begin{proof}
Fix $A\in\Arch_L$ and write
\[
e_A(x)=d_{\A}\{A(x),a_x^\star\}.
\]
For every pair $(x,y)$, the triangle inequality and the Lipschitz contract imply
\begin{align*}
d_{\A}(a_x^\star,a_y^\star)
&\le d_{\A}(a_x^\star,A(x))
   +d_{\A}(A(x),A(y))
   +d_{\A}(A(y),a_y^\star)\\
&\le e_A(x)+Ld_\X(x,y)+e_A(y).
\end{align*}
Therefore
\[
\bigl[d_{\A}(a_x^\star,a_y^\star)-Ld_\X(x,y)\bigr]_+
\le e_A(x)+e_A(y).
\]
After squaring and using $(u+v)^2\le2u^2+2v^2$, for every self-coupling $\pi$ of $P_X$,
\begin{align*}
&\E_\pi
\bigl[d_{\A}(a_X^\star,a_Y^\star)-Ld_\X(X,Y)\bigr]_+^2\\
&\qquad\le
2\E_\pi e_A^2(X)+2\E_\pi e_A^2(Y)
=4\E e_A^2(X),
\end{align*}
because both marginals of $\pi$ equal $P_X$.  Taking the supremum over $\pi$ and then the infimum over $A\in\Arch_L$ proves
\[
\inf_{A\in\Arch_L}\E e_A^2(X)
\ge\frac14\sup_{\pi\in\Pi(P_X,P_X)}
\E_\pi[\cdots]_+^2.
\]
Finally, the exchange inequality gives, for every $A$,
\[
\E\Def_X\{A(X)\}\ge\frac\mu2\E e_A^2(X).
\]
Taking the infimum over $A$ and combining the two bounds yields the coefficient $\mu/8$.
\end{proof}

%% file: chapters/ch13_coordination.tex
\chapter{Rectangularity, Coordination Tax, and Memory}
\label{ch:coordination}
\section{Conditional architectures and their rectangular hull}
\index{conditional architecture}

Consider a finite history tree.  A deployed law $Q$ is determined by conditional kernels
\[
  q_h(\cdot\mid h)
\]
at each history $h$.  Let $P^\star$ be an oracle law with kernels $p_h^\star$.  At inverse temperature $\beta$, the native defect is
\[
  D_\beta(Q;P^\star)
  =
  \frac1\beta\KL(Q\|P^\star).
\]
The architecture obstruction is
\[
  O_\beta(\Arch;P^\star)
  =
  \inf_{Q\in\Arch}D_\beta(Q;P^\star).
\]
Every deployed law satisfies
\[
  D_\beta(Q;P^\star)
  =
  O_\beta(\Arch;P^\star)
  +
  \implgap_{\beta,\Arch}(Q;P^\star).
\]

\paragraph{Rectangular hull.}
\index{rectangularity!rectangular hull}

The local kernel sets available inside $\Arch$ can be freed and pasted independently across histories.

\begin{definition}[Rectangular hull]
$\operatorname{Rect}(\Arch)$ is the class obtained by allowing every conditional kernel that occurs locally in some member of $\Arch$ to be chosen independently at every history where it is admissible.
\end{definition}

Write
\[
  C_i^\Arch(h)
  =\{q_i(\cdot\mid h):Q^q\in\Arch\}
\]
for the locally available kernels at stage $i$ and history $h$.

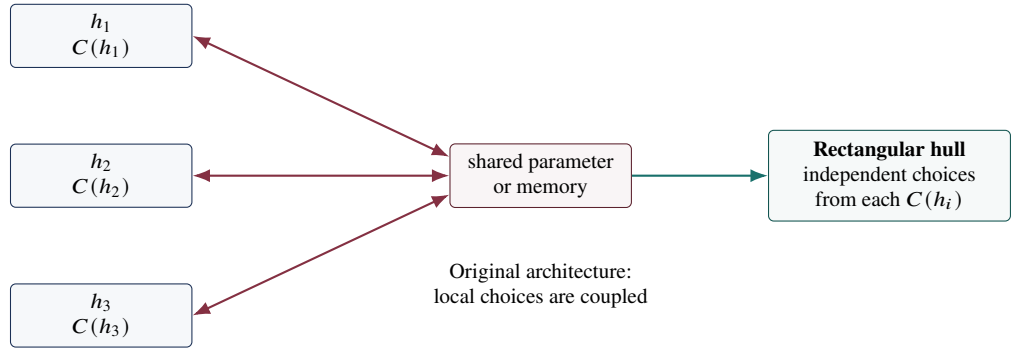
\begin{figure}[htbp]
\centering
\begin{tikzpicture}[>=Latex,node distance=10mm]
  \tikzset{
    hist/.style={draw=egblue!70!black,fill=egblue!4,rounded corners=2pt,
      minimum width=24mm,minimum height=8mm,align=center,font=\scriptsize},
    rect/.style={draw=egteal!75!black,fill=egteal!4,rounded corners=2pt,
      minimum width=32mm,minimum height=12mm,align=center,font=\scriptsize}
  }
  \node[hist] (h1) {$h_1$\\$C(h_1)$};
  \node[hist,below=of h1] (h2) {$h_2$\\$C(h_2)$};
  \node[hist,below=of h2] (h3) {$h_3$\\$C(h_3)$};
  \node[hist,right=34mm of h2,fill=egwine!5,draw=egwine!70!black] (shared) {shared parameter\\or memory};
  \node[rect,right=18mm of shared] (rect) {{\bfseries Rectangular hull}\\independent choices\\from each $C(h_i)$};
  \draw[<->,thick,egwine] (h1.east)--([yshift=2.5mm]shared.west);
  \draw[<->,thick,egwine] (h2.east)--(shared.west);
  \draw[<->,thick,egwine] (h3.east)--([yshift=-2.5mm]shared.west);
  \node[align=center,font=\scriptsize,below=6mm of shared] {Original architecture:\\local choices are coupled};
  \draw[->,thick,egteal] (shared.east)--(rect.west);
\end{tikzpicture}
\caption{Rectangularization preserves the locally available kernels but removes the cross-history constraint that forces them to share one parameterization or memory state.}
\label{fig:rectangular-hull}
\end{figure}

The rectangular hull removes cross-history coordination constraints while
preserving the local options.  Its obstruction obeys a Bellman recursion
because each history can then be optimized independently, conditional on
occupancy.  Rectangular uncertainty sets and robust Bellman recursions are
classical in robust dynamic programming and recursive multiple-priors models
\citep{Iyengar2005,NilimElGhaoui2005,EpsteinSchneider2003},
and tractable departures from rectangularity have also been studied
\citep{GoyalGrandClement2023}.  Here the object being rectangularized is not
an uncertainty set but a deployment architecture.  The loss created by forced
cross-history sharing is retained explicitly as a coordination tax.

\section{The coordination decomposition}
\index{rectangularity!coordination tax}
\index{rectangularity!coordination decomposition}

Since $\Arch\subseteq\operatorname{Rect}(\Arch)$,
\[
  O_\beta\{\operatorname{Rect}(\Arch);P^\star\}
  \le
  O_\beta(\Arch;P^\star).
\]
Define
\[
  \coordtax_\beta(\Arch;P^\star)
  =
  O_\beta(\Arch;P^\star)
  -
  O_\beta\{\operatorname{Rect}(\Arch);P^\star\}.
\]
\begin{sourcebox}[title={Source result; proof not reproduced here}]
Under a uniform positivity floor and compactness, COT proves the universal
converse: the original obstruction equals the rectangular Bellman value for
every full-support oracle if and only if $\Arch$ is rectangular.  The book
uses this classification as context but proves below only the Bellman and
coordination-residual decomposition needed by later chapters
\citep{COT}.
\end{sourcebox}

\begin{conditionbox}
The history tree is finite, every reference conditional kernel has full
support, and $\Arch$ is nonempty and compact, so its local kernel projections
are compact and the displayed minima are attained.  The rectangular hull
allows each history to choose its local rule independently while preserving
exactly the sets $C_i^\Arch(h)$.
\end{conditionbox}

\begin{theorem}[Rectangular Bellman and coordination-residual decomposition]
\label{thm:rectangular-decomposition}
Set $W_m=0$ and define backward values by
\[
  W_{i-1}(h)
  =
  \min_{q\in C_i^\Arch(h)}
  \left\{
    \frac1\beta\KL\{q\|p_i^\star(\cdot\mid h)\}
    +\sum_y q(y)W_i(h,y)
  \right\}.
\]
For a locally available kernel define
\[
  \rho_i(q;h)
  =
  \frac1\beta\KL\{q\|p_i^\star(\cdot\mid h)\}
  +\sum_yq(y)W_i(h,y)-W_{i-1}(h).
\]
Then
\[
  W_0=O_\beta\{\operatorname{Rect}(\Arch);P^\star\},
  \qquad
  \rho_i(q;h)\ge0.
\]
Every $Q^q\in\operatorname{Rect}(\Arch)$ satisfies the exact chain rule
\[
  D_\beta(Q^q;P^\star)
  =W_0+
  \sum_{i=1}^m
  \E_{Q^q}\rho_i\{q_i(\cdot\mid H_{i-1});H_{i-1}\}.
\]
Moreover,
\[
  \coordtax_\beta(\Arch;P^\star)
  =
  \min_{Q^q\in\Arch}
  \sum_{i=1}^m
  \E_{Q^q}\rho_i\{q_i(\cdot\mid H_{i-1});H_{i-1}\}.
\]
Consequently every $Q\in\Arch$ obeys
\[
  D_\beta(Q;P^\star)
  =
  O_\beta\{\operatorname{Rect}(\Arch);P^\star\}
  +\coordtax_\beta(\Arch;P^\star)
  +\implgap_{\beta,\Arch}(Q;P^\star).
\]
\end{theorem}

\begin{interpretationbox}
The exact defect field separates local conditional inadequacy from the extra cost of forcing one shared parameter, memory state, or network to coordinate all histories.  The coordination tax is measured in the native forward-KL/Bellman scale.  Rectangularization is an analytical relaxation; it is not itself the deployed architecture.
\end{interpretationbox}

\begin{proofroadmap}
Use the conditional KL chain rule to express every architecture loss as an expected sum of local Bellman residuals.  Minimize first over the rectangular hull to obtain the local obstruction, then add and subtract that value inside the original architecture infimum.  The chapter appendix gives the induction and exact decomposition.
\end{proofroadmap}

The theorem assigns a native cost to parameter sharing.  A shared network
couples decisions across histories; it is not merely a smaller list of local
functions.  Recent work studies when shared representations help in multitask
learning \citep{LiEtAl2025MultitaskRep}, how learned representations affect
offline transfer RL \citep{BoseDuFazel2025}, and how finite memory creates
return error in partially observed RL \citep{EberhardEtAl2025}.  Those results
motivate the same deployment pressure.  The theorem here isolates a narrower
quantity: the exact native cost of cross-history coupling after the locally
available kernels have been fixed.

Rectangularization is also distinct from operation C in the P/G/X/V/C
calculus of Chapter~\ref{ch:towers-pgxvc}.  It frees admissible historywise
choices; C keeps the declared marginals fixed and changes only their joint
coupling.

\paragraph{Local Bellman residuals.}
\index{Bellman residual}

The rectangular dynamic program generates a nonnegative residual at each history.  Under the occupancy induced by $Q$, the expected sum of local residuals equals
\[
  \coordtax+\implgap.
\]
This produces a diagnostic map: the global tax can be localized to histories where sharing is most expensive.

\paragraph{A shared-kernel example.}
\index{conditional policy!shared kernel}

Suppose the same kernel $q$ is used at several histories with strictly positive
oracle kernels $p_h^\star$ on a common finite output set.  Let the fixed
occupancy weights satisfy $w_h\ge0$ and
\[
  W=\sum_h w_h>0.
\]
Assume also that each $p_h^\star$ is locally admissible when histories are
freed, so the corresponding rectangular obstruction is zero.

The shared optimization problem is
\[
  \inf_{q\in\Delta(\mathcal Y)}
  \frac1\beta\sum_h w_h\KL(q\|p_h^\star).
\]
Define
\[
  Z_W=\sum_z\prod_h p_h^\star(z)^{w_h/W},
  \qquad
  q^\star(z)=Z_W^{-1}\prod_h p_h^\star(z)^{w_h/W}.
\]
Direct expansion gives the exact identity
\[
  \frac1\beta\sum_h w_h\KL(q\|p_h^\star)
  =
  \frac W\beta\KL(q\|q^\star)-\frac W\beta\log Z_W.
\]
Hence the minimum, and therefore the exact fixed-occupancy shared-kernel
coordination tax under the declared weights, is
\[
  -\frac W\beta\log Z_W.
\]
This is not the full endogenous coordination tax unless a separate occupancy
or fixed-point theorem shows that the weights $w_h$ are the occupancies induced
by the optimizing shared law.  When occupancies vary with $q$, they must remain
inside the architecture-level optimization.
Generalized H\"older gives $0<Z_W\le1$, so this tax is nonnegative; it
vanishes exactly when all oracle kernels carrying positive weight coincide.
When the weights are normalized, $W=1$, this reduces to the usual normalized
geometric-pooling formula \citep{GenestZidek1986}.

\section{Memory compression as base change}
\index{memory!compression}
\index{base change!memory compression}

Rectangularization removes cross-history coupling; memory compression changes
which histories are distinguishable before that coupling is assessed.  The
next identity treats this compression as a base change, and the source box
states which results are imported.

\begin{sourcebox}[title={Source result; proof not reproduced here}]
The memory base-change identity, canonical-fiber saturation criterion,
approximate-saturation bound, and dequantization refinements summarized in the
remainder of this chapter are established in COT.  They are not counted among
the principal book results; the chapter appendix proves the rectangular
Bellman and coordination-residual theorem above
\citep{COT}.
\end{sourcebox}

Within that imported scope, let a fine history $h$ map to a compressed memory
state $m=\rho(h)$.  The KL chain rule decomposes a fine law into a coarse memory
law and conditional fiber laws.  At architecture level,
\[
  O_{\rm fine}
  =
  \inf_{Q_M}
  \left\{
  O_{\rm coarse}(Q_M)
  +
  \Psi(Q_M)
  \right\},
\]
where $\Psi(Q_M)$ is the minimum fine conditional realization tax compatible with the coarse law.

A memory representation can preserve coarse objective values while failing to realize the canonical fine conditional kernels.  This is objective faithfulness without obstruction faithfulness.

\paragraph{Saturation and exact base change.}
\index{saturation}
\index{base change!exact}

Universal equality between fine and coarse architecture obstructions holds when the fine architecture is saturated by the canonical oracle fiber kernels.  Approximate saturation yields an additive upper bound.  This is stronger than saying that the coarse statistic is sufficient for one fixed oracle; it is an architecture-level statement over the declared class.

\section{Zero temperature}
\index{zero-temperature limit}

As $\beta\to\infty$, soft KL-regularized functionals may converge to hard costs.  For fixed architecture classes, uniform law-level dequantization passes to architecture minima.  For varying classes, $\Gamma$-convergence and inner/outer limit conditions are needed.

The zero-temperature limit can remain randomized.  Temperature removal alone does not force a deterministic architecture.  The admissible class and its closure determine the hard limit.

\begin{boundarybox}
Rectangularity is a property of conditional-kernel pasting.  It is not equivalent to network width, and it is not automatically restored by more training data.
\end{boundarybox}

\section*{Exercises}

\begin{exercise}
Derive the geometric-pooling solution for general weights and verify its
normalized-weight specialization.
\end{exercise}

\begin{exercise}
Give a nonrectangular architecture with two histories and compute its coordination tax explicitly.
\end{exercise}

\begin{exercise}
Construct a coarse memory map that preserves one-step action distributions but loses a future-relevant conditional distinction.
\end{exercise}

\input{chapter_appendices/ch13_proofs}

%% file: chapter_appendices/ch13_proofs.tex
\chapterproofappendix

\subsection*{Proof of the rectangular Bellman and coordination-residual decomposition}
\proofdependency{The finite history tree and strict positivity of the reference kernels make every local forward-KL term well defined.  Compactness gives existence of the local Bellman minima and of the architecture minimum.  The exact algebra uses the forward-KL chain rule.}
\begin{proof}
Let $Q^q$ be the path law generated by a conditional-kernel array $q$, and let $P^\star$ be the strictly positive oracle path law.  The chain rule for forward relative entropy gives
\[
\frac1\beta\KL(Q^q\|P^\star)
=
\sum_{i=1}^m
\E_{Q^q}
\left[
\frac1\beta\KL\{q_i(\cdot\mid H_{i-1})\|p_i^\star(\cdot\mid H_{i-1})\}
\right].
\]
In $\operatorname{Rect}(\Arch)$ the locally available kernel at every history can be selected independently.  Backward induction therefore defines values $W_m=0$ and
\[
W_{i-1}(h)
=
\min_{q\in C_i^\Arch(h)}
    \left\{
\frac1\beta\KL\{q\|p_i^\star(\cdot\mid h)\}
+
\sum_yq(y)W_i(h,y)
\right\},
\]
and yields
\[
W_0=O_\beta\{\operatorname{Rect}(\Arch);P^\star\}.
\]
Define the coordination tax by
\[
\coordtax_\beta(\Arch;P^\star)
=
O_\beta(\Arch;P^\star)
-
O_\beta\{\operatorname{Rect}(\Arch);P^\star\}.
\]
It is nonnegative because $\Arch\subseteq\operatorname{Rect}(\Arch)$.
For any $Q\in\Arch$, add and subtract $O_\beta(\Arch;P^\star)$:
\begin{align*}
D_\beta(Q;P^\star)
&=
O_\beta(\Arch;P^\star)
+
\{D_\beta(Q;P^\star)-O_\beta(\Arch;P^\star)\}\\
&=
O_\beta\{\operatorname{Rect}(\Arch);P^\star\}
+
\coordtax_\beta(\Arch;P^\star)
+
\implgap_{\beta,\Arch}(Q;P^\star).
\end{align*}
This is the claimed decomposition.

Define the local Bellman residual
\[
\rho_i(q;h)
=
\frac1\beta\KL\{q\|p_i^\star(\cdot\mid h)\}
+
\sum_yq(y)W_i(h,y)-W_{i-1}(h).
\]
It is nonnegative for every locally available kernel by Bellman optimality.
For every $Q^q\in\operatorname{Rect}(\Arch)$, expanding and telescoping the
value terms gives
\[
D_\beta(Q;P^\star)
=
W_0+
\sum_{i=1}^m\E_Q\rho_i\{q_i(\cdot\mid H_{i-1});H_{i-1}\}.
\]
Subtracting $W_0$ and minimizing over $Q\in\Arch$ identifies the minimum
expected residual sum with the coordination tax.  Adding and subtracting the
architecture minimum for an arbitrary $Q\in\Arch$ leaves exactly
$\implgap_{\beta,\Arch}(Q;P^\star)$, proving the final three-term decomposition.
\end{proof}

%% file: chapters/ch14_singular.tex
\chapter{Singular Fibers, Monodromy, and Catastrophe Taxes}
\label{ch:singular}
\index{oracle geometry!singular}
\index{discriminant}
\index{obstruction transfer!singular regime}

The regular theory assumes a unique oracle or a finite separated cover.  At a discriminant, branches collide, the vertical Hessian loses rank, and the oracle incidence map becomes stratified.  Distance-to-branch arguments can degenerate precisely where the topology changes.

Here monodromy is the permutation of locally tracked oracle branches after
continuation around a loop; when the declared output contract cannot absorb
that permutation, the resulting native loss is a catastrophe tax.
\index{topology!monodromy}

The singular theory begins from a normal form whose native residual remains meaningful even when curvature vanishes.

\begin{figure}[htbp]
\centering
\begin{tikzpicture}[scale=1.0,>=Latex]
  \tikzset{outcome/.style={rounded corners=2pt,align=center,font=\small,
    text width=42mm,minimum height=11mm}}
  \draw[thick,egblue] (0,0) circle (1.25);
  \foreach \ang/\lab in {0/$0$,90/$\pi/2$,180/$\pi$,270/$3\pi/2$}{
    \fill[egblue] ({1.25*cos(\ang)},{1.25*sin(\ang)}) circle (1.2pt);
  }
  \draw[->,thick,egwine] (0:1.55) arc (0:320:1.55);
  \node[font=\scriptsize,align=center] at (0,-2.0) {one circuit in input space};
  \node[outcome,draw=egwine!75!black,fill=egwine!5] (branch) at (4.7,0.7)
    {$v(2\pi)=-v(0)$\\or a branch permutation};
  \node[outcome,draw=egteal!75!black,fill=egteal!4] (quotient) at (4.7,-0.7)
    {$vv^\top$ or an unordered set\\returns to itself};
  \draw[->,thick,egwine] (1.7,0.7)--(branch.west);
  \draw[->,thick,egteal] (1.7,-0.7)--(quotient.west);
\end{tikzpicture}
\caption{Monodromy depends on the output contract. A point-valued representative can fail to return to itself after a loop, while a quotient-valued object such as a projector or unordered root set can remain globally well defined.}
\label{fig:monodromy-contract}
\end{figure}
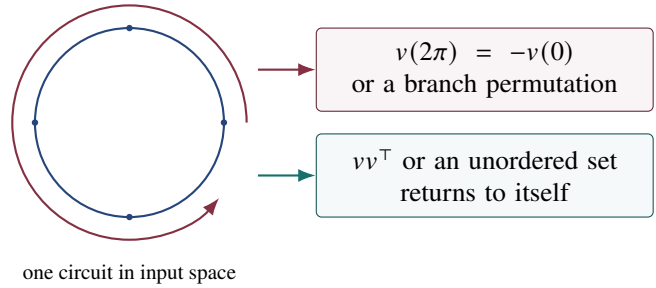

\section{The radical normal form}
\index{singularity!radical normal form}
\index{catastrophe tax}

Let
\[
  H_\theta(a)=|a^k-re^{i\ell\theta}|^2,
  \qquad
  a\in\C,
  \quad \theta\in S^1,
\]
with $k\ge2$, $r>0$, and $\ell\in\mathbb Z$.  Each fiber has $k$ zero-defect roots.  A continuous root selection exists exactly when $k$ divides $\ell$.

\begin{conditionbox}
The input is the circle, the deployment is a continuous single-valued complex function, the loss is uniform, $k\ge2$, and $r>0$.  The exponent $\ell$ determines the winding of the oracle fiber.  Changing to average risk, measurable sections, multiple charts, or quotient-valued outputs changes the problem.
\end{conditionbox}

\begin{theorem}[Exact radical catastrophe tax]
\label{thm:radical-tax}
\[
  \inf_{A\in C(S^1,\C)}
  \sup_{\theta\in S^1}
  |A(\theta)^k-re^{i\ell\theta}|^2
  =
  \begin{cases}
    0,&k\mid\ell,\\
    r^2,&k\nmid\ell.
  \end{cases}
\]
\end{theorem}

\begin{interpretationbox}
When $k$ does not divide $\ell$, no continuous point-valued selector can keep the residual strictly below $r^2$ everywhere, and the zero deployment attains exactly that value.  The result is a sharp worst-case architecture tax.  It does not by itself imply a positive unrestricted nonatomic average-risk lower bound.
\end{interpretationbox}

\begin{proofroadmap}
Assume a deployment has loss strictly below $r^2$.  Its image avoids zero, so its phase has an integer winding number.  The map $A^k$ then has winding divisible by $k$, but closeness inside the circle centered at the target forces winding $\ell$, a contradiction unless $k\mid\ell$.  The matching upper bound is $A\equiv0$.  The full homotopy argument is in the chapter appendix.
\end{proofroadmap}

The upper witness lies on the collision stratum.  The exact constant survives without strong convexity or branch separation and vanishes continuously as the target loop approaches the discriminant.

\section{Risk level, general loops, and degree}
\paragraph{Uniform versus average risk.}
\index{risk!uniform versus average}

For a nonatomic sampling law on the circle, the infimum average defect over continuous fields is zero even when the uniform tax is $r^2$.  A selector can follow one branch everywhere except on a narrow seam whose measure tends to zero.

Thus the theorem is a uniform deterministic architecture tax.  A positive average tax requires an additional regularity, lower-mass, or complexity condition preventing seam concentration.

\paragraph{General loops and degree.}
\index{topology!degree}
\index{topology!loop lifting}

For a nonvanishing target loop $z(\theta)$, winding incompatibility yields two-sided catastrophe bounds controlled by radial clearance from zero.  In higher dimensions, let $F:\R^d\to\R^d$ be positively homogeneous and consider its normalized map on the sphere.  If the target degree is incompatible with the degree of $F/|F|$, a constant-radius target sphere has an exact residual floor analogous to the radical case.

The important pattern is:
\[
\begin{aligned}
&\text{homotopy class incompatible with the output map}\\
&\hspace{2cm}\Longrightarrow
\text{native residual at least the distance to the discriminant}.
\end{aligned}
\]

\section{Beyond radicals: roots, eigenlines, and partial monodromy}
\paragraph{Polynomial roots and critical values.}
\index{polynomial roots}
\index{critical value}

Let $p$ be a complex polynomial and let a target loop avoid the critical-value discriminant.  Its roots form a covering with monodromy.  If the monodromy has no fixed branch, any continuous root-valued deployment must incur residual at least the target loop's distance to the critical-value set.  Near a ramification point of order $m$, analytic coordinates reduce the problem to the radical normal form.

Using discriminant complements, monodromy, and braid structure to obstruct
continuous algorithms has a classical topology-of-algorithms lineage
\citep{Smale1987,Vassiliev1988}.  The additional claim here is
objective-denominated: the native residual and clearance from the discriminant
turn that qualitative obstruction into a numerical loss floor.

\paragraph{Eigenline catastrophe.}
\index{eigenline catastrophe}
\index{spectral output}

Consider a real symmetric $2\times2$ matrix family around a conical crossing.  The top eigenline forms a projective bundle with odd holonomy.  An oriented continuous unit eigenvector must fail after one loop and pays an exact Rayleigh tax.  The rank-one projector
\[
  P=vv^\top
\]
is invariant under $v\mapsto-v$ and can remain continuous with zero defect.

This example demonstrates that the obstruction may be a representation artifact even when the underlying geometric object is perfectly regular.
It sits beside modern impossibility results for continuous canonicalization
and topological bottlenecks in learned representations
\citep{DymLawrenceSiegel2024,BatsonEtAl2021,EsmaeiliEtAl2023}.  The exact
Rayleigh tax here is a more specific statement in the native spectral loss,
not a consequence claimed from those adjacent results.

\paragraph{Missing branches versus monodromy.}
\index{topology!monodromy}
\index{oracle!missing branch}

Monodromy permutes branches that exist throughout a loop.  A missing branch is undefined on part of the path.  The two phenomena require partial rather than total transport maps.  Forcing every edge transport to be a permutation can manufacture spurious cycles and false global inconsistency.  The transport construction in Chapter~\ref{ch:transport-cycles} therefore treats transports as partial bijections and audits both fixed-point behavior and domain survival.
Partial maps and their synchronization are established in multi-object
matching; the distinction developed here is the oracle/task meaning assigned
to disappearance and domain survival, as documented with prior art in
Chapter~\ref{ch:transport-cycles}.

\section{Stability}
\index{singularity!stability}

The exact tax is stable under perturbations that preserve the homotopy class and maintain positive clearance from the discriminant.  When the target approaches the discriminant, the tax can vanish.  This is the correct singular behavior: topology remains incompatible, but the objective price of incompatibility becomes arbitrarily small.

\begin{boundarybox}
Topology alone does not supply a fixed numerical risk floor.  The price depends on the native defect and the target's geometric clearance from the singular set.
\end{boundarybox}

\section*{Exercises}

\begin{exercise}
Prove the radical theorem for $k=2$ and $\ell=1$ using only the argument principle.
\end{exercise}

\begin{exercise}
Construct a sequence of continuous fields whose average radical defect tends to zero under the uniform measure, while the maximum defect remains at least $r^2$.
\end{exercise}

\begin{exercise}
For a loop of eigenlines with sign holonomy, explain why a projector-valued output is a quotient repair rather than a higher-capacity approximation.
\end{exercise}

\input{chapter_appendices/ch14_proofs}

%% file: chapter_appendices/ch14_proofs.tex
\chapterproofappendix

\subsection*{Proof of the exact radical catastrophe tax}
\proofdependency{The proof uses continuity on the circle and the integer winding number of loops in $\C^\star$.  The lower bound is uniform; it does not imply a positive unrestricted average-risk lower bound.}
\begin{proof}
If $k\mid\ell$, write $\ell=km$ and define
\[
A(\theta)=r^{1/k}e^{\mathrm i m\theta}.
\]
This is a continuous loop and satisfies $A(\theta)^k=re^{\mathrm i\ell\theta}$, so the uniform defect is zero.

Assume now that $k\nmid\ell$.  Suppose, toward a contradiction, that a continuous field $A:S^1\to\C$ satisfies
\[
|A(\theta)^k-re^{\mathrm i\ell\theta}|<r
\qquad\text{for every }\theta.
\]
The inequality implies $A(\theta)^k\ne0$.  Consider the straight-line homotopy
\[
H_s(\theta)=(1-s)re^{\mathrm i\ell\theta}+sA(\theta)^k,
\qquad 0\le s\le1.
\]
For every $(s,\theta)$,
\[
|H_s(\theta)-re^{\mathrm i\ell\theta}|
=s|A(\theta)^k-re^{\mathrm i\ell\theta}|<r,
\]
so $H_s(\theta)$ lies in the open disk of radius $r$ centered at $re^{\mathrm i\ell\theta}$; that disk does not contain the origin.  Hence the loops $A^k$ and $re^{\mathrm i\ell\theta}$ are homotopic in $\C^\star$ and have the same winding number.  But
\[
\operatorname{wind}(A^k)=k\operatorname{wind}(A)
\quad\text{and}\quad
\operatorname{wind}(re^{\mathrm i\ell\theta})=\ell,
\]
which would imply $k\mid\ell$, a contradiction.  Therefore every continuous $A$ has some $\theta$ with residual magnitude at least $r$, and hence uniform squared defect at least $r^2$.

The constant field $A\equiv0$ has squared defect exactly
\[
|0-re^{\mathrm i\ell\theta}|^2=r^2
\]
for all $\theta$.  Thus the lower bound is attained.
\end{proof}

%% file: chapters/ch15_repairs.tex
\chapter{Atlases, Quotients, and Randomized Repair}
\label{ch:repairs}
\section{From obstruction diagnosis to a repair contract}
\index{repair!mechanism-matched}

A positive obstruction does not imply that the original task is impossible.  It means that the declared architecture contract is incompatible with the oracle geometry.  A repair changes the contract in a controlled way.

The principal repair families are:
\begin{enumerate}
  \item local atlases;
  \item quotient-valued outputs;
  \item set-valued outputs;
  \item probability-valued outputs or randomized point rules;
  \item memory refinement or rectangularization;
  \item nonlinear semantic projection;
  \item test-time local refinement.
\end{enumerate}

The correct repair is the smallest contract change that removes the certified obstruction while preserving downstream information.  This criterion rules out a flat menu of interchangeable fixes.  Some
obstructions arise because one representative cannot be chosen globally;
others arise from forced sharing, an invalid semantic interface, or an
insufficient deployment computation.  The remainder of the chapter follows
these mechanisms and then returns to the statistical problem of selecting
among their repairs.

\section{Local and symmetry-respecting repair}
\sectionmark{Local and symmetry-respecting repair}
\index{atlas!repair}
\index{repair!atlas}

Let $\{U_j\}_{j=1}^K$ cover $\X$.  On each chart $U_j$, suppose a continuous
oracle section $a_j^\star$ exists.  A hard atlas architecture consists of a
gate $g(x)$ and local experts $A_j(x)$.  The minimum number of local sections
needed to cover a fibration is closely related to sectional category and
Schwarz genus \citep{Schwarz1961}; the theorem below uses only the existence
of a declared cover and does not claim a minimal chart number.
Multi-chart representation learning and chart-based autoencoders provide a
separate machine-learning lineage \citep{SchonsheckChenLai2019,
KalatzisEtAl2021,StolbergLarsenSommer2021}.  In this chapter the equivalence
between a cover by at most $K$ local sections and a $K$-chart atlas is a
definitional inversion.  The theorem below is only the witness lemma that such
declared sections saturate the atlas class; it does not claim to invent
multi-chart models or to learn the cover \citep{SingularEG}.

\begin{conditionbox}
The input space is covered by declared charts, each chart admits a legitimate local zero-defect section, and the local expert class contains that section.  A measurable selector chooses one chart at deployment time.  The theorem is existential and does not assert that the cover, selector, or local experts can be estimated from finite data.
\end{conditionbox}

\begin{theorem}[Atlas saturation]
\label{thm:atlas-saturation}
Suppose the declared charts form a measurable cover, the hard-atlas grammar
contains a measurable gate $g$ satisfying $x\in U_{g(x)}$, and every local
expert class contains an exact oracle section on its chart.  Then the hard
atlas class has zero uniform obstruction:
\[
  \inf_{(g,A_1,\ldots,A_K)}
  \sup_x\Def_x\{A_{g(x)}(x)\}=0.
\]
\end{theorem}

\begin{interpretationbox}
A positive obstruction of a single global point-valued architecture can disappear after changing the architecture contract to an atlas.  The theorem explains why targeted multi-chart repair can be qualitatively different from merely widening one global network.  It does not establish minimal chart number or gating generalization.
\end{interpretationbox}

\begin{proofroadmap}
Choose on every chart its exact local oracle section and deploy the section selected by the measurable gate.  At each input the selected local section has zero native defect.  The chapter appendix states the short verification and its precise measurability boundary.
\end{proofroadmap}

The theorem is existential: it neither learns the cover nor proves that the
chosen chart count is minimal.

An atlas retains point-valued representatives but localizes where they must be
chosen.  When the labels themselves carry no downstream meaning, the more
economical repair is instead to change the output object.

\index{repair!quotient}
\index{representation!quotient-valued}

Suppose a symmetry group $G$ acts on the auxiliary fiber and the defect is invariant:
\[
  \Def_x(ga)=\Def_x(a).
\]
If the downstream task depends only on the orbit $[a]\in\A/G$, then the quotient is the natural output.  A global quotient section may exist even when no global representative section does.

Examples include:
\begin{itemize}
  \item eigenprojectors instead of oriented eigenvectors;
  \item unordered mixture components instead of labeled lists;
  \item subspaces instead of bases;
  \item equivalence classes of gauges instead of coordinates.
\end{itemize}

\index{repair!set-valued}
\index{repair!probability-valued}

For the radical family, the full root set is globally defined.  A set-valued output can therefore achieve zero defect.  Likewise, the uniform distribution over roots is invariant under monodromy.

\paragraph{Probability-valued versus randomized repair.}
\index{repair!randomized}
A probability-valued output returns a law $K_x$ and therefore requires a
task and loss defined on laws.  A randomized point-valued deployment instead
draws $A\sim K_x$ and, under expected native loss, pays
\[
  \int \Def_x(a)\,K_x(\dd a).
\]
The two contracts remain distinct even when they use the same kernel.  Neither
is the deterministic ambient average $\int a\,K_x(\dd a)$, when that average
is defined; the average may leave the oracle set.

These repairs change downstream semantics.  A task that needs one labeled root may still require a chart or additional side information.  A symmetric task can operate directly on the set or law.

\index{repair!unsafe averaging}

The quotient, set-valued, and probability-valued constructions also explain
why a superficially simpler aggregation can be unsafe.  A tempting response
to multiple local branches is to average representatives.  In the square-root
example, averaging $a$ and $-a$ gives zero, which lies on the discriminant and
pays the full catastrophe tax.  Ambient averaging can therefore destroy
oracle information.

Safe aggregation must respect the quotient or use a native barycenter under
the declared defect.  Thus atlas and quotient repair solve different problems: the former permits
several coherent local representatives, whereas the latter removes distinctions
that the task is not entitled to observe.  Neither authorizes an arbitrary
average in the ambient representation space.

\section{Repairing deployment mechanisms}
\index{repair!memory refinement}
\index{rectangularity!repair}

In sequential systems, the repair may be to enlarge the memory state so that histories requiring different local kernels no longer collide.  Alternatively, one can relax shared parameters and allow independent local kernels, moving toward the rectangular hull.

The benefit is measured by the coordination tax removed.  The cost is increased state, memory, communication, or generalization complexity.

Forced sharing is not the only mechanism-level failure.  A legal linear
calculation may also leave the semantic output set, in which case the repair
belongs at the validity interface rather than in the memory state.

\index{repair!semantic projection}

The graph-CDF example uses a different repair.  The raw high-order field is projected onto the valid semantic set.  This changes neither the output type nor the local oracle family, but it replaces an invalid linear mechanism by a nonlinear interface.

The projection is justified only because it is nonexpansive in the native risk geometry.  An arbitrary clipping or renormalization step requires its own theorem.

\index{repair!test-time refinement}

Deployment computation provides a third mechanism-level intervention.  A
semi-amortized system starts from a shared prediction and performs local
optimization at test time.  This can reduce implementation error and, if the
allowed iteration budget is part of the architecture, enlarge the effective
class.
Laboratory~\ref{sec:lab-mnist-amortization} audits this distinction on MNIST
by declaring the zero-step and refined deployment classes before evaluating
the held-out native objective.

Test-time refinement is not equivalent to atlas repair.  It may still follow the wrong labeled branch or remain trapped by a representation obstruction.  The repair must be matched to the certificate.

Memory refinement, semantic projection, and test-time optimization therefore
alter different clauses of the contract.  Their costs cannot be compared by
parameter count alone; each must be charged in the resource and loss scale
declared for deployment.

\section{Selection, validation, and repair frontiers}
\index{repair!selection}

Suppose a finite library $\{\Arch^{(1)},\ldots,\Arch^{(M)}\}$ contains candidate repairs.  Calibration data produce lower and upper obstruction certificates; tuning data train the candidates; test data compare their operational performance.  This separation avoids selecting and certifying the same architecture on the same random fluctuations.

\begin{principlebox}
A repair is justified by a chain of evidence: positive baseline obstruction, mechanism-specific diagnosis, zero or smaller repaired obstruction, and independent held-out improvement.
\end{principlebox}

\index{atlas!persistent frontier}

For atlas repairs, this decision can be recorded as a frontier rather than a
single winning chart count.  For tolerance $\tau$, define the minimum number
of charts required to attain uniform defect at most $\tau$.  The resulting
integer-valued atlas curve can jump.  Its inverse---the best loss attainable
with $K$ charts---is often more stable under perturbation and is the preferred
statistical object.

The frontier keeps the structural and statistical questions separate.  It
records what each chart budget can realize, while sample splitting determines
which point on that frontier is supported by the available evidence.

\section*{Exercises}

\begin{exercise}
For the square-root family, compare a two-chart repair, an unordered-pair output, and a uniform root-valued law.  Which downstream tasks distinguish them?
\end{exercise}

\begin{exercise}
Give a symmetric-orbit example in which ambient averaging leaves the oracle set and incurs positive native defect.
\end{exercise}

\begin{exercise}
Design a sample-split protocol for choosing among three architectures: single-chart, two-chart, and quotient-valued.
\end{exercise}

\input{chapter_appendices/ch15_proofs}

%% file: chapter_appendices/ch15_proofs.tex
\chapterproofappendix

\subsection*{Proof of atlas saturation}
\proofdependency{The cover must be genuine, the gate must route only to charts containing the input, and each local expert class must contain an exact local oracle section.  No learning or minimal-chart claim is involved.}
\begin{proof}
For each chart $U_j$, let $a_j^\star:U_j\to\A$ be an exact local oracle section contained in the declared expert class, so that
\[
\Def_x\{a_j^\star(x)\}=0
\qquad(x\in U_j).
\]
Because the charts cover $\X$, choose a measurable selector $g(x)$ with $x\in U_{g(x)}$; for a finite open cover on a standard Borel space, one may take the smallest chart index containing $x$.  Set $A_j=a_j^\star$.  Then for every $x$,
\[
\Def_x\{A_{g(x)}(x)\}=0.
\]
Therefore the supremum over $x$ is zero.  Since defects are nonnegative, no architecture can attain a negative value, and the infimum over the atlas class is exactly zero.
\end{proof}

%% file: part_notes/part04_history.tex
\parthistoricalnotes{}
\index{obstruction transfer!historical comparison}
\index{rectangularity!historical comparison}
\index{monodromy!historical comparison}

\subsection*{Three obstruction mechanisms, three literatures}

Regular transport bounds combine Lipschitz approximation, couplings, and
curvature exchange.  Rectangular Bellman recursion belongs to robust control
and recursive multiple-priors theory
\citep{EpsteinSchneider2003,Iyengar2005,NilimElGhaoui2005}.  Loop lifting,
winding, degree, discriminants, and Schwarz genus are classical topology and
singularity theory \citep{Hatcher2002,Milnor1997,Schwarz1961}.  Part IV does
not relabel these ingredients as discoveries of elimination geometry.

Its organizing distinction is the deployment quantifier.  A regular
architecture may be unable to track an oracle that varies too quickly; a
shared conditional architecture may be unable to paste locally available
kernels; and a continuous point-valued output may be unable to choose one
branch around a singular loop.  These failures demand different witnesses
and different repairs.

\begin{center}
\small
\begin{tabularx}{0.97\textwidth}{@{}p{0.18\textwidth}p{0.25\textwidth}p{0.25\textwidth}X@{}}
\toprule
Regime & Classical engine & Book/program quantity & Matched repair \\
\midrule
Regular variation & Lipschitz comparison and transport & native oracle-
variation lower certificate & more regularity budget, local refinement, or
additional charts \\
Coordination & rectangular Bellman recursion and KL chain rule & exact cost
of cross-history sharing in forward-KL units & memory refinement,
rectangularization, or a different sharing rule \\
Singular topology & winding, monodromy, degree, genus & uniform
objective-denominated catastrophe tax & atlas, quotient, set-valued, or
randomized output \\
\bottomrule
\end{tabularx}
\end{center}

The rectangular residual decomposition is a source-program architecture
specialization of classical rectangularity.  Atlas saturation is a direct
witness lemma.  The exact radical value $r^2$ is a useful quantitative
winding corollary in native-objective units; it is not presented as a new
topological mechanism.

%% file: chapters/ch16_resource_rd.tex
\chapter{Resource-Constrained Architecture Rate--Distortion}
\label{ch:resource-rd}
\section{The architecture as carrier and decoder}
\index{carrier!deployment}
\index{decoder}
\index{resource constraints!carrier--decoder factorization}

Here \emph{carrier} means the deployment information state passed to an
action decoder.  It is not the reduced slack carrier of
Section~\ref{sec:reduced-slack-carrier}, which removes representation gauge
before deployment is considered.  Section~\ref{sec:lift-target-visible}
supplies the conditional extraction gate from a quotient-reduced lift to a
deployment carrier; without such a gate the two remain distinct.

An architecture first produces a carrier $Z$ and then decodes it:
\[
  X\xrightarrow{\phi}Z\xrightarrow{g}A.
\]
The terminology deliberately recalls rate--distortion theory, where a source
is compressed subject to a fidelity criterion \citep{Shannon1959,CoverThomas2006}.
Here the distortion is not chosen independently of the learning problem: it is
the native objective defect, and the resource grammar may constrain both the
carrier and its decoder.
Across inputs, the oracle map $x\mapsto a^\star(x)$ determines which
distinctions a deployment carrier must preserve.  Within each fiber, the
native defect determines the price of merging those distinctions.  The
carrier need not retain all information in $x$; it need only be sufficient
for the oracle signature.  Conversely, width, state count, or cone size is
not effective capacity unless the retained state preserves those signatures
and makes them decodable.

A resource grammar indexed by $(m,r)$ restricts carrier size, number of labels, communication, coherence, smoothness, or decoder complexity.  Write $\mathfrak G_{m,r}$ for its declared class of admissible carrier--decoder pairs $(Z,g)$, and put
\[
  \mathcal Z_{m,r}=\{Z:\text{there exists }g\text{ with }(Z,g)\in\mathfrak G_{m,r}\}.
\]

The architecture can fail in two ways.
\begin{enumerate}
  \item The carrier erases distinctions among oracle signatures.
  \item The carrier retains the information, but the decoder class cannot realize the optimal readout.
\end{enumerate}

\section{Canonical Bregman carrier}
\index{Bregman geometry!canonical carrier}
\index{carrier!information loss}
\index{decoder!nonsaturation}

Reuse the dual oracle signature $S=\nabla\Phi\{a^\star(X)\}$ from
Section~\ref{sec:lift-target-visible}.  For a carrier $Z$, put
\[
  S_Z=\E(S\mid Z).
\]
The optimal unconstrained decoder is
\[
  a_Z^\dagger
  =
  \nabla\Phi^\star(S_Z).
\]
This follows from conditional Bregman projection; conditional expectation as
the optimal Bregman predictor and the associated centroid identity are
classical \citep{BanerjeeGuoWang2005}.  The resource split
below specializes that identity to a carrier--decoder grammar
\citep{ArchitectureRD}.

The carrier information loss is
\[
  \mathcal C_\Phi(Z)
  =
  \E D_\Phi\{a_Z^\dagger(Z)\|a^\star(X)\}.
\]
For a carrier admitted by the grammar, define decoder nonsaturation by
\[
  \Psi_{m,r}(Z)
  =
  \inf_{g:(Z,g)\in\mathfrak G_{m,r}}
  \E D_\Phi\{g(Z)\|a_Z^\dagger(Z)\}.
\]
The infimum is $+\infty$ if the carrier has no admissible decoder.  The
native architecture distortion is the best original objective defect among
admissible pairs,
\[
  \mathcal D_{\rm nat}(m,r)
  =
  \inf_{(Z,g)\in\mathfrak G_{m,r}}
  \E D_\Phi\{g(Z)\|a^\star(X)\}.
\]
The divergence orientation is chosen to match the exact Bregman Pythagorean identity.

\begin{conditionbox}
The native defect is generated by a Legendre potential on a common fiber, the oracle is unique, the dual signature is integrable, and conditional dual means are well defined.  The potential and pairing terms used in the conditional calculation are integrable for every finite-cost admissible decoder.  Resource constraints are encoded in the declared carrier and decoder classes.  The decomposition is an identity between infima and does not require attainment; a statement about one exact zero-distortion carrier--decoder pair additionally requires that such a pair attain the infimum.
\end{conditionbox}

\begin{theorem}[Canonical architecture rate--distortion decomposition]
\label{thm:architecture-rd}
Under the common-fiber Legendre assumptions,
\[
  \mathcal D_{\rm nat}(m,r)
  =
  \inf_{Z\in\mathcal Z_{m,r}}
  \bigl\{
  \mathcal C_\Phi(Z)+\Psi_{m,r}(Z)
  \bigr\}.
\]
\end{theorem}

\begin{interpretationbox}
The architecture distortion separates exactly into information lost by the carrier and nonsaturation of the resource-limited decoder.  When the infimum is attained, zero total distortion requires both task-relative sufficiency and a decoder capable of realizing the canonical conditional oracle.  A raw bit, dimension, or label count becomes meaningful only after an extraction theorem links it to these classes.
\end{interpretationbox}

\begin{proofroadmap}
Condition the Bregman defect on the carrier, insert the conditional dual-mean decoder, and use the Bregman Pythagorean identity.  Minimize the second term over decoders paired with that carrier and then over admissible carriers.  The chapter appendix proves the decomposition and its zero-distortion criterion.
\end{proofroadmap}

The formula is a second elimination over carriers.  Width or parameter count is
only one possible resource coordinate; the canonical retained state is the
conditional dual oracle signature.  The resource grammar is upstream of the
P/G/X/V/C calculus in
Chapter~\ref{ch:towers-pgxvc}: it declares the feasible architecture within
which an operation is performed.  Changing that grammar is not a sixth mode.

\section{Discrete and continuous carrier extraction}
\paragraph{Hard carrier extraction.}
\index{carrier!hard}

Suppose $Z$ takes at most $m$ labels and a weighted Potts coherence budget
limits label changes on a graph.  The inverse minimum $q$-cut profile gives
the exact number of usable labels under the budget.  On paths and trees this
capacity has a closed form; related cardinality-constrained and multiway-cut
algorithms are classical
\citep{GoldschmidtHochbaum1994,FredericksonSolisOba1998,EngelbergEtAl2007}.

Precisely, for a weighted graph $G=(V,E,w)$ let $c(G\setminus F)$ denote the
number of connected components after deleting $F\subseteq E$, and define
\[
  \lambda_q(G)
  =
  \min\left\{\sum_{e\in F}w_e:
    F\subseteq E,\ c(G\setminus F)\ge q\right\},
  \qquad \lambda_1(G)=0.
\]
The hard-label capacity under label budget $m$ and cut budget $r$ is
\[
  \kappa_G^{\rm hard}(m,r)
  =
  \min\left\{m,
    \max\{q:\lambda_q(G)\le r\}\right\}.
\]
It is exact for the declared Potts grammar: every feasible labeling is
constant on the components left by its cut set, and conversely those
components may be assigned distinct labels up to the cap $m$.

If oracle signatures are well separated by type, the visible cardinality bound is attained.  Strong convexity then converts limited label capacity into a quantization floor.

\paragraph{Continuous carriers.}
\index{carrier!continuous}
\index{metric entropy}

For a normalized continuous conic carrier with a Lipschitz readout and
quadratic graph coherence, metric entropy controls the number of
distinguishable oracle signatures at resolution $\epsilon$.  The matching
lower exponent requires a co-Lipschitz readout on an interior region and a
resource grammar admitting arbitrarily dense deployments in that region;
the finite-site capacity cap still applies.

This produces a resource-to-information map of the form
\[
  \log N(\epsilon)
  \lesssim
  d\log\left(1+\frac{C(m,r)}{\epsilon}\right).
\]
Strong defect growth converts a covering-radius lower bound into native distortion.

\section{Finite frontiers and phase diagrams}
\paragraph{The finite Bregman quantization frontier.}
\index{Bregman geometry!quantization}
\index{resource constraints!quantization frontier}

For a finite oracle set with weights $p_i$ and codebook size $q$, the optimal Bregman quantization error is
\[
  \min_{c_1,\ldots,c_q}
  \sum_i p_i\min_j D_\Phi(c_j\|a_i^\star).
\]
Within each assigned cell the optimal codeword is the dual-coordinate
centroid.  Separated signatures give a positive lower bound whenever $q$ is
smaller than the number of distinguishable types.  This is the usual vector
quantization problem in the native Bregman geometry
\citep{GrayNeuhoff1998,GrafLuschgy2000,FriedrichEtAl2008}.

\paragraph{Resource phase diagrams.}
\index{resource constraints!phase diagram}

On an $n$-vertex path whose edges all have the common weight $w>0$, with hard
routing and quadratic local loss, the effective carrier size is
\[
  q_{\rm eff}
  =
  \min\{m,n,1+\lfloor r/w\rfloor\}.
\]
The native distortion reduces to balanced scalar quantization.  The operational audit distance becomes a clipped discrete covering radius.  This yields an exact capacity--resolution phase diagram.

\section{An exact data-selection frontier}
\label{sec:exact-data-selection-frontier}
\index{data selection}
\index{resource constraints!data-selection budget}

The preceding examples constrain a carrier chosen from a fixed codebook.  A
different finite resource contract arises when the carrier must itself be a
small submultiset of the observed data and the decoder is fixed in advance.
This setting yields an exact distortion curve rather than only an upper or
lower rate.

Let $D=\{z_1,\ldots,z_N\}\subset\R^d$ be a finite multiset, and write
\[
  L_D(h)=\frac1N\sum_{i=1}^N\|h-z_i\|^2,
  \qquad
  \mu_D=\frac1N\sum_{i=1}^Nz_i,
  \qquad
  V_D=L_D(\mu_D).
\]
For an integer $n\ge1$, an $n$-point carrier stores indices
$i_1,\ldots,i_n$, with repetition
allowed, and the decoder returns
$\widehat\mu=n^{-1}\sum_{r=1}^nz_{i_r}$.  For $V_D>0$, define its normalized
worst-case native distortion by
\[
  \mathcal D_{\rm sel}(d,n)
  =
  \sup_{D\subset\R^d}
  \min_{i_1,\ldots,i_n}
  \frac{\|\widehat\mu-\mu_D\|^2}{V_D}.
\]
The bias--variance identity is the first certified elimination:
\[
  L_D(h)=V_D+\|h-\mu_D\|^2.
\]
Thus the usual worst-case loss ratio is exactly
$F(d,n)=1+\mathcal D_{\rm sel}(d,n)$; selection budget becomes a resource
coordinate measured directly in excess squared loss.

\begin{sourcebox}[title={Companion result; proof not reproduced here}]
The exact data-selection manuscript proves
\[
  \mathcal D_{\rm sel}(d,n)=\frac1{2n-1}
  \quad(1\le d\le3,\ n\ge1),
  \qquad
  \mathcal D_{\rm sel}(d,2)
  =\max\!\left\{\frac13,\frac{d-1}{2d}\right\}
  \quad(d\ge1).
\]
The first formula resolves the low-dimensional mean-estimation row of the
published data-selection problem; the second records its budget-two
dimension transition.  These statements are not counted among the book's
principal results \citep{HannekeEtAl2025DataSelection,Huang2026ExactDataSelection}.
\end{sourcebox}

The proof exposes why this is more than a cardinality bound.  A
variance-nonincreasing Carath\'eodory reduction first preserves the mean while
leaving at most $d+1$ support points.  For a centered reduced law
$\sum_i p_ix_i=0$ with variance $V=\sum_i p_i\|x_i\|^2$, suppose the
residual-capacity inequalities admit integer anchor counts $s_i$ with
$\sum_i s_i=n-1$ and
$(n-1)s_i\le n^2p_i$.  Put $u=\sum_i s_ix_i$,
$S=\sum_i s_i\|x_i\|^2$, and
\[
  q_i=\frac{n^2p_i-(n-1)s_i}{2n-1},
  \qquad
  \Gamma(s)=(n-1)S-\|u\|^2\ge0.
\]
For $J\sim q$, the remaining randomized point satisfies the exact ledger
\[
  \E\left\|\frac{u+x_J}{n}\right\|^2
  =
  \frac{V}{2n-1}
  -\frac{\Gamma(s)}{n^2(2n-1)}.
\]
Taking the best point in the finite support of $q$ eliminates the auxiliary
randomness and returns a legal deterministic selector.

Capacity failure requires a mechanism-specific repair.  At $n=3$, the
remaining four-atom region is closed by a finite correlated-count certificate.
For $n\ge4$, failure forces a heavy pair; merging those two atoms releases
variance
$\lambda\|x_1-x_2\|^2$ with
$\lambda\ge(2n-1)/(4n^2)$.  If $V'$ is the merged variance, randomized
nearest-integer splitting back to legal data points obeys
\[
  \frac{V'}{2n-1}
  +\frac{\|x_1-x_2\|^2}{4n^2}
  \le \frac{V}{2n-1}.
\]
The coarse-graining credit therefore pays the entire fiber-realization cost.
The all-dimensional budget-two column closes differently: a sharp pair
certificate after sparse reduction meets the centered regular-simplex lower
bound at the displayed dimension transition.
The theorem is logically self-contained rather than a corollary of the
general calculus; what transfers is the discipline of exposing a native
defect, eliminating only recoverable auxiliary objects, and closing every
pushforward--recovery step in one currency.

\section{Resource limits and operational boundaries}
\paragraph{Resource limits and foundation models.}
\index{foundation models!resource limits}
\index{resource constraints!foundation models}

The theory suggests several resource coordinates for large AI systems:
\begin{itemize}
  \item context length and memory state;
  \item number of experts or active routes;
  \item communication bandwidth among modules;
  \item test-time iteration budget;
  \item output vocabulary or structured state dimension;
  \item coherence imposed by parameter tying.
\end{itemize}

The relevant question is not whether a model has many parameters in total, but how much oracle-relevant distinction survives the declared carrier and can be extracted by the decoder.

\paragraph{What rate--distortion does not settle.}
\index{resource constraints!rate--distortion}
\index{claim boundary!rate--distortion}

A positive native distortion need not affect a downstream task.  The task may be invariant to the lost distinctions.  Operational visibility requires the transmission and exposure gates of Chapter~\ref{ch:operational-semantics}.

\section*{Exercises}

\begin{exercise}
For squared Euclidean defect, show that the canonical decoder from a carrier $Z$ is the conditional mean of the oracle.
\end{exercise}

\begin{exercise}
Compute the optimal two-codeword quantization of three equally weighted scalar oracle values $0,1,3$.
\end{exercise}

\begin{exercise}
Propose a resource grammar for a transformer with bounded context and a fixed number of active experts.  Which quantity might serve as the oracle signature?
\end{exercise}

\input{chapter_appendices/ch16_proofs}

%% file: chapter_appendices/ch16_proofs.tex
\chapterproofappendix

\subsection*{Proof of the canonical architecture rate--distortion decomposition}
\proofdependency{The oracle is unique and lies in a common open convex fiber; $\Phi$ is Legendre; the dual signature $S=\nabla\Phi(a^\star)$ is integrable; conditional means lie in the relative interior of $\operatorname{dom}\Phi^\star$; and the potential and pairing terms in the finite-cost conditional calculation are integrable.  These conditions make the canonical conditional decoder and every displayed conditional expectation well defined.}
\begin{proof}
Fix an admissible pair $(Z,g)\in\mathfrak G_{m,r}$ and write
$A=g(Z)$ and $a^\star=a^\star(X)$.  Put
\[
S=\nabla\Phi(a^\star),
\qquad
S_Z=\E(S\mid Z),
\qquad
 a_Z^\dagger=\nabla\Phi^\star(S_Z).
\]
Rather than subtracting two expected divergences, which could conceal an
$\infty-\infty$ expression, start from the finite algebraic difference of
their Bregman expansions:
\[
  \Phi(A)-\Phi(a_Z^\dagger)
  -\langle S,A-a_Z^\dagger\rangle.
\]
Conditioning this expression on $Z$ and using
$\E(S\mid Z)=S_Z=\nabla\Phi(a_Z^\dagger)$ gives
\[
  \Phi(A)-\Phi(a_Z^\dagger)
  -\langle S_Z,A-a_Z^\dagger\rangle
  =D_\Phi(A\|a_Z^\dagger).
\]
Combining this conditional identity with the two original Bregman
expansions and then integrating yields
\[
\E D_\Phi(A\|a^\star)
=
\E D_\Phi(a_Z^\dagger\|a^\star)
+
\E D_\Phi(A\|a_Z^\dagger).
\]
The first term is the carrier information loss $\mathcal C_\Phi(Z)$ and the
infimum of the second term over decoders paired with $Z$ in
$\mathfrak G_{m,r}$ is $\Psi_{m,r}(Z)$.  Taking first the decoder infimum
and then the carrier infimum yields
\[
\mathcal D_{\rm nat}(m,r)
=
\inf_{Z\in\mathcal Z_{m,r}}
\{\mathcal C_\Phi(Z)+\Psi_{m,r}(Z)\}.
\]

Both summands are nonnegative.  If an admissible pair $(Z,g)$ attains zero total defect, then
\[
\mathcal C_\Phi(Z)=0,
\qquad
\E D_\Phi\{g(Z)\|a_Z^\dagger\}=0.
\]
Strict convexity implies $g(Z)=a_Z^\dagger$ almost surely.  Legendre duality gives
\[
\mathcal C_\Phi(Z)
=
\E D_{\Phi^\star}(S\|S_Z),
\]
so the first equality holds exactly when $S=S_Z$ almost surely.  Conversely, an admissible pair with these two properties has zero native defect.  Attainment of the pair, rather than only the outer carrier infimum, is needed for this zero characterization.
\end{proof}

%% file: chapters/ch17_operational_semantics.tex
\chapter{Operational Semantics and Contextual Observability}
\label{ch:operational-semantics}
\section{From internal difference to task-observable kernels}
\index{operational semantics}
\index{operational visibility}

Two pipelines can have different internal kernels while producing the same value for every downstream task that is legally allowed to use them.  Conversely, two modules that agree at one base input can become distinguishable after insertion into a larger context.

\index{min-plus kernel}

Operational semantics asks when one pipeline can safely replace another under
a declared task and context grammar.  To make this question quantitative,
represent a finite elimination pipeline by the cost it exposes between its
boundary states.  Such a pipeline is represented by a min-plus kernel
$K(y,x)$: the least internal cost of transforming input state $x$ to output
state $y$.  Given upstream potential $f$ and downstream terminal cost $g$, the
optimized value is
\[
  \mathsf V_K(f,g)
  =
  \inf_{x,y}\{f(x)+K(y,x)+g(y)\}.
\]
A task contract $\Task$ is a declared collection of admissible pairs $(f,g)$.

This representation separates internal implementation from the values that a
legal observer can extract.  The next construction makes that separation
canonical for a fixed task contract.

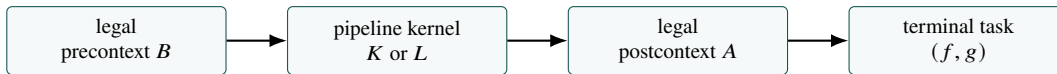
\begin{figure}[htbp]
\centering
\begin{tikzpicture}[>=Latex,node distance=8mm]
  \tikzset{obox/.style={draw=egteal!75!black,fill=egteal!4,rounded corners=2pt,
    minimum width=29mm,minimum height=9mm,align=center,font=\scriptsize}}
  \node[obox] (pre) {legal\\precontext $B$};
  \node[obox,right=of pre] (ker) {pipeline kernel\\$K$ or $L$};
  \node[obox,right=of ker] (post) {legal\\postcontext $A$};
  \node[obox,right=of post] (task) {terminal task\\$(f,g)$};
  \draw[->,thick] (pre)--(ker);
  \draw[->,thick] (ker)--(post);
  \draw[->,thick] (post)--(task);
\end{tikzpicture}
\caption{Contextual observability.  Two internal kernels are operationally equivalent only if no legal precontext, postcontext, and terminal task can distinguish them.}
\label{fig:contextual-observability}
\end{figure}

\section{Task envelopes and contextual completion}
\index{task envelope}
\index{task contract}

Define the task envelope
\[
  (\mathsf E_\Task K)(y,x)
  =
  \sup_{(f,g)\in\Task}
  \{\mathsf V_K(f,g)-f(x)-g(y)\}.
\]
The envelope is the pointwise least kernel that preserves all declared task values.
The construction is related in spirit to classical residuation and convex
conjugacy, while Attouch--Wets metrics concern convergence of epigraphs on
unbounded spaces rather than the finite sup-norm identity stated here
\citep{AttouchWets1986}.  Proposition~\ref{prop:task-envelope} is therefore
proved directly from this task envelope and should not be read as an
Attouch--Wets theorem \citep{OperationalEG}.
For two kernels of the same type, define the base task observational distance by
\[
  d_\Task(K,L)
  =
  \sup_{(f,g)\in\Task}
  \bigl|\mathsf V_K(f,g)-\mathsf V_L(f,g)\bigr|.
\]

\begin{conditionbox}
The state spaces and kernels are finite, and the set of allowed tasks is nonempty and declared in advance.  All kernels compared have the same input and output types, so the same task pairs apply.  No context closure is needed for this result.  The operational metric compares only differences visible to that task contract.
\end{conditionbox}

\begin{proposition}[Task-envelope isometry]
\label{prop:task-envelope}
The envelope preserves exactly the values $\mathsf V_K(f,g)$ for $(f,g)\in\Task$, and the task observational distance between $K$ and $L$ equals
\[
  \|\mathsf E_\Task K-\mathsf E_\Task L\|_\infty.
\]
\end{proposition}

\begin{interpretationbox}
The task envelope is the pointwise least kernel preserving all allowed task values, and sup-norm distance between envelopes equals the worst task-value discrepancy.  Internal kernel differences outside the envelope are operationally invisible.  Changing the task set changes the quotient.
\end{interpretationbox}

\begin{proofroadmap}
Use the variational definition of every task value to show the envelope lies below the original kernel and preserves all values.  Minimality follows by comparison with any other value-preserving kernel; the signed isometry follows by exchanging supremum and pointwise difference.  Complete details are in the chapter appendix.
\end{proofroadmap}

Task equivalence at the exposed boundary is not yet a compositional notion.
A module that is invisible in isolation may become visible after legal
preprocessing or postprocessing, so the task contract must be closed under
the deployment grammar.

\index{contextual closure}
\index{full abstraction}

A grammar $\Gram$ specifies legal pre- and post-compositions.  The base task set must be closed under pullback through every legal context.  Let
\[
  \operatorname{Cl}_\Gram\Task
\]
be the least context-stable task system containing $\Task$.  Define
\[
  \mathsf Q_{\Gram,\Task}K
  =
  \mathsf E_{\operatorname{Cl}_\Gram\Task}K.
\]
Equivalently, if $B$ and $A$ range over type-compatible legal pre- and
postcontexts, define
\[
  d_{\Gram,\Task}^{\rm ctx}(K,L)
  =
  \sup_{A,B}
  d_\Task(A\otimes K\otimes B,
          A\otimes L\otimes B).
\]
Identity contexts are included, so this contextual distance dominates the
base task distance.  A pseudometric $\rho$ on type-compatible kernels is
called \emph{task-adequate} when $d_\Task(K,L)\le\rho(K,L)$, and
\emph{grammar-nonexpansive} when
\[
  \rho(A\otimes K\otimes B,A\otimes L\otimes B)
  \le \rho(K,L)
\]
for every legal precontext $B$ and postcontext $A$.

\begin{conditionbox}
The context grammar contains identities, is closed under composition, and acts monotonically on finite kernels.  The closure operator takes the envelope over all legal contexts and terminal tasks.  These grammar assumptions are essential for congruence.
\end{conditionbox}

\begin{theorem}[Contextual completion and full abstraction]
\label{thm:contextual-full-abstraction}
The contextual distance is
\[
  d_{\Gram,\Task}^{\rm ctx}(K,L)
  =
  \|\mathsf Q_{\Gram,\Task}K-\mathsf Q_{\Gram,\Task}L\|_\infty.
\]
It is the least task-adequate grammar-nonexpansive pseudometric.  Its zero set is the greatest grammar congruence contained in base task equivalence, and $\mathsf Q_{\Gram,\Task}K$ is the pointwise least representative of the fully abstract quotient.
\end{theorem}

\begin{interpretationbox}
Two pipelines are equivalent exactly when no legal context followed by a legal task can distinguish them.  The contextual quotient is fully abstract: it identifies all and only operationally indistinguishable kernels.  This is task- and grammar-relative, not an absolute semantic identity.
\end{interpretationbox}

\begin{proofroadmap}
Take the supremum of terminal task envelopes over the context closure.  Show that equal closures imply equal values after every context; conversely use the definition of the closure to recover pointwise equality from universal contextual equivalence.  Closure under composition proves congruence.  The chapter appendix supplies the full argument.
\end{proofroadmap}

\section{From native defect to operational separation}
\index{operational visibility!native-to-operational transfer}

\index{operational visibility!transmission gate}
\index{operational visibility!exposure gate}
The preceding theorem supplies the operational quotient of pipeline space.
Suppose an architecture defect produces a nonnegative increment in a pipeline
kernel.  Two gates are needed.  The first is the \emph{transmission gate}: the native defect must actually
increase the operational kernel rather than being absorbed by another
internal choice.  The second is the \emph{exposure gate}: the closed task
contract must contain a context that selects the affected states.  Only then
does a positive native architecture floor imply a positive contextual
separation.

\begin{boundarybox}
Contextual observability is not an additional scalar tax to be blindly added to native distortion.  It is a gate or transform determining how much native difference is operationally visible.
\end{boundarybox}

\index{metric resolution}

The amount transmitted through these gates depends on the resolution of the
task family.  For Lipschitz task contracts, the envelope becomes a double
metric erosion.  If a defect spike is narrower than the task resolution, it
may be clipped or smoothed away.  The exact resolution profile records the
observable separation as task Lipschitz budgets vary.

Finite task families yield a visibility transform with an exact margin-clipped lower bound.  Compact metric task classes yield a resolution coefficient determined by task and kernel Lipschitz moduli.

Thus the operational image of a native defect is generally a profile rather
than a binary label.  At coarse task resolution a real internal seam can be
invisible; as the contract becomes more discriminating, the same seam can
produce a positive contextual distance.

\section{Ordering curvature and contextual witnesses}
\index{curvature!ordering}

For $n$ operations, each permutation produces a kernel $K_\pi$.  The contextual ordering curvature is the diameter
\[
  \max_{\pi,\sigma}
  d_{\Gram,\Task}^{\rm ctx}(K_\pi,K_\sigma).
\]
Pairwise swap budgets can bound the global curvature through an
inversion-weighted Kendall sum \citep{Kendall1938}.  Pairwise commutation at
one input does not imply contextual interchange coherence.

\index{memory!finite-memory screening}

The need for context is already visible in a minimal finite-memory witness.  A
two-state screening pipeline can be constructed so that every operation order
is invisible under a zero-input contract, yet a legal precontext creates a
value gap of one.  The full resolution profile is
\[
  \max\{0,\min(1,a-1,b)\}
\]
for task budgets $(a,b)$.  The example demonstrates why one fixed audit input is insufficient for safe module substitution.

\index{software semantics}

This conclusion parallels contextual equivalence and full abstraction in
programming-language semantics \citep{Milner1977,Plotkin1977}.  Here the
boundary objects are optimization kernels and the task contract is
quantitative, so the resulting quotient is both semantic and metric.

\section*{Exercises}

\begin{exercise}
For a finite state space and the full task contract containing all bounded potentials, show that the task envelope reconstructs the kernel exactly up to additive normalization allowed by the contract.
\end{exercise}

\begin{exercise}
Give two kernels that agree under a zero-input task but differ after precomposition with a legal context.
\end{exercise}

\begin{exercise}
Give an example in which a native architecture defect is invisible to a downstream task that is permutation invariant.
\end{exercise}

\input{chapter_appendices/ch17_proofs}

%% file: chapter_appendices/ch17_proofs.tex
\chapterproofappendix

\subsection*{Proof of task-envelope reconstruction and isometry}
\proofdependency{The kernel and state sets are finite, so all minima and maxima are attained.  The task contract may be arbitrary; no closure under contexts is needed for this first result.}
\begin{proof}
For every task $(f,g)$ and state pair $(x,y)$,
\[
\mathsf V_K(f,g)
\le f(x)+K(y,x)+g(y),
\]
so
\[
(\mathsf E_\Task K)(y,x)
=\sup_{(f,g)\in\Task}
\{\mathsf V_K(f,g)-f(x)-g(y)\}
\le K(y,x).
\]
Fix $(f_0,g_0)\in\Task$.  Its own term in the supremum gives
\[
f_0(x)+(\mathsf E_\Task K)(y,x)+g_0(y)
\ge \mathsf V_K(f_0,g_0)
\]
for every $(x,y)$.  Taking the minimum over $(x,y)$ gives
\(
\mathsf V_{\mathsf E_\Task K}(f_0,g_0)
\ge\mathsf V_K(f_0,g_0)
\), while $\mathsf E_\Task K\le K$ gives the reverse inequality.  Hence every task value is preserved.

Value preservation implies idempotence of $\mathsf E_\Task$, and monotonicity follows from monotonicity of the minimum.  If $M$ has the same task values as $K$, then
\[
M(y,x)
\ge \mathsf V_M(f,g)-f(x)-g(y)
=\mathsf V_K(f,g)-f(x)-g(y)
\]
for every task.  Taking the supremum proves $\mathsf E_\Task K\le M$, so the envelope is the pointwise least representative.

Write $K_\Task=\mathsf E_\Task K$ and $L_\Task=\mathsf E_\Task L$.  Minimum comparison and value preservation give
\[
\sup_{(f,g)\in\Task}
\{\mathsf V_K(f,g)-\mathsf V_L(f,g)\}
\le
\max_{x,y}\{K_\Task(y,x)-L_\Task(y,x)\}.
\]
Conversely, for fixed $(x,y)$, the difference of the two suprema defining $K_\Task$ and $L_\Task$ is bounded by the supremum of the task-value differences.  Maximizing proves equality of the signed quantities.  Repeating with $K$ and $L$ exchanged yields
\[
d_\Task(K,L)=\|K_\Task-L_\Task\|_\infty.
\]
\end{proof}

\subsection*{Proof of contextual completion and full abstraction}
\proofdependency{The grammar contains identity or formal empty contexts and is closed under typed composition.  Context closure is defined by pulling every base task back through every legal pre- and post-context.}
\begin{proof}
For a precontext $B$, postcontext $A$, and external task $(h,k)$, both
\[
\mathsf V_{A\otimes K\otimes B}(h,k)
\quad\text{and}\quad
\mathsf V_K(f_{B,h},g_{A,k})
\]
are the minimum of
\[
h(w)+B(x,w)+K(y,x)+A(z,y)+k(z)
\]
over $(w,x,y,z)$.  Thus contextual testing is exactly testing $K$ with the pullback-closed task system $\operatorname{Cl}_\Gram\Task$.  Applying the preceding task-envelope isometry gives
\[
d_{\Gram,\Task}^{\rm ctx}(K,L)
=
\|\mathsf Q_{\Gram,\Task}K-
  \mathsf Q_{\Gram,\Task}L\|_\infty.
\]

The supremum of absolute differences of scalar functionals is a pseudometric.  Identity contexts show task adequacy.  If fixed contexts are first placed around $K$ and $L$, any further context combines with them into another legal context; hence the metric is grammar-nonexpansive.

Let $\rho$ be any task-adequate grammar-nonexpansive pseudometric.  For every legal context,
\begin{align*}
d_\Task(A\otimes K\otimes B,A\otimes L\otimes B)
&\le \rho(A\otimes K\otimes B,A\otimes L\otimes B)\\
&\le \rho(K,L).
\end{align*}
Taking the supremum over contexts proves that the contextual metric is the least such pseudometric.

Its zero set is a congruence by grammar nonexpansiveness and lies inside base task equivalence by task adequacy.  Conversely, if $\sim$ is any grammar congruence contained in base task equivalence and $K\sim L$, then all contextual substitutions remain $\sim$-related and hence base-task equivalent.  Every contextual value difference is zero, so $K$ and $L$ have contextual distance zero.  This proves the greatest-sound-congruence property.

Finally, apply task-envelope reconstruction to the closed contract.  The envelope is deflationary, monotone, idempotent, preserves every closed task, and is the unique pointwise-least member of its equivalence class.  Equality of contextual envelopes is equivalent to zero contextual distance, completing the fully abstract quotient statement.
\end{proof}

%% file: chapters/ch18_composition.tex
\chapter{Composition, Base Change, and Dequantization}
\label{ch:composition}

Here \emph{base change} means passage between fine and coarse
representations, memories, or architecture levels.  It preserves obstruction
only under a proved commuting or exact decomposition theorem; equality of one
optimized objective value is not sufficient.
\index{base change}

\section{Conditional chain rules and architecture composition}
\index{composition!theorem requirement}

Candidate-level identities need not survive minimization over a restricted architecture.  A local certificate may also fail after conditioning, pushforward, or a zero-temperature limit.  We therefore distinguish several levels of composition, each of which can fail independently.

\begin{enumerate}
  \item Law-level or candidate-level defect identities.
  \item Architecture-level infima.
  \item Representation or base-change maps.
  \item Operational contextual composition.
  \item Temperature limits.
\end{enumerate}

\paragraph{Conditional KL chain rule.}
\index{KL divergence!conditional chain rule}

Let $Q_{XY}$ and $P_{XY}$ be joint laws.  The chain rule is
\[
  \KL(Q_{XY}\|P_{XY})
  =
  \KL(Q_X\|P_X)
  +
  \E_{Q_X}\KL(Q_{Y\mid X}\|P_{Y\mid X}).
\]
The conditional term is weighted by the trial law $Q_X$.  Replacing this weight by the oracle law without a theorem changes the identity.

The chain rule is the prototype of a vertical fiber decomposition: coarse defect plus expected conditional defect.

Entropy-regularized and KL-control formulations of stochastic control are an
established neighboring literature: path-integral and linearly-solvable
control, maximum-entropy inverse control, control as inference, and soft
actor--critic all exploit closely related exponential tilting or KL-penalized
objectives \citep{Kappen2005,Todorov2006,Todorov2009,ZiebartEtAl2010,
Levine2018,HaarnojaEtAl2018}.  Accordingly, the soft chain identity below is
not presented as a new control identity.  Its role here is to state the exact
normalization and common-loss-scale conditions under which local EG
certificates may be composed.

\paragraph{Soft kernels compose in one objective loss scale.}
\index{composition!soft kernels}
\index{native loss scale!composition}

Let $\mathcal X_0,\ldots,\mathcal X_m$ be standard Borel spaces.  For
$i=1,\ldots,m$, let $M_i(dx_i\mid x_{i-1})$ be a reference transition
kernel, let $K_i(y,x)$ be the cost of a transition from input $x$ to output
$y$, and let $G$ be a
terminal cost.  At inverse temperature $\beta>0$, define backward soft
values by
\begin{align}
  V_m^\beta&=G,\notag\\
  V_{i-1}^\beta(x)
  &=-\frac1\beta\log\int
  \exp\!\left[-\beta\{K_i(y,x)+V_i^\beta(y)\}\right]
  M_i(dy\mid x).
  \label{eq:book-soft-backward-values}
\end{align}
The normalized report
\[
  b_i^\beta(y,x)
  =K_i(y,x)+V_i^\beta(y)-V_{i-1}^\beta(x)
\]
induces the oracle Gibbs transition
\begin{equation}
  P_i^{\beta,\star}(dy\mid x)
  =e^{-\beta b_i^\beta(y,x)}M_i(dy\mid x).
  \label{eq:book-soft-oracle-transition}
\end{equation}

Fix $x_0$ and write
$M_{x_0}=\prod_{i=1}^m M_i(\cdot\mid X_{i-1})$.  A trial path law
$Q\ll M_{x_0}$ may depend on its whole past; denote its conditional laws by
$Q_i(\cdot\mid X_{0:i-1})$.  Define
\begin{align}
  \mathcal H_\beta(Q;x_0)
  &=\E_Q\!\left\{\sum_{i=1}^mK_i(X_i,X_{i-1})+G(X_m)\right\}
    +\frac1\beta\KL(Q\|M_{x_0}),
  \label{eq:book-soft-free-energy}\\
  \delta_i^\beta(Q_i;X_{0:i-1})
  &=\frac1\beta
  \KL\!\left(
  Q_i(\cdot\mid X_{0:i-1})
  \,\middle\|\,
  P_i^{\beta,\star}(\cdot\mid X_{i-1})
  \right).
  \label{eq:book-soft-local-certificate}
\end{align}

\begin{conditionbox}
All soft partition functions are finite and strictly positive.  The trial
path law is absolutely continuous with respect to the reference path law,
regular conditional laws exist, and the displayed costs and relative
entropies are integrable.  The reference kernels and inverse temperature are
fixed throughout the composition.
\end{conditionbox}

\begin{theorem}[Soft conditional chain theorem]
\label{thm:book-soft-conditional-chain}
Let
\[
  P_{x_0}^{\beta,\star}(dx_{1:m})
  =\prod_{i=1}^mP_i^{\beta,\star}(dx_i\mid x_{i-1}).
\]
Then every $Q\ll M_{x_0}$ satisfies
\begin{align}
  \mathcal H_\beta(Q;x_0)-V_0^\beta(x_0)
  &=\frac1\beta\KL(Q\|P_{x_0}^{\beta,\star})
  \label{eq:book-soft-global-kl}\\
  &=\sum_{i=1}^m
  \E_Q\bigl\{\delta_i^\beta(Q_i;X_{0:i-1})\bigr\}.
  \label{eq:book-soft-chain-sum}
\end{align}
In particular, the global free-energy defect is nonnegative and vanishes
exactly at the oracle path law, up to null histories.
\end{theorem}

\begin{interpretationbox}
Local conditional certificates can be added because every one is measured
against the same composed Gibbs objective.  The trial law need not be
Markov: history dependence is absorbed by its regular conditional laws.
The theorem does not license adding arbitrary local losses that use different
reference measures, temperatures, or target normalizations.
\end{interpretationbox}

\begin{proofroadmap}
Multiply the normalized Gibbs transitions to obtain the oracle path density
relative to the reference path law.  The reports telescope to total path
cost minus the initial soft value, yielding the first equality.  Apply the
relative-entropy chain rule to that product law for the second.  The chapter
appendix records both steps and the equality case.
\end{proofroadmap}

\section{Base change and data processing}
\index{base change!architecture}

Conditional chain rules live at the candidate-law level.  Passing them
through a representation map and an architecture infimum requires the
feasible fine fiber to remain visible.  The source box identifies the imported
results that justify this step.

\begin{sourcebox}[title={Source result; proof not reproduced here}]
The exact architecture base-change identity, canonical-fiber saturation
criterion, quantitative dequantization bounds, and temperature-dependent class
variants summarized in the remainder of this chapter come from COT and
\EGIII.  The chapter appendix proves only the soft conditional chain theorem
above; these additional variants are not counted as principal book results
\citep{COT,FrameworkPartIII}.
\end{sourcebox}

Within that imported scope, fix a base-change map from a fine deployment to a
coarse deployment.  Taking architecture infima gives an exact infimal theorem
only when the feasible fine fibers are tracked as a function of the coarse
law.  The fine obstruction is
\[
  O_{\rm fine}
  =
  \inf_{Q_{\rm coarse}}
  \left\{
  O_{\rm coarse}(Q_{\rm coarse})
  +
  \Psi(Q_{\rm coarse})
  \right\}.
\]
The fiber term depends on coarse occupancy.  Replacing it by a constant can be wrong.

Universal equality between fine and coarse obstructions requires canonical-fiber saturation.  Objective preservation alone is insufficient.

\paragraph{Pushforward and data processing.}
\index{data processing}
\index{pushforward}

A measurable map reduces relative entropy:
\[
  \KL(Q\|P)
  \ge
  \KL(T_\#Q\|T_\#P).
\]
The chain rule used above and this data-processing inequality belong to the
classical relative-entropy calculus \citep{CoverThomas2006,Csiszar1975}.  The
difference is a data-processing gap.  Equality can imply sufficiency or
recoverability under additional assumptions; at architecture level,
pushforward can remove an obstruction by quotienting irrelevant labels yet
create a fiber-realization tax when the fine architecture cannot reconstruct
the canonical conditional law.

\section{When conditional elimination destroys encoder semantics}
\label{sec:marton-markovity-case}
\index{Marton's inner bound}
\index{Markovity Conjecture}
\index{architecture obstruction!Marton dual}

The Markovity Conjecture asks whether one may impose $U-X-V$ without losing
the optimum of a Marton dual problem.  The useful EG question is more local:
\emph{what exactly is eliminated, which observables survive, and which carrier
semantics can be destroyed?}  For a two-receiver broadcast channel, write
\begin{equation}
  G_\Theta(p)
  =-\alpha H(Y)-(\lambda-\alpha)H(Z)
   +I(U;Y)+\lambda I(V;Z)-I(U;V)+\E[a_X],
  \label{eq:book-marton-dual}
\end{equation}
where $\Theta$ collects the channel and dual parameters.  Conditional
Markovization is the declared elimination
\[
  (\mathsf Mp)(u,v,x)=p(x)p(u\mid x)p(v\mid x).
\]
It removes dependence between $U$ and $V$ given $X$, while preserving the
$(U,X)$ and $(V,X)$ marginals.  Thus it preserves everything seen separately
by the two receivers, but it need not preserve the semantics
$H(X\mid U,V)=0$ of a deterministic encoder.  For such an encoder, the exact
native ledger is
\begin{equation}
  G_\Theta(\mathsf Mp)-G_\Theta(p)
  =\underbrace{I_p(U;V\mid X)}_{\text{dependence credit}}
   -\underbrace{H_{\mathsf Mp}(X\mid U,V)}_{\text{semantic ambiguity tax}}.
  \label{eq:book-credit-semantic-tax}
\end{equation}
The first term is the advertised reward for eliminating conditional
dependence.  The second is a typed realization loss: after productization, the
same pair $(u,v)$ may become compatible with several input symbols.  In EG
language,
\[
  \mathsf M\bigl\{p:H_p(X\mid U,V)=0\bigr\}
  \not\subseteq
  \bigl\{p:H_p(X\mid U,V)=0\bigr\}.
\]
The operation preserves visible marginals but can leave the deterministic
carrier class.  Rectangular symbol fibers are exactly the zero-tax case.

This ledger changed the search.  Instead of asking a local optimizer to find a
Markov point directly, the search looked for a nonrectangular deterministic
branch on which the semantic tax exceeds the dependence credit.  It then
separated three logically different gates,
\begin{equation}
  L_{\mathsf M}(p)
  \quad\longrightarrow\quad
  \Gamma_{\rm rect}(\pi,m)
  \quad\longrightarrow\quad
  \mathcal O_{\rm rect}(\Theta),
  \label{eq:book-marton-three-gates}
\end{equation}
namely loss under direct Markovization, the best rectangular map switch at
fixed cell masses, and loss after complete reoptimization over every
deterministic rectangular architecture.  Failure at either of the first two
gates is only a candidate certificate.  A counterexample requires the third.

This distinction also exposed the computational blind spot.  The ABCA branch
is a local maximum separated from the dominant rectangular branch by a finite
map-switch barrier.  Convergent local searches therefore fall into the
rectangular basin and report its value; sampling more starting points does not
certify that the other architecture is absent.  The finite-change
KL/Bregman ledger priced the branch switch, while the EG gate structure routed
the final task to branchwise interval certification rather than another local
optimization run.

\begin{sourcebox}[title={Computer-assisted source result}]
For an exact rational three-input dense channel,
\citet{LiuHuang2026Markovity} certifies
\[
  G_\Theta(p_{\rm ABCA})-
  \sup_{p\in\mathcal R_{\rm det}}G_\Theta(p)
  \ge 2.711224394247\times10^{-11}>0.
\]
Together with a strict-Jensen equality bridge, this proves that no global
optimizer satisfies $U-X-V$.  The result is not counted among the book's
principal results.
\end{sourcebox}

Here $\mathcal R_{\rm det}$ contains all finite deterministic encoders whose
nonempty fibers are rectangles.  Write
\[
\begin{aligned}
  F(\Theta)&=\sup_pG_\Theta(p), &
  F_{\rm rect}(\Theta)&=\sup_{p\in\mathcal R_{\rm det}}G_\Theta(p),\\
  \mathcal O_{\rm rect}(\Theta)&=F(\Theta)-F_{\rm rect}(\Theta).&&
\end{aligned}
\]
For every deterministic rectangular candidate $q$, the exact architecture
ledger is
\begin{equation}
  F(\Theta)-G_\Theta(q)
  =\underbrace{\mathcal O_{\rm rect}(\Theta)}_{\text{architecture error}}
   +\underbrace{F_{\rm rect}(\Theta)-G_\Theta(q)}_{\text{implementation error}}.
  \label{eq:book-marton-architecture-ledger}
\end{equation}
It shows why more optimization inside the rectangular class cannot repair the
result: it can remove the implementation term, never the architecture term.

Only a short proof bridge is needed after the interval certificate.  First, a
Markov-class cardinality lemma gives finite attainment.  Second, with $p_{UV}$
fixed, $G_\Theta$ is convex in $K=p(x\mid u,v)$.  Equality at a global maximum,
together with the strict marginal-entropy curvature of
$-\alpha H(Y)-(\lambda-\alpha)H(Z)$, forces the original maximizing kernel
itself to be deterministic whenever a positively weighted output sees every
input change.  This step uses neither Markovity nor full grid support.  For the
exact channel, $T_Y$ is nonsingular.  Finally, the published full-support
theorem \citep[Theorem~1]{GohariElGamalAnantharam2014Marton} completes the
active grid; determinism plus $U-X-V$ then makes every symbol fiber a
rectangle.  Thus the interval separation excludes the literal Markov optimum
as well.  The rectangle-cover and generalized-AND/XOR route is not part of
this proof.

\begin{boundarybox}
The word ``rectangular'' is used in two different directions.  In
Chapter~\ref{ch:coordination}, $\operatorname{Rect}(\mathsf{Arch})$ is an
\emph{enlargement} that permits independent pasting of admissible kernels.  In
the Marton problem, a rectangular deterministic map is a \emph{restriction}
on the symbol fibers of $f(U,V)$.  The common geometry is product closure, not
an identification of the two constructions.
\end{boundarybox}

The proof is classical information theory plus rigorous numerics; it does not
depend on EG terminology.  EG supplied the discovery and certification
architecture: type the eliminated relation, measure the native semantic tax,
separate candidate-level failure from architecture-level obstruction, and let
the failed gate decide which certificate to build.  That is the transferable
lesson.  The result refutes neither Marton's achievable region nor the
separate Additivity Conjecture.

\section{Hard limits and composition boundaries}
\paragraph{Zero temperature.}
\index{zero-temperature limit}
\index{dequantization}

A Gibbs defect has the form
\[
  \frac1\beta\KL(Q\|P_\beta^\star).
\]
As $\beta\to\infty$, the soft objective approaches hard excess cost under uniform dequantization conditions.  For a fixed finite class, minima converge.  For temperature-dependent architecture classes, one needs $\Gamma$-convergence and control of inner and outer limits.

A useful rate has the form
\[
  \frac{\Lambda}{\beta}
  +
  \operatorname{osc}(D_\infty)h_\beta,
\]
where $h_\beta$ measures class convergence.  Shared coordination taxes can dequantize at the sharper thermal rate when nested minima share the same perturbation.

\paragraph{Randomized hard limits.}
\index{randomization!hard limit}

The zero-temperature limit of a soft architecture need not be deterministic.  If the architecture closure contains randomized mixtures but not pure selectors, the hard limit can remain randomized.  Determinism is a property of the admissible class, not a consequence of low temperature alone.

\paragraph{Interchange curvature.}
\index{curvature!interchange}

Two operations may commute individually but fail in a larger context.  The
companion Foundations manuscript develops this failure as interchange
curvature \citep{FrameworkFoundations}.  Operational semantics refines the
question: which contexts and tasks can detect the noncommutation?

Examples include:
\begin{itemize}
  \item conditioning before versus after representation compression;
  \item coupling before versus after elimination;
  \item quantization before versus after task projection;
  \item temperature limit before versus after architecture restriction.
\end{itemize}

\paragraph{A composition checklist.}
\index{composition!checklist}

Before composing two certificates, verify:
\begin{enumerate}
  \item the defects use compatible orientation and units;
  \item the conditioning weight is correct;
  \item feasible fibers are nonempty and measurable;
  \item architecture infima commute only under saturation/rectangularity conditions;
  \item representation maps preserve the declared oracle object;
  \item limits are uniform or controlled by epi/$\Gamma$-convergence;
  \item downstream tasks expose the retained difference.
\end{enumerate}

\section*{Exercises}

\begin{exercise}
Derive the KL chain rule and identify the trial-law weighting of the conditional term.
\end{exercise}

\begin{exercise}
Give an example where a fine-to-coarse map preserves the optimal objective value but the fine architecture pays a positive realization tax.
\end{exercise}

\begin{exercise}
Construct a sequence of randomized soft minimizers whose zero-temperature limit remains randomized.
\end{exercise}

\input{chapter_appendices/ch18_proofs}

%% file: chapter_appendices/ch18_proofs.tex
\chapterproofappendix

\subsection*{Proof of the soft conditional chain theorem}
\proofdependency{The normalized transitions in \eqref{eq:book-soft-oracle-transition} must be probability kernels, the trial law must be absolutely continuous with respect to the reference path law, and all relative entropies must be well defined.  Standard Borel state spaces provide the required regular conditional laws.}
\begin{proof}
By multiplying the transition densities in
\eqref{eq:book-soft-oracle-transition},
\[
  \frac{dP_{x_0}^{\beta,\star}}{dM_{x_0}}(x_{1:m})
  =\exp\!\left\{-\beta\sum_{i=1}^m
    b_i^\beta(x_i,x_{i-1})\right\}.
\]
The backward values telescope:
\begin{align*}
  \sum_{i=1}^m b_i^\beta(x_i,x_{i-1})
  &=\sum_{i=1}^m
    \{K_i(x_i,x_{i-1})+V_i^\beta(x_i)
      -V_{i-1}^\beta(x_{i-1})\}\\
  &=\sum_{i=1}^mK_i(x_i,x_{i-1})+G(x_m)-V_0^\beta(x_0).
\end{align*}
Consequently,
\begin{align*}
  \frac1\beta\KL(Q\|P_{x_0}^{\beta,\star})
  &=\frac1\beta\KL(Q\|M_{x_0})
    +\E_Q\!\left\{\sum_{i=1}^m b_i^\beta(X_i,X_{i-1})\right\}\\
  &=\mathcal H_\beta(Q;x_0)-V_0^\beta(x_0),
\end{align*}
which proves \eqref{eq:book-soft-global-kl}.

The relative-entropy chain rule for the two path laws gives
\[
  \KL(Q\|P_{x_0}^{\beta,\star})
  =\sum_{i=1}^m\E_Q
  \KL\!\left(
    Q_i(\cdot\mid X_{0:i-1})
    \,\middle\|\,
    P_i^{\beta,\star}(\cdot\mid X_{i-1})
  \right).
\]
Divide by $\beta$ and use
\eqref{eq:book-soft-local-certificate} to obtain
\eqref{eq:book-soft-chain-sum}.  Each summand is nonnegative.  Their sum is
zero exactly when the conditional laws agree at every stage outside a
$Q$-null set; sequential factorization then gives
$Q=P_{x_0}^{\beta,\star}$.  The converse is immediate.
\end{proof}

%% file: part_notes/part05_history.tex
\parthistoricalnotes{}
\index{rate--distortion!historical comparison}
\index{full abstraction!historical comparison}

\subsection*{Three neighboring traditions}

The carrier--decoder language meets classical rate--distortion and
quantization \citep{Shannon1959,CoverThomas2006,GrayNeuhoff1998}; its exact
conditional split uses classical Bregman centroid geometry
\citep{BanerjeeGuoWang2005}.  Task envelopes use residuation
and idempotent projection \citep{CohenGaubertQuadrat2004}, while contextual
equivalence and full abstraction come from programming-language semantics
\citep{Milner1977,Plotkin1977}.  The soft composition theorem uses the Gibbs
variational formula and relative-entropy chain rule
\citep{Csiszar1975,CoverThomas2006}.

Part V's claim is not that these three traditions were previously absent.  It
states which interfaces must be declared before they can be combined.

\begin{center}
\small
\begin{tabularx}{0.96\textwidth}{@{}p{0.21\textwidth}p{0.32\textwidth}X@{}}
\toprule
Question & Classical object & Part V qualification \\
\midrule
What must the carrier retain? & source coding, quantization, conditional
prediction & retain the conditional dual oracle signature needed by the
native eliminated objective \\
What can a task observe? & contextual equivalence and full abstraction &
close a declared task class under a declared min-plus context grammar and use
its canonical envelope \\
When do local costs add? & Gibbs normalization and KL chain rule & use one
reference family, temperature, and telescoping normalization; arbitrary local
scores are not composable by declaration \\
\bottomrule
\end{tabularx}
\end{center}

The architecture rate--distortion theorem and the two operational theorems
are source-program realizations built on those classical cores.  The soft
conditional-chain theorem is a classical identity in the book's typed
notation.  In particular, contextual observability is a gate or transform,
not a third scalar added to carrier loss and decoder nonsaturation.

%% file: chapters/ch19_statistical_eg.tex
\chapter{Confidence Worlds and Statistical Elimination Geometry}
\label{ch:statistical-eg}
\chaptermark{Confidence Worlds and Statistical Geometry}
\index{statistical elimination geometry}
\index{statistical certification}
\index{confidence world}

Population elimination geometry assumes that the defect, oracle field,
architecture grammar, and operational kernel are known.  In data analysis
they are estimated.  Finite data do not select one population world; they
leave a set of worlds with inferential standing.  Structural conclusions must
therefore be transported through that set rather than evaluated only at a
point estimate or a favored posterior story.

The statistical layer has one basic chain:
\[
  \text{data}
  \longmapsto C_n
  \longmapsto J_{C_n}(q)
  \longmapsto \delta_{C_n}^q
  \longmapsto \text{authorized action}.
\]
The confidence world $C_n$ supplies validity, its identified image
$J_{C_n}(q)$ records the full query-relevant uncertainty, and the typed
certificate $\delta_{C_n}^q$ reports whether the declared boundary is
resolved.  Only after these objects are fixed should one ask how to acquire
more information or deploy a common witness.

\begin{figure}[htbp]
\centering
\begin{tikzpicture}[>=Latex,node distance=6mm]
  \tikzset{cbox/.style={draw=egblue!70!black,rounded corners=2pt,fill=egblue!4,
    minimum width=25mm,minimum height=9mm,align=center,font=\scriptsize}}
  \node[cbox] (data) {data};
  \node[cbox,right=of data] (world) {$C_n$\\confidence world};
  \node[cbox,right=of world] (image) {$J_{C_n}(q)$\\identified image};
  \node[cbox,right=of image] (cert) {$\delta_{C_n}^q$\\certificate};
  \draw[->,thick] (data)--(world); \draw[->,thick] (world)--(image); \draw[->,thick] (image)--(cert);
  \node[below=7mm of cert,font=\scriptsize,align=center] {resolved only if all covered\\worlds lie on one side};
\end{tikzpicture}
\caption{Finite-data certification transports a covered set of population worlds through a declared query before producing a terminal label.}
\label{fig:confidence-world-chain}
\end{figure}
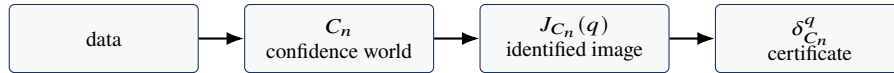

\section{Confidence worlds and finite-data permission}
\label{sec:confidence-worlds}
\index{confidence world!definition}
\index{confidence world!simultaneous}
\index{confidence world!anytime}
\index{coverage contract}

Let $\mathcal W$ be the universe of complete population worlds.  A world
specifies every population object needed by the declared query, including
nuisance components that may affect its sampling law or future experiments.

\begin{definition}[Confidence world]
A fixed-record confidence world is a data-dependent random set
$C_n\subseteq\mathcal W$ satisfying the declared coverage contract, for
example the uniform guarantee
\begin{equation}
  \inf_{w\in\mathcal W}
  \Pp_w\{w\in C_n\}
  \ge 1-\alpha.
  \label{eq:confidence-world-coverage}
\end{equation}
It is the set of worlds that remain entitled to participate in the current
certificate, not the set of worlds assigned the largest prior or posterior
mass.
\end{definition}

The construction is a world-level confidence set followed by projection, not
a new replacement for confidence-region theory.  Simultaneous confidence
regions and projection have a long statistical lineage
\citep{Scheffe1953,Dufour1997}; universal confidence sets for optimization
solutions provide an especially close formal neighbor \citep{Vogel2008}.
The term \emph{confidence world} records that the set must contain every
population component needed by the downstream architecture query.

Coverage is part of the object, not a footnote attached after a claim.  The
contract must say whether the guarantee is pointwise, uniform, or asymptotic
and whether it protects one query, a simultaneous query class, or a
data-selected query.  If adaptive stopping is allowed, the appropriate object
is an anytime confidence-world sequence satisfying
\begin{equation}
  \inf_{w\in\mathcal W}
  \Pp_w\{w\in C_t\text{ for every }t\ge0\}
  \ge 1-\alpha,
  \label{eq:anytime-confidence-world-coverage}
\end{equation}
with every protected query evaluated on the same time-uniform event.  This is
the confidence-sequence contract of time-uniform inference
\citep{HowardEtAl2021}, lifted to the complete-world object.

\begin{table}[htbp]
\centering
\small
\caption{The confidence-world construction contract.}
\label{tab:confidence-world-contract}
\begin{tabularx}{\textwidth}{p{0.25\textwidth}X}
\toprule
Declared item & What must be auditable\\
\midrule
coverage regime
  & Fixed-record, simultaneous, or anytime validity; pointwise, uniform, or
    asymptotic scope.\\
protected scope
  & The times, queries, subgroups, model components, and selection rules
    covered by one event.\\
population content
  & The nuisance, regularity, sampling, and identification assumptions encoded
    in a complete world.\\
set status
  & Whether the reported set is exact or an inner or outer approximation, and
    how that approximation changes honesty or decisiveness.\\
empty-set rule
  & $C_n=\varnothing$ is reported as model conflict $M$; it is not used for
    vacuous inference.\\
\bottomrule
\end{tabularx}
\end{table}

Confidence worlds may be built from parametric confidence regions,
distributionally robust sets, moment inequalities, partial-identification
regions, valid bootstrap bands, or simultaneous module-wise envelopes.  The
construction method may change; the coverage and reporting contract may not
be left implicit.
Conformal prediction \citep{VovkGammermanShafer2005} can supply a predictive
set or one component of such an envelope under its own exchangeability
contract; it is not an ancestor of, or substitute for, projection of a
population confidence world.

\section{Identified images, margins, and honest certificates}
\label{sec:identified-images-certificates}
\index{identified image}
\index{certificate!three-way rule}
\index{certificate!feasible}
\index{certificate!impossible}
\index{certificate!unresolved}
\index{certificate!margin}
\index{certificate!tolerance profile}
\index{model conflict}
\index{worldwise validity}

Before inspecting the data, a query contract declares the population
functional or truth map, its tolerance or action boundary, the observer,
deployment quantifier, admissible experiment class, and stopping rule.  For a
real-valued functional $\Gamma_q$, the confidence world induces the
identified image
\begin{equation}
  J_C(q)=\{\Gamma_q(w):w\in C\}.
  \label{eq:identified-image}
\end{equation}
This image, rather than a terminal color, is the complete finite-data answer
to the query.  The color is its projection relative to a declared boundary.
Projection of confidence regions and inference under partial identification
are established ideas \citep{Dufour1997,ImbensManski2004,
ChernozhukovHongTamer2007}.  The book's contribution is the typed
architecture/deployment interpretation of the projected image, not the
statistical operation of taking an image of a set.

\begin{definition}[Standard confidence certificate]
Let $q:\mathcal W\to\{F,I\}$ be a declared truth map.  For any
$C\subseteq\mathcal W$, define
\begin{equation}
  \delta_C^q=
  \begin{cases}
    M,&C=\varnothing,\\
    d,&q(C)=\{d\}\text{ for some }d\in\{F,I\},\\
    U,&|q(C)|\ge2.
  \end{cases}
  \label{eq:standard-confidence-certificate}
\end{equation}
Here $M$ records model conflict, while $U$ records genuine disagreement among
compatible worlds.  Neither state authorizes a binary population claim.
\end{definition}

For the threshold query
\[
  q(w)=F\quad\Longleftrightarrow\quad
  \Gamma_q(w)\le\eps_q,
\]
suppose the nonempty identified image is the interval
$J_C(q)=[L_C,U_C]$.  Then
\[
  \delta_C^q=
  \begin{cases}
    F,&U_C\le\eps_q,\\
    I,&L_C>\eps_q,\\
    U,&L_C\le\eps_q<U_C.
  \end{cases}
\]
The associated certificate margin is
\begin{equation}
  m_C(q)=
  \begin{cases}
    \eps_q-U_C,&\delta_C^q=F,\\
    L_C-\eps_q,&\delta_C^q=I,\\
    0,&\delta_C^q=U.
  \end{cases}
  \label{eq:certificate-margin}
\end{equation}
No margin is assigned to the model-conflict state.  A positive margin records
how far the entire identified image lies from a label change; the full map
$\eps\mapsto\delta_C^q(\eps)$ is the tolerance profile.  Reporting the image
and margin permits another stakeholder to audit a different declared
tolerance without reconstructing the population world set.

The threshold readout is also continuous with the classical equivalence and
noninferiority literature \citep{Blackwelder1982,BergerHsu1996,
PiaggioEtAl2006,Schuirmann1987,Lakens2017}.  The familiar two-one-sided-tests
procedure is a two-sided sibling of this set rule, not its source: here a
single resolved label is authorized only when the entire covered image lies
on the declared side of the boundary.

\begin{conditionbox}
The confidence world covers the complete population object under the declared
query and stopping contract.  For maximality, competing rules receive only
the same realized nonempty confidence world and truth map, take values in
$\{F,I,U\}$, and must be correct for every world in that set; they receive no
extra statistic or separate randomized error allowance.  The $M$ output is a
diagnostic for an empty set, not a population truth label.
\end{conditionbox}

\begin{theorem}[Honesty and maximal decisiveness]
\label{thm:three-way}
On the coverage event, every resolved output of the standard confidence
certificate is correct.  Among the comparator rules just declared, it is
maximally decisive: it returns the common truth label whenever all compatible
worlds agree and returns unresolved otherwise.
\end{theorem}

\begin{interpretationbox}
Unresolvedness is a mathematical consequence of partial identification, not
a failed optimizer or a request to guess.  Maximal decisiveness belongs to the
exact image $q(C)$.  If computation supplies only an outer enclosure of that
image, resolved outputs remain honest but an unresolved output may be
conservative.
\end{interpretationbox}

\begin{proofroadmap}
On the coverage event, a singleton truth image contains the true label.  If
two labels remain possible, either binary choice is false in one compatible
world, so no worldwise-correct comparator may decide.  The chapter appendix
proves both honesty and maximality.
\end{proofroadmap}

\paragraph{Three kinds of unresolvedness.}
Statistical unresolvedness means that compatible population worlds imply
opposite answers; computational unresolvedness means that the exact
identified image is defined but has not been computed; search unresolvedness
means that no witness has been found within the declared budget.  The remedies
differ: more computation cannot resolve the first, and more data cannot by
itself repair a misspecified grammar.

The compatible-world logic is rooted in partial identification
\citep{Manski2003}, with projection confidence procedures providing a nearby
inferential comparison \citep{KaidoMolinariStoye2016}.  The book's additional
step is to require one auditable architecture certificate under the declared
deployment quantifier.

As a running anchor, Section~\ref{sec:book-ucb-audit} constructs one 95\%
simultaneous confidence world for the 12 department--sex admission
probabilities in \texttt{UCBAdmissions}.  Its projection through the maximum
department difference is $[-0.0895,0.3109]$; relative to the tolerance $0.10$,
the standard certificate is $U$.  The full audit later shows why a tiny pooled
$p$-value does not replace this worldwise conclusion.

\section{Defect envelopes and architecture grammars}
\index{defect!confidence envelope}
\index{architecture!grammar}

The generic objects become structural by projecting a confidence world
through the defect field and the architecture grammar.  Suppose a
simultaneous event provides lower and upper defect envelopes
\[
  \underline D_x(a)
  \le
  D_x^\star(a)
  \le
  \overline D_x(a)
\]
for all relevant $(x,a)$.  Suppose also that the true architecture lies between inner and outer grammars
\[
  \underline\Arch_r
  \subseteq
  \Arch_r^\star
  \subseteq
  \overline\Arch_r
\]
for every resource budget $r$.

Simultaneous objective and feasible-set envelopes of this type are standard
tools in stochastic and robust optimization
\citep{Vogel2008,BerkEtAl2013}; the point here is to preserve their common
event through elimination.

Use monotone $\rho=P_n$ for empirical risk, or $\rho=P$ for
confidence-controlled population risk.
\begin{align*}
  \underline O(r)
  &=\inf_{A\in\overline\Arch_r}\rho(\underline D_A),\\
  O^\star(r)
  &=\inf_{A\in\Arch_r^\star}\rho(D_A^\star),\\
  \overline O(r)
  &=\inf_{A\in\underline\Arch_r}\rho(\overline D_A).
\end{align*}
The directions are deliberately reversed: a larger architecture and smaller defect produce a lower bound; a smaller architecture and larger defect produce an upper bound.

\begin{conditionbox}
One simultaneous confidence event must contain pointwise lower/upper defect
envelopes and inner/outer architecture-grammar inclusions for every resource
budget under study.  The same declared risk functional $\rho$ is used in the
lower, true, and upper frontiers and is monotone.  If the target is a
population frontier but $P$ is unknown, the confidence construction must
also justify the population risk operation; substituting $P_n$ silently is
not sufficient.  Pointwise intervals selected separately after looking at
the data are not sufficient.
\end{conditionbox}

\begin{theorem}[Honest elimination bracket]
\label{thm:honest-elimination-bracket}
On the simultaneous confidence event,
\[
  \underline O(r)
  \le
  O^\star(r)
  \le
  \overline O(r)
\]
for all declared resource budgets $r$ simultaneously.
\end{theorem}

\begin{interpretationbox}
The theorem converts uncertain defects and uncertain feasible sets into a simultaneous lower and upper bracket for the entire architecture frontier defined by $\rho$.  The width records statistical resolution rather than optimization failure.  Honesty is inherited from the common event; no assertion is made outside it, and an empirical frontier is not relabeled as a population frontier.
\end{interpretationbox}

\begin{proofroadmap}
Use monotonicity to compare each defect envelope with the true defect under
the same risk functional and use set inclusion to compare the three infima.
Because the event is simultaneous in the budget, the deterministic sandwich
holds for the whole frontier at once.  The chapter appendix gives the
ordered-infimum proof.
\end{proofroadmap}

At threshold $\tau$, the bracket gives the safe readout
\begin{align*}
\text{certified feasible}&:\quad \overline O\le\tau,\\
\text{certified impossible}&:\quad \underline O>\tau,\\
\text{unresolved}&:\quad \underline O\le\tau<\overline O.
\end{align*}
If $[\underline O,\overline O]$ is the exact identified image of the
frontier over $C_n$, this is the standard certificate above.  If it is only an
outer enclosure, the two resolved declarations remain honest, while the
unresolved region can also contain conservatism from the enclosure.

\section{Persistent atlases and operational transport}
\sectionmark{Persistent Atlases and Transport}
\paragraph{Persistent atlas number.}
\index{atlas!persistent number}

Uniformity over $r$ permits data-dependent resource selection without a second
pointwise argument.  For tolerance $\tau$, define the minimum chart number
using the convention that a ``$K$-chart deployment'' has at most $K$
nonempty charts; an atlas with fewer charts may be padded by empty or repeated
charts.  Thus feasibility is monotone in $K$.  Put
\[
  K_D(\tau)
  =
  \min\left\{K:
  \text{there exists a $K$-chart deployment with uniform defect}\le\tau
  \right\}.
\]
The integer-valued curve can jump under small perturbations.  Define the inverse frontier
\[
  L_D(K)=\inf\{\tau:K_D(\tau)\le K\}.
\]

The inversion between a cover by at most $K$ local sections and a $K$-chart
description is definitional.  It is not the theorem below.  Chart
autoencoders, multi-chart flows, and atlas generative models already use
multi-chart representations \citep{SchonsheckChenLai2019,KalatzisEtAl2021,
StolbergLarsenSommer2021}; the theorem concerns stability of the declared
loss frontier under a uniform defect perturbation.

The sharp interleaving and inverse-frontier stability theorem is developed
canonically in the Statistical EG companion \citep{StatisticalEG}.  The EGML
companion reuses that interface for its learning-facing atlas and one-sided
certificate ledger \citep{EGML}; the book records the mathematical statement
once, in the stronger extended-valued form below.

\begin{conditionbox}
The population and estimated defects are uniformly within $\epsilon$ on the complete incidence space, and the atlas number is defined from sublevel incidences using the ``at most $K$ nonempty charts'' convention above.  The atlas number and inverse frontier are extended-valued, with $+\infty$ used when no admissible finite-chart deployment exists at the declared level.
\end{conditionbox}

\begin{theorem}[Atlas interleaving and stable inverse]
\label{thm:atlas-stability}
If
\[
  \|D-\widehat D\|_\infty\le\epsilon,
\]
then the atlas curves are horizontally interleaved by $\epsilon$, and
\[
  L_{\widehat D}(K)\le L_D(K)+\epsilon,
  \qquad
  L_D(K)\le L_{\widehat D}(K)+\epsilon
\]
in the extended order for every $K$.  Whenever either inverse frontier is
finite, both are finite and therefore
\[
  |L_D(K)-L_{\widehat D}(K)|\le\epsilon.
\]
The constant one on the finite part is sharp.
\end{theorem}

\begin{interpretationbox}
The integer-valued minimum chart number can jump, but its sublevel filtration interleaves under uniform defect perturbation.  Consequently the inverse $K$-chart loss frontier is stable.  This is a stability theorem for achievable loss, not a consistency theorem for a particular greedy atlas.
\end{interpretationbox}

\begin{proofroadmap}
The uniform defect bound yields two inclusions between population and estimated sublevel incidences at thresholds shifted by $\epsilon$.  Monotonicity of chart number under incidence inclusion gives interleaving; invert the inequalities to obtain the loss-frontier bound.  Full details appear in the chapter appendix.
\end{proofroadmap}

The inverse chart-budget/loss frontier is therefore the statistically stable object.
This filtration-and-confidence viewpoint is adjacent to confidence sets for
persistence diagrams \citep{FasyEtAl2014}, but the object stabilized here is
an architecture loss frontier rather than a homological summary.
The underlying interleaving logic is the standard stability mechanism for
filtered objects \citep{deSilvaMunchStefanou2017,MemoliStefanouZhou2024}.

\paragraph{Nonexpansive operational transport.}
\index{operational visibility!nonexpansive transport}

Task envelopes and contextual closure are nonexpansive in sup norm.
Consequently, if internal kernels are simultaneously bracketed, observable
architecture frontiers inherit the same confidence event without a new
stochastic proof.  This is an important design principle: prove deterministic
nonexpansiveness once, then reuse the statistical envelope.

\subsection{Pointwise feasibility versus common deployment}
\index{common deployment}
\index{partial identification!common witness}

The standard confidence certificate asks whether each compatible population world
is feasible.  Deployment demands a stronger quantifier: one architecture
must work throughout the confidence world.  Let $C$ be a nonempty set of
compatible worlds, let $\Arch_R$ be the class admitted by resource budget
$R$, and let $\ell_\theta(A)\ge0$ be the declared audit loss.  At tolerance
$\eta$, put
\[
  \mathfrak F_R^\eta(\theta)
  =\{A\in\Arch_R:\ell_\theta(A)\le\eta\}.
\]
Define
\begin{align}
  V_{\rm lo}(C;R)
  &=\inf_{\theta\in C}\inf_{A\in\Arch_R}\ell_\theta(A),
  \label{eq:book-common-deployment-lo}\\
  V_{\rm pw}(C;R)
  &=\sup_{\theta\in C}\inf_{A\in\Arch_R}\ell_\theta(A),
  \label{eq:book-common-deployment-pw}\\
  V_{\rm cd}(C;R)
  &=\inf_{A\in\Arch_R}\sup_{\theta\in C}\ell_\theta(A).
  \label{eq:book-common-deployment-cd}
\end{align}

\begin{conditionbox}
The confidence world and resource class are nonempty and fixed by the
declared contract.  The same architecture must be deployed in every world;
world-specific architectures are not silently supplied by an oracle.  The
threshold equivalences below require finite sets or attainment of the
displayed extrema.  Randomized architectures count only if the declared
grammar admits them.
\end{conditionbox}

\begin{theorem}[Common-deployment quantifier theorem]
\label{thm:book-common-deployment}
For every nonempty $C$ and $\Arch_R$,
\[
  V_{\rm lo}(C;R)
  \le V_{\rm pw}(C;R)
  \le V_{\rm cd}(C;R).
\]
Under the attainment condition:
\begin{enumerate}
  \item $V_{\rm lo}>\eta$ exactly when every compatible world is
  individually impossible at budget $R$;
  \item $V_{\rm pw}\le\eta$ exactly when every compatible world is
  individually feasible;
  \item $V_{\rm cd}\le\eta$ exactly when
  $\bigcap_{\theta\in C}\mathfrak F_R^\eta(\theta)$ is nonempty.
\end{enumerate}
Both inequalities can be strict.  In particular, pointwise feasibility in
every compatible world need not provide a common deployable witness.
\end{theorem}

\begin{interpretationbox}
There are two distinct reasons not to deploy.  A confidence world may mix
feasible and impossible populations.  Alternatively, every world may be
individually feasible while the feasible sets have empty intersection.  The
second case is a common-witness conflict.  More data can remove it by
shrinking the confidence world; a router or robust architecture can remove
it only by changing the declared grammar.
\end{interpretationbox}

\begin{proofroadmap}
The inequalities follow from two relaxations: replacing a common architecture
by a world-specific one, and replacing a supremum over worlds by an infimum.
Under attainment, each threshold statement is a direct translation of the
corresponding feasible-set condition.  The strict-gap example below shows why
pointwise feasibility does not imply a common witness.  The chapter appendix
gives the full quantifier argument.
\end{proofroadmap}

\begin{table}[H]
\centering
\small
\caption{Four confidence-world deployment states under attainment.}
\begin{tabularx}{\textwidth}{p{0.23\textwidth}p{0.23\textwidth}X}
\toprule
State & Frontier condition & Meaning and authorized next step\\
\midrule
common-deployable
  & $V_{\rm cd}\le\eta$
  & The feasible sets have a common point; exhibit and deploy one common witness.\\
uniformly impossible
  & $V_{\rm lo}>\eta$
  & Every compatible world is infeasible; report impossibility or revise the budget or contract.\\
truth-mixed unresolved
  & $V_{\rm lo}\le\eta<V_{\rm pw}$
  & Some worlds are feasible and others are not; collect data that separates them.\\
common-witness conflict
  & $V_{\rm pw}\le\eta<V_{\rm cd}$
  & Every world is individually feasible but the feasible-set intersection is empty; shrink $C$ or enlarge the legal routing or robust-deployment grammar.\\
\bottomrule
\end{tabularx}
\end{table}

\begin{example}[A minimal common-witness conflict]
Let $C=\{\theta_0,\theta_1\}$, $\Arch_R=\{A_0,A_1\}$, and let the loss table be
\[
\begin{array}{c|cc}
  & \theta_0 & \theta_1\\ \hline
 A_0 & 0 & 1\\
 A_1 & 1 & 0
\end{array}.
\]
At tolerance $\eta=0$, each world has a zero-loss architecture, so
$V_{\rm pw}=0$, but every fixed architecture has worst-world loss one, so
$V_{\rm cd}=1$.  Pointwise feasibility is complete while common deployment
fails.
\end{example}

Suppose now that $(C_t)_{t\ge0}$ is an anytime-valid confidence sequence, so
one time-uniform event of probability at least $1-\alpha$ satisfies
$\theta\in C_t$ for every $t$.  At any stopping time $\tau$, every
$\mathcal F_\tau$-measurable witness in the feasible-set intersection is then
valid in the true world.  A fixed-time simultaneous confidence set alone does
not justify this optional-stopping conclusion.  Nor does a confidence world
license choosing a different witness for each still-compatible world and
presenting that family as one deployed system.

\subsection{Distribution-free prediction under conditional and shift contracts}
\index{conformal prediction!conditional coverage}
\index{covariate shift!prediction sets}

Prediction sets make the common-deployment quantifier visible without any
architecture terminology.  Under exchangeability, a conformal rule can
guarantee marginal coverage for a future observation.  That contract does not
imply exact conditional coverage at every covariate value over a rich
distribution class: distribution-free conditional guarantees can force
essentially uninformative sets unless the conditioning class, distribution
family, or target notion is restricted
\citep{BarberEtAl2021ConditionalCoverage}.  Pointwise existence of a short,
well-calibrated set in each compatible world therefore does not exhibit one
data-dependent set rule that is simultaneously short and conditionally valid
throughout the confidence world.

Covariate shift supplies a different repair only under a different declared
experiment.  Weighted conformal methods can recover a target-population
coverage statement when the conditional law of the response given covariates
is invariant and the source-to-target density ratio is known or estimated
well enough for the stated guarantee \citep{TibshiraniEtAl2019CovariateShift}.
The ratio is an information carrier, not a free optimizer improvement.  If the
conditional law also shifts, or target support lies outside source support,
reweighting does not close the contract.

The typed repair choices are consequently distinct: enlarge the output to a
set, allow abstention or an ``unresolved'' action, collect target-domain
calibration data, or restrict the target/conditioning domain.  Tuning the same
point predictor cannot by itself supply missing exchangeability, overlap, or
conditional information.

\begin{boundarybox}
This is an application of classical conformal coverage and impossibility
results, not a new conformal theorem.  Any use of the book's deployment
language must preserve the exact coverage level, conditioning sigma-field,
shift assumptions, set-size loss, and abstention semantics.
\end{boundarybox}

\subsection{Statistical ancestry: pointwise success is not uniform deployment}
\index{oracle property!pointwise versus uniform}
\index{superefficiency}

The common-deployment quantifier is not detached from classical statistics.
The Hodges--Le Cam superefficiency example already shows that an estimator may
have exceptional asymptotic risk at one fixed parameter while paying for that
behavior on a neighborhood that contracts with sample size
\citep{LeCam1953}.  H\'ajek's convolution and local asymptotic minimax theory
then made regularity and locally uniform risk comparison central
\citep{Hajek1970,Hajek1972}.  The lesson is not that pointwise results are
false.  It is that their quantifiers cannot be exchanged without proof.

Sparse estimation exhibits the same problem sharply.  An oracle property is
normally a statement at each fixed parameter: asymptotically, the procedure
behaves like an estimator that knew the true submodel in advance.
Leeb--P\"otscher show why such pointwise asymptotics can coexist with severe
nonuniformity and poor maximal risk \citep{LeebPotscher2005,LeebPotscher2008}.
Thus
\[
  \forall\theta\ \exists n_0(\theta)
  \quad\not\Rightarrow\quad
  \exists n_0\ \forall\theta.
\]
The deployment rule and sample size are fixed before the unknown world is
revealed; the second order of quantifiers is the relevant one.

\section[Pointwise-oracle sparse model selection]{A complete quantifier case: sparse model-selection procedures with pointwise oracle guarantees}
\label{sec:sparse-quantifier-case}
\index{sparse model selection!pointwise oracle guarantee}
\index{SCAD}
\index{MCP}
\index{minimax concave penalty}
\index{Adaptive LASSO}
\index{Adaptive Elastic Net}
\index{TLP}
\index{truncated $\ell_1$ penalty}
\index{best subset selection}
\index{BeSS}

The class at issue consists of support-selecting procedures whose oracle
guarantees are pointwise in the underlying parameter.  It includes
folded-concave penalization, pilot-adaptive weighted penalization with or
without quadratic stabilization, and capped or discrete $\ell_0$-type
selection.  SCAD and MCP, Adaptive LASSO and Adaptive Elastic Net, and TLP
with BSS/BeSS below represent these three mechanism classes; they are not an
exhaustive catalogue of sparse procedures.

Consider one coordinate of an orthogonal Gaussian regression experiment,
\[
  Z_j=\beta_j+\frac{\sigma}{\sqrt n}\varepsilon_j,
  \qquad \varepsilon_j\sim N(0,1),
\]
with every penalized objective normalized as
\[
  \frac12(Z_j-b)^2+\operatorname{pen}_n(b).
\]
This coordinate model exposes the thresholds exactly.  General designs add
design identifiability, pilot, and optimization conditions, but do not remove
the quantifier conflict.

\paragraph{A common moving-threshold calculation.}
Suppose a coordinate rule has an effective activation threshold $s_n$: it
sets $\widehat\beta_j=0$ whenever $|Z_j|\le s_n$, where $s_n\to0$ and
$\sqrt n s_n\to\infty$.  At the moving world $\beta_{j,n}=cs_n$, $0<c<1$,
one has $Z_j/s_n\to c$ in probability, so
$P_{\beta_{j,n}}(\widehat\beta_j=0)\to1$.  Consequently,
\[
 \begin{aligned}
 n\,\E_{\beta_{j,n}}(\widehat\beta_j-\beta_{j,n})^2
 &\ge n\beta_{j,n}^2
   P_{\beta_{j,n}}(\widehat\beta_j=0)\\
 &= (c^2+o(1))ns_n^2\longrightarrow\infty.
 \end{aligned}
\]
This is divergence of the $n$-scaled risk, not necessarily of the unscaled
mean squared error, which may still vanish because $s_n\to0$.  The failure is
nonuniform root-$n$ behavior rather than inconsistency.

\subsection{SCAD and MCP: folded-concave penalties leave moving transition bands}

For SCAD parameter $a>2$, the orthogonal coordinate solution is
\[
\widehat\beta_j^{\rm SCAD}=
\begin{cases}
\operatorname{sgn}(Z_j)(|Z_j|-\lambda_n)_+,
  & |Z_j|\le2\lambda_n,\\[2mm]
\displaystyle
\frac{(a-1)Z_j-\operatorname{sgn}(Z_j)a\lambda_n}{a-2},
  & 2\lambda_n<|Z_j|\le a\lambda_n,\\[3mm]
Z_j,& |Z_j|>a\lambda_n.
\end{cases}
\]
Under the usual oracle scaling $\lambda_n\to0$ and
$\sqrt n\lambda_n\to\infty$, every fixed nonzero coordinate eventually
enters the unpenalized region, while a fixed zero coordinate is deleted with
probability tending to one \citep{FanLi2001}.  But for the moving world
$\beta_{j,n}=c\lambda_n$, $0<c<1$, one has
$Z_j/\lambda_n\to c$ in probability.  The coordinate is therefore set to
zero with probability tending to one.  The common moving-threshold
calculation applies with $s_n=\lambda_n$, so the $n$-scaled SCAD risk diverges
along this moving world.
SCAD removes persistent shrinkage bias for fixed large signals; it relocates,
rather than eliminates, the price to a threshold band moving toward zero.

The minimax concave penalty (MCP) makes the same mechanism especially
transparent.  For concavity parameter $\gamma>1$, put
\[
 p_{\lambda_n,\gamma}^{\rm MCP}(t)=
 \begin{cases}
  \lambda_n t-\dfrac{t^2}{2\gamma},&0\le t\le\gamma\lambda_n,\\[2mm]
  \dfrac{\gamma\lambda_n^2}{2},&t>\gamma\lambda_n.
 \end{cases}
\]
Zhang's MC+ procedure couples MCP with the PLUS path algorithm; the scalar
calculation here concerns the global MCP minimizer itself
\citep{Zhang2010MCP}.  For the normalized objective above, that minimizer is
the firm-threshold rule
\[
 \widehat\beta_j^{\rm MCP}=
 \begin{cases}
  0,&|Z_j|\le\lambda_n,\\[1mm]
  \displaystyle
  \operatorname{sgn}(Z_j)\frac{|Z_j|-\lambda_n}{1-1/\gamma},
    &\lambda_n<|Z_j|\le\gamma\lambda_n,\\[3mm]
  Z_j,&|Z_j|>\gamma\lambda_n.
 \end{cases}
\]
Under the same scaling $\lambda_n\to0$ and
$\sqrt n\lambda_n\to\infty$, MCP deletes a fixed zero and eventually leaves
a fixed nonzero coordinate unpenalized.  Yet at
$\beta_{j,n}=c\lambda_n$, $0<c<1$, it deletes the coordinate with probability
tending to one.  Taking $s_n=\lambda_n$ in the common calculation gives the
same divergence of the $n$-scaled MCP risk.
SCAD and MCP have different transition maps, but both exchange fixed-signal
unbiasedness for a moving nonuniform band.

\subsection{Adaptive LASSO and Adaptive Elastic Net: weighted lockout and stabilization}

Use the same score as pilot and let $w_j=|Z_j|^{-\gamma}$, $\gamma>0$.
The coordinate solution is
\[
 \widehat\beta_j^{\rm AL}
 =\operatorname{sgn}(Z_j)
  \left(|Z_j|-\frac{\lambda_n}{|Z_j|^\gamma}\right)_+.
\]
Its effective threshold is
\[
 t_n=\lambda_n^{1/(\gamma+1)},
 \qquad
 \widehat\beta_j^{\rm AL}\ne0\Longleftrightarrow |Z_j|>t_n.
\]
In the present normalization, the classical pointwise oracle conditions can
be written
\[
 \sqrt n\lambda_n\to0,
 \qquad n^{(\gamma+1)/2}\lambda_n\to\infty.
\]
The first makes the penalty negligible at a fixed nonzero coordinate; the
second is equivalent to $\sqrt n t_n\to\infty$ and deletes a fixed zero
coordinate \citep{Zou2006AdaptiveLasso}.  At the moving world
$\beta_{j,n}=ct_n$, $0<c<1$, however, $Z_j/t_n\to c$, so deletion occurs with
probability tending to one.  Taking $s_n=t_n$ in the common calculation gives
divergence of the $n$-scaled Adaptive-LASSO risk.
Adaptive weighting resolves an important fixed-zero/fixed-nonzero conflict of
ordinary LASSO, but near zero the pilot amplifies its own noise.  The
nonuniform region becomes a pilot-dependent lockout region.

Adaptive Elastic Net adds quadratic stabilization to the same weighted
selection gate.  With a pilot $\widetilde\beta_j$, weight
$w_j=|\widetilde\beta_j|^{-\gamma}$, and $\rho_n\ge0$, its normalized
coordinate criterion is
\[
 \frac12(Z_j-b)^2+\frac{\rho_n}{2}b^2+\lambda_n w_j|b|.
\]
For a fixed realized weight, the minimizer before the conventional exterior
rescaling is
\[
 \widetilde\beta_j^{\rm AEN}
 =\frac{\operatorname{sgn}(Z_j)(|Z_j|-\lambda_nw_j)_+}{1+\rho_n},
 \qquad
 \widetilde\beta_j^{\rm AEN}\ne0
 \Longleftrightarrow |Z_j|>\lambda_nw_j.
\]
Exterior rescaling changes the nonzero magnitude, not this support gate.
The quadratic term can improve conditioning and stability among correlated
predictors, which is the additional role emphasized by
\citet{ZouZhang2009AdaptiveElasticNet}; it cannot reopen a coordinate already
excluded by the pilot-weighted $\ell_1$ term.  Correspondingly, the oracle
theorem requires a minimum-signal condition relative to the dimension and
tuning sequences.  A moving world that follows the frozen weighted threshold
retains the Adaptive-LASSO nonuniformity.  Moreover, if a pilot can equal zero,
an implementation must declare whether its adaptive weight is capped;
otherwise that pilot decision becomes an irreversible support lockout.

\subsection{TLP, BSS, and BeSS: capped or discrete selection cannot create information}

The truncated $\ell_1$ penalty (TLP) is a capped surrogate for discrete
$\ell_0$ selection.  In the present normalization, let
\[
 \operatorname{pen}^{\rm TLP}_n(b)
 =\mu_n\min\left\{\frac{|b|}{\tau_n},1\right\},
 \qquad \mu_n>0,\quad\tau_n>0,
\]
and write $a_n=\mu_n/\tau_n$.  Direct minimization gives two regimes, up to
arbitrary tie-breaking.  If $\tau_n\le\sqrt{\mu_n/2}$, TLP is exactly the
hard-threshold rule
\[
 \widehat\beta_j^{\rm TLP}
 =Z_j\mathbf1\{|Z_j|>\sqrt{2\mu_n}\}.
\]
If $\tau_n>\sqrt{\mu_n/2}$, it instead has a linear activation segment followed
by an unpenalized tail:
\[
 \widehat\beta_j^{\rm TLP}=
 \begin{cases}
  0,&|Z_j|\le a_n,\\[1mm]
  \operatorname{sgn}(Z_j)(|Z_j|-a_n),
    &a_n<|Z_j|\le\tau_n+a_n/2,\\[1mm]
  Z_j,&|Z_j|>\tau_n+a_n/2.
 \end{cases}
\]
Thus TLP interpolates between a capped shrinkage rule and exact coordinatewise
$\ell_0$ selection.  Shen, Pan, and Zhu develop it as a computational
surrogate for $\ell_0$ likelihood and state selection guarantees under an
explicit separation condition \citep{ShenPanZhu2012TLP}.  Let $v_n$ denote
its activation threshold, equal to $\sqrt{2\mu_n}$ in the first regime and
$a_n$ in the second.  Whenever pointwise-oracle tuning makes
$v_n\to0$ and $\sqrt n v_n\to\infty$ (with the flat-tail boundary also
tending to zero), the moving world $\beta_{j,n}=cv_n$, $0<c<1$, is deleted
with probability tending to one.  Taking $s_n=v_n$ in the common calculation
gives divergence of the $n$-scaled TLP risk.
The cap changes the method-induced transition geometry; it does not remove
the moving threshold band.

The one-dimensional $\ell_0$ objective
\[
 \frac12(Z_j-b)^2+\mu_n\mathbf1\{b\ne0\}
\]
has the global hard-threshold solution
\[
 \widehat\beta_j^{\rm BSS}
 =Z_j\mathbf1\{|Z_j|>u_n\},
 \qquad u_n=\sqrt{2\mu_n}.
\]
If $u_n\to0$ and $\sqrt n u_n\to\infty$, it has the same fixed-parameter
oracle pattern and the same failure of uniformity at
$\beta_{j,n}\asymp u_n$.  For one already frozen Adaptive-LASSO rule, choosing
$u_n/t_n\to0$ lets BSS recover signals in the corridor
\[
 u_n\ll|\beta_{j,n}|\lesssim t_n.
\]
That is a genuine rule-specific repair.  It is not a new information bound:
the Adaptive-LASSO sequence could itself be retuned, and both rules still use
the same observation.

For a general design, exact BSS additionally needs a subset-identifiability
margin, while an approximate solver must spend part of that margin on
optimization error.  BeSS searches subsets by a primal--dual active-set
algorithm and selects model size by an additional sequential or
golden-section rule \citep{WenEtAl2020BeSS}.  Even if BeSS is assumed to
return the exact global BSS solution, it removes only failures caused by
pilot weights, convex relaxation, or optimization.  Exact support recovery
still requires signal separation and design identifiability
\citep{GuoZhuFan2020BSS}.

\subsection{The information-limited core and the honest action}

Compare the local worlds
\[
 \beta_j^{(0)}=0,
 \qquad \beta_{j,n}^{(1)}=\frac h{\sqrt n}.
\]
Their Kullback--Leibler divergence is $h^2/(2\sigma^2)$ and does not diverge
with $n$.  Every support selector $\widehat S_n$ therefore obeys the two-point
testing bound
\[
 P_{\beta^{(0)}}(j\in\widehat S_n)
 +P_{\beta_n^{(1)}}(j\notin\widehat S_n)
 \ge 1-\operatorname{TV}
 \{P_{\beta^{(0)}},P_{\beta_n^{(1)}}\}>0.
\]
No support-selecting sparse procedure with only a pointwise oracle guarantee
can uniformly recover support on a parameter class containing these local
worlds; this includes every procedure above.  In a high-dimensional
multiple-coordinate problem, the corresponding separation scale is
typically enlarged to order $\sigma\sqrt{\log p/n}$, with constants depending
on the design and error criterion.

There are consequently two different blind regions.  A method-induced region,
such as pilot-weight lockout, can be narrowed by changing the tuning or by a
capped/discrete procedure such as TLP or BSS/BeSS.  An
information-theoretic indistinguishability region cannot be removed by any
selector using the same data.  A uniform support claim must either restrict
the parameter domain, for example to
\[
 \Theta_n(b_n)=
 \left\{\beta:\min_{j\in S(\beta)}|\beta_j|\ge b_n\right\},
\]
with its design margin, model-size rule, and optimization error declared, or
report a three-way action: stably nonzero, stably zero, or unresolved.

\begin{boundarybox}
This section proposes no new sparse-estimation theorem.  Pointwise-oracle
sparse model selection and the representative mechanisms analyzed here---SCAD
and MCP, Adaptive LASSO and Adaptive Elastic Net, TLP, hard
thresholding/BSS, and BeSS---have classical antecedents; so do the two-point
lower bound and BSS identifiability margins.  The book-level contribution is
to place method-induced repair and
information-limited nonrepair in one typed common-deployment interface,
thereby stating when a method change alters an algorithmic output and when it
can actually change an authorized deployment conclusion.
\end{boundarybox}

\section[Observational overlap and deployment conflict]{Main interface theorem family: observational overlap and deployment conflict}
\label{sec:deployment-conflict}
\index{observational overlap}
\index{deployment conflict}
\index{architecture repair!information-matched}

The sparse example raises a general question.  If every world has its own
low-risk architecture, can the data reliably select the required one at
deployment time?  Let the finite world set be
$\mathcal W=\{0,\ldots,m-1\}$ and $Y\sim P_w$.  Put
\[
 \mu=\frac1m\sum_{w=0}^{m-1}P_w,
 \qquad p_w=\frac{\dd P_w}{\dd\mu}.
\]
Let the deployment action space $\mathcal A$ be compact metric, let each
native loss $D_w(a)$ be lower semicontinuous in $a$, and define
\[
 F_\eta(w)=\{a\in\mathcal A:D_w(a)\le\eta\},
 \qquad
 \mathcal W(y)=\{w:p_w(y)>0\}.
\]

\begin{conditionbox}
The experiment laws, native losses, deployment action class, and tolerance
are declared together.  The exact statement concerns one measurable rule
chosen before the unknown world is revealed.  The quantitative statement
below concerns the declared binary action class $\{a_0,a_1\}$; enlarging the
action class can alter its conclusion.
\end{conditionbox}

\begin{theorem}[Deployment conflict under observational overlap]
\label{thm:book-deployment-conflict}
Under the preceding conditions:
\begin{enumerate}
\item A measurable deployment rule $\delta(Y)$ satisfying
\[
 D_w\{\delta(Y)\}\le\eta
 \qquad P_w\text{-a.s. for every }w
\]
exists if and only if
\begin{equation}
 \bigcap_{w\in\mathcal W(y)}F_\eta(w)\ne\varnothing
 \quad\text{for $\mu$-a.e. $y$ with $\mathcal W(y)\ne\varnothing$.}
 \tag{DC-E}\label{eq:deployment-conflict-exact}
\end{equation}
\item For two worlds and two candidate architectures, suppose
\[
 \Delta_0=D_0(a_1)-D_0(a_0)>0,
 \qquad
 \Delta_1=D_1(a_0)-D_1(a_1)>0.
\]
Every measurable rule taking values in $\{a_0,a_1\}$ satisfies
\begin{equation}
 \max_{i\in\{0,1\}}
 \E_i\!\left[D_i\{\delta(Y)\}-D_i(a_i)\right]
 \ge
 \frac{\Delta_0\Delta_1}{\Delta_0+\Delta_1}
 \{1-\operatorname{TV}(P_0,P_1)\}.
 \tag{DC-R}\label{eq:deployment-conflict-regret}
\end{equation}
When $\Delta_0=\Delta_1=\Delta$, the right side is
$\Delta\{1-\operatorname{TV}(P_0,P_1)\}/2$.  The same number is the exact
equal-prior mean-regret Bayes value, attained by a likelihood-ratio rule.
The maximum-regret lower bound is not claimed sharp for every asymmetric
experiment.
\end{enumerate}
\end{theorem}

\begin{proofroadmap}
The exact condition says that every observational fiber must contain one
action acceptable in all worlds that can produce that observation.  For two
actions, expected architecture regret is the cross-loss gap times the
corresponding testing error.  Le Cam's identity lower-bounds the sum of the
two errors by $1-\operatorname{TV}$; balancing the two gap-weighted errors
gives the displayed harmonic factor.  The chapter appendix supplies the
complete argument.
\end{proofroadmap}

\begin{corollary}[Repair trichotomy and invariance]
\label{cor:book-repair-trichotomy}
For the problem in \cref{thm:book-deployment-conflict}:
\begin{enumerate}
\item changing only the optimizer, while preserving $P_w$, $D_w$, and
$\mathcal A$, changes neither \eqref{eq:deployment-conflict-exact} nor the
bound \eqref{eq:deployment-conflict-regret};
\item an information or exposure repair changes the experiment $P_w$, and
may reduce overlap by increasing total variation;
\item an architecture or class repair changes $\mathcal A$ and the feasible
sets $F_\eta(w)$, and may create a common acceptable action without changing
the observations.
\end{enumerate}
An empirical case can therefore claim that this interface changed an
architecture choice only if it froze the experiment/information carrier,
native loss, action class, and compute contract before confirmation, and the
mechanism-matched repair beats equal-information, equal-compute controls on
a separately declared endpoint.
\end{corollary}

\begin{interpretationbox}
Between the local worlds $0$ and $h/\sqrt n$, all three representative
procedure families rewrite a support action using the same observation.  They do not
change $P_0$ or $P_1$ and hence cannot remove the information core of
\eqref{eq:deployment-conflict-regret}.  BSS can still repair an extra
Adaptive-LASSO lockout corridor for a frozen rule.  That is a method-level
repair, not an erasure of observational overlap.
\end{interpretationbox}

\begin{boundarybox}
The mathematical ingredients of \cref{thm:book-deployment-conflict} are a
measurable common-witness condition and Le Cam's two-point testing identity
\citep{LeCam1953,Tsybakov2009}; the information-repair interpretation is
adjacent to Blackwell comparison of experiments \citep{Blackwell1953}.  They
are not claimed here as a new decision-theoretic lower bound.  The book-level
increment is the typed interface that forces solver, information, and
architecture repair to leave different auditable traces.
\end{boundarybox}

\subsection{Treatment policies under observational overlap and deployment constraints}
\index{policy learning!observational overlap}
\index{causal inference!deployment contract}

Let $Y$ denote the observational calibration record used to select a policy,
let $\pi$ map deployment covariates to a treatment in $\{0,1\}$, and let
$D_w(\pi)$ be negative policy welfare or regret in causal world $w$.  Each
compatible world may have its own optimal policy.  If two calibration-data
laws $P_0,P_1$ overlap while the preferred policies $\pi_0,\pi_1$ reverse and
have positive cross-gaps, then \cref{thm:book-deployment-conflict} lower-bounds
the regret of every selector that must choose from $\{\pi_0,\pi_1\}$ using
only $Y$.  The statement concerns the frozen observational experiment and
policy class; it is not a causal identification theorem.

Orthogonal nuisance estimation, double/debiased machine learning,
heterogeneous-effect learners, and observational policy learning can improve effect or value
estimation under their identification and regularity assumptions
\citep{ChernozhukovEtAl2018DML,AtheyWager2021Policy}.  They do not by themselves
create positivity, measure an unobserved confounder, or transport effects to a
population unsupported by the data.  When opposite treatment preferences
remain observationally indistinguishable, replacing the learner while
preserving $(P_w,D_w,\mathcal A)$ leaves the deployment-conflict certificate
unchanged.

The certificate instead points to typed repairs.  Randomization, an additional
effect modifier, or a valid proxy changes the information experiment; a
referral/abstention option or a richer dynamic policy changes the action
class; restricting to an overlap population changes the deployment domain.
Any claim that the repair changed policy architecture must freeze these
choices and compare the mechanism-matched intervention against
equal-information, equal-compute controls on independent units.

\begin{boundarybox}
The potential-outcome identification conditions and policy-learning methods
cited here are classical.  This example only instantiates the book's
observational-overlap interface.  Without consistency, an appropriate
exchangeability condition, positivity, and a declared transport population,
the native welfare $D_w$ is not identified and the displayed audit cannot be
estimated from observational data alone.
\end{boundarybox}

\section[Active-set conflict and repair depth]{Flagship candidate family: active-set conflict, saturation, and repair depth}
\label{sec:active-set-depth-separation}
\index{active-set conflict modulus}
\index{architecture saturation!native loss}
\index{repair depth}
\index{amortization gap}

The preceding theorem separates information repair from architecture repair
abstractly.  The next family makes the architecture side computable for one
important inverse problem.  Fix a dictionary $D\in\R^{m\times k}$ of full
column rank and $\lambda>0$, and consider nonnegative sparse inference
\begin{equation}
 z^\star(x)=\argmin_{z\ge0}J_x(z),
 \qquad
 J_x(z)=\frac12\|Dz-x\|_2^2+\lambda\mathbf1^\top z.
 \label{eq:nnsc-oracle}
\end{equation}
Put $H=D^\top D$, $\kappa=\lambda_{\min}(H)$,
$L=\lambda_{\max}(H)$, and $q=1-\kappa/L\in[0,1)$.  Full column rank is stronger
than is needed for local active-set formulas, but it makes the oracle unique
and the global native loss strongly convex.

At an input $x$ with strict KKT support $S$, write $H_S=D_S^\top D_S$ and
\begin{align}
 z_S^\star(x)&=H_S^{-1}(D_S^\top x-\lambda\mathbf1),
 &z_{S^c}^\star(x)&=0,\nonumber\\
 Q_S&=H_S^{-1}D_S^\top,
 &P_S&=D_SQ_S,\label{eq:strict-cell-objects}\\
 s_\ell(x)&=D_\ell^\top\{D_Sz_S^\star(x)-x\}+\lambda,
 &&\ell\notin S.\nonumber
\end{align}
Strictness means $z_j^\star(x)>0$ for $j\in S$ and $s_\ell(x)>0$ for
$\ell\notin S$.  Let $V\subseteq\R^m$ be a declared nonzero linear
perturbation subspace, let $\Pi_V$ be its orthogonal projector, and let
$q_{S,j}^\top$ denote row $j$ of $Q_S$.  Define the fully data-computable
$V$-restricted radius
\begin{equation}
 r_{S,V}(x)=\frac12\min\left\{
  \min_{j\in S}\frac{z_j^\star(x)}{\|q_{S,j}^\top\Pi_V\|_2},
  \min_{\ell\notin S}
   \frac{s_\ell(x)}{\|D_\ell^\top(P_S-I)\Pi_V\|_2}
 \right\},
 \label{eq:strict-cell-radius}
\end{equation}
where a positive numerator divided by zero is $+\infty$ and the minimum over
an empty set is $+\infty$.  Write $r_S=r_{S,\R^m}$ for the unrestricted
case, and put
\[
 B_V(x,r)=\{x+u:u\in V,\ \|u\|_2\le r\}.
\]

\begin{conditionbox}
The oracle objective, nonnegative code domain, dictionary, penalty, and native
loss are fixed.  The shallow comparator class is exactly
$\mathcal F_1=\{x\mapsto[Wx+b]_+\}$, with one shared affine row per output
coordinate.  Two certificate centers have strict KKT supports and positive
radii no larger than \eqref{eq:strict-cell-radius}.  The perturbation subspace
and any declared input domain containing both restricted balls are fixed
before evaluation.  A broader ReLU network, a router, or an unrolled
optimizer is a different architecture class rather than a member of
$\mathcal F_1$.
\end{conditionbox}

\begin{theorem}[Certified active-set conflict modulus and one-pass saturation]
\label{thm:active-set-conflict-modulus}
Let $x_1,x_2$ be strict KKT inputs with supports $S_1,S_2$.  Fix finite radii
$0<r_i\le r_{S_i,V}(x_i)$ and write $B_i=B_V(x_i,r_i)$.  Then:
\begin{enumerate}
\item on the closed restricted ball $B_i$, the support remains $S_i$ and
\[
 z_{S_i}^\star(x)=z_{S_i}^\star(x_i)+Q_{S_i}(x-x_i),
 \qquad z_{S_i^c}^\star(x)=0;
\]
\item if a coordinate $j\in S_1\cap S_2$ has local oracle rows
$q_i^\top=q_{S_i,j}^\top$, define
\begin{align}
 \rho_i&=z_j^\star(x_i)-r_i\|q_i^\top\Pi_V\|_2,
 &h&=(r_1^{-1}+r_2^{-1})^{-1},\nonumber\\
 \Gamma_{12,j,V}&=\min\{\rho_1,\rho_2,
 h\|(q_1-q_2)^\top\Pi_V\|_2\}.
 \label{eq:active-set-conflict-modulus}
\end{align}
Every one-pass thresholded-affine encoder $f\in\mathcal F_1$ obeys
\begin{equation}
 \sup_{x\in B_1\cup B_2}\|f(x)-z^\star(x)\|_2
 \ge \Gamma_{12,j,V}.
 \label{eq:one-pass-code-floor}
\end{equation}
Consequently its worst-case native-loss gap satisfies
\begin{equation}
 \sup_{x\in B_1\cup B_2}
 \{J_x(f(x))-J_x(z^\star(x))\}
 \ge \frac\kappa2\Gamma_{12,j,V}^2.
 \label{eq:one-pass-native-floor}
\end{equation}
The certificate is nonvacuous exactly when the displayed conflict modulus is
positive.
\end{enumerate}
\end{theorem}

\begin{interpretationbox}
The obstruction is not merely that two active sets differ.  It is the
quantitative incompatibility of two oracle Jacobian rows along perturbations
that the deployment contract actually permits, on neighborhoods where the
shared coordinate must stay positive.  A single ReLU coordinate is then
forced to use one affine row on both neighborhoods, while the oracle uses
two.  The harmonic radius converts that mismatch into a code-error floor;
strong convexity converts the code floor into the declared native loss.  The
choice $V=\R^m$ recovers ordinary Euclidean balls.  A smaller $V$ can freeze
categorical or otherwise inadmissible coordinates, but the restricted balls
must still lie inside the declared input domain.
\end{interpretationbox}

Define one nonnegative proximal-gradient repair step by
\begin{equation}
 T_x(z)=\left[z-\frac1L(Hz-D^\top x+\lambda\mathbf1)\right]_+.
 \label{eq:nnsc-proximal-repair}
\end{equation}
This is a legal structural repair only when inference-time recurrence is
admitted by the deployment grammar and charged to its compute ledger.

\begin{theorem}[Repair contraction, native-loss upper envelope, and crossing budget]
\label{thm:active-set-repair-budget}
For every $x$ and $z\ge0$,
\begin{equation}
 \|T_x^t(z)-z^\star(x)\|_2
 \le q^t\|z-z^\star(x)\|_2.
 \label{eq:proximal-code-contraction}
\end{equation}
Let $K\subset\R^m$ be nonempty and compact, let
$f:K\to\R_+^k$ be bounded, and put
\begin{align*}
 E_0&=\sup_{x\in K}\|f(x)-z^\star(x)\|_2,\\
 B_K&=\sup_{x\in K}\|Hz^\star(x)-D^\top x+\lambda\mathbf1\|_2.
\end{align*}
Then the repaired architecture $f_t(x)=T_x^t\{f(x)\}$ satisfies
\begin{equation}
 \sup_{x\in K}\{J_x(f_t(x))-J_x(z^\star(x))\}
 \le B_Kq^tE_0+\frac L2q^{2t}E_0^2.
 \label{eq:proximal-native-envelope}
\end{equation}
For a native-loss tolerance $\eta>0$, set
\begin{equation}
 d_\eta=\frac{\sqrt{B_K^2+2L\eta}-B_K}{L}.
 \label{eq:nativeloss-code-target}
\end{equation}
Define the sufficient crossing depth for all boundary cases by
\begin{equation}
 t_\eta:=
 \begin{cases}
  0, & E_0\le d_\eta,\\[2pt]
  1, & E_0>d_\eta\ \text{and}\ q=0,\\[2pt]
  \left\lceil\dfrac{\log(E_0/d_\eta)}{-\log q}\right\rceil,
     & E_0>d_\eta\ \text{and}\ 0<q<1.
 \end{cases}
 \label{eq:repair-crossing-budget}
\end{equation}
Every integer depth $t\ge t_\eta$ guarantees native-loss gap at most $\eta$
on $K$.  The contraction factor $q$ is sharp for this fixed proximal
mechanism whenever the projection is locally inactive along a
$\kappa$-eigenvector of $H$.
\end{theorem}

\begin{corollary}[Certificate-driven architecture change]
\label{cor:active-set-architecture-change}
Take $K\supseteq B_1\cup B_2$ and suppose
$0<\eta<\kappa\Gamma_{12,j,V}^2/2$.  No architecture in $\mathcal F_1$ can
meet the native-loss tolerance uniformly on $K$.  If a declared initializer
$f\in\mathcal F_1$ has finite $E_0$ and the grammar admits at least $t_\eta$ applications of
\eqref{eq:nnsc-proximal-repair}, then $f_{t_\eta}$ does meet it.  Under the
same information carrier and native loss, the certificate therefore changes
the admissible architecture choice from a one-pass thresholded-affine encoder
to a recurrent or unrolled repair of sufficient depth.
\end{corollary}

\begin{proofroadmap}
Strict KKT margins survive the certified perturbation radius, yielding the
two affine oracle cells.  If a one-pass encoder approximated both restricted
balls more closely than $\Gamma_{12,j,V}$, its shared coordinate would be
positive on both balls.  Symmetric perturbations inside $V$ force the
restriction of its one affine row to lie within $e/r_i$ of each restricted
oracle row, contradicting the Jacobian conflict.
Strong convexity yields the native floor.  For repair, ReLU is nonexpansive
and $\|I-H/L\|_2=q$; expanding the quadratic objective around the KKT point
gives the native upper envelope and then the explicit crossing depth.  The
chapter appendix supplies all steps and boundary cases.
\end{proofroadmap}

\begin{boundarybox}
This family is an architecture-specific quantitative synthesis, not a claim
that shallow sparse encoders or proximal repair were previously unknown.
Gregor--LeCun introduced learned finite-depth approximations to sparse coding
\citep{GregorLeCun2010}; unfolded ISTA has established convergence theory
\citep{ChenLiuWangYin2018}; proximal-gradient and FISTA rates are classical
\citep{BeckTeboulle2009}; local sparse/piecewise-linear radii have close
neighbors in sparse local Lipschitz analysis \citep{MuthukumarSulam2023}.
Most importantly, \citet{ONeillGumranKlindt2025} already prove a global
amortisation gap for a one-layer linear--nonlinear sparse autoencoder and
show that inference-time optimization can improve sparse inference.  The
candidate increment here is narrower: an input-computable two-neighborhood
Jacobian conflict modulus, its native-loss floor, and a matched sufficient
repair-depth formula.  The exact combination was not located in the recorded
priority search, but independent priority is not established; the family is
therefore tagged \textsf{O}, not advertised as an independently confirmed
original theorem.
\end{boundarybox}

\subsection{A no-free-lunch theorem for hidden winding}
\index{hidden winding}
\index{no-free-lunch theorem!hidden winding}

\begin{sourcebox}[title={Source result; proof not reproduced here}]
The hidden-winding impossibility, exact mixed-integer degree-identification
formulation, local-spacing recovery condition, and matching random-design
rates summarized in this and the next section are established in the
Statistical Elimination Geometry manuscript.  They are not principal formal
results of this book, and their proofs are not in the chapter appendix
\citep{StatisticalEG}.
\end{sourcebox}

In the radical architecture, the one-chart obstruction is zero or one
according to a divisibility condition on the target degree after
normalization.  With finitely many sampled angles and no regularity or
coverage condition, a winding can be inserted into an unsampled interval.
The observed data can then be identical under feasible and impossible
population worlds.  Integer-wrap and phase-unwrapping formulations provide
the classical computational background
\citep{Itoh1982,GoldsteinZebkerWerner1988,Costantini1998,Yokoyama1997}.
Information-based complexity results for topological degree of
multidimensional Lipschitz maps provide a further neighboring lower-bound
tradition \citep{BoultSikorski1986}.  That cube problem is not identical to
the present one-dimensional circle-winding contract.

Therefore no uniformly valid binary procedure can be informative in general.  An honest procedure must sometimes return unresolved.

\section{Identification, hidden winding, and data-dependent architecture}
\sectionmark{Hidden Winding and Architecture}
\paragraph{Degree identification under regularity.}
\index{topology!degree identification}

Assume the target phase is $L$-Lipschitz and observed with bounded angular error.  The set of degrees compatible with the data can be described exactly by a finite mixed-integer program.  Divisibility of the entire identification set yields a valid radical certificate.

A local spacing condition can collapse the identification set to one degree.  Under uniform random design, upper and lower sample-complexity bounds match at order
\[
  L\{\log L+\log(1/\alpha)\}
\]
for the square-root parity problem.

The rate is governed by coverage of locations where an obstruction can hide, not only by pointwise estimation accuracy.
The limited novelty claim is the noisy exact mixed-integer identification
set together with the random-design logarithmic coverage surcharge.  The
bare fact that a degree-$L$ loop requires order-$L$ spatial resolution is not
claimed as new.

\paragraph{Sample splitting and data-dependent architecture.}
\index{sample splitting}
\index{architecture!data-dependent}

A practical OALI analysis can use three independent levels:
\begin{enumerate}
  \item calibration data estimate local oracles, transports, and obstruction certificates;
  \item tuning data choose among a finite repair library and fit architectures;
  \item held-out biological units evaluate final operational performance.
\end{enumerate}

This design avoids the need for a fully general random-set theorem in the first implementation, though such theory remains valuable for data efficiency.

\section*{Exercises}

\begin{exercise}
Verify the direction of the inner/outer architecture bracket.  Why does the lower bound use the outer grammar?
\end{exercise}

\begin{exercise}
Show that a sup-norm perturbation of the defect shifts the inverse $K$-chart loss frontier by at most the same amount.
\end{exercise}

\begin{exercise}
Construct two circle maps that agree on a finite sample but have different winding parity.
\end{exercise}

\input{chapter_appendices/ch19_proofs}

%% file: chapter_appendices/ch19_proofs.tex
\chapterproofappendix

\subsection*{Proof of honesty and maximal decisiveness}
\proofdependency{The confidence world must cover the true population world.
The query and its truth map are declared, and the exact truth image $q(C)$ is
available.  Compactness is unnecessary for the set-inclusion rule; it is
needed only if one replaces the identified image by attained lower and upper
endpoints.}
\begin{proof}
Fix a realized nonempty confidence world $C$.  If $q(C)=\{d\}$, then every
compatible world has truth label $d$.  On the event that the true world
belongs to $C$, the resolved output is therefore correct.  Thus any false
resolved declaration implies failure of confidence-world coverage, whose
probability is at most $\alpha$.

For maximal decisiveness, suppose $q(C)$ contains both $F$ and $I$.  An $F$
declaration is incorrect in a compatible $I$-world, while an $I$ declaration
is incorrect in a compatible $F$-world.  Hence no rule required to be correct
for every world in $C$ may make either terminal declaration.  The unresolved
output is forced.  If $q(C)$ is a singleton, the standard certificate reports
its common label.  It therefore declares in every and only every case in
which a worldwise-correct comparator can declare.  When $C=\varnothing$, the
separate output $M$ records model conflict and makes no population claim.
\end{proof}

\subsection*{Proof of the honest elimination bracket}
\proofdependency{One simultaneous event must contain both the pointwise defect envelopes and the inner/outer grammar inclusions for every resource budget under consideration.  The same declared risk functional is used for all three frontiers and must be monotone; translation equivariance is useful for perturbation bounds but not needed for the bracket itself.}
\begin{proof}
Fix a resource budget $r$.  On the simultaneous event, for every $A\in\Arch_r^\star$,
\[
\rho\{\underline D(A,\cdot)\}
\le
\rho\{D^\star(A,\cdot)\}
\le
\rho\{\overline D(A,\cdot)\}
\]
by monotonicity.  Moreover,
\[
\underline\Arch_r\subseteq\Arch_r^\star\subseteq\overline\Arch_r.
\]
Enlarging a feasible set can only decrease an infimum, while restricting it can only increase an infimum.  Hence
\[
\inf_{A\in\overline\Arch_r}\rho\{\underline D(A,\cdot)\}
\le
\inf_{A\in\Arch_r^\star}\rho\{D^\star(A,\cdot)\}
\le
\inf_{A\in\underline\Arch_r}\rho\{\overline D(A,\cdot)\}.
\]
The middle term is $O^\star(r)$.  Because the event is simultaneous in $r$, the same deterministic inequalities hold for every declared budget at once.  If the event has probability at least $1-\alpha$ uniformly over population worlds, the complete frontier bracket is uniformly honest at that level.
\end{proof}

\subsection*{Proof of atlas interleaving and inverse stability}
\proofdependency{The defects are continuous only to make the local-section formulation standard; the numerical interleaving itself uses the uniform sup-norm bound.}
\begin{proof}
If $D(x,a)\le\tau-\epsilon$, then
\[
\widehat D(x,a)
\le D(x,a)+\epsilon
\le\tau,
\]
so $E_D(\tau-\epsilon)\subseteq E_{\widehat D}(\tau)$.  Similarly,
$E_{\widehat D}(\tau)\subseteq E_D(\tau+\epsilon)$.  A local section of a smaller incidence remains a local section after inclusion into a larger incidence.  Since the minimum chart number decreases when the incidence enlarges,
\[
K_D(\tau+\epsilon)
\le K_{\widehat D}(\tau)
\le K_D(\tau-\epsilon),
\]
with the convention $K_D(s)=+\infty$ for $s<0$.

For the inverse frontier, consider any $K$-chart deployment.  Its maximum defect under $D$ and under $\widehat D$ differs by at most $\epsilon$.  Taking the infimum over all $K$-chart deployments gives
\[
L_{\widehat D}(K)
\le L_D(K)+\epsilon.
\]
Interchanging $D$ and $\widehat D$ gives
\[
L_D(K)\le L_{\widehat D}(K)+\epsilon.
\]
Both inequalities are meaningful in the extended order.  If either frontier
is finite, the two inequalities make the other finite as well, and subtraction
then gives
\[
|L_D(K)-L_{\widehat D}(K)|\le\epsilon.
\]
Sharpness on the finite part follows on singleton input and action spaces with
constant defects $D\equiv c$ and $\widehat D\equiv c+\epsilon$.
\end{proof}

\subsection*{Proof of the common-deployment quantifier theorem}
\proofdependency{The inequality chain itself needs only nonempty world and architecture sets.  Finiteness or attainment is used for the exact threshold equivalences, because an unattained infimum at the threshold need not supply a feasible witness.}
\begin{proof}
For the first inequality, write
\[
  m(\theta)=\inf_{A\in\Arch_R}\ell_\theta(A).
\]
Then
\[
  V_{\rm lo}(C;R)=\inf_{\theta\in C}m(\theta)
  \le
  \sup_{\theta\in C}m(\theta)=V_{\rm pw}(C;R).
\]
For every fixed $A\in\Arch_R$ and every $\theta\in C$,
\[
  \inf_{A'\in\Arch_R}\ell_\theta(A')
  \le \ell_\theta(A).
\]
Taking the supremum over $\theta$ and then the infimum over $A$ yields
\[
  V_{\rm pw}(C;R)
  \le
  \inf_{A\in\Arch_R}\sup_{\theta\in C}\ell_\theta(A)
  =V_{\rm cd}(C;R).
\]

Under attainment, $V_{\rm lo}>\eta$ says precisely that
$\ell_\theta(A)>\eta$ for every pair $(\theta,A)$, which is equivalent to
$\mathfrak F_R^\eta(\theta)=\varnothing$ for every $\theta$.  Likewise,
$V_{\rm pw}\le\eta$ says that each worldwise minimum is at most $\eta$,
so each feasible set is nonempty.  Finally, $V_{\rm cd}\le\eta$ says that
one attained architecture has loss at most $\eta$ simultaneously for every
world, exactly the condition
$\bigcap_{\theta\in C}\mathfrak F_R^\eta(\theta)\ne\varnothing$.

For strictness of the second inequality, take two worlds and two
deterministic architectures with loss table
\[
\begin{array}{c|cc}
 &\theta_0&\theta_1\\ \hline
 A_0&0&1\\
 A_1&1&0
\end{array}.
\]
Then $V_{\rm lo}=V_{\rm pw}=0$ but $V_{\rm cd}=1$: each world has a
zero-loss architecture, yet there is no common zero-loss witness.  With the
singleton architecture class $\{A_0\}$, one instead has
$V_{\rm lo}=0<V_{\rm pw}=V_{\rm cd}=1$, proving that the first inequality
can also be strict.  Adding randomized architectures may change these
values, but only when randomization belongs to the declared grammar.
\end{proof}

\subsection*{Proof of deployment conflict under observational overlap and the repair trichotomy}
\proofdependency{The world set is finite.  The experiment laws, native
losses, action class, and tolerance are common to the statement.  The exact
condition uses one measurable deployment rule.  The regret bound restricts
that rule to the two declared candidate actions.}
\begin{proof}
Suppose first that a deployment rule $\delta$ satisfies the tolerance in every
world.  For each $w$,
$\delta(y)\in F_\eta(w)$ on $\{p_w>0\}$ outside a $\mu$-null set.  There are
only finitely many worlds, so the union of these exceptional sets is still
$\mu$-null.  At every remaining $y$, the chosen action belongs to
$F_\eta(w)$ for every $w\in\mathcal W(y)$, proving necessity of
\eqref{eq:deployment-conflict-exact}.

Conversely, suppose the intersections in
\eqref{eq:deployment-conflict-exact} are nonempty almost everywhere.  The
support pattern $\mathcal W(y)$ takes only finitely many values.  For every
nonempty pattern $S\subseteq\mathcal W$ that occurs off the exceptional set,
choose one action
$a_S\in\bigcap_{w\in S}F_\eta(w)$ and set
$\delta(y)=a_{\mathcal W(y)}$.  Each set
$\{y:\mathcal W(y)=S\}$ is measurable because the densities $p_w$ are
measurable, so $\delta$ is measurable.  For a fixed world $w$, whenever
$p_w(y)>0$ one has $w\in\mathcal W(y)$ and hence
$\delta(y)\in F_\eta(w)$, outside the common null set.  This proves
sufficiency.

For the binary bound, put
\[
 e_0=P_0\{\delta(Y)=a_1\},
 \qquad
 e_1=P_1\{\delta(Y)=a_0\}.
\]
Le Cam's two-point testing identity gives
\[
 e_0+e_1\ge1-\operatorname{TV}(P_0,P_1)=:c.
\]
The expected architecture regrets in the two worlds are respectively
$\Delta_0e_0$ and $\Delta_1e_1$.  If their maximum is $M$, then
$e_0\le M/\Delta_0$ and $e_1\le M/\Delta_1$.  Therefore
\[
 c\le M\left(\frac1{\Delta_0}+\frac1{\Delta_1}\right),
\]
which rearranges to \eqref{eq:deployment-conflict-regret}.  With equal gaps,
the likelihood-ratio test minimizes $(e_0+e_1)/2$ and attains
$\{1-\operatorname{TV}(P_0,P_1)\}/2$, proving the stated equal-prior Bayes
claim.  The maximum-regret statement used only the preceding lower bound and
does not require the two errors to be equalizable.

For the corollary, changing only the optimizer preserves $P_w$,
$F_\eta(w)$, the gaps, and total variation.  Information repair changes the
experiment laws and can therefore change total variation.  Architecture
repair changes the action space and feasible-set intersections.  The frozen
empirical requirements are an identification contract keeping these three
interventions from being conflated, not an additional probabilistic
inequality.
\end{proof}

\subsection*{Proof of the active-set conflict and repair-depth family}
\proofdependency{The dictionary has full column rank, so $H=D^\top D$ is
positive definite and the constrained oracle is unique.  The lower bound is
specific to the one-pass class $[Wx+b]_+$ and to two symmetric balls in one
declared perturbation subspace on which a shared oracle coordinate is strictly
positive.  The repair bound charges each proximal application and uses the
fixed step $1/L$.}
\begin{proof}
We first prove the certified-cell statement.  At a strict KKT point with
support $S$, stationarity on the active coordinates gives
\[
 H_Sz_S^\star-D_S^\top x+\lambda\mathbf1=0,
\]
and hence the formula in \eqref{eq:strict-cell-objects}.  For a perturbation
$u\in V$, consider the candidate
\[
 \widetilde z_S=z_S^\star(x)+Q_Su,
 \qquad \widetilde z_{S^c}=0.
\]
It still satisfies active stationarity at $x+u$.  For every $j\in S$,
\[
 \widetilde z_j
 \ge z_j^\star(x)-\|q_{S,j}^\top\Pi_V\|_2\|u\|_2>0
\]
whenever $\|u\|_2\le r_{S,V}(x)$; the factor $1/2$ in
\eqref{eq:strict-cell-radius} makes the inequality strict.  For
$\ell\notin S$, the inactive KKT slack becomes
\[
 \widetilde s_\ell
 =s_\ell(x)+D_\ell^\top(P_S-I)u
 \ge s_\ell(x)-\|D_\ell^\top(P_S-I)\Pi_V\|_2\|u\|_2>0.
\]
The zero-denominator convention covers a slack or active coordinate that is
unchanged by $u$.  Thus the candidate satisfies the KKT conditions with the
same strict support.  Strict convexity makes it the unique oracle throughout
the closed restricted ball.  The conclusion remains valid for every smaller
positive radius.

Now fix a shared coordinate $j$ and abbreviate
$\Gamma=\Gamma_{12,j,V}$.  Suppose, for contradiction, that some
$f(x)=[Wx+b]_+$ has uniform code error $e<\Gamma$ on $B_1\cup B_2$.
On $B_i$ the oracle coordinate obeys
\[
 z_j^\star(x)
 \ge z_j^\star(x_i)-r_i\|q_i^\top\Pi_V\|_2=\rho_i.
\]
Because $e<\rho_i$, the $j$th output of $f$ is strictly positive throughout
both balls.  It is therefore the same affine function $w^\top x+b_j$ on
both balls, where $w^\top$ is row $j$ of $W$.

On $B_i$, the oracle coordinate is another affine function with row $q_i$.
For any unit vector $v\in V$, evaluate their difference at
$x_i+r_iv$ and $x_i-r_iv$.  Both absolute errors are at most $e$, so their
difference gives
\[
 2r_i|(w-q_i)^\top v|\le2e.
\]
Taking the supremum over unit $v\in V$ yields
$\|(w-q_i)^\top\Pi_V\|_2\le e/r_i$.  The triangle inequality now implies
\[
 \|(q_1-q_2)^\top\Pi_V\|_2
 \le e(r_1^{-1}+r_2^{-1}),
\]
or $e\ge h\|(q_1-q_2)^\top\Pi_V\|_2$, contradicting $e<\Gamma$.  Hence
\eqref{eq:one-pass-code-floor} holds, including the vacuous case
$\Gamma=0$.

The function $z\mapsto J_x(z)+\iota_{\R_+^k}(z)$ is
$\kappa$-strongly convex.  At its minimizer, the subgradient inequality gives
\[
 J_x(z)-J_x(z^\star(x))
 \ge\frac\kappa2\|z-z^\star(x)\|_2^2
 \qquad(z\ge0).
\]
Combining this pointwise inequality with the code-error floor proves
\eqref{eq:one-pass-native-floor}.

For the repair result, write
\[
 T_x(z)=\left[(I-H/L)z+D^\top x/L-\lambda\mathbf1/L\right]_+.
\]
The KKT conditions are equivalent to the fixed-point relation
$T_x(z^\star(x))=z^\star(x)$.  Coordinatewise projection onto the
nonnegative orthant is nonexpansive, while the spectrum of $H$ lies in
$[\kappa,L]$.  Therefore
\begin{align*}
 \|T_x(z)-z^\star(x)\|_2
 &\le\|(I-H/L)(z-z^\star(x))\|_2\\
 &\le(1-\kappa/L)\|z-z^\star(x)\|_2.
\end{align*}
Iteration proves \eqref{eq:proximal-code-contraction}.

Let $e_t=f_t(x)-z^\star(x)$ and let
$s(x)=Hz^\star(x)-D^\top x+\lambda\mathbf1$ be the KKT slack vector.  Direct
expansion of the quadratic objective gives
\[
 J_x(f_t(x))-J_x(z^\star(x))
 =\frac12e_t^\top He_t+s(x)^\top e_t.
\]
The slack is zero on active coordinates; on inactive coordinates
$z^\star=0$, $f_t\ge0$, and hence $s^\top e_t\ge0$.  Cauchy--Schwarz and the
largest eigenvalue bound give
\[
 J_x(f_t(x))-J_x(z^\star(x))
 \le\frac L2\|e_t\|_2^2+\|s(x)\|_2\|e_t\|_2.
\]
Taking suprema and using $\|e_t\|_2\le q^tE_0$ proves
\eqref{eq:proximal-native-envelope}.  The positive root of
$B_Kd+Ld^2/2=\eta$ is exactly $d_\eta$.  Thus
$q^tE_0\le d_\eta$ is sufficient.  If $E_0\le d_\eta$, depth zero already
works.  If $E_0>d_\eta$ and $q=0$, the contraction inequality makes the first
repaired iterate exact.  If $E_0>d_\eta$ and $0<q<1$, solving the same
inequality for integer $t$ gives the third branch of
\eqref{eq:repair-crossing-budget}.  Compactness and continuity of the unique
strongly convex solution map make $z^\star$ and $B_K$ bounded; boundedness of
the declared initializer makes $E_0$ finite.

Finally, suppose a KKT point is interior and a sufficiently small error is
parallel to a unit eigenvector $v$ of $H$ with eigenvalue $\kappa$, with the
entire segment remaining inside the positive orthant.  Projection is then
inactive and
\[
 T_x(z^\star+av)-z^\star=(I-H/L)av=qav.
\]
Hence no smaller uniform contraction coefficient is valid for this fixed
mechanism.  For the corollary, the initializer is explicitly restricted to
$\mathcal F_1$.  Applying the lower bound to
$B_1\cup B_2\subseteq K$ rules out that initializer class, while the upper
envelope at the well-defined depth $t_\eta$ certifies its recurrent repair.
\end{proof}

%% file: chapters/ch20_certificate_statistics.tex
\chapter{Certificate Statistics and Resolution Complexity}
\label{ch:certificate-statistics}
\section{The certificate record and four nonexchangeable objects}
\index{certificate statistics}
\index{certificate!auditable record}

An estimate or posterior alone does not record which population worlds remain compatible with the data, which decision boundary is being audited, or why the answer remains unresolved.  Certificate statistics treats the primary finite-information output as a typed record.

\begin{sourcebox}[title={Companion framework and proof provenance}]
The confidence-world record, resolution complexity, evidence-slack criterion,
and recursive quotient developed in this chapter are organized in the
Certificate Statistics companion manuscript \citep{CertificateStatistics}.
The chapter appendices give the complete proofs counted by this book; the
source citation records provenance rather than replacing those proofs.
\end{sourcebox}

A certificate record contains:
\begin{itemize}
  \item a declared query contract and operational boundary $B$;
  \item a confidence world $C_n$ with an explicit coverage contract;
  \item the identified image $J_{C_n}(q)$ and current certificate color;
  \item the certificate margin or tolerance profile when the query is ordered;
  \item evidence or witnesses supporting the color;
  \item a resolution profile indicating what information is still needed.
\end{itemize}

The record connects four objects without identifying their validity
contracts.

\index{certificate!population truth}
\index{certificate!confidence certificate}
\index{posterior credibility}
\index{deployment action}

\begin{center}
\small
\begin{tabularx}{0.98\textwidth}{p{0.25\textwidth}p{0.27\textwidth}X}
\toprule
Object & Evaluated on & What it can guarantee\\
\midrule
population truth $q(w)$
  & one fixed population world $w$
  & the task cell that is true at the population level\\
confidence certificate $\delta_{C_n}^q$
  & a covered confidence world $C_n$
  & worldwise error control for every resolved finite-data declaration\\
posterior color $\chi(b_n)$
  & a prior-relative belief state $b_n$
  & a summary of current posterior belief, not uniform validity over worlds\\
deployment action $a_n$
  & observer-visible information and the declared deployment grammar
  & operational loss control only when one admissible action or witness works
    across the required worlds\\
\bottomrule
\end{tabularx}
\end{center}

The four objects can constrain one another, but agreement of their displayed
labels does not make them interchangeable.  A resolved confidence certificate
is not yet a common deployment witness, and high posterior probability is not
by itself a worldwise certificate.

Reject-option classification and three-way decision theory are important
neighbors \citep{Chow1970,Yao2010}.  They optimize or summarize actions under
a probabilistic or loss-based contract.  Proposition~\ref{prop:posterior-not-worldwise}
isolates a different question: whether one terminal label is correct
simultaneously for every world retained by a confidence set.  The distinction,
not abstention itself, is the point.

\begin{conditionbox}
A prior and posterior are well defined for the positive statement.  For the separation, the model permits two observationally indistinguishable worlds with different truth labels.  No amount of repeated observation can identify those labels under the available experiment.
\end{conditionbox}

\begin{proposition}[Posterior credibility is not worldwise validity]
\label{prop:posterior-not-worldwise}
For any fixed posterior credibility level below one, there exist a prior and two observationally indistinguishable worlds with opposite certificate colors such that the posterior assigns high probability to one color at every time, while the minority world is reported incorrectly with probability one.  Under the available experiment family the corresponding worldwise resolution complexity is infinite.
\end{proposition}

\begin{interpretationbox}
Small posterior error is a prior-average statement.  It does not imply a uniform frequentist guarantee for every possible world: a low-prior but observationally indistinguishable world can be reported incorrectly with probability one.  This does not invalidate Bayesian credibility; it identifies a different validity contract.
\end{interpretationbox}

\begin{proofroadmap}
Use iterated expectation for the prior-average guarantee.  For the counterexample, place small prior mass on one of two observationally identical worlds with opposite labels; the posterior never moves, so the majority report remains credible but fails pointwise in the minority world.  The chapter appendix gives the construction.
\end{proofroadmap}

\subsection{Truth-map notation for the standard certificate}
\label{sec:truth-map-three-way}
\index{certificate!truth map}
\index{certificate!three-way rule}
\index{worldwise validity}

The preceding proposition warns against reading posterior mass as an honest
architecture certificate.  Chapter~\ref{ch:statistical-eg} defined the
standard confidence certificate for an arbitrary truth map.  Here we record
the compact notation needed by the resolution theory.  For a confidence
world $C_n$, write
\[
  c(C_n)=\{c(w):w\in C_n\}\subseteq\{F,I\},
\]
and set
\[
  \delta_{C_n}^{c}=
  \begin{cases}
  M,&C_n=\varnothing,\\
  d,&c(C_n)=\{d\}\text{ for }d\in\{F,I\},\\
  U,&|c(C_n)|>1.
  \end{cases}
\]

\begin{conditionbox}
The nonempty confidence world has uniform coverage, and a declared truth map
is evaluated on that world.  An empty set produces the model-conflict state
$M$, not a population declaration.  Simultaneous post-selection validity
requires one confidence event for the entire query class, not separately
chosen pointwise intervals.
\end{conditionbox}

\begin{corollary}[Truth-map form of three-way honesty]
\label{cor:certificate-honesty}
If $C_n$ covers the true world with probability at least $1-\alpha$, every
resolved output of $\delta_{C_n}^{c}$ is honest at level $1-\alpha$.  On a realized
nonempty confidence world, no other rule based on the same world can make
more declarations while preserving worldwise correctness.
\end{corollary}

\begin{interpretationbox}
This is Theorem~\ref{thm:three-way} written in the notation used below.
Unresolvedness means that the truth image is not a singleton; model conflict
means that no compatible world survived the declared model and confidence
construction.  The two states must not be conflated.
\end{interpretationbox}

\begin{proofroadmap}
Apply Theorem~\ref{thm:three-way} to the two cells of the truth partition.
One simultaneous coverage event gives the post-selection extension; the
chapter appendix records this reduction.
\end{proofroadmap}

\section{Resolution complexity}
\index{certificate!resolution complexity}
\index{characteristic information}
\index{characteristic time}
\index{confidence sequence}
\index{optional stopping}

Confidence sequences make the same statement valid under adaptive querying and
stopping.  Data-dependent hypothesis sets and random-set generalization bounds
address a related selection problem \citep{DupuisEtAl2024}; the object here is
different because the set represents covered population worlds rather than a
learned hypothesis class.  A current unresolved state contains compatible
worlds of opposite certificate colors.  The relevant information is therefore
not global parameter information but separation from the nearest world that
would reverse the declared conclusion.

Let $\Theta$ be a finite world set, let $q:\Theta\to\mathsf Q$ be the truth partition, and let $\mathcal E$ be a finite experiment set.  Experiment $e$ has law $P_{\theta,e}$ in world $\theta$.  Define
\begin{align}
 d_e(\theta,\lambda)
 &=\KL(P_{\theta,e}\|P_{\lambda,e}),
 &
 \operatorname{Alt}_q(\theta)
 &=\{\lambda:q(\lambda)\ne q(\theta)\},
 \label{eq:book-certificate-pairwise-kl}\\
 \mathcal I_X^{q,*}(\theta)
 &=\max_{w\in\Delta(\mathcal E)}
   \min_{\lambda\in\operatorname{Alt}_q(\theta)}
   \sum_{e\in\mathcal E}w_e d_e(\theta,\lambda),
 &
 T_X^{q,*}(\theta)
 &=\{\mathcal I_X^{q,*}(\theta)\}^{-1}.
 \label{eq:book-certificate-characteristic}
\end{align}
We use the convention $1/0=+\infty$.  The allocation $w$ is an experimental frequency, whereas the inner minimum is chosen by the least separated opposite-label world.

A sequential rule $(\tau,\widehat q)$ is \emph{$\alpha$-worldwise valid} when
\begin{equation}
 \sup_{\theta\in\Theta}
 \Pp_\theta\{\widehat q\ne q(\theta),\ \tau<\infty\}
 \le \alpha.
 \label{eq:book-worldwise-valid}
\end{equation}

\begin{conditionbox}
The experiment and world sets are finite; adaptive experiment choices are nonanticipating; the pairwise KL divergences are defined on common measurable spaces; and the standard sequential change-of-measure identity applies.  At the world under audit, $\E_\theta\tau<\infty$.  Pairwise identifiability is not assumed: if every allocation leaves an opposite label at zero KL rate, the theorem correctly returns an infinite lower bound.
\end{conditionbox}

\begin{theorem}[Worldwise resolution lower bound]
\label{thm:certificate-resolution-lower}
For $0<\alpha<1/2$, every $\alpha$-worldwise-valid rule with $\E_\theta\tau<\infty$ satisfies
\begin{equation}
 \boxed{
 \E_\theta\tau
 \ge
 T_X^{q,*}(\theta)\operatorname{kl}(1-\alpha,\alpha).}
 \label{eq:book-resolution-lower}
\end{equation}
Consequently,
\[
 \liminf_{\alpha\downarrow0}
 \frac{\E_\theta\tau}{\log(1/\alpha)}
 \ge T_X^{q,*}(\theta).
\]
\end{theorem}

\begin{interpretationbox}
The certificate logic determines the alternative set, and that set determines the sampling geometry.  The bound concerns expected observations needed for a worldwise terminal statement; it is not a generalization bound, posterior entropy, or deployment loss.  If the characteristic information is zero, additional observations through the same experiment family cannot produce a finite-time worldwise certificate.
\end{interpretationbox}

\begin{proofroadmap}
Compare the stopped transcript under $\theta$ with each opposite-label world $\lambda$.  Sequential change of measure lower-bounds the accumulated expected KL by the binary information needed to separate correct and incorrect terminal events.  Normalized expected experiment counts form one allocation in $\Delta(\mathcal E)$; the max--min definition then supplies the sharp information rate.  The chapter appendix carries out these steps.
\end{proofroadmap}

The max--min geometry is the established partition-identification lower-bound
mechanism \citep{KaufmannEtAl2016,GarivierKaufmann2016,
DegenneKoolen2019}.  Certificate statistics contributes the typed truth
partition and connects this rate to evidence and recursive carriers; it does
not rename generic pure exploration as a new theorem.

\subsection{AND/OR phase change}
\index{certificate!AND/OR phase change}

In a Gaussian multichannel threshold problem, feasibility may require verifying that every channel satisfies a constraint, while impossibility can be proved by one violated channel.  The optimal information geometry changes accordingly.

\begin{itemize}
  \item To certify feasibility, allocation must cover all near-boundary channels: an AND geometry.
  \item To certify impossibility, allocation may concentrate on one strongest obstruction: an OR geometry.
\end{itemize}

For example, sampling channel $i$ from $N(\mu_i,\sigma^2)$ and declaring feasibility when $\max_i\mu_i\le\eps$ gives, away from the boundary,
\begin{equation}
 T_X^{q,*}(\mu)
 =
 \begin{cases}
 2\sigma^2\displaystyle\sum_i(\eps-\mu_i)^{-2},
 &\max_i\mu_i<\eps,\\[0.8em]
 \displaystyle\frac{2\sigma^2}
 {\max_{i:\mu_i>\eps}(\mu_i-\eps)^2},
 &\max_i\mu_i>\eps.
 \end{cases}
 \label{eq:book-and-or-time}
\end{equation}
On the feasible side the optimal allocation equalizes the coordinate-wise KL rates.  On the impossible side it concentrates on a strongest violation.  Thus the asymmetry is intrinsic to the certificate, not to a particular algorithm.  We retain this closed form as a worked interpretation; the general exponential-family and partition-identification setting belongs to the pure-exploration literature \citep{KaufmannKoolenGarivier2018,JunejaKrishnasamy2019}.

\section{Evidence carriers and task-relative sufficiency}
\label{sec:certificate-evidence-carriers}
\index{carrier!evidence}
\index{sufficiency!task-relative}
\index{KL divergence!information loss}

For each experiment $e$, let $X_e$ denote the raw observation and let a parameter-independent Markov kernel produce retained evidence $Z_e=T_e(X_e)$.  Write $Q_{\theta,e}$ for the retained law and define
\begin{align}
 d_e^X(\theta,\lambda)
 &=\KL(P_{\theta,e}\|P_{\lambda,e}),
 &
 d_e^Z(\theta,\lambda)
 &=\KL(Q_{\theta,e}\|Q_{\lambda,e}),\\
 \ell_e(\theta,\lambda)
 &=\E_{Q_{\theta,e}}
   \KL\{P_{\theta,e}(\cdot\mid Z_e)
          \|P_{\lambda,e}(\cdot\mid Z_e)\}.
 \label{eq:book-evidence-loss}
\end{align}
The conditional KL chain rule gives the exact pairwise KL decomposition
\begin{equation}
 d_e^X(\theta,\lambda)
 =d_e^Z(\theta,\lambda)+\ell_e(\theta,\lambda).
 \label{eq:book-evidence-kl-decomposition}
\end{equation}
Consequently, if $\mathcal I_Z^{q,*}$ is defined from retained KL rates,
\begin{equation}
 \mathcal I_Z^{q,*}(\theta)\le\mathcal I_X^{q,*}(\theta),
 \qquad
 T_Z^{q,*}(\theta)\ge T_X^{q,*}(\theta).
 \label{eq:book-certificate-dpi}
\end{equation}

Pairwise KL loss is not yet the right sufficiency test.  Certificate sufficiency need not preserve every likelihood ratio; it need only avoid lowering the max--min floor.  At a fixed world $\theta$ and allocation $w$, put
\begin{align}
 G_X(w,\lambda)
 &=\sum_e w_e d_e^X(\theta,\lambda),
 &
 L(w,\lambda)
 &=\sum_e w_e\ell_e(\theta,\lambda),\\
 s_X(w,\lambda)
 &=G_X(w,\lambda)-\mathcal I_X^{q,*}(\theta).
 \label{eq:book-evidence-slack}
\end{align}
At a raw-optimal allocation, $s_X$ measures how far an alternative lies above the binding certificate floor.

\begin{conditionbox}
The raw and retained characteristic-information maxima are attained, the evidence kernel is the same in every world, and the displayed KL quantities are finite.  Attainment is needed only to choose an optimizer in the converse direction; an explicit allocation satisfying the inequality below is sufficient without a retained optimizer.
\end{conditionbox}

\begin{theorem}[Exact certificate preservation by slack]
\label{thm:certificate-evidence-slack}
The retained carrier preserves the raw characteristic information at $\theta$,
\[
 \mathcal I_Z^{q,*}(\theta)=\mathcal I_X^{q,*}(\theta),
\]
if and only if there is an allocation $w^\star\in\Delta(\mathcal E)$ such that
\begin{equation}
 \boxed{
 L(w^\star,\lambda)
 \le s_X(w^\star,\lambda)
 \quad
 \text{for every }\lambda\in\operatorname{Alt}_q(\theta).}
 \label{eq:book-evidence-slack-criterion}
\end{equation}
Every such $w^\star$ is raw-optimal.  In particular, every raw binding alternative at $w^\star$ has zero weighted KL loss.
\end{theorem}

\begin{interpretationbox}
Compression may discard nuisance information when nonbinding alternatives have enough excess separation to absorb the loss.  It may not discard any likelihood direction belonging to a binding opposite-label world.  This is weaker than classical parametric sufficiency and stronger than preserving only a posterior mean or present decision.
\end{interpretationbox}

\begin{proofroadmap}
Sum the chain-rule decomposition over experiments to obtain $G_Z=G_X-L$.  Slack domination keeps every retained alternative above the raw max--min floor, while data processing prevents it from exceeding that floor.  Conversely, a retained-optimal allocation at equality must keep every retained alternative above the common optimum; rearranging the same identity gives the slack inequalities.  The complete argument is in the chapter appendix.
\end{proofroadmap}

Locally, the score map projected to the nuisance-quotiented tangent space gives a task-relative information operator.  Its minimum modulus determines stability; inverse-square modulus controls sample inflation.  This modulus language follows the established analysis of ill-posed and semiparametric inverse problems \citep{BlundellChenKristensen2007,ChenReiss2011}; certificate statistics uses it only after restricting the tangent directions to those that can reverse the declared truth label.

\subsection{Average Bayesian loss versus worldwise complexity}
\index{Bayesian risk!versus worldwise complexity}
\index{worldwise validity}
\index{mutual information!task posterior}

Let $Q=q(\Theta)$ be the task label, let $X$ be the raw record, and let $Z$
be retained evidence generated from $X$.  Write
$\Pi_X^Q=\mathcal L(Q\mid X)$ and
$\Pi_Z^Q=\mathcal L(Q\mid Z)$.  Under the Markov relation
$Q\to X\to Z$, the standard conditional-information identity gives the
exact prior-predictive identity
\begin{equation}
  \E\KL(\Pi_X^Q\|\Pi_Z^Q)
  = I(Q;X\mid Z).
  \label{eq:book-posterior-identity}
\end{equation}
The left side measures how much the retained representation distorts the
full-data task posterior on average under the chosen prior.  It is therefore
a legitimate loss scale for Bayesian representation fidelity and expected
experimental utility \citep{CoverThomas2006,Lindley1956}.

It is not a worldwise certificate loss scale.  For a sharp separation, take two
worlds with
$X\mid\theta_0\sim\operatorname{Bernoulli}(1/4)$ and
$X\mid\theta_1\sim\operatorname{Bernoulli}(3/4)$, let $Z$ be constant, and
assign prior mass $r$ to $\theta_1$.  The posterior loss in
\eqref{eq:book-posterior-identity} is at most the binary entropy $h_2(r)$ and
therefore tends to zero as $r\downarrow0$.  Yet the retained laws are
identical, so $\mathcal I_Z^{q,*}=0$ and the worldwise characteristic time is
infinite, while the raw pair remains distinguishable.

\begin{boundarybox}
The Bayesian identity averages over worlds selected by a prior.  The certificate criterion
minimizes over opposite-label worlds and maximizes over experiment
allocations.  Both use KL, but their quantifiers make them nonexchangeable:
small posterior distortion neither authorizes stopping nor bounds worldwise
sample inflation.
\end{boundarybox}

\begin{table}[H]
\centering
\small
\caption{Three statistical objects that must not be conflated.}
\begin{tabularx}{\textwidth}{p{0.22\textwidth}p{0.30\textwidth}X}
\toprule
Object & Typical form & Quantifier, guarantee, and proper use\\
\midrule
Bayesian belief
  & posterior, Bayes risk, expected task information
  & Prior-average; evaluates expected task value and helps select the next experiment.\\
Certificate evidence
  & worldwise likelihood, max--min KL, characteristic time
  & Worst binding opposite-label world under fixed confidence; authorizes stopping and a terminal claim.\\
Recursive state
  & colored belief quotient, labeled update kernels, $N_h$
  & Closed under every future observation context; determines which states may be safely merged.\\
\bottomrule
\end{tabularx}
\end{table}

Posterior belief can guide design, worldwise evidence authorizes a
certificate, and recursive state preserves what future design will need.

\section{Recursive certificate posterior quotient}
\label{sec:certificate-recursive-quotient}
\index{carrier!recursive}
\index{belief state!recursive quotient}
\index{bisimulation!belief state}

The current posterior probability of certificate colors may be insufficient to choose the next experiment.  Two beliefs with the same current color probabilities can imply different optimal future measurements.

Assume now that hypotheses, experiments, observation alphabets, and the reachable belief set $\mathcal R\subseteq\Delta(\Theta)$ are finite.  The set $\mathcal R$ is closed under every positive-probability Bayes update.  Let $\chi:\mathcal R\to\mathsf C$ be the declared contract color.  It may record a present certificate permission, observer gate, or vector of immediate costs; fields omitted from $\chi$ are not protected by the quotient.

For experiment $e$ and observation $y$, define
\begin{align}
 p_e(y\mid b)
 &=\sum_{\theta}b(\theta)P_{\theta,e}(y),\\
 B_{e,y}b(\theta)
 &=\frac{b(\theta)P_{\theta,e}(y)}{p_e(y\mid b)},
 \label{eq:book-bayes-update}\\
 K_e\bigl(b;\{y\}\times A\bigr)
 &=p_e(y\mid b)\ind\{B_{e,y}b\in A\}.
 \label{eq:book-labelled-belief-kernel}
\end{align}
The label $y$ is retained: two states are not equivalent merely because they have the same unlabeled aggregate transition.

Start with $b\sim_0 b'$ when $\chi(b)=\chi(b')$.  Given $\sim_h$, define $b\sim_{h+1}b'$ when $b\sim_hb'$ and
\begin{equation}
 K_e(b;\{y\}\times A)
 =K_e(b';\{y\}\times A)
 \label{eq:book-recursive-refinement}
\end{equation}
for every experiment $e$, observation $y$, and $\sim_h$-class $A$.  Put
\[
 \sim_\infty=\bigcap_{h\ge0}\sim_h,
 \qquad
 N_h=|\mathcal R/\!\sim_h|.
\]

A recursive carrier $\phi:\mathcal R\to\mathsf S$ is \emph{exact} if the color factors through $\phi$ and, for every experiment, the joint law of the observation label and successor carrier state depends on $b$ only through $\phi(b)$.

\begin{conditionbox}
The model, observation alphabets, and reachable belief set are finite, and zero-probability Bayes branches are omitted.  Exactness preserves labeled observations and successor states for every declared experiment.  The result does not claim that the full posterior is minimal, nor that a color protecting only the present terminal action is sufficient for future acquisition.
\end{conditionbox}

\begin{theorem}[Recursive quotient and minimal exact state]
\label{thm:certificate-recursive-state}
The relations $\sim_h$ are decreasing equivalence relations and
\[
 N_0\le N_1\le\cdots\le N_\infty.
\]
If $b\sim_hb'$, every fixed observation-history policy using at most $h$ further observations induces the same law of observation labels and contract colors from $b$ and $b'$.  The limit $\sim_\infty$ is the largest color-respecting equivalence relation stable under every labeled kernel $K_e$.

The quotient map $b\mapsto[b]_\infty$ is exact, and every exact recursive carrier refines it:
\begin{equation}
 \phi(b)=\phi(b')
 \quad\Longrightarrow\quad
 b\sim_\infty b'.
 \label{eq:book-recursive-universal}
\end{equation}
Thus every exact deterministic implementation has at least $N_\infty$ reachable states.  If each of $L$ charts distinguishes at most $K$ states, then
\begin{equation}
 L\ge\left\lceil\frac{N_\infty}{K}\right\rceil.
 \label{eq:book-recursive-chart-bound}
\end{equation}
Signature refinement terminates after at most $N_\infty-N_0$ strict block increases, and every strict split returns an experiment, observation, and successor block witnessing the missing distinction.
\end{theorem}

The count $N_\infty$ concerns only the recursive belief carrier specified in
the theorem.  A concrete GLR test, deficit tracker, or D-tracking
implementation may keep likelihood sums, visit counts, or time indices in
separate and potentially unbounded auxiliary memory.  Those counters are not
included in $N_\infty$ unless the declared carrier grammar explicitly folds
them into the protected recursive state.

\begin{interpretationbox}
$N_0$ is the state capacity needed to preserve the present contract color.  $N_1$ also preserves which labeled observation can expose the next relevant distinction, and $N_\infty$ preserves every finite future acquisition context.  A recurrent state that is adequate now can therefore be structurally inadequate one observation later.
\end{interpretationbox}

\begin{proofroadmap}
Refinement signatures give decreasing equivalence relations.  Induction on the remaining horizon proves policy invariance.  Any stable color-respecting relation survives every refinement and is therefore contained in the limit.  Conversely, finiteness makes the refinement stabilize, so the limit itself defines a reduced labeled kernel.  The kernel relation induced by any exact carrier is stable and color-respecting, forcing it to refine the limit quotient; counting yields the state and chart bounds.
\end{proofroadmap}

This construction specializes strong belief bisimulation
\citep{CastroPanangadenPrecup2009,LarsenSkou1991} and realizes the finite
quotient by color-respecting partition refinement, connecting model
minimization and classical refinement algorithms
\citep{GivanEtAl2003,PaigeTarjan1987}.  Its role here is architectural: the
quotient cardinality is a task-relative recurrent capacity, and a split
witness tells a repair procedure which state distinction must be installed.

\subsection{CertTrack and certificate-aware design}
\index{CertTrack}
\index{certificate!adaptive design}
\index{Bayesian design!versus worldwise stopping}
\index{stopping rule!worldwise authorization}

A certificate-aware tracker assigns two different jobs to two different
loss scales.  At current belief $b$, the Bayesian task gain from experiment
$e$ is
\[
  G_e^q(b)=I_b\{q(\Theta);Y_e\}.
\]
This prior-average quantity may decide which informative experiment to run.
It cannot by itself decide when every opposite-color world has been excluded.
For retained evidence, define the $\rho$-optimal worldwise face
\begin{equation}
  \mathcal W_\rho^Z(\theta)
  =\left\{w\in\Delta(\mathcal E):
    \min_{\lambda\in\operatorname{Alt}_q(\theta)}
    \sum_e w_e d_e^Z(\theta,\lambda)
    \ge(1-\rho)\mathcal I_Z^{q,*}(\theta)
  \right\}.
  \label{eq:book-certificate-face}
\end{equation}
A posterior-guided design may maximize $\sum_e w_eG_e^q(b)$ inside this
face.  At $\rho=0$ it only breaks ties among certificate-optimal allocations;
with $\rho_t\downarrow0$ it may trade a vanishing fraction of the leading
worldwise rate for prior-relevant information.

Stopping remains separate.  It occurs only when a simultaneous confidence
world has one truth color, or when an anytime-valid likelihood-ratio process
crosses its declared boundary.  Such validity holds under any nonanticipating
design; it does not rely on posterior odds.  Forced exploration and deficit
tracking can then make empirical allocations follow the selected face
\citep{GarivierKaufmann2016,KaufmannKoolenGarivier2018}.

\begin{boundarybox}
Bayesian task information chooses where to look; a confidence sequence or
likelihood-ratio martingale authorizes what may be said.  Adding the two
objectives would invent an exchange rate between prior-average utility and
least-favorable-world progress.
\end{boundarybox}

Under finite-model regularity, the resulting tracker attains the optimal
almost-sure leading constant.  Expected-time equality additionally requires
uniform integrability; no posterior-odds optional-stopping argument is being
used.

\subsection{A compressed \texttt{UCBAdmissions} audit}
\label{sec:book-ucb-audit}
\index{UCBAdmissions audit}
\index{aggregation!department audit}
\index{certificate!tolerance profile}

The R \texttt{UCBAdmissions} table records 4526 applicants to six large UC
Berkeley departments in 1973, classified by department, sex, and admission
outcome \citep{RDataUCB}.  It is a standard example in which aggregation
changes the apparent association \citep{BickelEtAl1975}.  Pooling departments
gives 1198 admissions among 2691 male applicants and 557 among 1835 female
applicants: a male-minus-female admission-rate difference of $0.1416$.  A
two-sided Fisher exact test of that pooled association gives
$p=4.84\times10^{-22}$.

That calculation does not answer every department-level question.  As a
worked certificate, declare
\begin{equation}
  d_j=p_{j,M}-p_{j,F},
  \qquad
  \Gamma=\max_{j\in\{A,\ldots,F\}}d_j,
  \qquad
  \Gamma\le0.10.
  \label{eq:book-ucb-query}
\end{equation}
The ten-percentage-point tolerance is illustrative, not an ethical or causal
criterion.  Construct two-sided exact Clopper--Pearson intervals for the 12
department--sex cell probabilities at marginal error $0.05/12$, and project
their Bonferroni rectangle onto each difference $d_j$.  This gives a 95\%
simultaneous confidence world.

\begin{table}[htbp]
\centering
\caption{Department admission-rate differences and 95\% simultaneous exact
intervals, in percentage points.}
\label{tab:book-ucb-audit}
\begin{tabular}{crrrr}
\toprule
Dept. & $n_M$ & $n_F$ & Estimate & Simultaneous interval\\
\midrule
A & 825 & 108 & $-20.35$ & $[-34.32,-2.96]$\\
B & 560 &  25 & $ -4.96$ & $[-33.14,31.09]$\\
C & 325 & 593 & $  2.86$ & $[-10.47,16.35]$\\
D & 417 & 375 & $ -1.84$ & $[-15.67,11.99]$\\
E & 191 & 393 & $  3.83$ & $[-11.57,19.83]$\\
F & 373 & 341 & $ -1.14$ & $[ -8.95, 6.58]$\\
\bottomrule
\end{tabular}
\end{table}

Every department point estimate is below ten points, but the simultaneous
intervals for B--E cross that boundary, and no lower endpoint exceeds it.
The honest certificate for \eqref{eq:book-ucb-query} is therefore
$U$.  Department B is the resolution bottleneck: its upper excursion is
largest, driven by only 25 female applicants.

The same confidence world gives the entire tolerance profile.  Because it is
rectangular, the identified image of $\Gamma$ has endpoints
\[
  L_\Gamma=\max_jL_j=-0.0895,
  \qquad
  U_\Gamma=\max_jU_j=0.3109,
\]
to the displayed precision.  Hence, for the query $\Gamma\le\eps$,
\begin{equation}
  \delta(\eps)=
  \begin{cases}
    I,&\eps<-0.0895,\\
    U,&-0.0895\le\eps<0.3109,\\
    F,&\eps\ge0.3109,
  \end{cases}
  \label{eq:book-ucb-profile}
\end{equation}
up to endpoint rounding.  Thus a pooled $p$-value near $10^{-22}$ coexists
with unresolved department tolerances across a forty-point range.

\begin{center}
\begin{tabularx}{0.96\textwidth}{>{\raggedright\arraybackslash}p{0.27\textwidth}
                                  >{\raggedright\arraybackslash}X}
\toprule
readout & authorized action\\
\midrule
pooled association & report the pooled disparity for that declared query;
do not transport it into a department-level verdict\\
department certificate & return $U$, withhold a binary department verdict,
and direct further resolution effort toward the widest binding interval\\
\bottomrule
\end{tabularx}
\end{center}

All displayed quantities are deterministic functions of the public
contingency table.  No regression model, asymptotic normal approximation, or
prior enters this audit.  A more structured model may narrow the confidence
world, but it would constitute an additional declared assumption rather than
a reinterpretation of the same certificate.

\begin{interpretationbox}
The pooled test and the department certificate are not contradictory: they
have different truth maps.  The certificate retains the pooled association,
withholds an unsupported department-level verdict, and directs new
information toward the bottleneck beginning with Department B female
applicants.  The table alone does not identify causal discrimination,
qualifications, selection, or historical mechanisms; those require a larger
world and additional assumptions.
\end{interpretationbox}

\section*{Exercises}

\begin{exercise}
Construct a confidence world containing both feasible and impossible parameter values.  Show that every honest binary rule must abstain or make an error on some compatible world.
\end{exercise}

\begin{exercise}
Compute the max--min information allocation for a two-channel Gaussian OR certificate.
\end{exercise}

\begin{exercise}
Construct a raw-optimal allocation with one binding and one nonbinding alternative.  Determine how much KL the evidence carrier may discard in each direction without increasing characteristic time.
\end{exercise}

\begin{exercise}
Give two posterior distributions with the same current probability of feasibility but different optimal next experiments.  Identify the first refinement signature that separates them.
\end{exercise}

\input{chapter_appendices/ch20_proofs}

%% file: chapter_appendices/ch20_proofs.tex
\chapterproofappendix

\subsection*{Proof that posterior credibility is not worldwise validity}
\proofdependency{The positive statement is an application of iterated expectation.  The negative statement requires no identifiability: two worlds may have identical observation laws but different truth labels.}
\begin{proof}
If
\[
\Pp_{b_0}\{q(\Theta)\ne\widehat q\mid H_\tau\}\le\eta
\quad\text{almost surely},
\]
then taking expectation with respect to the prior-predictive law yields
\[
\Pp_{b_0}\{q(\Theta)\ne\widehat q\}
=
\E_{b_0}
\Pp_{b_0}\{q(\Theta)\ne\widehat q\mid H_\tau\}
\le\eta.
\]
This is a prior-average statement.

For the separation, take two worlds $\theta_F$ and $\theta_I$ with different truth labels and identical laws for every observation under every available experiment.  Assign prior masses $1-\eta$ and $\eta$, respectively.  Since the likelihood ratio is identically one, the posterior never changes.  Reporting the majority label has posterior error $\eta$ at every time and therefore satisfies the credibility criterion.  Conditional on the minority world, however, the report is wrong with probability one.  Thus no nontrivial uniform worldwise error guarantee follows from posterior credibility alone.  Moreover, every experiment has zero KL divergence between the two opposite-label worlds, so the characteristic information for their truth partition is zero and the corresponding worldwise resolution complexity is infinite.
\end{proof}

\subsection*{Proof of the truth-map form of three-way honesty}
\proofdependency{Theorem~\ref{thm:three-way} proves the claim for an arbitrary
truth map on a nonempty confidence world.  The simultaneous-query clause
requires one coverage event for the whole query class; the empty-set output
$M$ is diagnostic and is not a resolved truth claim.}
\begin{proof}
Apply Theorem~\ref{thm:three-way} to the two cells
$c^{-1}(F)$ and $c^{-1}(I)$.  A singleton image $c(C_n)$ is the
agreeing case of that theorem, and a two-label image is its unresolved case;
honesty and maximal decisiveness follow immediately.  If one confidence
world covers the complete population object simultaneously over a query
class, the same reduction holds for every query on that event and hence for
any data-selected query.
\end{proof}

\subsection*{Proof of the worldwise resolution lower bound}
\proofdependency{We use the standard sequential change-of-measure inequality: for a stopped transcript, any event $A$ measurable at the stopping time, and any pair of worlds $\theta,\lambda$,
\[
 \sum_e \E_\theta N_e(\tau)
 \KL(P_{\theta,e}\|P_{\lambda,e})
 \ge
 \operatorname{kl}\{\Pp_\theta(A),\Pp_\lambda(A)\}.
\]
This follows by applying data processing to the likelihood ratio of the stopped experiment.}
\begin{proof}
Fix $\lambda\in\operatorname{Alt}_q(\theta)$ and take
\[
 A_\theta=\{\tau<\infty,\ \widehat q=q(\theta)\}.
\]
Because $\E_\theta\tau<\infty$, the rule stops almost surely under $\theta$.  Worldwise validity therefore gives
\[
 \Pp_\theta(A_\theta)\ge1-\alpha.
\]
Since $q(\lambda)\ne q(\theta)$, the same event is an error under $\lambda$, so
\[
 \Pp_\lambda(A_\theta)\le\alpha.
\]
For $0<\alpha<1/2$, monotonicity of binary relative entropy on these two ranges and sequential change of measure yield
\begin{equation}
 \sum_e \E_\theta N_e(\tau)d_e(\theta,\lambda)
 \ge \operatorname{kl}(1-\alpha,\alpha).
 \label{eq:proof-resolution-change-measure}
\end{equation}

Define the normalized expected counts
\[
 \bar w_e
 =\frac{\E_\theta N_e(\tau)}{\E_\theta\tau}.
\]
They are nonnegative and sum to one.  Divide
\eqref{eq:proof-resolution-change-measure} by $\E_\theta\tau$ and minimize over $\lambda\in\operatorname{Alt}_q(\theta)$:
\[
 \min_{\lambda\in\operatorname{Alt}_q(\theta)}
 \sum_e\bar w_e d_e(\theta,\lambda)
 \ge
 \frac{\operatorname{kl}(1-\alpha,\alpha)}
      {\E_\theta\tau}.
\]
The left side is no larger than its maximum over all allocations, namely $\mathcal I_X^{q,*}(\theta)$.  Rearrangement gives
\[
 \E_\theta\tau
 \ge
 \frac{\operatorname{kl}(1-\alpha,\alpha)}
      {\mathcal I_X^{q,*}(\theta)}
 =T_X^{q,*}(\theta)\operatorname{kl}(1-\alpha,\alpha).
\]
If the characteristic information is zero, the preceding inequality forces $\E_\theta\tau=+\infty$, contradicting the assumed finite expectation; hence no such finite-time rule exists.  Finally,
$\operatorname{kl}(1-\alpha,\alpha)/\log(1/\alpha)\to1$, which proves the asymptotic statement.
\end{proof}

\subsection*{Proof of exact certificate preservation by slack}
\proofdependency{The evidence kernel is parameter-independent, so the KL chain rule for $(X_e,Z_e)$ has raw KL on the left.  The converse uses attainment of the retained max--min problem.}
\begin{proof}
For every allocation $w$ and alternative $\lambda$, summing
\eqref{eq:book-evidence-kl-decomposition} gives
\begin{equation}
 G_Z(w,\lambda)
 =G_X(w,\lambda)-L(w,\lambda).
 \label{eq:proof-evidence-decomposition}
\end{equation}
Because $L\ge0$, every retained alternative rate is no larger than its raw counterpart.  Taking the alternative minimum and allocation maximum proves
$\mathcal I_Z^{q,*}(\theta)\le\mathcal I_X^{q,*}(\theta)$.

Suppose first that $w^\star$ satisfies
\eqref{eq:book-evidence-slack-criterion}.  For every opposite-label $\lambda$,
\begin{align*}
 G_Z(w^\star,\lambda)
 &=G_X(w^\star,\lambda)-L(w^\star,\lambda)\\
 &\ge G_X(w^\star,\lambda)-s_X(w^\star,\lambda)\\
 &=\mathcal I_X^{q,*}(\theta).
\end{align*}
Thus $\mathcal I_Z^{q,*}(\theta)\ge\mathcal I_X^{q,*}(\theta)$.  Data processing supplies the reverse inequality, so equality holds.  Moreover,
\[
 \min_\lambda G_X(w^\star,\lambda)
 \ge
 \min_\lambda G_Z(w^\star,\lambda)
 \ge \mathcal I_X^{q,*}(\theta).
\]
By definition of the raw maximum, equality must hold and $w^\star$ is raw-optimal.

Conversely, assume
$\mathcal I_Z^{q,*}(\theta)=\mathcal I_X^{q,*}(\theta)$ and choose a retained-optimal allocation $w^\star$.  Then every opposite-label alternative obeys
\[
 G_Z(w^\star,\lambda)
 \ge\mathcal I_Z^{q,*}(\theta)
 =\mathcal I_X^{q,*}(\theta).
\]
Rearranging \eqref{eq:proof-evidence-decomposition} gives
\[
 L(w^\star,\lambda)
 \le G_X(w^\star,\lambda)-\mathcal I_X^{q,*}(\theta)
 =s_X(w^\star,\lambda),
\]
which is the required slack domination.  The preceding raw-optimality argument applies again.  Finally, if $\lambda$ is binding in the raw problem at $w^\star$, then $s_X(w^\star,\lambda)=0$.  Since $L\ge0$, slack domination forces $L(w^\star,\lambda)=0$.
\end{proof}

\subsection*{Proof of the recursive quotient and minimal exact state}
\proofdependency{All sets are finite.  This avoids measurable-quotient issues and makes the decreasing partition sequence stabilize after finitely many strict refinements.  Policies are fixed observation-history rules, not rules separately tailored to the two initial beliefs.}
\begin{proof}
Equality of $\chi$ is an equivalence relation, so $\sim_0$ is an equivalence relation.  Suppose $\sim_h$ is one.  Within each $\sim_h$-class, associate to $b$ the finite signature
\[
 \operatorname{sig}_h(b)
 =\bigl[K_e(b;\{y\}\times A)\bigr]_{e,y,A\in\mathcal R/\!\sim_h}.
\]
The relation $\sim_{h+1}$ is equality of these signatures inside a $\sim_h$-class.  It is therefore an equivalence relation and refines $\sim_h$.  The quotient cardinalities are consequently nondecreasing.

We prove finite-horizon invariance by induction.  At horizon zero, related beliefs have the same color.  Suppose the claim holds at horizon $h$ and let $b\sim_{h+1}b'$.  A fixed observation-history policy selects the same first experiment at the two empty histories.  Equation~\eqref{eq:book-recursive-refinement} gives the same probability to every observation label and successor $\sim_h$-class.  Conditional on such a label and class, the induction hypothesis gives the same law for the remaining $h$ observations and colors.  Summing over the first step proves the claim at horizon $h+1$.

Call an equivalence relation $R$ stable when it respects $\chi$ and gives equal mass, for every $e$ and $y$, to every $R$-class.  We show that every such $R$ is contained in every $\sim_h$.  Color respect gives $R\subseteq\sim_0$.  If $R\subseteq\sim_h$, each $\sim_h$-class is a union of $R$-classes.  Stability on the finer $R$-classes can therefore be summed to give equality on every $\sim_h$-class, so $R\subseteq\sim_{h+1}$.  Induction yields $R\subseteq\sim_\infty$.

Because $\mathcal R$ is finite, the decreasing equivalence relations stabilize: for some $H$, $\sim_H=\sim_{H+1}=\sim_\infty$.  The defining equality at the fixed point makes $\sim_\infty$ itself stable.  Hence it is the largest stable color-respecting equivalence relation.  Stability also makes the color decoder and every reduced labeled kernel on $\mathcal R/\!\sim_\infty$ well defined, proving exactness of the quotient map.

Let $\phi$ be any exact recursive carrier and define
$bR_\phi b'$ when $\phi(b)=\phi(b')$.  Color factorization makes $R_\phi$ color-respecting.  Factorization of the joint observation--successor law makes it stable.  Maximality therefore implies
$R_\phi\subseteq\sim_\infty$, which is exactly
\eqref{eq:book-recursive-universal}.  Thus distinct $\sim_\infty$-classes require distinct carrier states, giving at least $N_\infty$ states.  If $L$ charts expose at most $K$ states each, they expose at most $LK$ distinct chart--state pairs, and $LK\ge N_\infty$ proves
\eqref{eq:book-recursive-chart-bound}.

Starting from $N_0$ blocks, every strict signature-refinement round increases the block count by at least one and no round can exceed $N_\infty$.  Hence there are at most $N_\infty-N_0$ strict increases.  When a block splits, two of its beliefs have unequal signatures; one coordinate of that inequality supplies the promised experiment $e$, observation $y$, and successor block $A$.
\end{proof}

%% file: part_notes/part06_history.tex
\parthistoricalnotes{}
\index{confidence world!historical comparison}
\index{identified image!historical comparison}
\index{certificate!three-way rule!historical comparison}

\subsection*{Confidence world: the name is local, the confidence-set principle is not}

A confidence world $C_n$ is a random confidence set for a complete population
world.  Its mathematical coverage contract is not a new species of
confidence set.  In particular, \citet{Dufour1997} observed that a covered
joint set can be mapped through any function $g$: the image $g(C_n)$ remains
a valid confidence set because membership of the true joint parameter implies
membership of its image.  The book's identified image $J_{C_n}(q)$ is this
projection principle applied to a declared structural query.

The closest neighboring objects nevertheless live on different axes.

\begin{center}
\small
\begin{tabularx}{0.98\textwidth}{@{}p{0.19\textwidth}p{0.24\textwidth}p{0.25\textwidth}X@{}}
\toprule
Lineage & Primary object & Main question & Relation to $C_n$ \\
\midrule
Manski partial identification \citep{Manski2003} & population identified set
$\Theta_I(P)$ & what is determined even with the population law known? & a
population identification object, not the finite-sample random set itself \\
Dufour projection \citep{Dufour1997} & covered joint confidence set and its
image $g(C_n)$ & how is coverage preserved for a transformation or subvector?
& direct antecedent of the identified-image validity argument \\
Set inference \citep{ChernozhukovHongTamer2007,AndrewsSoares2010} & estimators
and confidence regions in partially identified or moment-inequality models &
when do set estimators and confidence procedures have consistency, coverage,
and useful power? & supplies model-specific construction and asymptotic theory
that the abstract coverage definition alone does not provide \\
Vogel random optimization \citep{Vogel2008} & simultaneous random objective
and constraint information & how can feasible sets, optimal values, and
solution sets be covered together? & direct neighbor of the honest
inner/outer architecture bracket \\
Confidence world (this book) & complete population object, including nuisance
and the sampling/future-experiment structure needed by the declared query &
which structural conclusions and actions are authorized on one coverage
event? & an enriched bookkeeping and interface convention built on the
preceding confidence-set principles \\
\bottomrule
\end{tabularx}
\end{center}

Coverage alone gives no contraction theorem: the vacuous choice
$C_n\equiv\mathcal W$ satisfies a coverage lower bound.  Convergence to an
identified set, Hausdorff consistency, nonconservativeness, or power requires
additional assumptions and a concrete construction, as in the set- and
moment-inequality literatures just cited.  Part VI therefore claims transport
of a declared coverage event, not automatic consistency of every confidence
world.

\subsection*{The three-way rule and the model-conflict state}

For a nonempty $C_n$, the rule is set unanimity:
\[
q(C_n)=\{F\}\Rightarrow F,\qquad
q(C_n)=\{I\}\Rightarrow I,\qquad
|q(C_n)|>1\Rightarrow U.
\]
This follows in one line from confidence-set projection and is historically
adjacent to equivalence and noninferiority testing, where a confidence set
must lie wholly inside or outside a declared margin
\citep{Blackwelder1982,BergerHsu1996,PiaggioEtAl2006}.  The book does not
claim that the $F/I/U$ threshold logic is new.  Its increment is to type the
query as a structural architecture claim, preserve the whole identified
image and tolerance profile, connect $U$ to resolution complexity, and keep
authorization separate from posterior belief and deployment.

The symbol $M$ is book terminology for an empty confidence world.  Empty-set
diagnostics are not without precedent: \citet{Dufour1997} explicitly notes
that an empty valid confidence set can be interpreted as rejection of the
model or overidentifying restrictions.  The book's contribution is to prevent
that diagnostic from being confused with either a feasible or impossible
population truth label.

\subsection*{What the theorem-level audit leaves as genuine increments}

The worldwise resolution lower bound and recursive quotient import
partition-identification and probabilistic-bisimulation machinery.  The
posterior/worldwise counterexample is standard.  The atlas inverse and common-
deployment results are program-specific specializations.  The exact
evidence-slack criterion asks when compression preserves one max--min
certificate floor rather than an entire parametric experiment.  Its proof is
a short KL data-processing and max--min argument, so the book treats it as a
useful specialization rather than a new information-theoretic mechanism; see
Appendix~\ref{app:principal-result-audit}.

%% file: chapters/ch21_certified_learning.tex
\chapter{Toward a Certified Structural Learning Theory}
\label{ch:certified-learning}
\index{structural learning theory}
\index{learning bound!four-component target}

A structural learning theorem should not hide distinct mechanisms inside one
undifferentiated excess-risk bound.  It should report lower or upper
certificates for the risk components that the data and structural analysis can
actually identify.

Let $\widehat\Arch$ be a possibly data-dependent architecture selected on calibration data, and let $\widehat A$ be trained on an independent training split.  On a held-out test split, the desired statement has the form
\[
  \underline{\modelgap}
  +
  \underline\archgap
  \le
  R(\widehat A)-R^\star
  \le
  \overline\modelgap
  +
  \overline\archgap
  +
  \overline\gengap
  +
  \overline\optgap.
\]
Every term should have a declared source.

\begin{sourcebox}[title={Companion interface and proof provenance}]
The typed-carrier synthesis and no-compensation interface are developed in
the second EGML companion \citep{EGMLII}.  This chapter gives the complete
book proofs of its principal interface statements and keeps the statistical,
recursive, deployment, and operational loss scales separate.
\end{sourcebox}

\section{Three nonexchangeable carrier contracts}
\index{carrier!deployment}
\index{carrier!evidence}
\index{carrier!recursive}
\index{no-compensation theorem}

Chapter~\ref{ch:certificate-statistics} has already fixed the evidence-carrier
contract in Section~\ref{sec:certificate-evidence-carriers} and the recursive
carrier contract in Section~\ref{sec:certificate-recursive-quotient}.  We
reuse those definitions unchanged.  The additional interface here is the
deployment carrier $H_{\rm dep}$: in the carrier--decoder factorization of
Chapter~\ref{ch:resource-rd}, it maps operating inputs to the state used by an
action decoder.
When $H_{\rm dep}$ is induced by a nonnative lift, it enters this chapter only
after passing the quotient-faithful extraction gate of
Section~\ref{sec:lift-target-visible}.

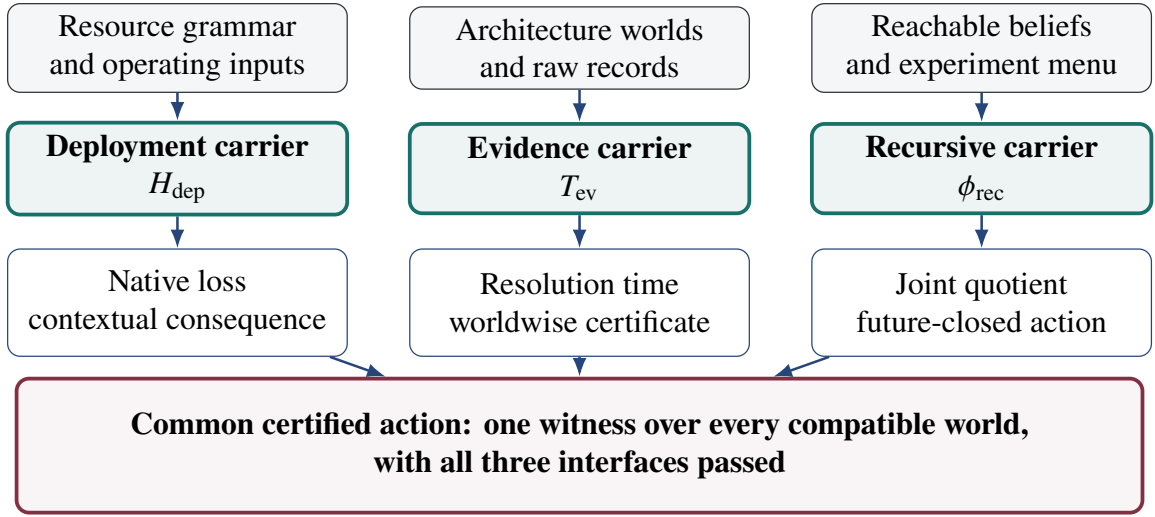
\begin{figure}[H]
\centering
\resizebox{0.98\textwidth}{!}{%
\begin{tikzpicture}[
  source/.style={draw=egdark,rounded corners,align=center,fill=eggray,
    minimum width=38mm,minimum height=9mm,font=\small},
  carrier/.style={draw=egteal,very thick,rounded corners,align=center,
    fill=egteal!8,minimum width=38mm,minimum height=10mm,font=\small\bfseries},
  obligation/.style={draw=egblue,rounded corners,align=center,fill=white,
    minimum width=38mm,minimum height=12mm,font=\small},
  gate/.style={draw=egwine,very thick,rounded corners,align=center,
    fill=egwine!5,minimum width=126mm,minimum height=15mm,font=\small\bfseries},
  arr/.style={-{Latex[length=2.1mm]},thick,draw=egblue}]
\node[source] (dep-source) at (-4.5,0)
  {Resource grammar\\and operating inputs};
\node[source] (ev-source) at (0,0)
  {Architecture worlds\\and raw records};
\node[source] (rec-source) at (4.5,0)
  {Reachable beliefs\\and experiment menu};

\node[carrier] (dep-carrier) at (-4.5,-1.35)
  {Deployment carrier\\$H_{\rm dep}$};
\node[carrier] (ev-carrier) at (0,-1.35)
  {Evidence carrier\\$T_{\rm ev}$};
\node[carrier] (rec-carrier) at (4.5,-1.35)
  {Recursive carrier\\$\phi_{\rm rec}$};

\node[obligation] (dep-out) at (-4.5,-2.85)
  {Native loss\\contextual consequence};
\node[obligation] (ev-out) at (0,-2.85)
  {Resolution time\\worldwise certificate};
\node[obligation] (rec-out) at (4.5,-2.85)
  {Joint quotient\\future-closed action};

\node[gate] (common) at (0,-4.45)
  {Common certified action: one witness over every compatible world,\\
   with all three interfaces passed};

\draw[arr] (dep-source)--(dep-carrier);
\draw[arr] (dep-carrier)--(dep-out);
\draw[arr] (ev-source)--(ev-carrier);
\draw[arr] (ev-carrier)--(ev-out);
\draw[arr] (rec-source)--(rec-carrier);
\draw[arr] (rec-carrier)--(rec-out);
\draw[arr] (dep-out)--(common);
\draw[arr] (ev-out)--(common);
\draw[arr] (rec-out)--(common);
\end{tikzpicture}%
}
\caption{The three architecture chains after finite-information closure.
Each has its own domain, exactness obligation, and reporting scale; they
meet only at a common certified action.}
\label{fig:three-carrier-gates}
\end{figure}

Placed side by side, the three exactness obligations have different domains
and loss scales:
\begin{align*}
  \text{deployment:}\quad
  &a^\star=d_{\rm dep}\circ H_{\rm dep},\\
  \text{evidence:}\quad
  &T_Z^{q,*}(\theta)=T_X^{q,*}(\theta),\\
  \text{recursion:}\quad
  &\text{the declared color and every observation-labeled update}
    \text{ factor through }\phi_{\rm rec}.
\end{align*}
The first is charged in native loss, the second in expected observations,
and the third in reachable states or charts.  They form a typed audit vector,
not three terms of one excess-risk sum.  Figure~\ref{fig:three-carrier-gates}
makes the three chains and their common terminal gate visible before the
no-compensation theorem is stated.

The surrounding foundations are classical: statistical decision functions,
sufficiency and comparison of experiments, treatment rules, and MDP state
abstraction all precede this synthesis
\citep{Wald1950,Bahadur1954,Blackwell1951,Blackwell1953,
Manski2004Treatment,LiWalshLittman2006}.  The claimed contribution is the
typed interface and its explicit common-witness gate, not those component
theories.

\begin{conditionbox}
The evidence and recursive statements use a finite world set, finite
experiment menu, and finite reachable belief set closed under
positive-probability Bayes updates.  The evidence carrier is independent of
the unknown world, the terminal task $q$ is declared in advance, and
$0<\alpha<1/2$.  The deployment branch uses a finite positive-mass input
problem with a common compact action space.  Its native defects are
finite and lower-semicontinuous in the action and vanish only at their respective
oracle actions.
\end{conditionbox}

\begin{theorem}[Typed no-compensation]
\label{thm:book-typed-no-compensation}
Each of the following failures survives unlimited resources in the other two
carrier branches.
\begin{enumerate}
  \item \emph{Evidence blindness.}  If opposite-label worlds $\theta$ and
  $\lambda$ satisfy $Q_{\theta,e}=Q_{\lambda,e}$ for every experiment $e$,
  no retained-evidence rule can stop almost surely and have worldwise error
  at most $\alpha$ at both worlds, even with exact recursive and deployment
  carriers.
  \item \emph{Recursive collision.}  If
  $\phi_{\rm rec}(b)=\phi_{\rm rec}(b')$ for beliefs in different classes
  of the minimal recursive quotient from
  Theorem~\ref{thm:certificate-recursive-state}, a finite labeled future
  context distinguishes them.  No decoder on that fixed recurrent state can
  realize the exact reduced process, even with raw evidence and an
  unrestricted deployment decoder.
  \item \emph{Deployment collision.}  If
  $H_{\rm dep}(x)=H_{\rm dep}(x')$ while
  $a_x^\star\ne a_{x'}^\star$, no decoder on that fixed carrier realizes
  both oracle actions.  Under the compact-separation assumptions above, when
  both inputs have positive mass the best such decoder has strictly positive
  native defect even with raw evidence and identity recursive state.
\end{enumerate}
Consequently, none of the three exactness predicates by itself implies either
of the other two.  This refutes bare implications and automatic
compensation; it does not rule out cross-branch reasoning from a conjunction
of hypotheses, a proved factorization theorem, or an optimization problem
with an explicitly declared exchange rate.
\end{theorem}

The evidence-blind branch is the zero-information endpoint of the classical
two-world change-of-measure method \citep{KaufmannEtAl2016}; the theorem
keeps that statistical obstruction typed separately from deployment and
recursive-state failures.

A minimal memory-split stress test makes the boundary especially sharp.  Let
a fair history bit $H$ be erased by a constant baseline memory before a
binary action is chosen.  In one world the binary target is an independent
fair coin; in another it is $H$ passed through a binary symmetric channel.
Every baseline-executable policy then has the same retained action--reward
law in both worlds, although retaining $H$ improves value only in the second
world.  If a temporary memory split is forbidden, the architecture query has
zero characteristic information.  If the split is allowed diagnostically,
choosing the action $A=H$ reduces the query to testing
$\operatorname{Bern}(1/2)$ against
$\operatorname{Bern}(1-\eta)$.  Charging for the diagnostic changes
information per unit cost, not the statistical object.  Thus operational
exposure is an experiment-menu gate: under a resettable finite contract it is
absorbed by active sequential experiment design
\citep{Chernoff1959,NaghshvarJavidi2013}, while the history quotient remains
within state-abstraction theory \citep{LiWalshLittman2006}.  The companion
go/no-go audit records the exact finite calculation; it is not an additional
principal result.

\begin{interpretationbox}
More parameters in a deployment head cannot recover a likelihood ratio
erased by the sensor; more samples after a recurrent collision cannot restore
the two histories that were merged; and a perfect population certificate
does not manufacture a missing action decoder.  A failed gate therefore
authorizes repair only on its own typed branch.
\end{interpretationbox}

\begin{proofroadmap}
Evidence blindness follows because every adaptive retained transcript has
the same law in the two opposite worlds.  Recursive collision is witnessed
at the first finite refinement depth separating the two beliefs.  Deployment
collision follows because one decoder value cannot equal two distinct oracle
actions; lower semicontinuity and compactness turn this incompatibility into
a strictly positive separation margin.  Identity choices on the remaining branches establish
non-compensation.  The chapter appendix supplies all three arguments.
\end{proofroadmap}

\subsection{Routed experts and external memory under bounded observability}
\index{mixture of experts!routing carrier}
\index{retrieval-augmented generation!information carrier}

In a sparsely routed expert system, let $H_{\rm dep}(x)$ be exactly the state
visible to the router and let the action be an expert index, or a legal
top-$k$ expert set.  A per-input oracle can choose the best expert for every
token or query, yet this pointwise family need not factor through one
measurable gate on $H_{\rm dep}$.  If two positive-mass query classes collide
in the router-visible state and reverse their preferred experts, the
deployment-collision branch of \cref{thm:book-typed-no-compensation} gives a
positive native defect.  More router optimization with the same carrier and
action grammar cannot remove that collision.  Sparse expert routing is the
representative mechanism, not the name of the problem class
\citep{ShazeerEtAl2017,FedusEtAl2022Switch}.

The possible repairs act on different contracts.  Giving the router an
additional feature or retrieved evidence changes its information carrier;
top-$k$ routing, a shared residual expert, or an abstention/fallback path
changes the action class; adding experts without making the distinguishing
information visible may change neither.  A larger context window changes a
charged resource budget and must be analyzed as such.  Retrieval-augmented
generation is particularly revealing: retrieval changes the experiment and
external-memory carrier available to the generator, rather than merely
optimizing the same closed-book decoder \citep{LewisEtAl2020RAG}.

A credible architecture-choice claim therefore freezes the router-visible
information, retrieval corpus and index, expert/action library, native task
loss, and compute/latency budget before confirmation.  It then tests the
certificate-directed carrier or class repair against equal-information and
equal-compute controls on an independent endpoint.  Without those controls,
an improved routed or retrieval system does not identify which typed repair
caused the gain.

\begin{boundarybox}
This is a theorem-interface example, not a new MoE or RAG theorem and not a
new experiment.  The obstruction applies only to the frozen carrier and
legal action class; changing either may be the correct repair and also changes
the proposition being tested.
\end{boundarybox}

\section{Typed architecture realization}
\label{sec:typed-architecture-realization}
\index{architecture!typed realization}
\index{carrier!joint recursive}
\index{common deployment!typed realization}

The no-compensation theorem is the negative half of the interface.  The
positive half explains how the deterministic deployment chain, statistical
authorization, and recursive continuation meet without identifying their
loss scales.

Let $q_{\rm arch}:\Theta\to\mathsf Q_{\rm arch}$ be an architecture-action
color.  It may record feasibility, a witness class, or a repair permission,
but it must distinguish worlds requiring different terminal deployments.
Let $C_t$ be an anytime confidence world for the same population model.
Apply the raw and retained characteristic-information definitions of
Chapter~\ref{ch:certificate-statistics} to this color, and denote the
resulting characteristic times by

\[
  T_X^{\rm arch,*}(\theta)
  \quad\text{and}\quad
  T_Z^{\rm arch,*}(\theta).
\]

For recursive control, color each reachable retained-evidence state $u$ by
\begin{equation}
  \zeta(u)
  =\bigl(c_{\rm dep}(u),c_{\rm cert}(u),c_{\rm ctx}(u)\bigr),
  \label{eq:book-joint-architecture-color}
\end{equation}
where the entries record the deployment output, certificate permission, and
sound contextual equivalence class.  Applying the labeled-kernel refinement
of Section~\ref{sec:certificate-recursive-quotient} to \(\zeta\) gives a
stable joint quotient; write $N_{\rm joint}$ for its number of classes.

\begin{conditionbox}
The world, experiment, observation, and reachable-state sets are finite; the
retained-evidence channel is independent of the unknown world; and the
confidence worlds are simultaneously valid under adaptive stopping.  The
architecture color is declared before the terminal decision, the contextual
component of \(\zeta\) is a sound operational congruence, and the deployment
loss, resource grammar, and task contract are the same ones used in
Chapters~\ref{ch:resource-rd}, \ref{ch:operational-semantics}, and
\ref{ch:statistical-eg}.  The terminal time $\tau$ is a stopping time for the
retained-evidence filtration, and $A_\tau$ is one
$\mathcal F_\tau$-measurable selector taking values in the declared resource
class.  It may use the full retained certificate state; if the grammar instead
requires the action to depend only on $q_{\rm arch}$, the corresponding
color-to-action factorization is an additional hypothesis.  The evidence-slack
equivalence additionally uses the attainment conditions of
Theorem~\ref{thm:certificate-evidence-slack}.  No common-witness membership
is implied by these standing conditions; confidence-world authorization
requires the separate gate displayed in the theorem below.
All positive-composition conclusions are asserted on one declared
time-uniform event of probability at least $1-\alpha$.  If the retained-evidence
and confidence-world guarantees are constructed on separate events with
errors $\alpha_{\rm ev}$ and $\alpha_{\rm cw}$, use their intersection and
require $\alpha_{\rm ev}+\alpha_{\rm cw}\le\alpha$.
\end{conditionbox}

\begin{theorem}[Typed architecture realization interface]
\label{thm:book-typed-architecture-realization}
Under the preceding conditions, the evidence and recursive branches impose
the following necessary bounds.  The confidence-world conclusion is a
separate conditional interface.
\begin{enumerate}
  \item Every $\alpha$-worldwise-valid system that resolves $q_{\rm arch}$
  from retained evidence has expected stopping time satisfying
  \begin{equation}
    \E_\theta\tau
    \ge
    T_Z^{\rm arch,*}(\theta)
    \operatorname{kl}(1-\alpha,\alpha).
    \label{eq:book-typed-architecture-time}
  \end{equation}
  Evidence processing gives
  $T_Z^{\rm arch,*}\ge T_X^{\rm arch,*}$, with equality exactly under the
  certificate-slack criterion for the binding opposite-action worlds.

  \item Every exact shared recursive carrier preserving \(\zeta\) and every
  observation-labeled update has at least $N_{\rm joint}$ reachable states.
  If each of $L$ charts distinguishes at most $K$ states, then
  \begin{equation}
    L\ge\left\lceil\frac{N_{\rm joint}}{K}\right\rceil.
    \label{eq:book-joint-architecture-chart-bound}
  \end{equation}

  \item For an adapted, measurable terminal selector $A_\tau$, the declared
  confidence-world authorization gate is the common-witness condition
  \begin{equation}
    A_\tau
    \in
    \bigcap_{\lambda\in C_\tau}
    \mathfrak F_R^\eta(\lambda).
    \label{eq:book-typed-common-witness}
  \end{equation}
  If this gate is verified, then on the simultaneous coverage event the
  witness satisfies the declared tolerance in the true world.  Items 1 and 2
  do not imply the gate.

  \item Conditional on the same common-witness gate, resource capacity and native loss remain governed
  by the carrier--decoder decomposition of Chapter~\ref{ch:resource-rd}; a claimed
  operational consequence follows only through the transmission and exposure
  gates of Chapter~\ref{ch:operational-semantics}.
\end{enumerate}
For the positive composition direction, an anytime-valid retained-evidence certificate, an exact joint
recursive carrier, an adapted witnessed common deployment, and the declared
native and contextual gates compose into an anytime-valid common certified
action.
\end{theorem}

\begin{interpretationbox}
The deployment chain answers what can be executed, the evidence chain what
can be authorized from finite information, and the recursive chain what must
remain distinguishable after future updates.  They meet at one action but do
not become one loss: extra state cannot repair a blind sensor, extra samples
cannot manufacture a decoder, and pointwise feasibility cannot replace a
common witness.  The common-witness membership is a declared authorization
hypothesis, not a consequence of worldwise correctness of the architecture
color or of the recursive state bound.  The $N_{\rm joint}$ bound concerns the recursive carrier; it
does not assert that separately feasible evidence, recursive, and deployment
carriers fit simultaneously inside one resource budget.  That assertion
belongs to a declared joint grammar.  Likewise, the common-witness route may
select $A_\tau$ from the full retained certificate state; a rule of the form
$A_\tau=D_{\rm act}\{q_{\rm arch},H_{\rm dep}\}$ requires a separate
color-to-action factorization and is not implied by the other gates.
Moreover, $N_{\rm joint}$ counts the protected recursive carrier classes, not
auxiliary likelihood sums, clocks, experiment counts, or tracking variables;
those may require separate or unbounded memory unless the joint grammar
explicitly includes them.
\end{interpretationbox}

\begin{proofroadmap}
Apply the resolution lower bound and evidence-slack theorem of
Chapter~\ref{ch:certificate-statistics} to $q_{\rm arch}$.  Apply its
recursive quotient theorem to the joint color \(\zeta\).  For the conditional
deployment branch, assume the displayed common-witness gate and apply
Theorem~\ref{thm:book-common-deployment}; the native and
contextual clauses are the interfaces of Chapters~\ref{ch:resource-rd} and
\ref{ch:operational-semantics}.  Apply the common-witness implication at the
adapted stopping time on the single time-uniform coverage event.  Their
conjunction proves the positive composition without adding the four
loss scales.  The chapter appendix gives the formal reduction.
\end{proofroadmap}

\subsection{Radical instance: a genuine common witness}
\label{sec:radical-common-witness}
\index{common deployment!radical instance}
\index{radical normal form!common witness}

The radical loop makes the common-witness quantifier concrete.  Let \(C_t\)
be any confidence world of continuous target loops
\(z_\lambda:S^1\to\C\) satisfying

\[
  \inf_{\lambda\in C_t}\inf_{\theta\in S^1}
  |z_\lambda(\theta)|\ge \rho>0.
\]

Freeze the repaired operating contract before the terminal decision: the
deployed input contains \(u=z_\lambda(\theta)\), or a carrier from which \(u\)
is exactly recoverable, and the native defect of a point output is

\[
  D_u(a)=|a^k-u|^2.
\]

For a symmetric task, the full-root rule

\[
  W_k(u)=\{a\in\C:a^k=u\}
\]

is one world-independent set-valued architecture.  If its loss is the largest
native defect of a returned element, write
\(\mathfrak F_{\mathrm{set}}^0(\lambda)\) for the zero-tolerance feasible set
under this repaired grammar.  Then

\[
  W_k\in
  \bigcap_{\lambda\in C_t}
  \mathfrak F_{\mathrm{set}}^0(\lambda).
\]

If a local point representative is required, cover \(\C\setminus\{0\}\) by
the two slit domains

\[
  U_0=\C\setminus(-\infty,0],
  \qquad
  U_1=\C\setminus[0,\infty).
\]

Each domain has a continuous logarithm and hence a local root
\(\psi_j(u)=\exp\{\operatorname{Log}_j(u)/k\}\).  A fixed measurable router chooses an
available domain and retains its chart label.  The resulting two-chart rule
\(A_{\mathrm{atl}}(u)=\bigl(j(u),\psi_{j(u)}(u)\bigr)\) is again independent of
\(\lambda\).  Writing \(\mathfrak F_{\mathrm{atl}}^0(\lambda)\) for the
corresponding zero-tolerance feasible set,

\[
  A_{\mathrm{atl}}\in
  \bigcap_{\lambda\in C_t}
  \mathfrak F_{\mathrm{atl}}^0(\lambda).
\]

Thus the confidence world need not collapse to a singleton before either
repaired grammar has an actual common witness.  Degree identification still
decides whether the original \(\theta\)-indexed one-chart baseline is ruled
out.  It does not, by itself, exhibit a common one-chart root: even if every
compatible loop has degree divisible by \(k\), their worldwise root fields may
be different.

This construction does not silently solve the original contract of
Theorem~\ref{thm:radical-tax}.  If the deployed input is only \(\theta\) and
does not expose \(z_\lambda(\theta)\), neither rule above is admissible.  If
the task requires one globally labeled root, the set-valued rule changes the
output semantics and the atlas rule changes the routing grammar.  In that
case one must instead shrink \(C_t\) and exhibit a single admissible
\(\theta\mapsto a(\theta)\) in the intersection.  The distinction is exactly
the difference between a mechanism-matched repair and a world-specific oracle
family disguised as one deployment.

This interface closes the statistical upgrade of the architecture theory.
Chapter~\ref{ch:oali-workflow} turns the first failed gate into a typed return
path for data acquisition, evidence repair, state refinement, deployment
expansion, or contextual revision.

\section{Finite-library comparison and data-dependent selection}
\paragraph{A finite-library version.}
\index{architecture!finite library}
\index{held-out validation!finite library}

Suppose calibration data choose a finite candidate library
\[
  \mathcal L=\{\Arch^{(1)},\ldots,\Arch^{(K)}\},
\]
from which one trained architecture $\widehat A$ is selected without using
the test sample.  Fix $M$ predeclared baselines $A_1,\ldots,A_M$, also
without inspecting the test sample.  Let the test-unit loss lie in $[0,B]$.
For $b=1,\ldots,M$ define
\[
  \widehat\Delta_b
  =
  \frac1{n_{\rm te}}
  \sum_{i=1}^{n_{\rm te}}
  \{\ell(A_b;Z_i)-\ell(\widehat A;Z_i)\}.
\]
Hoeffding's inequality and a union bound give, with probability at least
$1-\delta$, simultaneously for all $b=1,\ldots,M$,
\[
  \Delta_b
  \ge
  \widehat\Delta_b
  -
  B\sqrt{\frac{2\log(M/\delta)}{n_{\rm te}}}
\]
for the stated range convention.

\begin{conditionbox}
All fitted candidates, the selected $\widehat A$, and all baselines are fixed before the independent test sample is inspected.  The test observations are independent at the true experimental-unit level, and each paired loss difference lies in $[-B,B]$.  The candidate library has size $K$, while $M$ is the finite number of predeclared baseline comparisons covered by the union bound.
\end{conditionbox}

\begin{theorem}[Finite-library held-out improvement]
\label{thm:heldout-improvement}
With probability at least $1-\delta$, simultaneously for every
$b=1,\ldots,M$,
\[
  \Delta_b
  \ge
  \widehat\Delta_b
  -B\sqrt{\frac{2\log(M/\delta)}{n_{\rm te}}}.
\]
Consequently, on that event, if
\[
  \min_b\widehat\Delta_b
  >
  B\sqrt{\frac{2\log(M/\delta)}{n_{\rm te}}},
\]
then $\Delta_b>0$ for every baseline, so the selected architecture has
smaller population loss than every predeclared baseline.
\end{theorem}

\begin{interpretationbox}
The theorem turns held-out paired improvements into simultaneous population lower bounds against all predeclared baselines.  In multi-cell or repeated-measure applications the biological unit, not the cell, is the concentration unit.  The result does not cover adaptively added baselines without an adjusted selection argument.
\end{interpretationbox}

\begin{proofroadmap}
Apply Hoeffding's inequality to each paired loss difference using range length $2B$, choose a tolerance with tail probability $\delta/M$, and take a union bound over the library.  The chapter appendix verifies the constant and conditional-on-development formulation.
\end{proofroadmap}

The independent unit is the biological sample, task, or trajectory, not the individual cell or time point when those observations are dependent within units.

\paragraph{Architecture certificate plus held-out gain.}
\index{held-out validation!predictive gain}
\index{certificate!architecture plus gain}

A convincing OALI conclusion combines two statements.

\paragraph{Structural statement.}
Calibration data yield
\[
  \underline\Obs(\Arch_0)>0
\]
for the baseline architecture and a small upper witness for the repair.

\paragraph{Predictive statement.}
Independent test data show lower operational loss for the repair.

Either statement alone is weaker.  A held-out gain without a certificate may be generic architecture search.  A certificate without held-out gain may be operationally irrelevant.

\paragraph{Data-dependent classes.}
\index{architecture!data-dependent}

If the same data define the atlas and fit the experts, uniform fixed-class theory is insufficient.  Three routes are available.

\begin{enumerate}
  \item \textbf{Sample splitting.}  Simple, transparent, but less data efficient.
  \item \textbf{Finite-library conditioning.}  Condition on calibration output and validate a predeclared finite set.
  \item \textbf{Random-set complexity.}  Bound generalization of the data-dependent set of architectures or optimizer trajectories.
\end{enumerate}

PAC--Bayesian bounds for random hypothesis sets provide one general route to
the third option \citep{DupuisEtAl2024}.  The current OALI program uses the
first two routes.  A theorem specialized to learned atlases and their
structural certificates remains open.

\paragraph{Certification-aware architecture selection.}
\index{architecture!certificate-aware selection}

Ordinary model selection minimizes estimated predictive risk.  Certification-aware selection uses a two-dimensional criterion:
\[
  \bigl(
  \text{held-out risk},
  \text{unresolved obstruction interval}
  \bigr).
\]
An architecture with slightly lower empirical loss but an unresolved or unstable certificate may be less scientifically auditable than a robustly certified alternative.

\section{Confidence, stopping, and the learning-theory frontier}
\paragraph{Structural and operational confidence.}
\index{confidence!structural versus operational}

A population obstruction can be positive but hidden from the chosen task.  Conversely, a small native defect can be amplified by a sensitive downstream decision.  A complete confidence statement may therefore bracket both native and observable frontiers.

The deterministic nonexpansiveness of task envelopes allows one confidence event to propagate to the operational audit when kernel errors are controlled.

\paragraph{Stopping rules for architecture research.}
\index{stopping rule!architecture research}

Certificate statistics suggests a principled stopping rule.  Continue data collection while the confidence world contains both:
\begin{itemize}
  \item a world where the baseline is saturated; and
  \item a world where the baseline has a practically meaningful obstruction.
\end{itemize}
Stop when the certificate becomes decisive or when the cost of resolution exceeds the value of the architecture decision.

\paragraph{What would constitute a new learning theory.}
\index{structural learning theory!completion criteria}

A mature structural learning theory would provide:
\begin{enumerate}
  \item objective-native architecture lower bounds;
  \item resource-indexed upper witnesses;
  \item honest finite-data transport;
  \item data-dependent generalization;
  \item optimization residual certificates;
  \item task-relative operational interpretation;
  \item automated, mechanism-specific repair.
\end{enumerate}

The current program closes several of these interfaces in finite, convex, graph, and singular normal forms.  It does not yet supply one universal theorem for arbitrary deep networks.

\section*{Exercises}

\begin{exercise}
Derive the finite-library held-out bound with a two-sided confidence interval and compare constants.
\end{exercise}

\begin{exercise}
Explain why treating individual cells as independent test units can invalidate a cytometry held-out claim.
\end{exercise}

\begin{exercise}
Design a stopping rule for collecting new calibration instances near a suspected monodromy seam.
\end{exercise}

\input{chapter_appendices/ch21_proofs}

%% file: chapter_appendices/ch21_proofs.tex
\chapterproofappendix

\subsection*{Proof of typed no-compensation}
\proofdependency{The evidence branch is a finite adaptive experiment with a parameter-independent retained channel.  The recursive branch uses the finite labeled-kernel refinement of Chapter~\ref{ch:certificate-statistics}.  The deployment branch has positive mass on the colliding inputs, a common compact action space, and finite lower-semicontinuous native defects whose zero sets are their distinct oracle actions.}
\begin{proof}
For evidence blindness, fix opposite-label worlds $\theta$ and $\lambda$
with $Q_{\theta,e}=Q_{\lambda,e}$ for every experiment.  Induction on time
shows that every retained transcript has the same law in the two worlds:
conditional on a common transcript, a nonanticipating rule chooses the same
distribution over experiments, and the next retained observation has the
same conditional law.  The conclusion remains true at a stopping time.  If
$A$ is the event that the rule reports $q(\theta)$, almost-sure stopping and
worldwise correctness would require
\[
  \Pp_\theta(A)\ge1-\alpha,
  \qquad
  \Pp_\lambda(A)\le\alpha.
\]
Equality of transcript laws makes the two probabilities equal, contradicting
$\alpha<1/2$.

For a recursive collision, let $b$ and $b'$ be merged by
$\phi_{\rm rec}$ but separated by the stable relation $\sim_\infty$ of
Theorem~\ref{thm:certificate-recursive-state}.  Because the reachable set is
finite, the refinement stabilizes after finitely many steps, so there is a
least $h$ with $b\not\sim_h b'$.  If $h=0$, their declared colors differ.
Otherwise they agree through depth $h-1$, but some experiment, observation
label, and $\sim_{h-1}$ successor block has different probability from the
two beliefs.  This is a finite future context with different reduced laws.
A decoder receiving the same recurrent state cannot reproduce both laws.

For a deployment collision, every decoder $d$ satisfies
\[
  d\{H_{\rm dep}(x)\}=d\{H_{\rm dep}(x')\}.
\]
Write $p_x,p_{x'}>0$ for the two input masses and define, on the common
action space,
\[
  \delta_{x,x'}
  =
  \min_a\{p_x\Def_x(a)+p_{x'}\Def_{x'}(a)\}.
\]
The minimum is attained by compactness and lower semicontinuity.  If it were
zero, nonnegativity would force one action to have zero defect at both
inputs, hence to equal both distinct oracle actions.  Therefore
$\delta_{x,x'}>0$, and every decoder on the collided carrier has expected
native defect at least this margin.  This conclusion is unchanged if the
evidence carrier is the identity and the recursive state retains the full
belief.

In each construction, choose identity carriers on the other two branches.
Conversely, finite product components can place any one collision beside
exact versions of the other interfaces.  Thus exactness of either remaining
branch cannot compensate for the failed one, and the three predicates are
pairwise logically independent.
\end{proof}

\subsection*{Proof of the typed architecture realization interface}
\proofdependency{The architecture action color is a specialization of the
truth map in Chapter~\ref{ch:certificate-statistics}; the joint color uses the
same finite labeled-kernel refinement.  The terminal deployment statement
uses the single time-uniform confidence event of
Chapter~\ref{ch:statistical-eg}; the stopping time is adapted and the terminal
architecture is an $\mathcal F_\tau$-measurable selector in the declared
resource class.  Membership in the displayed common-witness intersection is
a separate authorization hypothesis, not a consequence of worldwise color
resolution or recursive-state preservation.  The native and contextual
conclusions retain their original contracts.  No joint resource-budget
conclusion or color-to-action factorization is inferred.  If statistical
components are constructed separately, their error budgets are combined by
a union bound before this common event is invoked.}
\begin{proof}
Use $q=q_{\rm arch}$ in Theorem~\ref{thm:certificate-resolution-lower}.
Because the terminal rule is worldwise valid under the retained laws, that
theorem gives \eqref{eq:book-typed-architecture-time} with
$T_Z^{\rm arch,*}$.  Applying data processing and
Theorem~\ref{thm:certificate-evidence-slack} to the same opposite-action
partition gives $T_Z^{\rm arch,*}\ge T_X^{\rm arch,*}$ and its equality
criterion.

Now use \(\zeta\) as the contract color in
Theorem~\ref{thm:certificate-recursive-state}.  Its stable quotient is the
joint quotient by definition, so the universal refinement property gives at
least $N_{\rm joint}$ states; counting chart--state pairs gives
\eqref{eq:book-joint-architecture-chart-bound}.

For the conditional deployment branch, assume the terminal selector passes
the authorization gate \eqref{eq:book-typed-common-witness}.  On the
time-uniform coverage event, the true world belongs to $C_t$ for every
$t$, hence to $C_\tau$ at the adapted stopping time.  Measurability of the
selector makes $A_\tau$ a legitimate terminal action, and membership in
\eqref{eq:book-typed-common-witness} places
$A_\tau$ in the true world's feasible set.  This is exactly the common-
deployment implication of Theorem~\ref{thm:book-common-deployment}.  The
carrier--decoder decomposition then evaluates the witness in the declared native
loss scale, and the transmission and exposure gates determine which part is
visible under the declared contextual contract.  Thus simultaneous passage
of the evidence, recursive, common-witness, native, and contextual gates
produces the claimed common certified action on the coverage event, whose
complement has probability at most \(\alpha\).
\end{proof}

\subsection*{Proof of finite-library held-out improvement}
\proofdependency{The candidate architectures and baselines must be fixed before the independent test sample is inspected.  The independent units, not lower-level observations nested inside them, are the concentration units.  The loss difference must be bounded in an interval of length $2B$.}
\begin{proof}
For baseline $b$, let
\[
X_i^{(b)}
=
\ell\{A_b;Z_i\}-\ell\{\widehat A;Z_i\},
\qquad -B\le X_i^{(b)}\le B,
\]
and write
\[
\widehat\Delta_b=\frac1{n_{\rm te}}\sum_{i=1}^{n_{\rm te}}X_i^{(b)},
\qquad
\Delta_b=\E X_i^{(b)}.
\]
Hoeffding's inequality for a variable with range length $2B$ gives
\[
\Pp\{\Delta_b<\widehat\Delta_b-t\}
\le
\exp\left(-\frac{n_{\rm te}t^2}{2B^2}\right).
\]
Set
\[
t=B\sqrt{\frac{2\log(M/\delta)}{n_{\rm te}}}.
\]
The right-hand side becomes $\delta/M$.  A union bound over the $M$ predeclared baselines yields, with probability at least $1-\delta$,
\[
\Delta_b
\ge
\widehat\Delta_b
-
B\sqrt{\frac{2\log(M/\delta)}{n_{\rm te}}}
\qquad\text{for every }b.
\]
If the smallest empirical improvement exceeds the common tolerance, every population improvement is positive.
\end{proof}

%% file: chapters/ch22_oali_workflow.tex
\chapter{Obstruction-Aware Learning and Inference}
\label{ch:oali-workflow}
\index{obstruction-aware learning and inference}
\index{certificate-guided intervention}

The preceding chapters separate population structure from finite-sample
evidence.  Obstruction-Aware Learning and Inference (\OALI) is proposed as a
falsifiable experimental and audit protocol, not as an established general
architecture-selection theory.  Its premise is deliberately
restrictive: a structural repair should be attempted only after a baseline
deployment contract has been declared, a mechanism-specific obstruction has
been diagnosed, and the available data support a certificate strong enough to
justify changing that contract.

\section{From diagnosis to controlled intervention}

The basic workflow is
\[
\boxed{
\begin{gathered}
\text{local oracle estimation}
\to
\text{structural audit}
\to
\text{certificate}
\\
\to
\text{mechanism-matched repair}
\to
\text{training}
\to
\text{independent validation}
\end{gathered}}
\]

\begin{figure}[htbp]
\centering
\begin{tikzpicture}[>=Latex,node distance=8mm]
  \tikzset{w/.style={draw=egteal!70!black,fill=egteal!4,rounded corners=2pt,
    minimum width=25mm,minimum height=8mm,align=center,font=\scriptsize}}
  \node[w] (o) {local oracle};
  \node[w,right=of o] (a) {structural audit};
  \node[w,right=of a] (c) {certificate};
  \node[w,below=of c] (r) {matched repair};
  \node[w,left=of r] (tr) {training};
  \node[w,left=of tr] (v) {independent\\validation};
  \draw[->,thick] (o)--(a); \draw[->,thick] (a)--(c); \draw[->,thick] (c)--(r);
  \draw[->,thick] (r)--(tr); \draw[->,thick] (tr)--(v);
\end{tikzpicture}
\caption{The proposed OALI workflow.  Arrows denote audit stages rather than theorem implications.  A certificate motivates a repair candidate but does not establish saturation, mechanism exclusivity, or held-out improvement.}
\label{fig:oali-workflow}
\end{figure}
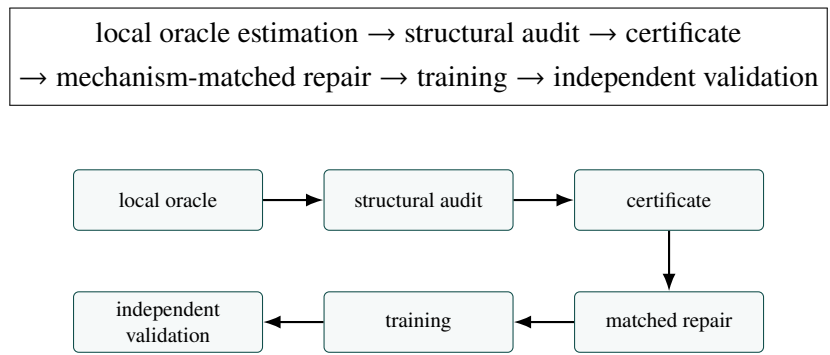

Before fitting the repaired system, one declares:
\begin{itemize}
\item the local objective and native defect;
\item the baseline deployment class and output semantics;
\item the population aggregation and statistical unit;
\item the resource and regularity constraints;
\item the downstream endpoint and independent validation protocol.
\end{itemize}

The local oracle stage estimates the relevant fiber, branch multiplicity, and
uncertainty.  Related work on amortized inference separates per-instance
optimization from the limitations of a shared inference mechanism
\citep{CremerLiDuvenaud2018,MargossianBlei2024,MarinoEtAl2018}.  OALI uses a
more restrictive intervention rule: it changes the deployment contract only
after the failure has been typed and certified.  Some gaps therefore call for
better local optimization rather than structural repair.

The structural audit then determines how local oracle objects
correspond across inputs: through point distances, permutations, partial
bijections, quotient relations, or set-valued correspondences.  Only after
this representation has been stabilized does the procedure compute a
structural certificate such as a transport lower bound, a coordination
obstruction, a cycle holonomy, a resource lower bound, or a three-way
feasible/impossible/unresolved certificate.

The distinction between \OALI\ and \OAI\ is operational.  \OALI\ is the
general workflow above.  \OAI\ is a concrete instance designed for
multibranch oracle families, using partial transport, cycle consistency, and
atlas construction.

\section{The repair must match the diagnosed mechanism}
\index{repair!mechanism-matched}

A positive certificate does not by itself say ``use a larger model.''  The
diagnosed mechanism determines which part of the deployment contract should
change.

\begin{table}[H]
\centering
\caption{Structural mechanisms and corresponding repair families.}
\small
\begin{tabularx}{\textwidth}{p{0.30\textwidth}X}
\toprule
Diagnosed mechanism & Candidate structural intervention\\
\midrule
excess oracle variation & additional charts, local experts, or test-time refinement\\
label monodromy & atlas, quotient, set-valued, or randomized output\\
missing branches & partial transport, abstention, or variable-cardinality output\\
coordination obstruction & rectangularization, memory refinement, or revised sharing\\
carrier information loss & richer carrier or task-relative sufficient state\\
decoder nonsaturation & stronger decoder with the carrier audit retained\\
probability or shape invalidity & projection or nonlinear validity repair\\
operational invisibility & no repair for the declared downstream task\\
optimization/reachability only & increase the optimization budget without a structural impossibility claim\\
\bottomrule
\end{tabularx}
\end{table}

The design principle is minimal intervention.  Change only the interface named
by the certificate whenever possible, while keeping parameter count,
computation, training protocol, and data usage comparable to the original
baseline.  This makes it possible to distinguish a structural repair from a
generic increase in flexibility.

A repaired class can still be implemented badly.  Atlas experts can collapse
onto the same branch; a quotient decoder can reintroduce an arbitrary
canonicalization; a gate can use information unavailable at deployment.
Training diagnostics should therefore retain the structural audit and report,
as applicable, chart occupancy, within-chart native defect, transport
consistency, gate stability, resource use, and numerical residuals.

\section{A falsifiable saturation--repair--validation chain}
\index{structural saturation}
\index{negative control}
\index{held-out validation!independent}

Training saturation is not an architecture theorem and is not implied by a
positive certificate.  Misspecification,
optimization failure, data scarcity, regularization, or an insensitive metric
can all produce plateaus.  A stronger structural interpretation requires the
following sequence.

\begin{enumerate}
\item A calibration-stage certificate predicts a positive obstruction under a
fixed deployment contract.
\item Increasing capacity, sample size, or optimization effort without
changing that contract does not remove the diagnosed native gap once those
ordinary errors have been controlled.
\item A minimal mechanism-matched change reduces the obstruction or native
defect in the direction predicted by the theory.
\item On independent statistical units, the repaired system improves the
predeclared endpoint relative to parameter- and compute-matched baselines.
\item Random partitions, generic mixtures, or other negative controls do not
stably reproduce the improvement; in no-obstruction controls the structural
procedure does not trigger unnecessary complexity.
\end{enumerate}

This sequence is intentionally falsifiable.  Failure of any step prevents the
full architecture-choice conclusion, although earlier steps may retain their
narrower meaning.  No experiment in the present edition completes all five
steps; the sequence is a proposed validation standard for future work.

Laboratory~\ref{sec:lab-mnist-amortization} illustrates only the certificate
and repair-comparison stages for amortized variational inference.  Its data,
width, and training-effort curves continue to improve at their tested
endpoints and therefore do not establish saturation.  Its Monte Carlo
reference and evaluation draws are reused, and it lacks compute-matched and
negative-control repairs.  It consequently does not validate the complete
chain or justify a general architecture choice.  Its endpoint is the sampled
native variational objective, not a downstream classification benefit.

\section{When a certificate implies population improvement}

The final comparison requires a common loss scale.  Suppose the baseline class
has a certified lower bound $\widehat O$ relative to the local oracle, the
repaired system has oracle and implementation error at most
$\epsilon_{\rm rep}$, and the required finite-sample tolerances sum to
$\Delta$.

\begin{conditionbox}
The baseline lower certificate and the repaired-system upper certificate must
hold on one declared event, refer to the same population target, and be
expressed in the same native or explicitly converted loss scale.  The
statistical and implementation terms may not be counted twice.
\end{conditionbox}

\begin{theorem}[Population improvement template]
\label{thm:population-improvement-template}
Under the stated certificate event, the population improvement of the repaired
system over the baseline architecture class is at least
\[
  \widehat O-\epsilon_{\rm rep}-\Delta.
\]
In particular, the improvement is strictly positive whenever the right-hand
side is positive.
\end{theorem}

\begin{interpretationbox}
A structural diagnosis becomes an actionable guarantee only when the
certified baseline floor exceeds the combined cost of learning and
implementing the repair.  A nonpositive right-hand side does not prove that
the repair fails; it means that the current evidence is insufficient to prove
strict population improvement.  Nor does monodromy or any other internal
obstruction alone imply improvement in an unrelated downstream metric.
\end{interpretationbox}

\begin{proofroadmap}
Write the population difference between the best baseline and the repaired
system as the baseline-to-oracle gap minus the repair-to-oracle gap, then
substitute the certified lower and upper bounds.  The chapter appendix gives
the exact subtraction.
\end{proofroadmap}

The theorem also fixes the role of independent validation.  OALI is not
identified by the use of experts, charts, or routing; those devices are common
architectural choices.  Its distinguishing claim is conditional: a declared
baseline contract is first shown to have a typed structural failure, and the
repair is selected because it targets that failure.  Capacity-matched random
partitions, wider single models, additional test-time computation, and
alternative repairs are therefore required controls rather than optional
ablations.

\section*{Exercises}

\begin{exercise}
Write a complete \OALI\ contract for amortized inference in a symmetric
two-component mixture.  State which observation units are reserved for
diagnosis and which for independent validation.
\end{exercise}

\begin{exercise}
Design a parameter- and compute-matched negative control for an atlas repair.
What result would weaken the claim that cycle structure is the source of the
gain?
\end{exercise}

\begin{exercise}
Give three ways an implementation can destroy a theoretically valid
zero-obstruction repair.
\end{exercise}

\input{chapter_appendices/ch22_proofs}

%% file: chapter_appendices/ch22_proofs.tex
\chapterproofappendix

\subsection*{Proof of the population improvement template}
\proofdependency{The baseline certificate must be a valid lower bound for every baseline architecture on the same event.  The repaired-risk upper bound must charge oracle approximation, implementation, and generalization errors only once.}
\begin{proof}
Let $R_{\rm base}^\star=\inf_{A\in\Arch_{\rm base}}R(A)$ and let $R_{\rm oracle}$ be the local-oracle risk.  The certified baseline event states
\[
R_{\rm base}^\star-R_{\rm oracle}\ge\widehat O.
\]
Suppose the repaired architecture satisfies
\[
R(\widehat A_{\rm rep})-R_{\rm oracle}
\le\epsilon_{\rm rep}+\Delta,
\]
where $\epsilon_{\rm rep}$ contains the declared oracle/implementation residuals and $\Delta$ contains the independent generalization tolerances.  Subtracting the second inequality from the first gives
\begin{align*}
R_{\rm base}^\star-R(\widehat A_{\rm rep})
&=
\{R_{\rm base}^\star-R_{\rm oracle}\}
-
\{R(\widehat A_{\rm rep})-R_{\rm oracle}\}\\
&\ge
\widehat O-\epsilon_{\rm rep}-\Delta.
\end{align*}
The population improvement is therefore positive whenever the final lower bound is positive.
\end{proof}

%% file: chapters/ch23_transport_cycles.tex
\chapter{Partial Transport, Cycle Holonomy, and Recovery}
\label{ch:transport-cycles}
The OALI workflow in Chapter~\ref{ch:oali-workflow} treats structural audit
and mechanism-matched repair at the procedural level.  This chapter
specializes that interface to finite branch sets and fixes the
partial-bijection model used by the recovery and holonomy theorems.

\section{Local branch sets}
\index{transport!local branch set}
\index{transport!partial bijection}

At graph vertex $v$, let
\[
  \mathcal B_v=\{1,\ldots,K_v\}
\]
index estimated local oracle branches.  Cardinalities may differ because components appear, disappear, split, or remain unresolved.

For edge $e=(u,v)$, a transport is a partial bijection
\[
  T_e:D_e\subseteq\mathcal B_u
  \longrightarrow
  R_e\subseteq\mathcal B_v.
\]
The inverse transport is used on the reverse orientation.

When all branch sets have the same size and every transport is total, this
problem is related to angular and permutation synchronization
\citep{Singer2011,PachauriKondorSingh2013}.  Partial correspondences,
cycle-consistent multi-matching, and partial permutation synchronization are
also established topics
\citep{ChenGuibasHuang2014,ZhouZhuDaniilidis2015,
BernardEtAl2019,LiShiLerman2022}.  The book does not claim to invent partial
maps.  Its distinction is semantic: disappearance and unresolved oracle
branches remain observable, and domain survival is carried into the
task-level audit rather than being filled by dummy permutations.

That semantic distinction is not, by itself, a novelty claim.  On finite
fibers, every partial bijection can be extended to a total permutation on a
larger fiber while a validity mask forbids dummy deployment states.  This
preserves the original global sections exactly and places the resulting
statistical problem next to masked synchronization and finite constraint
satisfaction.  A partial-domain learning theorem would need an additional
operational or statistical ingredient that is not preserved by this
totalization.

\section{Partial matching objective}
\index{transport!assignment margin}
\index{transport!partial matching}

Let $c_{ij}^{uv}$ be the cost of matching branch $i$ at $u$ to branch $j$ at $v$.  An unmatched penalty allows branches to disappear.  A structural partial matching minimizes total matched and unmatched cost.

The relevant assignment margin is the gap between the optimal structural matching and the second-best distinct structural matching.  Dummy-label permutations are not distinct structures.

\begin{definition}[Structural assignment margin]
If $C_e(T)$ is the edge objective and $T_e^\star$ is its unique optimal partial bijection, define
\[
  \Gamma_e
  =
  \min_{T\ne T_e^\star}
  \{C_e(T)-C_e(T_e^\star)\}.
\]
\end{definition}

An implementation that excludes only one optimal matched edge can overestimate this margin by missing alternatives that keep all optimal pairs and add another pair.  Exact second-best enumeration or an equivalent optimization is required.

\section{Noisy recovery}
\index{transport!noisy recovery}

Assume true branch representatives $a_{vk}^\star$ and estimates $\widehat a_{v,\pi_v(k)}$ satisfy
\[
  d(\widehat a_{v,\pi_v(k)},a_{vk}^\star)
  \le
  \varepsilon_v
\]
for unknown local relabelings $\pi_v$.

For metric matching costs, changing representatives perturbs the cost of any matched pair by at most $\epsilon_u+\epsilon_v$.  A matching uses at most $\min(K_u,K_v)$ pairs.

\begin{conditionbox}
Each local oracle fiber is finite and estimated up to an unknown local permutation.  The matched-pair cost is the ambient metric,
$c_{ij}^{uv}=d(a_{ui}^\star,a_{vj}^\star)$.  The population partial matching has a unique structural optimum after dummy permutations are quotiented out, the unmatched penalty is fixed, and representative errors are bounded by $\varepsilon_v$.  The strict assignment margin must dominate twice the worst total cost perturbation.
\end{conditionbox}

\begin{theorem}[Noisy partial-transport recovery]
\label{thm:partial-transport-recovery}
If
\[
  \Gamma_e
  >
  2\min(K_u,K_v)(\varepsilon_u+\varepsilon_v),
\]
then the estimated optimal partial matching is
\[
  \widehat T_e
  =
  \pi_vT_e^\star\pi_u^{-1}.
\]
The matched, unmatched, domain, and range structures are recovered exactly up to local relabeling.
\end{theorem}

\begin{interpretationbox}
Under margin separation, all matched pairs, missing branches, domains, and ranges are recovered exactly up to local relabeling.  The theorem deliberately outputs no forced match when the margin condition fails.  It applies to richer costs once a uniform structural-cost perturbation radius replaces the metric bound.
\end{interpretationbox}

\begin{proofroadmap}
Bound the perturbation of every matched-pair distance by the two endpoint radii, sum over at most $\min(K_u,K_v)$ pairs, and invoke a finite-class argmin stability lemma.  The chapter appendix includes the lemma and the exact structural proof.
\end{proofroadmap}

\section{Path transport, holonomy, and global sections}
\index{transport!path transport}
\index{transport!cycle holonomy}

For a path $p=e_1\cdots e_m$, define the partial composition
\[
  T_p=T_{e_m}\circ\cdots\circ T_{e_1}
\]
on the branches for which every intermediate image is defined.

For a cycle $c$ based at root $r$, the holonomy is
\[
  H_c=T_c:D(H_c)\subseteq\mathcal B_r\to\mathcal B_r.
\]
A branch is:
\begin{itemize}
  \item fixed if $H_c(k)=k$;
  \item moved if $H_c(k)$ is defined and different from $k$;
  \item undefined if it disappears along the cycle.
\end{itemize}

\paragraph{Global sections.}
\index{global section}
\index{topology!cycle consistency}

A \emph{global labeled section} is a tuple
\[
  s=(s_v)_{v\in V},\qquad s_v\in\mathcal B_v,
\]
such that, for every oriented edge $e=(u,v)$, one has
$s_u\in D_e$ and $T_e(s_u)=s_v$.  Thus the definition is intrinsic to the edge
transports and does not depend on a root or spanning tree.

Fix a spanning tree and let $p_v$ be the unique tree path from root $r$ to
vertex $v$.  The next theorem
characterizes the global labeled sections by the root branches that survive
every path and are fixed by every fundamental-cycle holonomy.

\begin{conditionbox}
The graph is finite and connected, reverse transports are partial inverses, a root and spanning tree are fixed, and partial compositions are evaluated on their natural domains.  A global section must use one common root branch that survives every path and satisfies every chord constraint.
\end{conditionbox}

\begin{theorem}[Cycle-space characterization]
\label{thm:cycle-space}
The set of global labeled sections is in bijection with
\[
  \mathcal S_r
  =
  \left\{
  k:
  k\in\bigcap_vD(T_{p_v}),
  \quad H_c(k)=k\text{ for every fundamental cycle }c
  \right\}.
\]
\end{theorem}

\begin{interpretationbox}
Global labeled sections are exactly the common fixed branches of the fundamental holonomies that survive all root paths.  The count and the missing-versus-moved distinction are invariant under local label permutations.  Checking each cycle for some fixed point is insufficient because different cycles may fix different branches.
\end{interpretationbox}

\begin{proofroadmap}
Transport a candidate root branch along the spanning tree and use each chord holonomy to verify the remaining edges.  Conversely, recover the root branch from any global section.  Conjugate all path transports and holonomies under local relabeling.  The complete bijection and recovery proof is in the chapter appendix.
\end{proofroadmap}

Checking each cycle only for the existence of some fixed branch is insufficient.  Different cycles may fix different branches while the common intersection is empty.
Cycle-space diagnostics are naturally related to combinatorial Hodge methods
\citep{JiangEtAl2011}; the theorem above additionally carries the domains of
partial maps, so an undefined branch is not conflated with nontrivial
permutation holonomy.
Local-to-global consistency and global sections have a broad applied-sheaf
and cellular-sheaf lineage \citep{Curry2014,Robinson2017,
HansenGhrist2019}.  Constraint-network theory likewise long predates this
example and already separates local consistency from global satisfiability
\citep{Montanari1974,Freuder1978}.  The theorem's scoped contribution is the
spanning-tree/fundamental-cycle characterization for the declared
partial-bijection domains and its gauge-invariant branch accounting.
This local-to-global consistency question is also adjacent to phase unwrapping,
where locally observed phase increments must integrate around cycles
\citep{Itoh1982,GoldsteinZebkerWerner1988,Costantini1998}.  Partial branch
transport has the extra possibility that a label becomes undefined, so phase
closure alone cannot represent the full obstruction here.

\paragraph{Gauge invariance.}
\index{transport!gauge invariance}

Under local relabeling $\pi_v$,
\[
  \widehat T_{uv}
  =
  \pi_vT_{uv}\pi_u^{-1}.
\]
Path transports transform similarly, and root holonomies satisfy
\[
  \widehat H_c
  =
  \pi_rH_c\pi_r^{-1}.
\]
Therefore fixed/moved/undefined counts, existence of global sections, and the number of global sections are gauge invariant.

\section{Multi-cycle witness, atlas construction, and uncertainty}
\index{cycle space}
\index{transport!multi-cycle witness}

The \OAI{} v0.5 companion construction has 28 vertices, 30 edges, and cycle
rank three.  Each fundamental cycle individually has at least one fixed
branch, yet the common global-section set is empty.  Two cycles also contain
undefined branches.  A two-chart atlas covers all graph cells.
The specific instance is a diagnostic illustration; the logical phenomenon
that individually consistent local constraints may have no common global
solution is not claimed as novel.

\begin{figure}[t]
\centering
\includegraphics[width=0.84\textwidth]{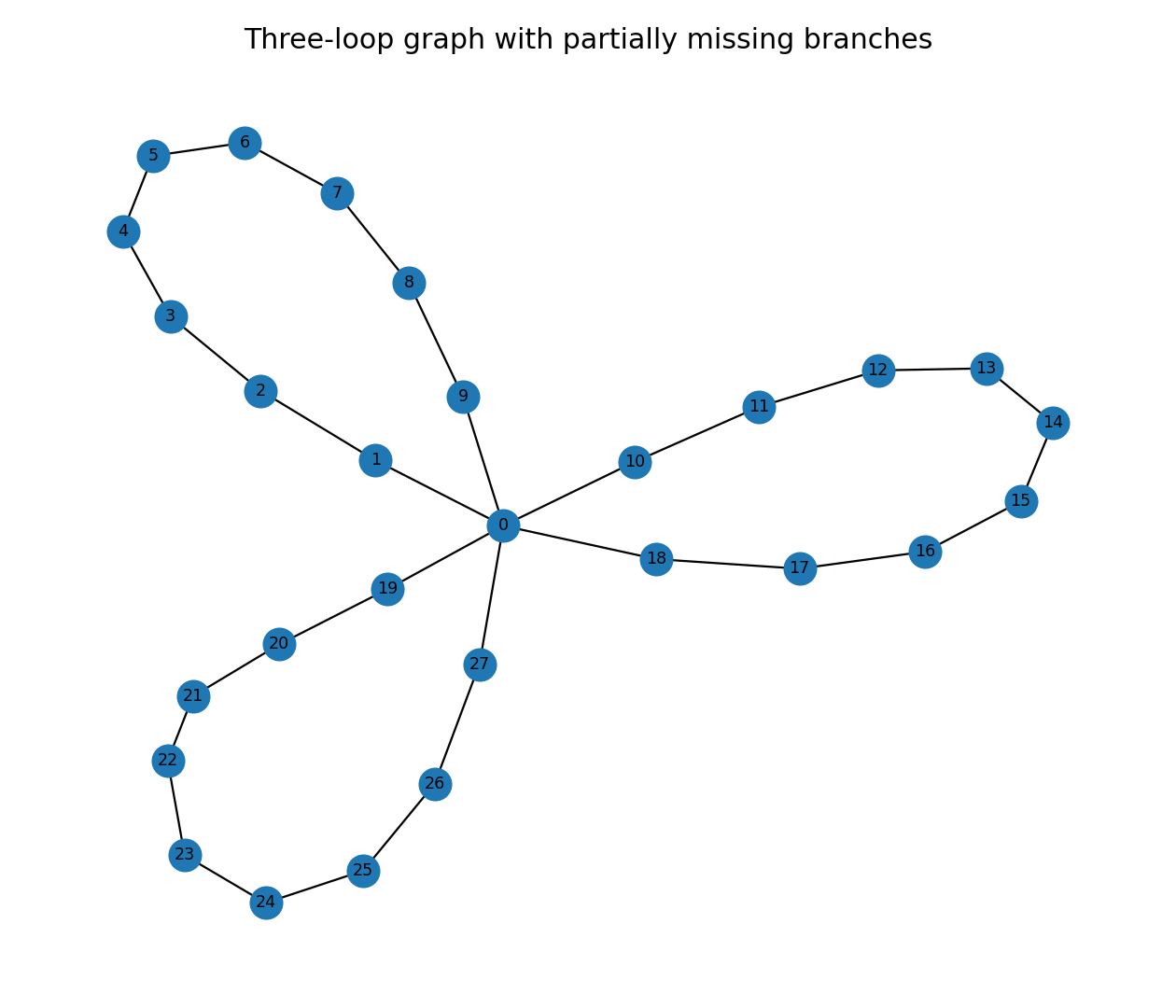}
\caption{Synthetic multi-cycle graph illustrating partial transports, cycle-space obstruction, and atlas coverage.}
\label{fig:multicycle-graph}
\end{figure}

\paragraph{Atlas construction on a graph.}
\index{atlas!graph construction}

This is an analytic companion construction, not a real-data scientific
result.  A chart is a connected subgraph on which a consistent local labeling
exists.  A greedy construction removes obstructed cycle edges until the
remaining subgraphs admit sections and checks vertex and edge coverage; it
does not claim globally minimal chart number.  Unlike the radical-loop audit
below, this release ships no executable artifact for the multi-cycle example.

\paragraph{Statistical uncertainty.}
\index{transport!statistical uncertainty}

The exact recovery theorem reduces the statistical bridge to local representative radii and assignment margins.  If the margin condition fails, the correct output is unresolved transport rather than a forced edge match.

\section{Deterministic code-path audit of the radical repair}
\label{sec:radical-oali-validation}
\index{radical normal form!\OAI{} validation}
\index{common deployment!radical validation}

Section~\ref{sec:radical-common-witness} exhibited the object that the generic
workflow still owed: one fixed repaired architecture that works throughout a
confidence world, rather than a different root field for every compatible
loop.  A small executable audit checks the corresponding formulas and code
paths.  It is not an independent fitted-model or held-out population
validation.

The frozen square-root family is

\[
  z(\theta)
  =
  r\exp\!\left[
    i\{\ell\theta+\beta+\gamma\sin(\theta+\omega)\}
  \right],
  \qquad \theta\in[0,2\pi].
\]
The audit draws $\ell$ uniformly from $\{-3,\ldots,3\}$ and draws
$r,\beta,\gamma,\omega$ independently and uniformly from
\[
  [0.75,1.25],\qquad[-\pi,\pi],\qquad[-0.35,0.35],
  \qquad[0,2\pi],
\]
respectively.  Writing
$\phi(\theta)=\ell\theta+\beta+\gamma\sin(\theta+\omega)$ gives
\[
  |\phi'(\theta)|
  =|\ell+\gamma\cos(\theta+\omega)|
  \le |\ell|+|\gamma|
  \le 3.35,
\]
so the stated public constant is $L_{\max}=3.35$ throughout the sampling
family.

The deployed input contains \(u=z(\theta)\), the loss is
\(|a^2-u|^2\), and a repaired point output retains its chart label.  The two
slit-plane charts and their router are fixed before any loop is drawn.  They
receive \(u\), not the hidden tuple
\((\ell,r,\beta,\gamma,\omega)\), so the same atlas is evaluated in every
world.

For the certificate check, 1,024 equally spaced contexts are observed with
exact complex target values.  The public uniform bound
\(|\phi'|\le L_{\max}=3.35\) keeps every adjacent phase increment below
\(\pi\); the certificate code does not receive the generating values of
\(\ell\) or \(\gamma\).  Summing principal increments therefore gives a
singleton degree identification set.  One seeded pool of 240 generated loops
exercises the routine, the fixed atlas, and the full-root-set rule.

The script also constructs a \emph{worldwise oracle one-chart witness}.  It
reads the generator's degree and phase, returns an exact root in even degree,
and returns zero in odd degree.  It is therefore an analytic benchmark for the
zero and \(r^2\) identities, not a deployable common one-chart baseline.  No
parameter is fitted, so splitting the generated loops into calibration and
test subsets would create a false impression of statistical independence
doing inferential work.

\begin{table}[H]
\centering
\small
\caption{Radical-loop common-witness deterministic code-path audit.}
\label{tab:radical-oali-audit}
\begin{tabularx}{\textwidth}{
  >{\raggedright\arraybackslash}p{0.31\textwidth}
  >{\raggedright\arraybackslash}p{0.18\textwidth}
  >{\raggedright\arraybackslash}X}
\toprule
Check & Audit units & Result\\
\midrule
degree certificate
  & 240 loops
  & all singleton sets recovered the generating degree; the certified adjacent-increment bound was \(0.02056<\pi\)\\
\addlinespace[0.25em]
fixed two-chart common witness
  & 240 loops
  & maximum recorded native defect \(1.79\times10^{-30}\)\\
\addlinespace[0.25em]
full-root-set common witness
  & 240 loops
  & maximum recorded native defect \(1.79\times10^{-30}\)\\
\addlinespace[0.25em]
even-degree oracle identity
  & 101 loops
  & maximum worldwise oracle-witness defect \(5.55\times10^{-31}\)\\
\addlinespace[0.25em]
odd-degree oracle identity
  & 139 loops
  & maximum residual from the analytic \(r^2\) identity \(6.67\times10^{-16}\)\\
\addlinespace[0.25em]
collapsed-root ablation
  & 240 loops
  & averaging the two roots returns zero; maximum residual from the analytic \(r^2\) identity \(6.67\times10^{-16}\)\\
\addlinespace[0.25em]
router trace
  & 240 loops
  & chart identifiers are aggregated from the per-context router outputs, not asserted as a literal constant\\
\bottomrule
\end{tabularx}
\end{table}

The numerical residuals in Table~\ref{tab:radical-oali-audit} are
floating-point checks of analytic identities.  There is no Hoeffding bound:
the script fits nothing, and the odd-degree benchmark already uses generator
information to instantiate its worldwise oracle witness.  The run therefore
audits implementation paths and artifact integrity; it neither estimates a
population improvement nor replaces the structural proofs.

A random seam is deliberately not called a negative control.  Moving a valid
branch cut is a gauge choice in this exact family and yields another correct
atlas.  The mechanism-destroying ablation is instead to erase the retained
root distinction: averaging \(a\) and \(-a\) produces zero and pays the full
\(r^2\) catastrophe tax.  The even-winding control checks the other direction:
the procedure must not report a one-chart obstruction when continuation
closes.

The complete standard-library-only entry point, unit tests, per-loop output,
summary, and content hashes are under
\path{reproducibility/radical_loop/}.  This is a deterministic code-path
audit with exact target values and a known regularity bound.  It is not
independent evidence for a fitted architecture and does not validate noisy
degree recovery, a learned router, neural optimization, population
generalization, or an application effect.  Those remain separate gates rather
than being inferred from the synthetic closure.

\section[Closing the radical loop]{Closing the radical loop: from native defect to validated repair}
\label{sec:radical-seven-step-closure}
\index{radical normal form!seven-step closure}
\index{certification ladder!radical loop}
\index{obstruction-aware learning and inference!closed-loop validation}

The radical example is the book's longitudinal audit, whereas the six
laboratories in Chapters~\ref{ch:laboratories} and
\ref{ch:applied-laboratories} are a basis and its application-level
recombinations.  Table~\ref{tab:radical-seven-step-closure} records the
logical dependency order: Part III closes the defect-consistency and
lift-validity gates, and Part IV then formulates the covering-space
obstruction.  The remaining rows add task meaning, finite-data permission,
and certified intervention.
Every row therefore passes forward a typed object together with the gate that
keeps it valid.

\begin{table}[H]
\centering
\footnotesize
\caption{The radical loop as a seven-step certification chain.}
\label{tab:radical-seven-step-closure}
\begin{tabularx}{\textwidth}{
  >{\raggedright\arraybackslash}p{0.205\textwidth}
  >{\raggedright\arraybackslash}p{0.335\textwidth}
  >{\raggedright\arraybackslash}X}
\toprule
Stage and location & Certified object & Handoff and non-negotiable boundary\\
\midrule
\textbf{1. Native defect}\newline
Parts I--IV; Chs.~\ref{ch:deployability}, \ref{ch:conjugate-lifts}, and
\ref{ch:singular}
  & With continuous single-valued point output and uniform loss frozen,
    \(D_\theta(a)=|a^k-z(\theta)|^2\) is the objective-native loss scale and
    \(\inf_aD_\theta(a)=0\).
  & The same defect prices constructive upper witnesses and converse lower
    certificates.  A convenient parameter distance cannot replace it.\\
\addlinespace[0.35em]
\textbf{2. Lift}\newline
Parts III--IV; Chs.~\ref{ch:lift-complexity} and \ref{ch:singular}
  & The radical question is whether
    \(z:S^1\to\C^\star\) lifts through
    \(p_k(a)=a^k\); a lift exists exactly when \(k\mid\deg z\).
  & This covering lift is not permission for a target-calling auxiliary
    coordinate.  Any carrier that exposes \(u=z(\theta)\) must be declared
    and pass the lift-admissibility audit.\\
\addlinespace[0.35em]
\textbf{3. Architecture obstruction}\newline
Part IV; Ch.~\ref{ch:singular}
  & The failed lift becomes an exact native floor: for the round loop,
    Theorem~\ref{thm:radical-tax} gives \(0\) when \(k\mid\ell\) and \(r^2\)
    otherwise.
  & The claim is for the declared continuous, one-chart, point-valued class
    under uniform risk.  Average risk, measurable output, an atlas, or a
    quotient is a different contract.\\
\addlinespace[0.35em]
\textbf{4. Visibility}\newline
Parts IV--V; Chs.~\ref{ch:repairs} and
\ref{ch:operational-semantics}
  & A labeled root exposes monodromy; the full root set or an appropriate
    quotient can erase it.  Legal contexts decide whether that distinction
    remains task-observable.
  & Visibility is a semantic gate, not another scalar tax.  A change of
    output meaning or context class must be declared before certification.\\
\addlinespace[0.35em]
\textbf{5. Confidence world}\newline
Part VI; Ch.~\ref{ch:statistical-eg}
  & Finite data determine a compatible degree set and hence a
    feasible--impossible--unresolved certificate; regularity and coverage can
    collapse that set to one degree.
  & Hidden winding may not be guessed.  Without adequate regularity or
    coverage, unresolved is the honest result, and every conclusion must hold
    throughout \(C_t\).\\
\addlinespace[0.35em]
\textbf{6. Common witness}\newline
Part VII; Ch.~\ref{ch:certified-learning}
  & When \(C_t\) stays away from zero and deployment exposes \(u\), the fixed
    rule \(W_k(u)\), or the fixed two-chart atlas with retained chart label,
    belongs to every world's repaired feasible set.
  & Worldwise divisibility alone does not produce one common
    \(\theta\mapsto a(\theta)\).  If \(u\) is hidden or a global labeled root
    is required, these witnesses are inadmissible.\\
\addlinespace[0.35em]
\textbf{7. \OAI{} code-path audit}\newline
Part VII; Chs.~\ref{ch:oali-workflow} and \ref{ch:transport-cycles}
  & The failed one-chart gate returns the atlas repair; its charts and router
    are fixed and exercised on one seeded loop pool, with even-degree,
    odd-degree, and collapsed-root identity checks.
  & This exact synthetic audit checks formulas and code paths.  It is not a
    held-out performance experiment and does not validate noisy recovery,
    learned routing, neural optimization, or application efficacy.\\
\bottomrule
\end{tabularx}
\end{table}

The loop closes only if Step 7 tests the same common witness authorized in
Step 6.  That witness is evaluated in the native loss scale fixed in Step 1,
against the obstruction diagnosed in Step 3, after the lift and visibility
gates of Steps 2 and 4 and the finite-data permission of Step 5.  No later
step may retroactively change an earlier contract.  Chapter~\ref{ch:applied-laboratories}
recombines these mechanisms in broader systems; it is not an eighth arrow in
the audit.

Part VIII is not an eighth inferential step.  It stress-tests the seven certification arrows.
The chain reopens at a named gate if uniform risk is replaced by average risk,
clearance from the discriminant vanishes, regularity or coverage is absent,
output semantics change, or a deterministic identity audit is relabeled as
held-out evidence.  In those cases the correct response is a narrower claim, a
new certificate, or unresolvedness---not a declaration that the loop has
closed.

\section*{Exercises}

\begin{exercise}
Construct three cycle holonomies on four root branches such that every cycle has a fixed point but their common fixed-point set is empty.
\end{exercise}

\begin{exercise}
Prove gauge invariance of the global-section count.
\end{exercise}

\begin{exercise}
For two local branch sets of sizes two and three, enumerate all structural partial matchings and compute the exact second-best margin for a chosen cost matrix.
\end{exercise}

\input{chapter_appendices/ch23_proofs}

%% file: chapter_appendices/ch23_proofs.tex
\chapterproofappendix

\subsection*{A finite-class stability lemma}
\proofdependency{The lemma uses only finiteness of the candidate set and a strict population margin.  It is the deterministic device that turns a uniform cost perturbation bound into exact structural recovery.}

\begin{lemma}[Finite-class argmin stability]
Let $\mathcal T$ be finite.  Suppose $C:\mathcal T\to\mathbb R$ has a unique minimizer $\tau^\star$ and margin
\[
  \Gamma=\min_{\tau\ne\tau^\star}\{C(\tau)-C(\tau^\star)\}>0.
\]
If
\[
  \sup_{\tau\in\mathcal T}|\widehat C(\tau)-C(\tau)|\le r
  \qquad\text{and}\qquad
  \Gamma>2r,
\]
then $\tau^\star$ is the unique minimizer of $\widehat C$.
\end{lemma}

\begin{proof}
For any $\tau\ne\tau^\star$,
\begin{align*}
\widehat C(\tau)-\widehat C(\tau^\star)
&=\{\widehat C(\tau)-C(\tau)\}
 +\{C(\tau)-C(\tau^\star)\}
 +\{C(\tau^\star)-\widehat C(\tau^\star)\}\\
&\ge -r+\Gamma-r
=\Gamma-2r>0.
\end{align*}
Thus every competitor has strictly larger estimated cost.
\end{proof}

\subsection*{Proof of noisy partial-transport recovery}
\proofdependency{The local fibers are finite; the population structural optimum is unique after dummy-label permutations are quotiented out; each estimated representative lies within its declared radius after a local relabeling; and the pair cost is the ambient metric while the unmatched penalty is held fixed.}
\proofboundary{The conclusion is exact only under a strict margin.  When the margin is not separated from the perturbation radius, the theorem requires an unresolved edge rather than a forced matching.}

\begin{proof}
Fix an edge $e=\{u,v\}$ and align the estimated labels with the population labels through $\pi_u$ and $\pi_v$.  Let $\tau$ be any injective partial matching.  Its population cost is
\[
  C_e(\tau)
  =\sum_{k\in D(\tau)}
    d(a_{uk}^\star,a_{v,\tau(k)}^\star)
   +\lambda_e\{K_u+K_v-2|D(\tau)|\}.
\]
The estimated version of the same structural matching, written in the estimated labels, is $\pi_v\tau\pi_u^{-1}$.  The unmatched term is identical in the two costs.  For each matched pair, the triangle inequality gives
\begin{align*}
&\left|
 d(\widehat a_{u,\pi_u(k)},\widehat a_{v,\pi_v(\tau(k))})
 -d(a_{uk}^\star,a_{v,\tau(k)}^\star)
\right|\\
&\qquad\le
 d(\widehat a_{u,\pi_u(k)},a_{uk}^\star)
 +d(\widehat a_{v,\pi_v(\tau(k))},a_{v,\tau(k)}^\star)
 \le \varepsilon_u+\varepsilon_v.
\end{align*}
Because a partial matching contains at most
$m_e=\min(K_u,K_v)$ pairs,
\[
  \left|
  \widehat C_e(\pi_v\tau\pi_u^{-1})-C_e(\tau)
  \right|
  \le m_e(\varepsilon_u+\varepsilon_v)
\enspace\text{for every }\tau.
\]
Apply the finite-class stability lemma with
$r=m_e(\varepsilon_u+\varepsilon_v)$.  The assumed inequality
\[
  \Gamma_e>2m_e(\varepsilon_u+\varepsilon_v)
\]
preserves the strict ordering of the population optimum against every structural competitor.  Hence
\[
  \widehat\tau_e=\pi_v\tau_e^\star\pi_u^{-1}
\]
is the unique estimated optimum.  Conjugation by the two local permutations transports the graph of the matching bijectively, so its domain, range, matched pairs, and unmatched branches are all recovered up to the unavoidable local labels.
\end{proof}

\subsection*{Proof of the cycle-space characterization}
\proofdependency{The graph is finite and connected; reverse edge transports are partial inverses; a root and spanning tree have been fixed; and all compositions are interpreted on their natural partial domains.}
\proofboundary{It is not enough that each fundamental cycle possess some fixed branch.  A global section requires one root branch that survives every tree path and is fixed by every chord holonomy simultaneously.}

\begin{proof}
Let $\mathcal T$ be the fixed spanning tree and $r$ its root, and let $p_v$
be the unique tree path from $r$ to $v$.

\paragraph{From an admissible root branch to a global section.}
Take $k\in\mathcal S_r$ and define
\[
  s_v=T_{p_v}(k),\qquad v\in V.
\]
The domain condition in the definition of $\mathcal S_r$ guarantees that every
$s_v$ is defined.  Compatibility on every tree edge is automatic because the
$T_{p_v}$ are tree-path compositions.

Now let $e=(u,v)$ be a chord, oriented from $u$ to $v$, and let $c_e$ be its
fundamental cycle.  The root holonomy is
\[
  H_{c_e}=T_{p_v}^{-1}\circ T_e\circ T_{p_u}.
\]
Since $H_{c_e}(k)=k$, all terms in this composition are defined and
\[
  T_{p_v}^{-1}\{T_e(T_{p_u}(k))\}=k.
\]
Applying $T_{p_v}$ on its domain yields
\[
  T_e(s_u)=s_v.
\]
Thus the tuple $(s_v)_{v\in V}$ is compatible with every edge and is a global labeled section.

\paragraph{From a global section to an admissible root branch.}
Conversely, let $s=(s_v)_{v\in V}$ be a global section.  Compatibility along the unique tree path from $r$ to $v$ implies inductively that
\[
  s_v=T_{p_v}(s_r),
\]
so $s_r$ survives every tree path.  Compatibility on a chord $(u,v)$ gives
\[
  T_e\{T_{p_u}(s_r)\}=T_{p_v}(s_r).
\]
Applying $T_{p_v}^{-1}$ shows $H_{c_e}(s_r)=s_r$.  Hence
$s_r\in\mathcal S_r$.  The two constructions are inverse because a global section is uniquely determined by its root value.  This proves the bijection.

Because the set of global sections is defined without reference to a root or spanning tree, its existence and cardinality are independent of those choices.

\paragraph{Gauge invariance.}
Let local relabelings $\pi_v$ act on the edge transports by
\[
  \widehat T_e=\pi_vT_e\pi_u^{-1}.
\]
For a one-edge tree path the corresponding relation for path transport is immediate.  If it holds for the path to $u$, composition with the next edge gives
\[
  \widehat T_{p_v}
  =\widehat T_e\widehat T_{p_u}
  =\pi_vT_e\pi_u^{-1}
    \pi_uT_{p_u}\pi_r^{-1}
  =\pi_vT_{p_v}\pi_r^{-1}.
\]
Induction proves this relation for every vertex.  Substitution into the root-holonomy formula yields
\begin{align*}
\widehat H_{c_e}
&=(\widehat T_{p_v})^{-1}\widehat T_e\widehat T_{p_u}\\
&=\pi_rT_{p_v}^{-1}\pi_v^{-1}
  \pi_vT_e\pi_u^{-1}
  \pi_uT_{p_u}\pi_r^{-1}\\
&=\pi_rH_{c_e}\pi_r^{-1}.
\end{align*}
Conjugation by $\pi_r$ is a bijection between domains, undefined complements, fixed-point sets, and moved sets.  It also sends $\mathcal S_r$ to the estimated admissible set.  Therefore all their cardinalities and the existence of a global section are gauge invariant.

Finally, on the simultaneous edge-recovery event, every estimated edge transport is exactly a gauge conjugate of its population counterpart.  The preceding path and holonomy identities then show that the complete estimated cycle audit---including undefined branches, fixed branches, moved branches, and the common global-section set---is recovered exactly up to root relabeling.
\end{proof}

%% file: chapters/ch24_applied_laboratories.tex
\chapter{Four Applied Laboratories}
\label{ch:applied-laboratories}
\input{reproducibility/ch24/results/mnist_amortization/generated_results}
\input{reproducibility/ch24/results/digits_sparse_repair_r2_equal_info/generated_results}
\input{reproducibility/ch24/results/diabetes_sparse_repair_r1/generated_results}
\input{reproducibility/ch24/results/diabetes_sparse_repair_r1/theorem_alignment}
\section{How the laboratories close the loop}
\index{applied laboratories}
\index{obstruction-aware learning and inference!applications}

Chapter~\ref{ch:laboratories} isolated four pure mechanism directions:
oracle survival under coarse-graining, semantic repair for graph-indexed
probability fields, coordination under shared conditional deployment, and
resource-indexed expansion from one-pass prediction to local refinement.
Chapters~\ref{ch:conjugate-lifts} and~\ref{ch:composition} supplied exact and
soft base-change decompositions; Chapters~\ref{ch:oali-workflow}
and~\ref{ch:transport-cycles} supplied a workflow and its transport machinery.
The examples below are neither one-to-one sequels nor new theories.  Each
recombines several mechanisms inside a more complex scientific contract and
occupies a different point on the certification ladder.

\begin{decompositionbox}[title=From pure mechanisms to composite laboratories]
\textbf{Amortized and sparse inference: one-pass sharing plus test-time
refinement.}  The MNIST study is a five-run held-out audit whose scaling
curves do not establish saturation.  A frozen digits control upgrade compares
repairs at equal native information and equal computation.  A separately
frozen patient-level diabetes benchmark carries the same contract to a
scientific dataset, preserves a failed WDBC precursor in the evidence ledger,
and changes the deployed architecture only after all confirmatory gates pass.

\textbf{Posterior-family Monte Carlo: carrier reuse plus certified base
change.}  A shared augmentation or particle cloud may discard target-specific
simulation work only if the requested posterior functional remains
recoverable.  Uncorrected sharing pays stationary-law defect, exact kernel
correction pays rejection and mixing variance, and exact-ratio reuse pays
finite-particle error through overlap.  Current evidence is an exactly
enumerable Gibbs family plus a fixed analytic Gaussian-posterior pilot.

\textbf{Multi-sample cytometry: carrier survival plus graph compatibility.}
Local summaries must retain rare populations while partial transports align
the surviving branches and pass cycle audits.  Current evidence is a
synthetic smoke test plus a confirmatory protocol.

\textbf{Sequential systems: dynamic coarse-graining plus coordination.}
Memory merges histories while one shared system must coordinate the
distinctions that remain across occupancy and downstream contexts.  Exact
finite interfaces are available; broad systems extensions remain open.
\end{decompositionbox}

The comparison enforces one editorial rule: an analytic mechanism check, a
working software path, and a completed real-data claim are different kinds
of evidence.
It also keeps the operation label separate from the scientific contract:
P/G/X/V/C records how a declared field, surrogate, objective, or joint law is
handled, while an atlas or quotient changes the contract itself.

\section{Laboratory I: one-pass inference audits}
\sectionmark{Laboratory I: one-pass inference}
\index{MNIST!amortization audit}
\index{amortized inference!finite-grammar audit}
\index{test-time refinement!MNIST audit}
\subsection{MNIST amortization: frozen but incomplete}
\label{sec:lab-mnist-amortization}

This laboratory asks a deliberately narrow real-data question.  Fix a trained
VAE decoder and the diagonal-Gaussian variational family.  For image \(x\),
let \(J_x(a)\) be the negative evidence lower bound at variational parameter
\(a\), and let \(\widetilde a_x\) be a high-accuracy imagewise refit.  The
audited native defect of a deployed encoder \(A\) is
\[
  \widehat\Def_x\{A(x)\}
  =J_x\{A(x)\}-J_x(\widetilde a_x).
\]
The baseline contract \(\Arch_0\) is a frozen finite library of trained
one-pass encoders.  The repaired contract \(\Arch_T\) starts from the selected
encoder and permits \(T\) imagewise optimization steps.  Because test-time
steps are declared as a resource coordinate, the comparison
\(\Arch_0\subset\Arch_T\) is a deployment-contract expansion rather than a
claim that the same computation was merely implemented more carefully.

\paragraph{Frozen design and ordinary-error controls.}
Before the confirmatory results were generated, the protocol froze five
decoder-training seeds, a 55,000-image fitting split, a 5,000-image validation
split, and the same \(\MNISTTestImages\) held-out official test images per
run.  Validation alone selects among seven one-pass recipes varying training
sample size, epoch count, and hidden width; the official test images do not
select the encoder.  The imagewise reference starts from both the best
library output and the standard-normal variational parameter, continues each
start by 800 Adam and 30 L-BFGS steps, and retains the lower objective.  Every
run passed the predeclared reference-dominance and gradient-residual gates,
with no fallback invocation.

Table~\ref{tab:ch24-mnist-audit} compares changes inside the one-pass contract
with a test-time resource expansion.  Increasing data, epochs, and width
reduces the mean gap throughout the tested ranges.  The curves therefore do
not establish a plateau or structural saturation; observing a positive gap at
the largest tested setting is not enough.  The
validation-selected encoder is the width-512 recipe in four runs and the
width-256 recipe in one.

\begin{table}[H]
\centering
\caption{Prospectively frozen MNIST audit.  The first three rows report mean
test amortization gaps in nats across five independently trained decoders.
The final two rows are the seed-level primary endpoints; parentheses contain
the seed standard deviation and confidence bounds are one-sided 95\% Student
\(t\) bounds.}
\label{tab:ch24-mnist-audit}
\small
\begin{tabularx}{\textwidth}{@{}
  >{\raggedright\arraybackslash}p{0.20\textwidth}
  >{\raggedright\arraybackslash}p{0.30\textwidth}
  >{\raggedright\arraybackslash}p{0.23\textwidth}
  >{\raggedright\arraybackslash}X@{}}
\toprule
Audit component & Frozen settings & Held-out result & Interpretation\\
\midrule
Training data
  & 10,000 / 30,000 / 55,000 images
  & \(\MNISTDataGaps\) nat
  & continued improvement; no saturation claim\\
Training effort
  & 4 / 12 / 24 epochs
  & \(\MNISTEpochGaps\) nat
  & continued improvement; no saturation claim\\
Encoder width
  & 64 / 256 / 512 hidden units
  & \(\MNISTWidthGaps\) nat
  & continued improvement; no saturation claim\\
Selected one-pass gap
  & validation-selected finite library
  & \(\MNISTGapMean\) (\(\MNISTGapSD\)) nat; lower bound \(\MNISTGapLower\)
  & passes \(>0.5\)-nat gate\\
Eight-step residual ratio
  & uniform \(T=8\) refinement
  & \(\MNISTRefineMean\) (\(\MNISTRefineSD\)); upper bound \(\MNISTRefineUpper\)
  & passes \(<0.80\) in \(\MNISTSeedPasses/\MNISTSeeds\) runs\\
\bottomrule
\end{tabularx}
\end{table}

\paragraph{Observed gap and refinement comparison.}
The selected one-pass gap averages \(\MNISTGapMean\) nats, with a one-sided
95\% lower bound of \(\MNISTGapLower\).  Uniform refinement leaves residual-gap
ratios \(\MNISTRefinementCurve\) after \(T=1,4,8,16\) steps.  At the
predeclared eight-step endpoint it therefore removes about
\(\MNISTRemovedPercent\%\) of the measured defect.  Of
\(\MNISTCurvatureTotal\) audited encoder--reference segments,
\(\MNISTCurvaturePositive\) have positive minimum curvature on the declared
seven-node grid, and all \(\MNISTBracketCovered\) such paths satisfy the local
curvature bracket.  This is a local diagnostic, not a global-convexity claim.

These results establish that additional imagewise optimization lowers the
sampled variational objective relative to the selected one-pass encoder.  They
do not identify one-pass sharing as the dominant mechanism.  The same Monte
Carlo draw collection is used to construct the numerical reference, guide
refinement and curvature calculations, and evaluate the final gaps; hence the
reported endpoint is a fixed sample-average objective rather than an
independently evaluated population ELBO.  The study also lacks a
compute-matched generic optimizer, random or generic mixture controls, and an
independent downstream endpoint.  It is therefore an incomplete OALI
laboratory, not a successful certificate--saturation--repair validation.
It is not an OAI atlas experiment: no branch transport, holonomy, or chart
selection is involved.  Adaptive allocation has mean
adaptive/uniform ratio \(\MNISTAdaptiveMean\), but its one-sided upper bound
\(\MNISTAdaptiveUpper\) does not establish a stable one-percent advantage over
uniform allocation; that secondary claim is withheld.

\paragraph{Reproducibility record.}
The book snapshot and compact machine-readable summaries are under
\path{reproducibility/ch24/}; within that directory,
\path{verify_mnist_audit.py} checks the archived seed summaries, primary
decisions, generated TeX numbers, frozen-source hashes, and shipped checksums.
Full retraining remains a separate, explicitly requested run; it is not
hidden inside the normal book build.

\begin{boundarybox}
This result is conditional on the frozen decoder, diagonal-Gaussian local
family, seven-recipe one-pass library, optimization protocol, fixed MNIST
audit set, and reused Monte Carlo sample-average objective.  It establishes a
positive measured gap for the validation-selected encoder and shows that
imagewise refinement reduces that same measured objective.  It does not
establish saturation of the one-pass class, a mechanism-exclusive structural
obstruction, independent population-ELBO improvement, superiority to
compute-matched generic alternatives, improvement in classification or
another downstream task, or a population confidence interval for all
handwritten digits.  Accordingly it cannot by itself justify an architecture
choice.
\end{boundarybox}

\subsection{Handwritten digits: equal-information, equal-compute sparse repair}
\label{sec:lab-digits-sparse-repair}
\index{sparse coding!frozen architecture repair}
\index{architecture repair!compute-matched validation}

The preceding MNIST result must remain incomplete: its archived outputs cannot
retroactively supply independent Monte Carlo evaluation or a saturation
certificate.  The R1 digits protocol first closed a narrower frozen workflow,
but its generic control did not receive the native gradient used by the
proximal update.  It matched multiply--accumulates, not information.  Protocol
\texttt{DIGITS-SPARSE-OALI-R2-EQUAL-INFO-2026-08-17} froze a new control, the
common carrier, compute budget, fitted models, numerical gates, code,
environment record, and hashes before the control saw confirmatory outputs.
A fit-free acceptor then scored the frozen artifacts.  R2 reuses the R1 image
split, so it is a strict control upgrade rather than a fresh replication.

The data are the 1,797 handwritten $8\times8$ images shipped with
\texttt{scikit-learn}.  A nonnegative 32-atom dictionary $D$, learned on the
60\% training split and then frozen, defines the imagewise oracle
\[
 J_x(z)=\frac12\lVert Dz-x\rVert_2^2+0.02\lVert z\rVert_1,
 \qquad z\ge 0.
\]
The baseline architecture is one thresholded-affine pass
$z_0(x)=[xW+b]_+$.  Ridge regularization and training-set compensation are
selected on a disjoint 20\% development split.  Both repairs receive the same
native carrier
\[
 \{z_0,\ L^{-1}\nabla J_x(z_0),\ D^\top D,\lambda,L\}.
\]
The mechanism-matched repair performs two proximal-gradient steps.  The
learned control first computes the same native gradient and then applies a
$64\!\to\!32\!\to\!32$ residual MLP.  Two native-gradient evaluations for
the targeted repair and one gradient plus the MLP for the generic control
each cost exactly \DigitsRepairMACs{} multiply--accumulates.  A two-step
fixed-coordinate permutation of the gradient supplies an equal-information,
equal-compute mechanism-destroying control.

\paragraph{Why the saturation statement is structural.}
For a strict active set $S$, the nonnegative Lasso oracle is locally affine,
with active-coordinate Jacobian
\[
 (D_S^\top D_S)^{-1}D_S^\top.
\]
In contrast, coordinate $j$ of a thresholded-affine encoder has one fixed
derivative whenever it is active.  The frozen audit searched for two strict
KKT neighborhoods sharing an active coordinate but having incompatible
oracle Jacobian rows.  It found \DigitsStrictPoints{} strict points among
\DigitsSparseN{} confirmatory images and a maximum shared-row conflict of
\DigitsJacobianConflict.  Hence no one fixed thresholded-affine map can agree
with the oracle on both certified neighborhoods.  This conclusion is about
the declared one-pass family; it is not a lower bound for arbitrary ReLU
networks.

\begin{table}[H]
\centering
\caption{Digits R2 equal-information, equal-compute control upgrade.  Bounds
are one-sided 95\% paired bootstrap bounds over \DigitsSparseN{} confirmatory
images.  The table groups all 12 frozen gates; every gate passed.}
\label{tab:ch24-digits-sparse-closure}
\small
\begin{tabularx}{\textwidth}{@{}
  >{\raggedright\arraybackslash}p{0.20\textwidth}
  >{\raggedright\arraybackslash}p{0.25\textwidth}
  >{\raggedright\arraybackslash}p{0.24\textwidth}
  >{\raggedright\arraybackslash}X@{}}
\toprule
Link & Frozen gate & Confirmatory result & Decision role\\
\midrule
Protocol integrity (3)
 & protocol ID, source hashes, and confirmatory completeness
 & all three passed
 & the contract predates scoring\\
Operational gap (3)
 & gap lower bound $>0.004$, oracle dominance, proximal residual near zero
 & gap \DigitsBaselineGap; lower \DigitsBaselineGapLower
 & positive fitted-baseline defect; not a class infimum\\
Structural saturation (1)
 & strict-KKT Jacobian conflict $>0.05$
 & conflict \DigitsJacobianConflict
 & one fixed pass cannot hit both certified neighborhoods\\
Targeted repair (1)
 & residual-ratio upper bound $<0.25$
 & ratio \DigitsRepairRatio; upper \DigitsRepairRatioUpper
 & two-step mechanism-matched reduction\\
Equal-information/equal-compute control (1)
 & repair/generic upper bound $<0.50$
 & ratio \DigitsRepairGenericRatio; upper \DigitsRepairGenericUpper
 & rules out extra native information or computation alone\\
Negative control (1)
 & permuted-gradient lower bound $>1.05$
 & ratio \DigitsNegativeRatio; lower \DigitsNegativeLower
 & destroying the mechanism reverses the gain\\
Operational endpoints (2)
 & SSE upper change $<0$; accuracy change $>-0.02$
 & SSE \DigitsDistortionChange{} (upper \DigitsDistortionUpper); accuracy
 \DigitsBaselineAccuracy{} to \DigitsRepairAccuracy
 & no fidelity or recognition trade\\
\bottomrule
\end{tabularx}
\end{table}

The frozen acceptor returned
\texttt{ACCEPT\_ONE\_PASS\_PLUS\_TWO\_STEP\_PROXIMAL\_EQUAL\_INFO}.  It
trained no model and read only the predeclared thresholds, frozen hashes, raw
per-image table, and sealed summary.  Under the same information carrier and
deployment computation, the decision changes from a one-pass encoder to that
encoder plus two proximal steps.  The baseline gap is an operational defect
of one fitted rule, not the class infimum $\mathfrak O$; class-level failure
of exact saturation comes from the Jacobian conflict, not from the positive
gap alone.

\begin{boundarybox}
This closure is restricted to the frozen digits split, dictionary,
nonnegative $\ell_1$ objective, thresholded-affine grammar, and
\DigitsRepairMACs-MAC envelope.  R2 removes R1's clearest extra-information
explanation but uses images already scored in R1.  It is a frozen real-
benchmark control upgrade, not a fresh replication, external-person audit, or
scientific-domain validation.  Proximal gradient, active-set geometry, sparse
coding, and algorithm unrolling remain classical.
\end{boundarybox}

\subsection{Scientific case: patient-level diabetes progression repair}
\label{sec:lab-diabetes-sparse-repair}
\index{diabetes!patient-level architecture repair}
\index{architecture repair!scientific validation}

To test whether the same structural certificate can change an architecture
choice on independent scientific units, protocol
\texttt{DIABETES-SPARSE-OALI-R1-2026-08-17} was frozen prospectively.  The
data contain ten baseline variables and one-year disease-progression responses
for 442 patients.  Fixed 60/20/20 patient splits separate fitting,
development, and confirmation.  The fitting patients alone determine the
min--max transform, an eight-atom nonnegative dictionary, one-pass ridge
encoder, equal-information generic control, and downstream ridge progression
predictor.  The \DiabetesSparseN{} confirmatory patients were scored once,
after all source files, fitted models, thresholds, environment records, and
ten hashes were frozen.

The native objective is again nonnegative $\ell_1$ sparse reconstruction.
Here $m=10$ and $k=8$, so one native gradient costs
$mk+k^2=144$ multiply--accumulates and two proximal steps cost
\DiabetesRepairMACs{} MAC.  The generic control receives the identical
$(z_0,L^{-1}\nabla J)$ carrier and applies a $16\!\to\!6\!\to\!8$ MLP; its
one gradient plus MLP also costs \DiabetesRepairMACs{} MAC.  The two-step
permuted-gradient control has the same budget.  The scientific endpoint is
not the reconstruction objective under another name: it is disease-
progression MSE from a predictor fitted only on the training patients.

\begin{table}[H]
\centering
\caption{Frozen patient-level diabetes confirmation.  One-sided 95\% paired
bootstrap bounds treat the patient as the independent unit.  All 12 frozen
gates passed.}
\label{tab:ch24-diabetes-sparse-closure}
\small
\begin{tabularx}{\textwidth}{@{}
  >{\raggedright\arraybackslash}p{0.20\textwidth}
  >{\raggedright\arraybackslash}p{0.24\textwidth}
  >{\raggedright\arraybackslash}p{0.25\textwidth}
  >{\raggedright\arraybackslash}X@{}}
\toprule
Link & Frozen gate & Confirmatory result & Architecture role\\
\midrule
Protocol and numerical quality (5)
 & ID, ten hashes, 89 rows, oracle residual, oracle dominance
 & all five passed
 & excludes post-hoc refitting and a numerical pseudo-oracle\\
Operational gap (1)
 & baseline-gap lower bound $>10^{-4}$
 & gap \DiabetesBaselineGap; lower \DiabetesBaselineGapLower
 & positive native defect of the fitted one-pass rule\\
Structural saturation (1)
 & strict-KKT conflict $>0.02$
 & \DiabetesStrictPoints{} points; conflict \DiabetesJacobianConflict
 & one thresholded-affine pass cannot hit both certified neighborhoods\\
Targeted repair (1)
 & residual-ratio upper bound $<0.30$
 & ratio \DiabetesRepairRatio; upper \DiabetesRepairRatioUpper
 & large reduction in native defect\\
Equal-information/equal-compute control (1)
 & target/generic upper bound $<0.50$
 & ratio \DiabetesRepairGenericRatio; upper \DiabetesRepairGenericUpper
 & carrier and MAC alone do not explain the gain\\
Mechanism negative control (1)
 & target/permuted upper bound $<0.50$
 & ratio \DiabetesRepairNegativeRatio; upper \DiabetesRepairNegativeUpper
 & correct gradient coordinates matter\\
Independent endpoints (2)
 & SSE-change upper bound $<0$; progression-MSE change upper bound $<100$
 & SSE \DiabetesDistortionChange{} (upper \DiabetesDistortionUpper); MSE
 \DiabetesBaselineMSE{} to \DiabetesRepairMSE{} (change upper
 \DiabetesMSEChangeUpper)
 & native improvement without sacrificing the frozen scientific endpoint\\
\bottomrule
\end{tabularx}
\end{table}

The fit-free acceptor returned
\path{ACCEPT_DIABETES_ONE_PASS_PLUS_TWO_STEP_PROXIMAL_EQUAL_INFO}.
In the language of \cref{thm:book-deployment-conflict}, the observation law
and information carrier are held fixed while the deployable mechanism class
is enlarged inside the same compute envelope.  The theorem family is not a
post-hoc explanation of a favorable number: it dictated which actions,
information, regret scale, and controls had to be frozen before the final
choice changed.

\paragraph{Post-confirmation alignment with the quantitative depth theorem.}
The preceding R1 decision predates
\cref{thm:active-set-conflict-modulus,thm:active-set-repair-budget}; it cannot
be relabeled as a prospective test of them.  A separate fit-free audit now
recomputes their quantities from the sealed dictionary, oracle, one-pass
encoder, and confirmatory records.  The post-confirmation geometry file
freezes the binary sex coordinate, admits perturbations only in the other
nine coordinates, restricts them to the training-minmax box, and requires
the two centers to have the same frozen value.  Of the exact KKT centers,
\DiabDepthEligiblePoints{} have positive restricted radii inside that box.
The deterministic rule selects, among those pairs with different supports
and a shared active coordinate, the pair maximizing the restricted theorem
floor.  It selects confirmatory rows \DiabDepthLeftRow{} and
\DiabDepthRightRow{} (dataset rows \DiabDepthLeftDatasetIndex{} and
\DiabDepthRightDatasetIndex{}), shared atom \DiabDepthAtom{}, with radii
\DiabDepthLeftRadius{} and \DiabDepthRightRadius{}.  The restricted
Jacobian-row conflict is
\DiabDepthConflict{}, giving
\[
 \Gamma=\DiabDepthGamma,
 \qquad
 \frac{\kappa}{2}\Gamma^2=\DiabDepthNativeFloor,
 \qquad
 \kappa=\DiabDepthKappa.
\]
Thus every one-pass thresholded-affine encoder---not merely the fitted
baseline---exceeds native-loss tolerance
$\eta=\DiabDepthTolerance$ somewhere on the union of the two certified
restricted balls.  For the frozen baseline, an analytic restricted-ball
envelope gives
$q=\DiabDepthContraction$ and a sufficient repair depth of
\DiabDepthSufficientSteps{} proximal steps at that tolerance.  This is a
literal theorem-driven architecture recommendation: one pass is ruled out,
while the declared recurrent mechanism at the certified depth is guaranteed
to cross the native-loss threshold.

The new calculation also prevents two attractive overclaims.  The restricted
balls respect the declared box grammar and freeze the binary coordinate, but
the audit does not prove that every point in them lies in the support of the
patient population.  The lower bound is therefore uniform for the declared
input grammar, not a distributional lower bound over attainable patients.
The R1
two-step architecture was accepted on patient-average endpoints and
equal-information/equal-compute controls; it is not guaranteed to cross this
stronger uniform two-ball tolerance.  Conversely, the 24-step guarantee was
not a frozen R1 endpoint and is not charged against the 288-MAC comparison.
The two conclusions answer different deployment contracts and are reported
separately.

\paragraph{A failed precursor remains in the evidence ledger.}
The earlier frozen WDBC-R1 experiment passed 11 of 12 gates but failed its
preregistered requirement that a permuted gradient be worse than no repair.
The residual-ratio one-sided 95\% lower bound was $0.615$, below $1.05$.
Although the targeted repair beat the permuted repair, the all-gates acceptor
returned \texttt{REJECT\_ARCHITECTURE\_REPAIR\_CLAIM}.  Before touching its
fresh confirmatory patients, the diabetes protocol replaced that mechanism
question by a direct targeted/permuted comparison.  The WDBC failure was not
deleted or retrospectively reclassified.

\begin{boundarybox}
This is a prospectively frozen patient-level validation on a scientific
dataset, with a fit-free hash-checking acceptor separated from model fitting.
It is not an external-person or external-institution replication.  Disease-
progression MSE is a benchmark endpoint, not evidence of clinical utility,
treatment benefit, or causality.  The case instantiates the theorem's typed
contract and changes one frozen architecture decision; it does not prove the
Le Cam/Blackwell ingredients or establish a universal architecture-choice
theory.  The active-set depth certificate is a post-confirmation theorem-
alignment audit, not an independently frozen validation of the new theorem
family.  A future claim that the family itself changed a prospectively
accepted architecture must freeze its pair-selection rule, uniform tolerance,
repair depth, and compute budget before opening new confirmation units.
\end{boundarybox}

\section[Laboratory II: certified Monte Carlo reuse]{Laboratory II: certified Monte Carlo reuse\texorpdfstring{\\}{ }across posterior families}
\sectionmark{Laboratory II: Monte Carlo reuse}
\index{Monte Carlo!reuse across posterior families}
\index{posterior family!shared simulation}
\label{sec:lab-monte-carlo-reuse}

\subsection{The reuse contract}

Let \(\{\pi_t:t\in\mathcal T\}\) be posterior laws on a common latent or
parameter space, and let
\[
  \mu_t(f)=\int f(z)\,\pi_t(\dd z)
\]
be the declared oracle.  Independent simulation at every \(t\) preserves the
contract but ignores family structure.  A shared Monte Carlo carrier is useful
only when it retains what is needed to recover \(\mu_t(f)\): the relevant
functional, the source identity, and either a certified conditional-law
approximation or an evaluable source-to-target density ratio.  A large bank
without those oracle distinctions is storage, not effective capacity.

This is the applied counterpart of the first laboratory in
Chapter~\ref{ch:laboratories}.  Coarse-graining may safely forget the
random-number history and other target-irrelevant details of a source cloud,
but it may not forget the source law or information needed for target base
change.  The horizontal posterior family says which distinctions must
survive; the vertical Monte Carlo defect says what is paid when they do not.

\subsection{Three contracts and three diagnostic scales}
\index{Monte Carlo!three diagnostic scales}

For a two-block Gibbs family, write
\[
 P_t(x,A)=\int q_t(\dd z\mid x)r_t(A\mid z),\qquad
 \bar q_t=\sum_s\alpha_{ts}q_s,
\]
and let \(\widetilde P_t\) insert \(\bar q_t\) while retaining the target
update \(r_t\).  With
\[
 \epsilon_t=\sup_x\TV\{\bar q_t(\cdot\mid x),q_t(\cdot\mid x)\},
 \qquad \delta(P_t)\le\gamma_t<1,
\]
the elementary kernel-perturbation recursion gives
\[
 \TV(\widetilde\mu_t,\mu_t)
 \le \frac{\epsilon_t}{1-\gamma_t},\qquad
 |\widetilde\mu_t f-\mu_t f|
 \le \operatorname{osc}(f)\frac{\epsilon_t}{1-\gamma_t}
\]
whenever the two invariant laws exist \citep{Mitrophanov2005}.  If the
conditional field is twice differentiable and
\(m_{1,t}=\sum_s\alpha_{ts}(s-t)\), then
\(\epsilon_t\le L_1|m_{1,t}|+(L_2/2)\sum_s\alpha_{ts}(s-t)^2\).
One-sided reuse is therefore generically first order in the sharing radius;
a centered operator cancels the first-order term and is second order under
the stated smoothness.  The cancellation is a property of the sharing
operator, not a generic consequence of a smooth posterior functional.

An exact Metropolis--Hastings correction changes the relevant component.  It preserves
the target law when the acceptance ratio is exact, but stronger coupling can
appear as rejection and autocorrelation.  It should not be ranked against an
uncorrected kernel using stationary defect alone.

The third contract uses a particle bank.  From independent source clouds
\(Z_{sj}\sim\pi_s\), define
\[
 w_{ts}=\frac{\dd\pi_t}{\dd\pi_s},\qquad
 \widehat\mu_{t\leftarrow s}(f)
 =\frac{\sum_j w_{ts}(Z_{sj})f(Z_{sj})}
        {\sum_j w_{ts}(Z_{sj})},
 \qquad
 \widehat N_{ts}^{\rm eff}
 =\frac{(\sum_jw_{tsj})^2}{\sum_jw_{tsj}^2}.
\]
Exact ratios remove deterministic localization bias.  Finite self-normalized
importance sampling still has \(O(N_s^{-1})\) bias, while its leading
variance is controlled by
\[
 \rho_{ts}=\E_s w_{ts}^2
 =1+\chi^2(\pi_t\|\pi_s),
 \qquad
 \widehat N_{ts}^{\rm eff}/N_s\ \longrightarrow\ \rho_{ts}^{-1}.
\]
Thus an ESS gate screens usable overlap and requests a fresh target cloud
when no source passes; ESS is a diagnostic and pooling heuristic, not a
test-function-specific oracle \citep{AgapiouEtAl2017,ElviraMartinoRobert2022}.

\subsection{What the fixed pilots establish}
\index{effective sample size!posterior-family gate}

The Gibbs pilot uses an exactly enumerable binary augmentation family.  Its
\(2\times2\) kernels provide invariant laws without Monte Carlo error over nine
sharing radii.  Log--log slopes are fitted on the six smallest radii; 300
independent chains at radius 0.1 serve only as an implementation check.  The
observed slopes in Table~\ref{tab:ch24-monte-carlo-loss-scales} recover the
predicted first- and second-order regimes, and every exact stationary bias lies
below its declared perturbation bound.

The particle pilot generates 240 covariates and 41 analytic local Gaussian
posteriors.  The oracle is \(\Pr_t(\theta>0)\); 21 alternating targets supply
independent anchor clouds.  Across 300 replications with seed 20260714, the
low-cost bank uses 96 particles per anchor, whereas the independent baseline
uses 96 at every target.  The equal-fresh-draw bank uses 187 particles per
anchor, 3,927 draws in total, and attains MSE
\(0.132\times10^{-3}\).  The low-cost and equal-budget banks additionally
perform 82,656 and 161,007 density-ratio evaluations per replication.
Mean empirical ESS tracks
\(\{1+\chi^2(\pi_t\|\pi_s)\}^{-1}\) with correlations 0.9896 in the smooth
family and 0.9891 after an imposed mean discontinuity.  Under that negative
control, no inspected cross-discontinuity source passes the mean ESS gate and
all equally local same-side sources pass.

\begin{table}[H]
\centering
\caption{Three diagnostic scales for posterior-family Monte Carlo reuse.  The final
column reports only the fixed analytic pilots described above.}
\label{tab:ch24-monte-carlo-loss-scales}
\small
\begin{tabularx}{\textwidth}{@{}
  >{\raggedright\arraybackslash}p{0.18\textwidth}
  >{\raggedright\arraybackslash}p{0.18\textwidth}
  >{\raggedright\arraybackslash}p{0.24\textwidth}
  >{\raggedright\arraybackslash}X@{}}
\toprule
Mechanism & Target contract & Coupling price and certificate & Reproducible evidence\\
\midrule
Uncorrected shared Gibbs
  & perturbed unless \(\epsilon_t=0\)
  & stationary defect \(\epsilon_t/(1-\gamma_t)\); conditional and kernel discrepancy
  & exact binary slopes: 1.006 one-sided, 2.006 centered; all 18 evaluated perturbation bounds pass\\
\addlinespace[0.35em]
MH-corrected coupled move
  & exact under the declared ratio
  & rejection and mixing variance; acceptance and autocorrelation
  & comparator only; no numerical claim in this laboratory\\
\addlinespace[0.35em]
Exact-ratio particle bank
  & population base change is exact; finite-\(N\) self-normalization remains
  & \(\chi^2\) overlap, ratio evaluations, memory, and ESS-triggered refresh
  & smooth-family MSE \(0.221\times10^{-3}\) versus \(0.658\times10^{-3}\), with 2,016 versus 3,936 fresh draws\\
\bottomrule
\end{tabularx}
\end{table}

\paragraph{Reproducibility record.}
The single entry point is \path{reproducibility/ch24/run_all.py}.  The exact
source snapshot, tests, frozen protocol, and claim audit are vendored under
\path{vendor/mcmc_aux/} beside that entry point; the source-of-truth result is
\path{results/monte_carlo_reuse/summary.json}.  Seed, configuration, costs,
all 300-replication summaries, and pass/fail thresholds are machine-readable,
and \path{SHA256SUMS.txt} identifies every artifact.  The default run uses no
file outside the reproduction directory.

\begin{boundarybox}
The Gibbs pilot validates a finite-state localization calculation; the
particle pilot validates exact-ratio reuse and its overlap gate in a
low-dimensional conjugate family.  Fresh draws and ratio evaluations are
reported separately, so lower MSE per fresh draw is not a wall-clock claim.
No nonlinear state-space, long-trajectory, adaptive-source, or
application-scale superiority result is asserted.
\end{boundarybox}

\section{Laboratory III: multi-sample cytometry}
\index{cytometry!multi-sample alignment}
\label{sec:lab-cytometry}

\subsection{Alignment with missing populations}
\index{cytometry!missing populations}
\index{transport!partial bijection}

In multi-sample cytometry, cells are nested inside biological samples and
organized into local populations.  The problem is not only clustering but
whether those populations correspond across samples and whether one global
labeling is scientifically coherent.  Cardinalities can change through
biological absence, rare-population sampling failure, local split or merge,
marker drift, or ambiguous matching.  Full permutation synchronization
assumes away these events; the partial transports of
Chapter~\ref{ch:transport-cycles} retain them as explicit outcomes.

The OALI pipeline is:
\begin{enumerate}
  \item estimate local populations and uncertainty on calibration samples;
  \item build a scientifically justified sample graph;
  \item estimate partial edge transports and their assignment margins;
  \item audit cycle holonomy and global-section existence;
  \item construct charts only when the obstruction is stable;
  \item fit within-chart transfer rules and evaluate them on held-out
  biological samples.
\end{enumerate}

\subsection{Reproducible smoke test and numerical limits}

The fixed smoke test generates ten samples, 39 marker coordinates, and 24
designed populations.  Biological absence is simulated by dropping
populations sample by sample, with a higher drop probability for four rare
populations.  The sample graph has 25 edges and hence cycle rank 16.  After
within-sample standardization, rectangular Hungarian assignments are retained
only below the predeclared distance threshold 2.40; retained partial
transports are then audited on the non-tree edges and propagated to the
reference sample.  The local components and their labels are generated, so
this check does not evaluate cell clustering.

Table~\ref{tab:ch24-cytometry-smoke} prevents the phrase ``recovers the
designed labels'' from hiding two different conclusions.  Edge matching is
nearly exact, but two false accepted matches produce two inconsistent
fundamental cycles and substantially weaken rare-population recovery.  The
pipeline path is executable; the rare-population endpoint is not yet strong.

\begin{table}[tbp]
\centering
\caption{Synthetic cytometry smoke test with 95\% intervals.  Sample-level
endpoints use 5,000 bootstrap resamples of the ten samples; discrete edge and
cycle endpoints use Wilson intervals.}
\label{tab:ch24-cytometry-smoke}
\small
\begin{tabularx}{\textwidth}{@{}Xrrl@{}}
\toprule
Endpoint & Estimate & 95\% interval & Unit and method\\
\midrule
Sample-level macro-F1
  & 0.9304 & (0.9171, 0.9443) & 10 samples; bootstrap\\
Rare-population recall
  & 0.4683 & (0.3419, 0.5709) & 10 samples; bootstrap\\
Unresolved-cell rate
  & 0.0129 & (0.0104, 0.0155) & 10 samples; bootstrap\\
Edge-match precision
  & 0.9959 & (0.9851, 0.9989) & 486 matches; Wilson\\
Edge-match recall
  & 1.0000 & (0.9921, 1.0000) & 484 shared branches; Wilson\\
Missing-branch detection recall
  & 0.9623 & (0.9070, 0.9852) & 106 missing branches; Wilson\\
Cycle-transport agreement
  & 0.9920 & (0.9714, 0.9978) & 251 transports; Wilson\\
\bottomrule
\end{tabularx}
\end{table}

\paragraph{Reproducibility record.}
The code snapshot is identified by SHA-256 prefix
\texttt{f1c69873bf997112}; the full hash, configuration, package versions,
counts, and intervals are in
\path{reproducibility/ch24/results/cytometry/cytometry_run.json}.  The checked
environment is Python 3.11.5, NumPy 1.24.3, and SciPy 1.11.1.  Seed 20260806
generates the actual data entry
\path{reproducibility/ch24/results/cytometry/generated_components.csv}; seed
20260807 fixes the bootstrap.  The directory-level
\path{reproducibility/ch24/SHA256SUMS.txt} checks every source and result
artifact.

No real-data cytometry result is reported.  The predeclared entry point for a
future confirmation is the \texttt{Samusik\_all\_SE()} object in
\texttt{HDCytoData} (or its corresponding \texttt{flowSet}), with original
files at FlowRepository accession
\texttt{FR-FCM-ZZPH} \citep{SamusikEtAl2016,WeberSoneson2019}.  That resource
contains 841,644 cells from ten mice, 39 markers, and 24 manually gated
populations.  These matching dimensions do not turn the synthetic table into
a Samusik result.

The confirmatory protocol therefore separates biological samples into
calibration, tuning, and final test sets.  Tuning selects graph density,
unmatched penalties, and a predeclared repair.  Final endpoints include
macro-F1, rare-population recall, unresolved-branch rate, and runtime.
Baselines include pooled clustering, one-reference Hungarian matching, tree
propagation, full synchronization, generic mixtures or random atlases, and
\OAI{} partial transport with cycle audit.

\subsection{The scientific claim boundary}
\index{claim boundary!cytometry}

A structural result would require a stable nontrivial obstruction or
unresolved region on calibration data.  The same analysis should predict a
plateau for the global labeling, identify the cycles or missing branches to be
repaired, and yield held-out improvement that random cuts, wider pooled
models, or tree-only propagation do not reproduce.  Required negative
controls include sample-level bootstrap, graph perturbation, cluster-number
and penalty sensitivity, transport margins, and label or graph randomization.
If the certificate is unstable, the conclusion remains unresolved.

The same contract can arise in multi-site extremes, hospital demand states,
ecological communities, single-cell transcriptomics, and drifting sensor
populations.  Each application still needs its own native oracle and
scientific endpoint; an atlas is not a result merely because it can be fit.

\section{Laboratory IV: policies, world models, and memory}
\index{conditional policy!world model}
\index{world model!memory}
\label{sec:lab-sequential}

\subsection{Compressed histories and the four-component decomposition}
\index{memory!compressed history}
\index{four-component risk decomposition!sequential decisions}

At a history $h_t$, a local oracle may be a conditional policy, posterior
belief, predictive kernel, planning auxiliary, or control variate.  A
deployed system compresses history to $z_t=\rho(h_t)$ and reuses one policy,
transition, or value architecture.  Subject to a justified exchange into
the control objective, the decomposition has the form
\[
\begin{aligned}
\text{control regret}
={}&\text{local model error}
 +\text{state/architecture obstruction}\\
 &+\text{generalization}
 +\text{planning/optimization error}.
\end{aligned}
\]
Local predictive error is not automatically control regret; an operational
exposure theorem is required.

\subsection{Coordination, recursion, and context}

A shared policy couples decisions across histories.  Rectangularization
compares it with independent local conditional choices, but occupancy is
endogenous: an early shared action changes which later histories are seen.
A memory state is sufficient only when histories it merges have identical
task-relevant conditional futures.  Preserving the present action or
one-step prediction need not preserve the next experiment, a changed reward,
multi-step planning, an intervention, or a safety constraint.  The minimal
recursive quotient of Chapter~\ref{ch:certificate-statistics} supplies the
finite exact model of this requirement.

Test-time computation is itself a resource coordinate.  With
\[
  \Arch_T
  =\{\text{deployments realizable with at most $T$ test-time steps}\},
\]
the frontier $T\mapsto\Obs_P(\Arch_T)$ separates an enlarged effective
class from better optimization inside a fixed class.

\subsection{Witness-directed memory refinement}
\index{memory!witness-directed refinement}

An obstruction-aware procedure estimates local conditional signatures,
finds histories merged by the present memory, computes a coordination or
carrier certificate, splits only the binding states, and validates on
held-out trajectories.  A failure to certify preservation of a
safety-relevant distinction is reported as unresolved together with the
experiment that could separate it; it is not silently converted into a
state merge.

Exact finite-history results support this design.  Continuous states and
actions, learned occupancies, nonstationarity, approximate local kernels,
data-dependent representations, nonconvex training, and long-horizon
exposure remain outside the present closure.

\paragraph{What the four laboratories jointly show.}
\index{applied laboratories!comparative lessons}

The MNIST audit records a positive sample-average amortization gap for a
selected finite library and shows that additional imagewise computation
reduces it.  Because saturation, independent Monte Carlo evaluation, matched
generic baselines, and negative controls are absent, it does not isolate a
structural mechanism or complete the proposed OALI chain.  The posterior-family
interface tests carrier reuse under approximate and
exact base change; its finite-state and conjugate pilots establish the
declared mechanisms, but not application-scale speed or superiority.
Cytometry couples carrier survival to graph compatibility; its smoke test
verifies the software path while exposing a rare-branch weakness that
independent scientific validation must address.  The sequential interface
makes coarse-graining dynamic and couples it to shared coordination.
Together they show how the Chapter~\ref{ch:laboratories} mechanisms compose
without claiming that one architecture or scalar score is universally best.

\section*{Exercises}

\begin{exercise}
For a contracting Gibbs kernel with a one-step perturbation bounded by
\(\epsilon\), derive the finite-time recursion and its stationary
\(\epsilon/(1-\gamma)\) limit.  State exactly where existence of an invariant
law is used.
\end{exercise}

\begin{exercise}
Design an ESS-gated particle bank for a one-parameter posterior family.
Report fresh draws, density-ratio evaluations, memory, refreshes, integrated
MSE, and a discontinuity control; explain why fresh draws alone do not prove
a runtime gain.
\end{exercise}

\begin{exercise}
Explain why a cell-level train/test split can overstate generalization across
biological samples, and give the corresponding sample-level split.
\end{exercise}

\begin{exercise}
Propose a negative control that distinguishes a biologically meaningful
atlas from a graph-partition artifact.
\end{exercise}

\begin{exercise}
Construct two histories with identical one-step action distributions but
different optimal two-step decisions.  Identify the recursive refinement
witness.
\end{exercise}

\begin{exercise}
Propose a task-relative oracle signature for a latent world model shared by
several reward functions, and state which changed contexts it must preserve.
\end{exercise}

%% file: reproducibility/ch24/results/mnist_amortization/generated_results.tex
\providecommand{\MNISTSeeds}{5}
\providecommand{\MNISTTestImages}{512}
\providecommand{\MNISTDataGaps}{5.084 / 3.016 / 2.642}
\providecommand{\MNISTEpochGaps}{4.296 / 3.053 / 2.642}
\providecommand{\MNISTWidthGaps}{3.778 / 2.642 / 2.527}
\providecommand{\MNISTGapMean}{2.493}
\providecommand{\MNISTGapSD}{0.126}
\providecommand{\MNISTGapLower}{2.373}
\providecommand{\MNISTRefineMean}{0.354}
\providecommand{\MNISTRefineSD}{0.027}
\providecommand{\MNISTRefineUpper}{0.380}
\providecommand{\MNISTSeedPasses}{5}
\providecommand{\MNISTRefinementCurve}{0.728 / 0.479 / 0.354 / 0.243}
\providecommand{\MNISTRemovedPercent}{64.6}
\providecommand{\MNISTCurvatureTotal}{320}
\providecommand{\MNISTCurvaturePositive}{311}
\providecommand{\MNISTBracketCovered}{311}
\providecommand{\MNISTAdaptiveMean}{0.982}
\providecommand{\MNISTAdaptiveUpper}{0.991}

%% file: reproducibility/ch24/results/digits_sparse_repair_r2_equal_info/generated_results.tex
\newcommand{\DigitsSparseN}{360}
\newcommand{\DigitsBaselineGap}{0.00831}
\newcommand{\DigitsBaselineGapLower}{0.00768}
\newcommand{\DigitsRepairGap}{0.00111}
\newcommand{\DigitsRepairRatio}{0.134}
\newcommand{\DigitsRepairRatioUpper}{0.141}
\newcommand{\DigitsGenericGap}{0.00512}
\newcommand{\DigitsRepairGenericRatio}{0.218}
\newcommand{\DigitsRepairGenericUpper}{0.233}
\newcommand{\DigitsNegativeRatio}{3.049}
\newcommand{\DigitsNegativeLower}{2.887}
\newcommand{\DigitsDistortionChange}{-0.00871}
\newcommand{\DigitsDistortionUpper}{-0.00780}
\newcommand{\DigitsBaselineAccuracy}{0.9500}
\newcommand{\DigitsRepairAccuracy}{0.9528}
\newcommand{\DigitsJacobianConflict}{0.869}
\newcommand{\DigitsStrictPoints}{359}
\newcommand{\DigitsRepairMACs}{6,144}

%% file: reproducibility/ch24/results/diabetes_sparse_repair_r1/generated_results.tex
\newcommand{\DiabetesSparseN}{89}
\newcommand{\DiabetesBaselineGap}{0.000478}
\newcommand{\DiabetesBaselineGapLower}{0.000304}
\newcommand{\DiabetesRepairGap}{0.0000480}
\newcommand{\DiabetesRepairRatio}{0.100}
\newcommand{\DiabetesRepairRatioUpper}{0.149}
\newcommand{\DiabetesGenericGap}{0.000407}
\newcommand{\DiabetesRepairGenericRatio}{0.118}
\newcommand{\DiabetesRepairGenericUpper}{0.187}
\newcommand{\DiabetesRepairNegativeRatio}{0.0456}
\newcommand{\DiabetesRepairNegativeUpper}{0.0670}
\newcommand{\DiabetesDistortionChange}{-0.000256}
\newcommand{\DiabetesDistortionUpper}{-0.0000672}
\newcommand{\DiabetesBaselineMSE}{3035.46}
\newcommand{\DiabetesRepairMSE}{3011.38}
\newcommand{\DiabetesMSEChange}{-24.08}
\newcommand{\DiabetesMSEChangeUpper}{-2.10}
\newcommand{\DiabetesJacobianConflict}{0.601}
\newcommand{\DiabetesStrictPoints}{89}
\newcommand{\DiabetesRepairMACs}{288}

%% file: reproducibility/ch24/results/diabetes_sparse_repair_r1/theorem_alignment.tex
\newcommand{\DiabDepthLeftRow}{17}
\newcommand{\DiabDepthRightRow}{68}
\newcommand{\DiabDepthLeftDatasetIndex}{61}
\newcommand{\DiabDepthRightDatasetIndex}{172}
\newcommand{\DiabDepthAtom}{6}
\newcommand{\DiabDepthEligiblePoints}{77}
\newcommand{\DiabDepthLeftRadius}{0.0410702}
\newcommand{\DiabDepthRightRadius}{0.120586}
\newcommand{\DiabDepthConflict}{0.389454}
\newcommand{\DiabDepthGamma}{0.0119313}
\newcommand{\DiabDepthNativeFloor}{3.55849\times10^{-5}}
\newcommand{\DiabDepthKappa}{0.499944}
\newcommand{\DiabDepthLipschitz}{1.90733}
\newcommand{\DiabDepthContraction}{0.737882}
\newcommand{\DiabDepthTwoContraction}{0.54447}
\newcommand{\DiabDepthTolerance}{3.20264\times10^{-5}}
\newcommand{\DiabDepthSufficientSteps}{24}

%% file: part_notes/part07_history.tex
\parthistoricalnotes{}
\index{OALI!historical comparison}
\index{repair!historical comparison}

\subsection*{OALI is a typed workflow, not a new generic architecture family}

Mixtures of experts, routing, recurrent memories, active acquisition,
matching, and held-out comparison all have independent literatures.  OALI
does not claim ownership of those components.  Its methodological position is
that a repair should follow a typed obstruction witness and should be judged
on an independent population endpoint after the mechanism-specific change.

\begin{center}
\small
\begin{tabularx}{0.97\textwidth}{@{}p{0.22\textwidth}p{0.29\textwidth}X@{}}
\toprule
Neighbor & Shared operation & OALI distinction \\
\midrule
Architecture search or mixture of experts & compare or route among candidate
models & first identify whether the failed branch is deployment, evidence,
recursive state, or contextual exposure; expand that branch only \\
Active learning and experiment design & select the next informative
observation & resolve a declared population certificate with worldwise
authorization; posterior utility may guide acquisition but does not authorize
stopping \\
Graph synchronization and assignment \citep{Singer2011,PachauriKondorSingh2013}
& align local labels across edges & preserve partial domains, missing
branches, and common fixed-section requirements rather than force a total
permutation \\
Held-out model comparison & concentration on an independent test sample &
predeclare the experimental unit and baselines, and combine gain with the
separate structural certificate \\
\bottomrule
\end{tabularx}
\end{center}

\subsection*{Ownership of the formal results}

Typed no-compensation and typed realization are book/program interface
syntheses.  The finite-library theorem is Hoeffding plus a union bound, and
the population-improvement theorem is certificate subtraction.  Noisy
partial-transport recovery is finite-class argmin stability.  The cycle-space
theorem specializes classical holonomy and graph consistency to partial
maps.  Their distinctive value lies in the typed end-to-end contract and the
missing-versus-moved branch semantics, not in rebranding their classical
engines.

%% file: chapters/ch25_quantum.tex
\chapter[Quantum Elimination Geometry]{Quantum Elimination Geometry and Noncommutative Rigidity}
\label{ch:quantum}
\index{quantum theory!noncommutative boundary}
\index{claim boundary!quantum analogy}

Classical elimination relies heavily on disintegration: a joint law splits
into a marginal law and conditional kernels, and relative entropy obeys an
exact chain rule.  In a quantum system, partial trace plays the role of
marginalization, but no universal classical conditional state has the same
properties.

The quantum theory therefore tests which parts of elimination geometry are genuinely structural and which depend on commutativity.

Strong-coupling thermodynamics and Hamiltonians of mean force are established
subjects \citep{TalknerHanggi2020,CresserAnders2021,
TrushechkinEtAl2022}.  This chapter does not claim the mean-force construction
itself.  It asks the narrower intervention question: when can one
field-independent coarse Hamiltonian reproduce the reduced Gibbs operator for
every declared external field?

\section{Universal Gibbs base change}
\index{quantum theory!Gibbs base change}
\index{base change!quantum}

Let $R=R^\star$ act on a finite-dimensional bipartite Hilbert space $\mathcal H_A\otimes\mathcal H_B$.  Fix $\beta>0$.  Ask whether there exist $K=K^\star$ on $A$ and $c>0$, independent of every external Hermitian field $F$, such that
\[
  \operatorname{Tr}_B
  e^{-\beta(F\otimes I_B+R)}
  =
  c e^{-\beta(F+K)}
  \qquad\text{for all }F.
\]

Completely positive dilations and the sufficiency/recovery theory for quantum
channels provide the structural background
\citep{Stinespring1955,Petz1988Sufficiency,MosonyiPetz2004,JencovaPetz2006}.  Equality
structures for quantum entropy inequalities supply a closely related
rigidity tradition \citep{HaydenJozsaPetzWinter2004}.  The result below is a
finite-dimensional Hamiltonian-language specialization of the established
factorization and exponential-family sufficiency theory, with a self-contained
proof included here; it is not claimed as a new quantum-information theorem.

\begin{conditionbox}
The system and environment are finite dimensional, $\beta>0$, all Gibbs states are faithful, and one fixed system Hamiltonian correction $K$ and scalar $c$ must work for every Hermitian system field $F$.  The quantifier includes noncommuting fields; restricting it to a commuting sector leads to a weaker controlled-interaction theorem.
\end{conditionbox}

\begin{theorem}[Universal Gibbs base-change rigidity]
\label{thm:quantum-base-change}
The identity holds for every Hermitian $F$ if and only if
\[
  R=K\otimes I_B+I_A\otimes L+\lambda I_{AB}
\]
for some Hermitian $L$ and scalar $\lambda$.
\end{theorem}

\begin{interpretationbox}
Universal exact Gibbs elimination is rigid: the bipartite Hamiltonian must contain no genuine interaction, apart from system, environment, and scalar terms.  One field or one commuting family is not enough to force this conclusion.  The scalar partition-function shadow for all fields already contains the same rigidity.
\end{interpretationbox}

\begin{proofroadmap}
Pass from operator base change to scalar pressure equality, take convex conjugates to obtain universal constrained relative-entropy saturation, use one Petz recovery channel for every marginal to force a product Gibbs reference, and take the Hermitian logarithm to recover an additive Hamiltonian.  The chapter appendix proves each intermediate lemma and the full equivalence.
\end{proofroadmap}

Jen\v{c}ov\'a and Petz characterize sufficiency for quantum exponential
families and the factorization of sufficient coarse-grainings
\citep[Theorems~5 and~7]{JencovaPetz2006}.  Taking the coarse observable map
to be $X\mapsto X\otimes I_B$ and allowing a basis of all Hermitian system
fields gives the product-state core of the theorem.  The proof below records
the same implication directly through pressure duality and the Petz map; its
scalar-shadow formulation is a convenient corollary-level packaging.

\paragraph{Proof architecture.}
\index{quantum theory!rigidity proof}

Thus universal exact base change over the full noncommutative system algebra
excludes genuine interaction.  The proof proceeds through several rigidities.
\begin{enumerate}
  \item Equality of partition functions for all $F$ identifies the constrained Gibbs--Fenchel dual.
  \item Equality in Umegaki data processing yields one recovery channel for the entire system-state family.
  \item A recovery extension that leaves every input state unchanged has product form.
  \item The Hamiltonian must therefore separate into system, environment, and scalar terms.
\end{enumerate}

The scalar partition-function shadow already suffices: it remembers the interaction when tested against all noncommuting fields.

\section{Commuting sectors}
\index{quantum theory!commuting sector}

The universal theorem used all Hermitian interventions and therefore forced
product structure.  Restricting interventions to a commuting algebra changes
the answer; the source box marks that shift and the provenance of the sector
classification.

\begin{sourcebox}[title={Source result; proof not reproduced here}]
The commuting-sector classification, replacement and recoverability identities,
and lossless-classicalization variants summarized in the remainder of the chapter
are established in the Quantum Elimination Geometry manuscript.  The chapter
appendix gives a complete proof only of the universal noncommutative rigidity
theorem above \citep{QuantumEG}.
\end{sourcebox}

Under this restricted contract, if admissible external fields lie in a
commuting sector algebra, genuine controlled interaction can be compatible
with exact base change.  The necessary and sufficient condition is block
diagonal structure over sectors, with each block's partial Gibbs operator
scalar on the corresponding system sector.

For a maximal abelian algebra, this is a controlled Hamiltonian.  The contrast is sharp:
\begin{itemize}
  \item full noncommutative field variation forces product structure;
  \item classical sector variation permits controlled interaction.
\end{itemize}

\section{Replacement, measurement, and operational meaning}
\paragraph{The replacement identity.}
\index{quantum theory!replacement identity}

For a general interaction, a trial-state Umegaki defect decomposes into:
\begin{enumerate}
  \item a coarse Gibbs defect;
  \item a nonnegative data-processing gap;
  \item a signed Hamiltonian-of-mean-force anomaly.
\end{enumerate}

The anomaly records external-field dependence of the effective coarse Hamiltonian.  Universal recovery maps convert the data-processing gap into a quantitative recoverability certificate.  In the rigid product case, the anomaly vanishes and the gap becomes an exact conditional relative entropy.

\paragraph{Measurement and lossless classicalization.}
\index{quantum theory!measurement}
\index{quantum theory!lossless classicalization}

A fixed measurement $M$ satisfies data processing,
\[
  D(\rho\|\sigma)
  \ge
  \KL(M\rho\|M\sigma).
\]
Entropy monotonicity under completely positive maps and its equality/recovery
theory provide the classical background
\citep{Lindblad1975,Petz1986,Petz1988Sufficiency,MosonyiPetz2004}.  The
measurement preserves the Umegaki defects of an entire faithful family
exactly if and only if the family is jointly recoverable.  One recovery
channel followed by copying the classical outcome would broadcast the family.
By the no-broadcasting theorem \citep{BarnumEtAl1996}, pairwise commutativity
is necessary; it is also sufficient through a common eigenbasis measurement.

\begin{boundarybox}
Measured KL is generally a lower certificate, not an exact replacement for quantum relative entropy.  Classicalization is lossless for a faithful family only under a commutativity condition.
\end{boundarybox}

\paragraph{Why this is relevant to AI.}
\index{scientific AI!quantum boundary}

The quantum results provide a general warning.  A coarse or measured representation can preserve one scalar objective while failing to preserve an entire family of defects under intervention.  Exact base change under all contexts is a strong rigidity statement.

The same logic appears classically in world models and representation learning: preserving predictions at one task is weaker than preserving all downstream conditional decompositions.

\section*{Exercises}

\begin{exercise}
Verify the universal base-change identity in the product Hamiltonian case.
\end{exercise}

\begin{exercise}
Explain why one recovery channel for all states in a family, combined with classical copying, produces a broadcasting channel.
\end{exercise}

\begin{exercise}
Construct a controlled two-qubit Hamiltonian that satisfies exact base change for diagonal system fields but not for arbitrary noncommuting fields.
\end{exercise}

\input{chapter_appendices/ch25_proofs}

%% file: chapter_appendices/ch25_proofs.tex
\chapterproofappendix

\subsection*{Preparatory Gibbs--Fenchel duality}
\proofdependency{All Hilbert spaces are finite dimensional and all Gibbs references are faithful.  These hypotheses ensure that Umegaki relative entropy is finite on every trial state, constrained minima are attained, and ordinary finite-dimensional Fenchel--Moreau duality applies.}

Let $\rho$ be a faithful state on $\mathcal H_A\otimes\mathcal H_B$.  Define
\[
  P_\rho(X)=\log\Tr_{AB}\exp(\log\rho+X\otimes I_B)
\]
and
\[
  I_\rho(\alpha)
  =\min_{\omega:\Tr_B\omega=\alpha}D(\omega\|\rho).
\]
For a faithful state $\sigma$ on $A$, put
\[
  P_\sigma(X)=\log\Tr_A\exp(\log\sigma+X).
\]

\begin{lemma}[Gibbs variational identity]
For a faithful state $\rho$ and Hermitian $Y$ on the same space,
\[
  \log\Tr e^{\log\rho+Y}
  =\max_\omega\{\Tr(\omega Y)-D(\omega\|\rho)\}.
\]
The maximizer is unique and equals
$\omega_Y=e^{\log\rho+Y}/\Tr e^{\log\rho+Y}$.
\end{lemma}

\begin{proof}
Let $Z=\Tr e^{\log\rho+Y}$.  Since
$\log\omega_Y=\log\rho+Y-\log Z$,
\[
  D(\omega\|\omega_Y)
  =D(\omega\|\rho)-\Tr(\omega Y)+\log Z.
\]
The nonnegativity of Umegaki relative entropy, with equality only at
$\omega=\omega_Y$, gives the claimed variational formula and uniqueness.
\end{proof}

\begin{proposition}[Constrained pressure duality]
For every faithful bipartite state $\rho$,
\[
  P_\rho(X)=\max_\alpha\{\Tr(\alpha X)-I_\rho(\alpha)\},
\]
\[
  I_\rho(\alpha)=\sup_X\{\Tr(\alpha X)-P_\rho(X)\}.
\]
For a faithful system state $\sigma$,
\[
  D(\alpha\|\sigma)=\sup_X\{\Tr(\alpha X)-P_\sigma(X)\}.
\]
\end{proposition}

\begin{proof}
Apply the Gibbs variational identity with $Y=X\otimes I_B$ and group trial states by their $A$-marginal:
\begin{align*}
P_\rho(X)
&=\max_\omega\{\Tr(\omega_A X)-D(\omega\|\rho)\}\\
&=\max_\alpha\{\Tr(\alpha X)-I_\rho(\alpha)\}.
\end{align*}
The constrained functional $I_\rho$ is proper, convex, and lower semicontinuous on the finite-dimensional state space, so Fenchel--Moreau duality gives the second formula.  Applying the same Gibbs variational identity directly on $A$ gives the formula for $D(\alpha\|\sigma)$.
\end{proof}

\subsection*{Universal saturation of the constrained extension problem}
\proofdependency{The converse uses equality in Umegaki data processing for the fixed channel $\Tr_B$ and the fixed faithful reference $\rho$.  The associated Petz recovery map is therefore the same for every marginal $\alpha$.}

\begin{lemma}[A universal right inverse of partial trace is product]
Let $\mathcal E:\mathcal L(\mathcal H_A)\to
\mathcal L(\mathcal H_A\otimes\mathcal H_B)$ be a quantum channel satisfying
\[
  \Tr_B\mathcal E(X)=X
  \qquad\text{for every }X.
\]
Then there exists a fixed state $\tau_B$ such that
\[
  \mathcal E(X)=X\otimes\tau_B
  \qquad\text{for every }X.
\]
\end{lemma}

\begin{proof}
Use the Choi representation.  Let $J_{\mathcal E}\ge0$ be the Choi matrix of $\mathcal E$ on
$A_{\rm out}\otimes B\otimes A_{\rm in}$.  The condition
$\Tr_B\circ\mathcal E=\mathrm{id}_A$ is equivalent to
\[
  \Tr_B J_{\mathcal E}=J_{\mathrm{id}}
  =|\Omega\rangle\langle\Omega|,
\]
where $|\Omega\rangle$ is a maximally entangled vector between $A_{\rm out}$ and $A_{\rm in}$.  The partial trace of the positive operator $J_{\mathcal E}$ has rank one.  Hence the support of $J_{\mathcal E}$ is contained in
$\operatorname{span}\{|\Omega\rangle\}\otimes\mathcal H_B$.  Indeed, if
$\Pi_\Omega$ is the support projection of
$|\Omega\rangle\langle\Omega|$ and $P=I-\Pi_\Omega$, then
\[
  \Tr\{(P\otimes I_B)J_{\mathcal E}\}
  =\Tr\{P\,\Tr_BJ_{\mathcal E}\}=0.
\]
Positivity implies $(P\otimes I_B)J_{\mathcal E}(P\otimes I_B)=0$ and also
eliminates the off-diagonal support blocks.  Therefore
\[
  J_{\mathcal E}=|\Omega\rangle\langle\Omega|\otimes\tau_B
\]
for some positive $\tau_B$.  Trace preservation of $\mathcal E$ normalizes $\tau_B$ to unit trace.  The Choi inversion formula now gives
$\mathcal E(X)=X\otimes\tau_B$.
\end{proof}

\begin{proposition}[Universal constrained saturation]
For faithful $\rho$ with $\rho_A=\Tr_B\rho$, the following are equivalent:
\begin{enumerate}
\item $I_\rho(\alpha)=D(\alpha\|\rho_A)$ for every state $\alpha$ on $A$;
\item $\rho=\rho_A\otimes\tau_B$ for some state $\tau_B$.
\end{enumerate}
\end{proposition}

\begin{proof}
If $\rho=\rho_A\otimes\tau_B$, choose the feasible extension
$\omega=\alpha\otimes\tau_B$.  Additivity of relative entropy gives
\[
  I_\rho(\alpha)
  \le D(\alpha\otimes\tau_B\|\rho_A\otimes\tau_B)
  =D(\alpha\|\rho_A).
\]
Data processing under partial trace gives the reverse inequality, so equality holds.

Conversely, for every $\alpha$ choose a constrained minimizer $\omega_\alpha$.  The assumed equality says
\[
  D(\omega_\alpha\|\rho)=D(\alpha\|\rho_A).
\]
Equality in data processing for partial trace and the fixed faithful reference $\rho$ implies exact recovery by the Petz map $\mathcal R_\rho$:
\[
  \omega_\alpha=\mathcal R_\rho(\alpha).
\]
Since $\Tr_B\omega_\alpha=\alpha$, the same channel satisfies
$\Tr_B\mathcal R_\rho(\alpha)=\alpha$ for every state and hence, by linearity, for every operator.  The preceding lemma gives
$\mathcal R_\rho(X)=X\otimes\tau_B$.  Evaluating at $X=\rho_A$ and using
$\mathcal R_\rho(\rho_A)=\rho$ yields
\[
  \rho=\rho_A\otimes\tau_B.
\]
\end{proof}

\subsection*{Proof of universal Gibbs base-change rigidity}
\proofdependency{Fix $\beta>0$, a bipartite Hermitian $R$, and a system Hermitian $K$.  The quantifier is over every Hermitian system field $F$, not merely a commuting subalgebra.  This universal noncommutative quantifier is the source of rigidity.}
\proofboundary{The result does not assert that an interacting Hamiltonian can never have a simple reduced Gibbs state at one field.  It asserts that one field-independent correction $K$ cannot work after every Hermitian intervention unless the Hamiltonian is additive.}

\begin{proof}
Define
\[
  Z_R(F)=\Tr_{AB}e^{-\beta(F\otimes I_B+R)},
  \qquad
  Z_K(F)=\Tr_Ae^{-\beta(F+K)},
\]
with normalized references
\[
  \rho=\frac{e^{-\beta R}}{Z_R(0)},
  \qquad
  \sigma=\frac{e^{-\beta K}}{Z_K(0)}.
\]
We prove the equivalence through six formulations.

\paragraph{Operator base change implies normalized and scalar base change.}
Suppose
\[
  \Tr_Be^{-\beta(F\otimes I_B+R)}
  =c e^{-\beta(F+K)}
\enspace\text{for all }F.
\]
Taking the full trace gives $Z_R(F)=cZ_K(F)$.  Dividing the operator identity by that scalar identity gives equality of the normalized reduced Gibbs state and the $K$-shifted system Gibbs state.

\paragraph{Normalized base change implies scalar base change.}
Assume only equality of the normalized reduced states.  Put
\[
  \phi(F)=\log Z_R(F)-\log Z_K(F).
\]
For a Hermitian direction $G$, differentiation of the trace exponential and cyclicity of trace give
\[
  d\log Z_R(F)[G]=-\beta\Tr_A(\alpha_FG),
  \qquad
  d\log Z_K(F)[G]=-\beta\Tr_A(\sigma_FG),
\]
where $\alpha_F$ is the reduced bipartite Gibbs state and $\sigma_F$ is the system Gibbs state.  The normalized identity gives $\alpha_F=\sigma_F$, hence $d\phi(F)[G]=0$ for all $F,G$.  The real vector space of Hermitian matrices is connected, so $\phi$ is constant.  Thus
$Z_R(F)=cZ_K(F)$ with $c=e^{\phi(0)}$.

\paragraph{Scalar base change implies equality of constrained duals.}
For Hermitian $X$, set $F=-X/\beta$.  The scalar identity and its value at zero imply
\[
  P_\rho(X)=\log\frac{Z_R(-X/\beta)}{Z_R(0)}
  =\log\frac{Z_K(-X/\beta)}{Z_K(0)}
  =P_\sigma(X).
\]
Taking convex conjugates in the constrained pressure-duality proposition gives
\[
  I_\rho(\alpha)=D(\alpha\|\sigma)
  \qquad\text{for every }\alpha.
\]

\paragraph{Equality of constrained duals forces a product reference.}
The state $\rho$ is feasible for the constraint $\alpha=\rho_A$, so
$I_\rho(\rho_A)=0$.  The dual identity yields
$D(\rho_A\|\sigma)=0$, hence $\rho_A=\sigma$.  We therefore have
\[
  I_\rho(\alpha)=D(\alpha\|\rho_A)
  \qquad\text{for every }\alpha.
\]
Universal constrained saturation gives a faithful state $\tau_B$ such that
\[
  \rho=\sigma\otimes\tau_B.
\]

\paragraph{A product reference forces an additive Hamiltonian.}
Write
\[
  \tau_B=\frac{e^{-\beta L}}{\Tr e^{-\beta L}}
\]
with $L=-\beta^{-1}\log\tau_B$.  From
\[
  \frac{e^{-\beta R}}{Z_R(0)}
  =\frac{e^{-\beta K}}{Z_K(0)}
   \otimes
   \frac{e^{-\beta L}}{\Tr e^{-\beta L}},
\]
we obtain
\[
  e^{-\beta R}
  =q\,e^{-\beta(K\otimes I_B+I_A\otimes L)}
\]
for a positive scalar $q$.  Both sides are positive definite.  Uniqueness of the Hermitian logarithm gives
\[
  R=K\otimes I_B+I_A\otimes L+\lambda I_{AB}
\]
for a real scalar $\lambda$.

\paragraph{Additivity implies operator base change.}
Conversely, if the last display holds, the two tensor summands commute and
\begin{align*}
\Tr_Be^{-\beta(F\otimes I_B+R)}
&=e^{-\beta\lambda}
  \Tr_B\left(e^{-\beta(F+K)}\otimes e^{-\beta L}\right)\\
&=e^{-\beta\lambda}\Tr(e^{-\beta L})e^{-\beta(F+K)}.
\end{align*}
This is universal operator base change with
$c=e^{-\beta\lambda}\Tr(e^{-\beta L})$.  All formulations, and hence the stated equivalence, follow.
\end{proof}

%% file: chapters/ch26_boundaries.tex
\chapter{Boundaries of Structural Explanation}
\label{ch:boundaries}
\index{claim boundary}
\index{structural explanation!boundary}

A structural theory becomes uninformative if every failure can be labeled an
obstruction after the fact.  The claims in this book therefore depend on five
declared contracts.  The local objective must generate the defect; the
representation must be intrinsic to the target; the deployment class must
match the system actually used; the relevant risk level must be stated; and
finite data must support the structural conclusion being reported.

\section{Failures that invalidate the structural object}

The first failure occurs before architecture analysis.  A reported discrepancy
may not arise from a certified elimination.  Locally assembled increments can
be nonintegrable, oriented incorrectly, or measured in an arbitrary parameter
metric.  In that case the architecture analysis must stop until the defect
system has been derived globally or repaired by an integrability theorem.

A second failure is an artificial representation.  Unrestricted exact lifts
permit dummy and target-calling coordinates, so geometric complexity can be
created by representation rather than by the declared target.  Slack
anchoring, quotient reduction, and an admissible lift language are therefore
prerequisites for intrinsic complexity claims.

A third failure is architecture misspecification.  A positive obstruction is
conditional on the deployment contract.  A lower bound for continuous
point-valued outputs is irrelevant if the deployed system is allowed external
anchors, discontinuities, additional memory, or variable-cardinality outputs.

Finally, a locally optimal variational family need not be a scientifically
adequate model.  A zero local defect only establishes optimality inside the
declared local family.  Model misspecification remains a separate source of
population risk.

\section{Interfaces that require additional theorems}

Several stronger conclusions do not follow automatically from a valid
architecture obstruction.

\begin{enumerate}
\item A positive worst-case topological cost need not imply a positive average
population risk; mass and regularity assumptions are required.
\item A native defect need not be visible to the downstream task; an exposure
or risk-transfer theorem is required.
\item A deterministic population obstruction need not be identifiable from
finite data; a valid statistical certificate is required.
\item Lift or extension size does not imply runtime complexity without an
explicit computation model, encoding, conditioning, and precision analysis.
\item A zero-temperature limit need not become deterministic if the closure of
the admissible class still contains randomized laws.
\item Classical conditional identities do not automatically pass through
noncommutative marginalization; quantum base change requires additional
product or commuting structure.
\end{enumerate}

These are not technical footnotes.  They define the levels at which claims in
the book are valid.

\section{Alternative explanations and falsification}

Poor training can imitate structural saturation.  Additional capacity can
imitate a mechanism-matched repair.  An unstable branch match can imitate
monodromy.  A rare topological seam can be irrelevant to average risk.
Consequently, a proposed structural explanation should be weakened or rejected
when any of the following persists under appropriate controls:
\begin{enumerate}
\item the structural lower certificate is zero or statistically unresolved;
\item stronger optimization removes the gap without changing the deployment
contract;
\item parameter- and compute-matched generic enlargements reproduce the
claimed repair gain;
\item the diagnosed structure is unstable under resampling or reasonable
changes in the audit graph;
\item the native defect is operationally invisible to the predeclared
downstream task;
\item the independent scientific endpoint does not improve;
\item the claimed obstruction disappears after replacing an arbitrary label or
lift by an intrinsic representation.
\end{enumerate}

\begin{boundarybox}
Performance saturation is a diagnostic signal, not a proof of architecture
obstruction.  The structural claim requires an independent native-risk
certificate and a repair whose effect follows the diagnosed mechanism.
\end{boundarybox}

\section{Conditions that must remain visible}

The current theorem system depends on several assumptions that should not be
hidden by streamlined exposition: compact metric structure where balls and
Lipschitz maps are used; explicit form-domain conditions in continuous-spectrum
arguments; well-defined certified convex sets on general cochain complexes;
exact second-best structural matching margins; biological-unit rather than
cell-level independence in confirmatory applications; and full-rank
assumptions for finite-temperature quantum identities.

These conditions define where the present theory stops.  The next chapter
treats extensions beyond them as research problems rather than completed
results.

\section*{Exercises}

\begin{exercise}
For three results in the book, identify the smallest change in assumptions that
would make the stated conclusion invalid or uninterpretable.
\end{exercise}

\begin{exercise}
Explain why a positive uniform radical obstruction does not imply a positive
average risk without additional mass or regularity conditions.
\end{exercise}

\begin{exercise}
Design an ablation that separates atlas structure from the benefit of extra
capacity.
\end{exercise}

%% file: chapters/ch27_agenda.tex
\chapter{A Falsifiable Research Program for Structural Learning}
\label{ch:agenda}
\chaptermark{Structural Learning Research Program}
\index{structural learning theory!research program}
\index{elimination geometry!research program}

The long-term value of elimination geometry does not depend on adoption of its
full vocabulary.  It depends on reusable results for a recurring problem:
local optima are available, but a shared deployment contract may not realize
them simultaneously.  A mature theory should quantify the resulting floor,
determine whether finite data can certify it, and predict a minimal
intervention whose benefit survives independent validation.

No theorem in the present book achieves that program in complete generality.
The proved results close specific interfaces in convex, graph, conditional,
singular, finite-sample, and quantum settings.  The remaining questions are
best organized by the points at which the current certification chain can
still fail.

\section{Data-dependent and computation-dependent deployment}

The statistical learning literature already provides tools for data-dependent
classes, including PAC--Bayesian bounds for random hypothesis sets
\citep{DupuisEtAl2024}.  Lifelong and multitask representation learning also
develop complexity measures for shared representations
\citep{WangZhangVinayak2026}.  The open question here is narrower: how should
those tools interact with a class whose structure was itself selected because
a structural certificate diagnosed a specific failure?

A repaired deployment class is often generated from data:
\[
  S_{\rm cal}
  \longmapsto
  \widehat{\mathcal C}
  \longmapsto
  \widehat\Arch.
\]
The relevant complexity is then not that of a fixed class but of a random
atlas, quotient, memory refinement, or routing structure.  A useful theory
should control chart number, overlap, transport uncertainty, gate stability,
and selection effects without discarding the structural information that
generated the class.

Computation creates a related ambiguity.  For a family
$\Arch_{m,T,\mathrm{Alg}}$, additional computation can have three distinct
effects:
\begin{enumerate}
\item it can reduce implementation error inside the same effective class;
\item it can enlarge the class of functions realized at test time;
\item it can restrict the reachable subset through optimizer dynamics or
implicit bias.
\end{enumerate}
A computation-dependent architecture theory should separate these effects
rather than treating test-time iteration as either pure optimization or pure
capacity by convention.

\section{Population effects of singularity and composition}

Singular normal forms often give exact worst-case costs, while average-risk
lower bounds require additional control of how failure can concentrate near a
seam.  A central open problem is a phase diagram relating regularity of the
deployment class, lower-mass assumptions on the population, and the smallest
achievable average defect.

Coordination theory faces a complementary extension problem.  The current
exact results are strongest for finite conditional structures.  Extending
rectangularity, memory obstruction, and coordination bounds to continuous
states, learned occupancies, world models, and function approximation requires
joint control of the deployment restriction and the state distribution induced
by the learned system.

Both directions ultimately require task-relative semantics.  A representation
should retain distinctions that are necessary for the local oracle and visible
to the downstream task, while quotienting distinctions that are irrelevant to
both.  The goal is not maximal reconstruction fidelity but minimal structural
information sufficient for realizability and decision.

\section{Scientific AI and foundation-model contracts}

Scientific applications should be chosen where forced global reuse has a
substantive consequence.  The structural object must change a scientific or
decision endpoint; an atlas, quotient, or certificate is not a contribution by
itself.  Candidate domains include multi-site environmental distributions,
multi-batch cell populations, sensor systems with branch-specific drift, and
shared latent-state models across institutions or operating regimes.

At foundation-model scale, the same questions arise through routing,
retrieval, memory, shared representations, and structured outputs.  The
hypotheses should remain concrete.  A shared representation may force
incompatible local solutions.  A memory bottleneck may merge histories that
require different actions.  A routing contract may create coordination
obstruction.  A quotient or set-valued output may remove an artificial
labeling problem.  Each claim must be tied to a native objective and a
falsifiable structural prediction.

\section{Success criteria and stopping rules}

The program would become a useful branch of learning theory if it yields tools
that are valuable independently of the surrounding terminology.  Examples
would include:
\begin{enumerate}
\item computable architecture-obstruction certificates used before model
scaling;
\item sharp saturation theorems under explicit representation or memory
contracts;
\item automatic atlas, quotient, or memory repairs with reproducible held-out
gains against matched controls;
\item statistically valid unresolved certificates that guide additional data
collection;
\item resource--risk frontiers that determine when more representation,
memory, or communication is worth its cost.
\end{enumerate}

The failure criteria are equally important.  The program should be narrowed if
most certificates are vacuous or uncomputable.  It should also be narrowed
when the examples reduce to known approximation or topology without new
objective-level consequences; when generic mixtures or extra parameters
reproduce OALI gains; when realistic samples leave the certificates
unresolved; or when the proposed structural distinctions do not change
independent scientific decisions.

Partial domains supply one concrete stopping-rule example.  For finite branch
sets, a partial bijection can be completed to a total permutation on an
enlarged fiber, with masks excluding dummy deployment states.  The global
section and repair problems then become masked synchronization or finite
constraint problems.  In the minimal noisy cycle experiment, deciding whether
one anomalous domain forces a seam costs order \(\log |E|\) observations per
edge, but the matching lower bound is exactly sparse anomalous-coordinate
detection.  Therefore partiality, domain survival, and a chart decision do not
alone constitute a new learning-theoretic mechanism.

Operational exposure supplies a second stopping-rule example rather than an
escape.  In the minimal memory-conflict experiment, every policy executable
through the baseline memory is exactly blind to two worlds with opposite
resource-adjusted architecture choices.  Forbidding a diagnostic split makes
worldwise certification impossible.  Allowing a temporary split turns it
into a controlled sensing action, after which the lower bound is the standard
max--min KL resolution bound of Chapter~\ref{ch:certificate-statistics}; a
priced split gives the same information calculation per unit cost.  Thus, for
finite resettable diagnostics, operational exposure changes the experiment
menu but does not create a new learning-theoretic rate.  This closes the
partial-domain and operational-memory flagship routes.  Reopening either
requires an explicit hard
family with a minimax term that cannot be absorbed into controlled sensing,
partial observability, automata discrimination, or ordinary representation
selection, not merely a larger memory example.

Chapter~\ref{ch:statistical-eg} opens a different, narrower route.  For
nonnegative sparse inference it turns strict active-set Jacobian conflict into
a computable one-pass native-loss floor and pairs that lower certificate with
a sufficient proximal repair depth.  This result can change an architecture
choice at a declared uniform tolerance, but it is not yet a settled flagship:
its ingredients have close antecedents, including the global sparse-
autoencoder amortisation gap of \citet{ONeillGumranKlindt2025}, independent
priority is unresolved, and the current diabetes alignment was computed after
confirmation.  The next stopping rule is therefore exact: obtain an
independent proof/priority review and freeze the conflict-pair rule, uniform
tolerance, repair depth, and compute ledger before opening new scientific
confirmation units.

These criteria suggest a practical research sequence:
\[
\begin{gathered}
\text{local oracle}
\rightarrow
\text{native loss}
\rightarrow
\text{deployment contract}
\rightarrow
\text{structural certificate}
\\
\rightarrow
\text{finite-sample certification}
\rightarrow
\text{minimal repair}
\rightarrow
\text{independent validation}
\end{gathered}
\]
The sequence should stop when the native defect has no demonstrated relation
to the downstream task, when the diagnosed obstruction occupies negligible
population mass, when matched generic baselines explain the gain, or when the
certificate is too unstable to support a structural decision.

The resulting mission is deliberately narrower than a universal theory of
learning:
\begin{quote}
\emph{When local optima are available, determine whether a shared deployment
contract can realize them, quantify the native risk forced when it cannot,
certify that conclusion from finite data, and identify the smallest
structural change whose benefit survives independent validation.}
\end{quote}

\section*{Research exercises}

\begin{exercise}
Formulate a computation-dependent architecture obstruction for a fixed-point
network with $T$ test-time iterations, separating class enlargement from
within-class optimization.
\end{exercise}

\begin{exercise}
Propose an average-risk theorem for a singular oracle family under a Lipschitz
deployment constraint and a lower-mass population condition.
\end{exercise}

\begin{exercise}
Design a nonresettable certificate-aware experiment for deciding whether a
shared memory state should be split.  State which diagnostic interventions
are reversible, include a negative control and a stopping rule, and identify
the proposed lower-bound term that is not already a controlled-sensing cost.
\end{exercise}

%% file: part_notes/part08_history.tex
\parthistoricalnotes{}
\index{quantum theory!historical comparison}

\subsection*{Quantum antecedents and the rigidity specialization}

Umegaki relative entropy, data processing, Petz equality/recovery, quantum
sufficiency, and Hamiltonians of mean force are mature theories
\citep{Umegaki1962,Lindblad1975,Petz1986}.  Part VIII uses them in their
established roles.  It does not claim a new data-processing equality theorem,
a new recovery map, or a new general theory of open-system equilibrium.

The displayed Hamiltonian statement is narrower: if one field-independent
unnormalized Gibbs base-change law must hold for \emph{every} Hermitian
external field, then the bipartite Hamiltonian is additive and contains no
genuine interaction.  A result-level priority audit found that this content
is already implied by the factorization of sufficient quantum
coarse-grainings and the explicit sufficiency criterion for quantum
exponential families \citep[Theorems~5 and~7]{JencovaPetz2006}.  The exact
Hamiltonian wording is useful, but it is a direct finite-dimensional
specialization rather than a defensible standalone originality claim.

\begin{center}
\small
\begin{tabularx}{0.94\textwidth}{@{}p{0.24\textwidth}p{0.31\textwidth}X@{}}
\toprule
Statement & Status & Boundary \\
\midrule
Universal Gibbs base-change rigidity & direct sufficiency/factorization
specialization with a complete book proof & finite dimensions, faithful
Gibbs states, one fixed correction, and all noncommuting external fields \\
Commuting-sector classification & source result summarized, proof left in
the companion paper & allows controlled interactions because the intervention
algebra is smaller \\
Recoverability and measurement limits & classical quantum-information core
with source specializations & measurement generally yields a lower
certificate, not an exact classical defect ledger \\
\bottomrule
\end{tabularx}
\end{center}

The final boundary chapters are therefore part of the contribution audit:
they distinguish a transparent specialization proved under a sharp
noncommutative quantifier from analogies, imported sector results, and open
extensions.

%% file: appendices/appA_convex.tex
\chapter{Convex Analysis and Bregman Geometry}
\index{convex analysis}
\index{Bregman geometry}
\label{app:convex}

This appendix collects the convex-analytic facts used throughout the book.  It is not intended to replace a full text on convex analysis \citep{Rockafellar1970}.

\section{Convex conjugacy}
\index{Fenchel conjugacy}

For a proper lower-semicontinuous convex function $\Phi$ on a finite-dimensional vector space,
\[
  \Phi^\star(u)=\sup_a\{\langle u,a\rangle-\Phi(a)\}.
\]
Fenchel--Young gives
\[
  \Phi(a)+\Phi^\star(u)\ge\langle u,a\rangle,
\]
with equality exactly when $u\in\partial\Phi(a)$.

If $\Phi$ is Legendre, $\nabla\Phi$ and $\nabla\Phi^\star$ are inverse maps on the interiors of their domains.

\section{Bregman divergence}
\index{Bregman geometry!divergence}

For differentiable $\Phi$,
\[
  D_\Phi(a\|b)
  =
  \Phi(a)-\Phi(b)-\langle\nabla\Phi(b),a-b\rangle.
\]
It is nonnegative but generally asymmetric and does not satisfy the triangle inequality.

The three-point identity is
\[
  D_\Phi(a\|c)
  =
  D_\Phi(a\|b)+D_\Phi(b\|c)
  +
  \langle a-b,\nabla\Phi(b)-\nabla\Phi(c)\rangle.
\]
This identity underlies Bregman Pythagorean identities and centroid formulas.

\section{Conditional Bregman projection}
\index{Bregman geometry!conditional projection}

Let $A^\star$ be an integrable random oracle and $Z$ a carrier.  Under regularity,
\[
  \argmin_{g(Z)}\E D_\Phi\{g(Z)\|A^\star\}
  =
  \nabla\Phi^\star\left(\E[\nabla\Phi(A^\star)\mid Z]\right),
\]
where $\Phi^\star$ denotes the convex conjugate and the right-hand side is interpreted through $\nabla\Phi^\star$.  This is the canonical carrier decoder used in Chapter~\ref{ch:resource-rd}.

The reverse orientation has a different centroid.  When the conditional
primal mean lies in the admissible domain,
\[
  \argmin_{g(Z)}\E D_\Phi\{A^\star\|g(Z)\}
  =
  \E(A^\star\mid Z).
\]

\section{Strong convexity and smoothness}
\index{strong convexity}
\index{smoothness}

If $\Phi$ is $\mu$-strongly convex,
\[
  D_\Phi(a\|b)\ge\frac\mu2\|a-b\|^2.
\]
If $\nabla\Phi$ is $L$-Lipschitz,
\[
  D_\Phi(a\|b)\le\frac L2\|a-b\|^2.
\]
These inequalities exchange native defects and geometric distances.

\section{Metric projection}
\index{metric projection}

For a nonempty closed convex set $C$ in a Hilbert space, the metric projection $P_C$ is firmly nonexpansive:
\[
  \|P_Cu-P_Cv\|^2
  \le
  \langle P_Cu-P_Cv,u-v\rangle.
\]
In particular,
\[
  \|P_Cu-v\|\le\|u-v\|
\]
for $v\in C$.

\section{Primal--dual gaps}
\index{primal--dual gap}

For a convex program with strong duality, a primal feasible point $x$ and dual feasible point $y$ satisfy
\[
  0\le f(x)-f^\star\le f(x)-g(y).
\]
The primal--dual gap is therefore a certified objective defect.  Turning the gap into runtime requires a numerical complexity model.

%% file: appendices/appB_graphs.tex
\chapter{Graph Operators, Min-Plus Algebra, and Partial Maps}
\index{graph theory}
\label{app:graphs}

\section{Weighted graph Laplacians}
\index{graph Laplacian}

For an undirected weighted graph with weights $w_{ij}\ge0$, the Laplacian is
\[
  L_{ii}=\sum_jw_{ij},
  \qquad
  L_{ij}=-w_{ij}\quad(i\ne j).
\]
It is positive semidefinite and
\[
  f^\top Lf
  =
  \frac12\sum_{i,j}w_{ij}(f_i-f_j)^2.
\]

\section{Spectral filters}
\index{spectral filter}

For a Borel function $s$ on the spectrum of $L$, define $S=s(L)$ by spectral calculus.  The effective dimension for filtered squared error is often $\operatorname{tr}(S^2)$.

A Markov filter must preserve constants and positivity.  Positive semidefiniteness alone does not imply entrywise nonnegativity.

\section{Min-plus algebra}
\index{min-plus algebra}

On $\R\cup\{+\infty\}$ define
\[
  a\oplus b=\min(a,b),
  \qquad
  a\otimes b=a+b.
\]
Kernel composition is
\[
  (K\otimes L)(z,x)
  =
  \inf_y\{K(z,y)+L(y,x)\}.
\]
Associativity is the algebraic form of Bellman recursion and obstruction-tower composition.

\section{Cycle rank}
\index{cycle space}

For a connected finite graph,
\[
  \beta_1=|E|-|V|+1
\]
is the cycle-space dimension.  A spanning tree plus one fundamental cycle for every non-tree edge gives a basis.

\section{Partial bijections}
\index{transport!partial bijection}

A partial bijection $T:D\subseteq A\to R\subseteq B$ is a bijection between its domain and range.  Composition is defined only when intermediate values survive.  The inverse is a partial bijection $T^{-1}:R\to D$.

Partial transports form an inverse semigroup rather than a permutation group.  This distinction is essential when branches disappear.

%% file: appendices/appC_topology.tex
\chapter{Covering Spaces, Degree, and Monodromy}
\index{topology}
\label{app:topology}

This appendix summarizes the topology used in the singular and atlas chapters.  Standard references include \citet{Hatcher2002} and \citet{Schwarz1961}.

\section{Covering spaces}
\index{topology!covering space}

A map $p:E\to X$ is a covering if every $x\in X$ has a neighborhood $U$ whose inverse image is a disjoint union of sheets, each mapped homeomorphically to $U$.  Local oracle branches in the regular separated regime form such sheets.

A loop in $X$ lifts to a path in $E$.  The endpoint of the lifted path defines
a permutation of the fiber: the monodromy action.  A global labeled section
exists only if the monodromy fixes a branch consistently.

\section{Winding number}
\index{topology!winding number}

For a nonvanishing loop $z:S^1\to\C^\star$,
\[
  \operatorname{wind}(z)
  =
  \frac{1}{2\pi i}\int_{S^1}\frac{z'(\theta)}{z(\theta)}\dd\theta
\]
when differentiable, with the usual topological extension.  Winding is invariant under homotopies avoiding zero and satisfies
\[
  \operatorname{wind}(z^k)=k\operatorname{wind}(z).
\]

\section{Topological degree}
\index{topology!degree}

For a continuous map $f:S^{d-1}\to S^{d-1}$, the degree is an integer invariant under homotopy.  Composition multiplies degrees.  Degree incompatibility gives higher-dimensional analogues of the radical obstruction.

\section{Schwarz genus and atlas number}
\index{atlas!Schwarz genus}

For a fibration $p:E\to X$, the Schwarz genus is the minimum number of open sets covering $X$ on each of which a continuous section exists.  It is a topological lower bound on chart number in exact atlas repair.

Statistical atlas numbers depend additionally on the native defect and tolerance.  They need not equal the topological genus away from zero tolerance.

\section{Discriminants}
\index{discriminant}

A discriminant is the set where the oracle fiber changes type: roots collide, eigenvalues cross, or the vertical Hessian loses rank.  Away from the discriminant, branches may form a covering.  Near it, local ramification normal forms replace separated-cover geometry.

%% file: appendices/appD_statistics.tex
\chapter{Statistical Tools for Honest Certification}
\index{statistical certification!technical tools}
\label{app:statistics}

\section{Uniform concentration}
\index{concentration!uniform}

If a loss class is bounded in $[0,B]$ and finite with size $M$, Hoeffding's inequality gives
\[
  \sup_{f\in\mathcal F}|P_nf-Pf|
  \le
  B\sqrt{\frac{\log(2M/\delta)}{2n}}
\]
with probability at least $1-\delta$.

For infinite classes, Rademacher complexity, covering numbers, stability, or PAC-Bayes tools may be used.  The correct class is the actual deployment class, potentially conditional on calibration data.

\section{Sample splitting}
\index{sample splitting}

Let calibration, training, tuning, and test samples be independent at the level of the scientific unit.  Conditioning on earlier splits turns a data-dependent candidate into a fixed object for the test analysis.  This is the simplest route to honest OALI validation.

\section{Confidence sets and three-way decisions}
\index{confidence world}
\index{certificate!three-way rule}

If $C_n$ covers the true world with probability $1-\alpha$, then any
declaration shared by all worlds in $C_n$ is valid on the coverage event.  If
compatible worlds disagree, the maximally decisive honest output is
unresolved.  If $C_n=\varnothing$, the output is model conflict rather than a
vacuous declaration.

\section{Confidence sequences}
\index{confidence sequence}

A confidence-world sequence $\{C_t\}_{t\ge1}$ satisfies
\[
  \Pp_w\{w\in C_t\text{ for all }t\}\ge1-\alpha.
\]
It permits adaptive stopping and querying.  Likelihood-ratio martingales and nonnegative supermartingales are standard constructions \citep{HowardEtAl2021,Ville1939}.

\section{Testing lower bounds}
\index{testing lower bound}

Le Cam, Fano, and Assouad arguments convert indistinguishable alternatives into risk lower bounds.  Certificate resolution uses a related but boundary-specific alternative set: only opposite-certificate worlds are binding.

\section{Clustered data}
\index{clustered data}
\index{sampling unit}

When cells, time points, or repeated observations are nested within a biological unit, the independent test sample size is the number of units, not the number of lower-level observations.  Resampling and confidence intervals should respect this hierarchy.

%% file: appendices/appE_dependency.tex
\chapter{Validation and Dependency Map}
\index{validation and dependency map}
\label{app:dependency}

The book is ordered by logical validation.  A structural lower bound is not
interpreted until the native defect, the integrability of its defect system, and the
admissibility of every nonnative carrier have been established.  The main
spine is an audit workflow, not a chain of theorem implications:
\[
\begin{aligned}
\text{define elimination}&\ \rightsquigarrow\ \text{derive a native defect}\\
&\ \rightsquigarrow\ \text{audit integrability and carrier admissibility}\\
&\ \rightsquigarrow\ \text{choose an architecture-obstruction theorem}\\
&\ \rightsquigarrow\ \text{add resource and operational assumptions}\\
&\ \rightsquigarrow\ \text{construct a statistical certificate}\\
&\ \rightsquigarrow\ \text{test saturation, repair, and validation}.
\end{aligned}
\]
Here \(\rightsquigarrow\) means ``next audit gate.''  It does not mean that
the object on the left mathematically implies the object on the right.  Each
transition may require a new theorem, a new modeling assumption, or an
independent experiment.  In particular, a positive obstruction does not imply
empirical saturation, and neither one implies that a proposed repair improves
a held-out endpoint.

\section{Constructive spine}
\begin{enumerate}
\item Conjugate defect identity. \item P/G/X/V/C distinctions.
\item Exactification and approximate-jet certificates.
\item Graph-CDF validity repair and CRPS risk decomposition.
\end{enumerate}

\section{Validity spine}
\begin{enumerate}
\item Flat integrability audits assembled local defect reports.
\item Hodge and period terms diagnose and repair nonintegrable defect fields.
\item Lift Complexity excludes dummy and target-calling carriers.
\item Target-visible reduction and quotient-faithful extraction connect an admissible lift to deployment capacity when a model-specific gate is proved.
\end{enumerate}

\section{Architecture spine}
\begin{enumerate}
\item Architecture obstruction is the second elimination.
\item EOT gives regular metric and flow lower bounds.
\item COT gives rectangularity and coordination decompositions.
\item Singular EG gives exact catastrophe taxes when regular assumptions fail.
\item Atlas, quotient, set-valued, and randomized repairs change the declared deployment contract.
\end{enumerate}

\section{Resources, semantics, and composition}
\begin{enumerate}
\item Resource rate--distortion separates carrier loss from decoder nonsaturation.
\item Operational semantics determines contextual visibility.
\item Foundations and Part V control composition, base change, and limits.
\end{enumerate}

\section{Statistical and intervention spine}
\begin{enumerate}
\item Simultaneous defect and grammar envelopes bracket architecture frontiers.
\item Persistent atlas inverse frontiers are stable.
\item Three-way decisions encode honest unresolvedness.
\item Certificate statistics controls resolution effort and evidence compression.
\item Typed realization joins finite-information authorization, recursive closure, and one common deployment witness without identifying their different units.
\item Sample-split held-out bounds validate repaired architectures.
\end{enumerate}

\section{No circularity in the graph-CRPS result}
The book-proved CDF validity repair and exact CRPS risk decomposition do not use the
graph-universal probability-validity classification.  The sharp path upper and
lower minimax theorem is summarized from \EGI\ and audited in \EGII; its full
proof is not reproduced here and is logically independent of the
probability-validity classification.

\section{No circularity in exactification}
The constructive exactification theorem restores target tangency without
using the converse theorem.  Target descent follows only after the separate
candidate-wise defect-remainder or acceptance condition is verified.  The
converse later shows that first-order target preservation forces the
defect-jet form modulo a flat term.

%% file: appendices/appF_glossary.tex
\chapter{Controlled Glossary of Structural Terms}
\label{app:glossary}

This glossary fixes the book's working vocabulary.  It is distinct from the
front-matter symbol table, which records notation, and from the thematic
index, which records every substantive occurrence.  The boundary clause in
each definition prevents a nearby concept from being silently substituted.
The gateway is the first systematic treatment, not necessarily the first
mention.

\begingroup
\small
\renewcommand{\arraystretch}{1.12}
\setlength{\LTpre}{0.6em}
\setlength{\LTpost}{0pt}
\begin{longtable}{>{\raggedright\arraybackslash}p{0.25\textwidth}
                  >{\raggedright\arraybackslash}p{0.55\textwidth}
                  >{\raggedright\arraybackslash}p{0.11\textwidth}}
\caption{Controlled glossary of book-level terms.}
\label{tab:controlled-glossary}\\
\toprule
Term & Working definition and boundary & Gateway\\
\midrule
\endfirsthead
\multicolumn{3}{@{}l}{\itshape Table~\thetable\ continued}\\[0.3em]
\toprule
Term & Working definition and boundary & Gateway\\
\midrule
\endhead
\midrule
\multicolumn{3}{r@{}}{\itshape Continued on next page}\\
\endfoot
\bottomrule
\endlastfoot

\multicolumn{3}{@{}l}{\color{egblue}\bfseries A--C}\\[0.2em]

\textbf{Architecture class}\index{architecture!class} &
A declared set $\Arch$ of deployable fields satisfying the output,
measurability, sharing, regularity, and resource contracts.  It is not the
training algorithm or merely the subset that one optimizer happens to reach.
& Ch.~\ref{ch:certified-systems}\\

\textbf{Architecture grammar}\index{architecture!grammar} &
The rules that generate legal representations, fields, compositions, and
resource budgets.  In finite-data work the grammar itself may be uncertain;
it is not automatically fixed by a model name or parameter count.
& Ch.~\ref{ch:certified-systems}\\

\textbf{Architecture obstruction}\index{obstruction!architecture} &
The infimum of native defect over a declared architecture class under a stated
risk aggregation rule.  It is a class-level realizability floor, not the
implementation gap of one trained deployment.
& Ch.~\ref{ch:second-elimination}\\

\textbf{Atlas}\index{atlas} &
A cover of the instance space by domains admitting legal local sections or
experts, together with an admissible routing rule.  Adding an atlas changes
the deployment contract; it is not merely widening one global model.
& Ch.~\ref{ch:repairs}\\

\textbf{Base change}\index{base change} &
Passage between fine and coarse representations, memories, or architecture
levels.  It preserves obstruction only under a proved commuting or exact
decomposition theorem; information loss alone does not make the change exact.
& Ch.~\ref{ch:composition}\\

\textbf{Bregman defect}\index{Bregman geometry!defect} &
The oriented Bregman divergence generated by a conjugate elimination.  Its
orientation and scale come from the objective, so an arbitrary symmetric
distance is not an interchangeable substitute.
& Ch.~\ref{ch:conjugate-lifts}\\

\textbf{Carrier}\index{carrier} &
The deployment information state $Z$ passed to a decoder or readout.  It is
distinct from the quotient-reduced slack carrier used to audit a conic lift;
identifying the two requires an interface theorem.
& Ch.~\ref{ch:resource-rd}\\

\textbf{Certificate}\index{certificate} &
A typed finite-data record containing a confidence world, a declared boundary,
a structural color, supporting evidence, and a resolution profile.  It is
more than a point estimate, posterior probability, or diagnostic score.
& Ch.~\ref{ch:certificate-statistics}\\

\textbf{Certificate margin}\index{certificate!margin} &
The distance from the complete identified image to the declared action
boundary when the certificate is resolved.  A positive margin quantifies
stability of the color; the tolerance profile records the color across all
candidate boundaries.
& Ch.~\ref{ch:statistical-eg}\\

\textbf{Certified elimination}\index{elimination!certified} &
An exact triple $(H,J,\Def)$ with $J(x)=\inf_a H_x(a)$ and
$\Def_x(a)=H_x(a)-J(x)\ge0$.  ``Certified'' refers to identity in the original
objective; attainment of the infimum is not required.
& Ch.~\ref{ch:certified-systems}\\

\textbf{Certified learning system}\index{learning system!certified} &
The complete declared package of population, auxiliary fibration, objective,
defect, architecture, representation/resource grammar, task contract, and
statistical experiment.  No obstruction claim is absolute outside this
package.
& Ch.~\ref{ch:certified-systems}\\

\textbf{Common deployment}\index{deployment!common} &
One legal architecture that works simultaneously across the indexed inputs or
population worlds.  It is stronger than pointwise feasibility
$\forall x\,\exists A_x$ and cannot be inferred by exchanging quantifiers.
& Ch.~\ref{ch:statistical-eg}\\

\textbf{Confidence world}\index{confidence world} &
A data-dependent set $C_n\subseteq\mathcal W$ that contains the true
population world under its declared fixed-record, simultaneous, or anytime
coverage contract.  It is the set of worlds retaining inferential standing,
not a set of high-posterior stories.
& Ch.~\ref{ch:statistical-eg}\\

\textbf{Context closure}\index{contextual closure} &
Closure of the declared task set under all legal pre- and post-compositions.
It yields a task- and grammar-relative fully abstract quotient, not an
absolute identity between internal pipelines.
& Ch.~\ref{ch:operational-semantics}\\

\textbf{Coordination tax}\index{rectangularity!coordination tax} &
The excess obstruction caused by requiring one shared conditional mechanism,
measured relative to the rectangular hull that may paste all locally available
kernels independently.
& Ch.~\ref{ch:coordination}\\

\textbf{Curvature}\index{curvature!structural} &
A quantified failure of local operations, transports, or elimination orders
to commute or integrate.  Square, ordering, and interchange curvatures live at
different interfaces and are not one universal scalar tax.
& Ch.~\ref{ch:integrability}\\

\multicolumn{3}{@{}l}{\color{egblue}\bfseries D--H}\\[0.2em]

\textbf{Decoder nonsaturation}\index{decoder!nonsaturation} &
The residual defect remaining after a carrier $Z$ is fixed because the legal
decoder class cannot realize the optimal readout.  It is separate from
information already erased by the carrier.
& Ch.~\ref{ch:resource-rd}\\

\textbf{Declared-loss scale rule}\index{declared-loss scale rule} &
Two nonnegative quantities may be added only when an exact identity, infimal
decomposition, or exchange theorem converts them into the same declared
objective loss scale.
& Ch.~\ref{ch:certified-systems}\\

\textbf{Defect}\index{defect!native} &
The exact residual $\Def_x(a)=H_x(a)-J(x)$ produced by elimination.  A metric,
regularizer, or surrogate discrepancy is not a defect unless a theorem links
it to this residual.
& Ch.~\ref{ch:certified-systems}\\

\textbf{Defect envelope}\index{defect!confidence envelope} &
Simultaneous lower and upper bounds on the population defect over a confidence
world.  Together with an architecture-grammar envelope it brackets an entire
obstruction frontier, not just one fitted architecture.
& Ch.~\ref{ch:statistical-eg}\\

\textbf{Elimination}\index{elimination} &
Optimization of an auxiliary object while retaining the optimized value and
its exact residual.  It is not automatically marginalization, projection, or
conditioning unless the declared objective makes those operations coincide.
& Ch.~\ref{ch:certified-systems}\\

\textbf{Elimination tower}\index{elimination!tower} &
A nested sequence of fine-to-coarse eliminations whose realization costs
compose by exact or infimal decompositions.  It separates choice of a coarse state
from the cost of realizing its fine fiber.
& Ch.~\ref{ch:towers-pgxvc}\\

\textbf{Evidence carrier}\index{carrier!evidence} &
A retained statistic used for future certification.  Sufficiency is relative
to the certificate truth partition and its binding alternatives, not
necessarily to the full parametric model.
& Ch.~\ref{ch:certificate-statistics}\\

\textbf{Exactification}\index{exactification} &
Subtraction of the relevant value or defect jet to restore target touching and
derivatives of a surrogate.  Exactification restores the interface; target
descent still needs a separate acceptance or remainder condition.
& Ch.~\ref{ch:exactification}\\

\textbf{Extension complexity}\index{extension complexity} &
The smallest size of a representation in a declared lift family.  Without an
exchange theorem and computational model, it is neither native defect nor an
unrestricted runtime lower bound.
& Ch.~\ref{ch:lift-complexity}\\

\textbf{Fiber realization tax}\index{fiber realization tax} &
The least vertical defect required to realize a chosen coarse state by an
admissible fine object.  It is the fine-fiber term in a coarse/fine obstruction
decomposition.
& Ch.~\ref{ch:certified-systems}\\

\textbf{Four-component risk decomposition}\index{four-component risk decomposition} &
The separation of model approximation, architecture obstruction,
generalization, and optimization or implementation components relative to
compatible reference risks.  The components do not become interchangeable
merely because they appear in one algebraic identity.
& Ch.~\ref{ch:four-components}\\

\textbf{Gauge invariance}\index{transport!gauge invariance} &
Invariance of a structural conclusion under legal changes of coordinates,
factor gauges, or local branch labels.  Raw representatives may change while
cycle type, section count, or quotient information remains fixed.
& Ch.~\ref{ch:transport-cycles}\\

\textbf{Global section}\index{global section} &
A tuple $(s_v)_{v\in V}$ with $s_v\in\mathcal B_v$ such that every oriented
edge transport is defined at $s_u$ and satisfies $T_{uv}(s_u)=s_v$.  Root-path
survival and common fixedness under fundamental holonomies characterize this
edgewise definition; a separate fixed branch for each cycle does not establish
one common global section.
& Ch.~\ref{ch:transport-cycles}\\

\textbf{Hodge repair}\index{integrability!Hodge repair} &
Projection of an observed edge defect field into exact, flat, and certified
components, separating local curvature, harmonic periods, and the final
nonnegative/touching constraint.
& Ch.~\ref{ch:integrability}\\

\textbf{Holonomy}\index{transport!cycle holonomy} &
The composite branch transport around a cycle.  For partial transports it may
fix, move, or lose a branch; under relabeling it changes by conjugacy rather
than as an absolute labeled permutation.
& Ch.~\ref{ch:transport-cycles}\\

\multicolumn{3}{@{}l}{\color{egblue}\bfseries I--O}\\[0.2em]

\textbf{Identified image}\index{identified image} &
The query-relevant projection $J_C(q)=\{\Gamma_q(w):w\in C\}$ of a
confidence world.  It is the complete finite-data range for the declared
functional; a certificate color is only its projection relative to an action
boundary.
& Ch.~\ref{ch:statistical-eg}\\

\textbf{Implementation gap}\index{implementation gap} &
The defect of a chosen deployment above the infimum within its declared
architecture class.  Better training may reduce this term; it does not by
itself reduce the architecture obstruction.
& Ch.~\ref{ch:certified-systems}\\

\textbf{Integrability}\index{integrability} &
The condition that locally reported elimination increments arise from one
global potential.  Face flatness suffices only on the appropriate simply
connected complex; global periods must otherwise also be audited.
& Ch.~\ref{ch:integrability}\\

\textbf{Lift admissibility}\index{lift!admissibility} &
Target-faithfulness of a nonnative lifted representation after slack,
minimal-face, quotient, and gauge reduction.  Finite auxiliary dimension by
itself does not make a lift intrinsic.
& Ch.~\ref{ch:lift-complexity}\\

\textbf{Local solvability}\index{local solvability} &
Existence and well-posedness of the pointwise oracle problem at each input,
including the native cost of deviating from that oracle.  It does not imply
that one shared deployment can realize all local optima simultaneously.
& Ch.~\ref{ch:deployability}\\

\textbf{Global realizability}\index{global realizability} &
Existence of one rule in the declared deployment class that realizes the
relevant local oracle family simultaneously.  It is conditional on the
representation, sharing, memory, regularity, and resource contracts.
& Ch.~\ref{ch:deployability}\\

\textbf{Finite-sample certifiability}\index{finite-sample certifiability} &
The ability of the declared statistical experiment to distinguish
realizability, nonrealizability, and unresolvedness with stated error control.
It is a property of the evidence available about the population structure,
not of the population obstruction alone.
& Ch.~\ref{ch:deployability}\\

\textbf{Model conflict}\index{model conflict} &
The typed output $M$ produced when the declared confidence world is empty.
It signals conflict among data, uncertainty construction, and model
restrictions; it does not authorize vacuous feasible or impossible claims.
& Ch.~\ref{ch:statistical-eg}\\

\textbf{Monodromy}\index{topology!monodromy} &
The permutation or partial transformation of local oracle branches induced by
lifting loops around a singular or multiply connected region.  It is distinct
from a branch that genuinely disappears.
& Ch.~\ref{ch:singular}\\

\textbf{Native loss scale}\index{native loss scale} &
The units of the objective-generated defect.  Parameter distance, topological
degree, communication, runtime, and task regret remain different units
until an exchange theorem connects them.
& Ch.~\ref{ch:conjugate-lifts}\\

\textbf{Obstruction-Aware Inference (OAI)}\index{obstruction-aware inference} &
The concrete prototype and software line used for flow certificates, partial
transport, cycle audit, and atlas construction.  OAI v0.1--v0.5 is an
implementation path inside the broader OALI workflow, not a second name for
that workflow.
& Chs.~\ref{ch:oali-workflow}--\ref{ch:transport-cycles}\\

\textbf{Obstruction-aware learning and inference (OALI)}\index{obstruction-aware learning and inference} &
The workflow that declares the contract, estimates local oracles, builds
transports, computes typed certificates, selects a mechanism-matched repair,
and validates it independently.  It is not generic mixture-of-experts routing.
& Ch.~\ref{ch:oali-workflow}\\

\textbf{Operational visibility}\index{operational visibility} &
The portion of an internal difference that some legal task in some legal
context can expose.  A positive native defect need not produce positive
downstream separation without transmission and exposure gates.
& Ch.~\ref{ch:operational-semantics}\\

\textbf{Oracle fiber}\index{oracle!fiber} &
The zero-defect set $\mathcal O(x)$ at one instance; the oracle field is the
family of these fibers over $x$.  A fiber may be set-valued even when no legal
global point-valued selection exists.
& Ch.~\ref{ch:certified-systems}\\

\multicolumn{3}{@{}l}{\color{egblue}\bfseries P--W}\\[0.2em]

\textbf{Partial transport}\index{transport!partial bijection} &
A partial bijection between local branch sets that permits unmatched branches
to appear, disappear, or remain unresolved.  It avoids manufacturing a forced
correspondence when no branch match is justified.
& Ch.~\ref{ch:transport-cycles}\\

\textbf{Persistent atlas number}\index{atlas!persistent number} &
The minimum chart count needed to achieve a stated uniform defect tolerance.
The integer count may jump, while its inverse chart-budget/loss frontier is
stable under uniform perturbation.
& Ch.~\ref{ch:statistical-eg}\\

\textbf{P/G/X/V/C calculus}\index{P/G/X/V/C calculus} &
A contract-based classification of pointwise elimination, external
globalization, defect-jet exactification, variational coupling, and
fixed-marginal coupling.  The label is determined by what changes and what is
held fixed, not by the numerical algorithm alone.
& Ch.~\ref{ch:towers-pgxvc}\\

\textbf{Population and uniform obstruction}\index{obstruction!population versus uniform} &
Infima of expected defect and worst-case defect, respectively.  A small
population seam can make the first zero while the second remains positive;
neither aggregation contract may replace the other silently.
& Ch.~\ref{ch:certified-systems}\\

\textbf{Quotient repair}\index{repair!quotient} &
Replacement of labeled representatives by the task-relevant equivalence class,
such as an eigenprojector or unordered mixture.  It removes label artifacts
only when the downstream contract is invariant to the quotient.
& Ch.~\ref{ch:repairs}\\

\textbf{Randomized repair}\index{repair!randomized} &
A deployment kernel used either as a probability-valued output or to draw a
point-valued action.  These are different task contracts and neither is the
deterministic average of representatives; the applicable loss must be declared
on the law or on the sampled action.
& Ch.~\ref{ch:repairs}\\

\textbf{Rectangular hull}\index{rectangularity!rectangular hull} &
The conditional architecture formed by independently pasting every locally
available kernel.  It is a comparison class used to isolate coordination, not
automatically a legal shared deployment.
& Ch.~\ref{ch:coordination}\\

\textbf{Resolution complexity}\index{certificate!resolution complexity} &
The information or sample effort needed to separate a world from the nearest
compatible world with the opposite certificate label.  It is not a generic
generalization bound or posterior entropy.
& Ch.~\ref{ch:certificate-statistics}\\

\textbf{Resource rate-distortion}\index{resource constraints!rate--distortion} &
The best native defect attainable under a declared carrier/decoder resource
budget, often decomposed into carrier information loss and decoder
nonsaturation.  Parameter count alone does not define the frontier.
& Ch.~\ref{ch:resource-rd}\\

\textbf{Saturation}\index{saturation} &
Zero architecture obstruction under a declared risk contract.  Exact
saturation additionally requires an attained zero-defect section; a zero
unattained infimum gives approximation but not exact realization.
& Ch.~\ref{ch:second-elimination}\\

\textbf{Set-valued oracle}\index{oracle!set-valued} &
An oracle whose correct output is a set, orbit, projector, subspace, or law
rather than one labeled point.  Forcing a representative can introduce an
artificial obstruction.
& Ch.~\ref{ch:certified-systems}\\

\textbf{Structural assignment margin}\index{transport!assignment margin} &
The objective gap between the best partial branch matching and the second-best
structurally distinct matching after dummy-label permutations are removed.
It controls exact recovery under perturbation.
& Ch.~\ref{ch:transport-cycles}\\

\textbf{Target-calling lift}\index{lift!target-calling} &
An exact lift that appends coordinates computed directly from the target or an
arbitrarily prescribed field.  It can manufacture fake coherence and is
excluded by intrinsic lift admissibility.
& Ch.~\ref{ch:lift-complexity}\\

\textbf{Target-visible quotient}\index{carrier!target-visible quotient} &
The conditional dual oracle signature $R_U=\E(S\mid U)$ retained by a
deployment carrier.  It is the coarsest quotient preserving optimal native
Bregman defect; its visible cardinality is not determined by cone size without
an extraction theorem.
& Ch.~\ref{ch:lift-complexity}\\

\textbf{Task envelope}\index{task envelope} &
The pointwise least operational kernel preserving every value in the declared
task family.  Internal differences removed by the envelope are invisible to
that task contract, though another contract may expose them.
& Ch.~\ref{ch:operational-semantics}\\

\textbf{Three-way certificate}\index{certificate!three-way rule} &
The feasible/impossible/unresolved rule obtained by evaluating a structural
truth map over one simultaneous confidence world.  It declares a color only
when every compatible world agrees.  An empty confidence world is handled
separately as model conflict $M$.
& Ch.~\ref{ch:statistical-eg}\\

\textbf{Touching}\index{defect!touching} &
Vanishing of the native defect, and when required its relevant derivatives, at
a declared oracle witness.  Nonnegativity without touching does not certify
the intended eliminated target.
& Ch.~\ref{ch:certified-systems}\\

\textbf{Transport}\index{transport} &
A declared rule for comparing or propagating local oracle states across
inputs, graph edges, or representations.  It may be partial, set-valued, or
gauge-dependent; path consistency must be proved rather than assumed.
& Ch.~\ref{ch:transport-cycles}\\

\textbf{Unresolved}\index{certificate!unresolved} &
The honest finite-data state in which the confidence world contains opposite
structural truth labels.  It is an authorized conclusion about present
resolution, not a failed optimizer or an instruction to guess.
& Ch.~\ref{ch:statistical-eg}\\

\textbf{Worldwise validity}\index{worldwise validity} &
Uniform correctness of a certificate for every population world under the
declared experiment.  It is stronger than prior-average posterior credibility
and can require infinite effort when opposite labels are observationally
indistinguishable.
& Ch.~\ref{ch:certificate-statistics}\\

\end{longtable}
\endgroup

%% file: appendices/appG_source_map.tex
\chapter{Source Manuscript Crosswalk}
\index{source manuscript crosswalk}
\label{app:source-map}

This monograph reorganizes the source manuscripts by concept rather than publication order.

\begin{longtable}{p{0.27\textwidth}p{0.66\textwidth}}
\toprule
Source & Primary book locations\\
\midrule
Elimination Geometry I & Chapters 2 and 5--8; native defect, P/G/X/V/C, exactification, graph-CDF/CRPS, and model-specific examples\\
Elimination Geometry II & Chapters 7--9 and boundary material in Chapter 26; exactification converse, probability-validity gates, integrability, and representation limits\\
Elimination Geometry III & Chapter 18; soft conditional chains, pushforward, base change, conditioning, and zero temperature\\
Foundations of Elimination Geometry & Chapters 2, 6, and 18; defect fibrations, obstruction towers, and interchange coherence\\
Elimination Obstruction Transfer & Chapters 11--12; second elimination, regular obstruction transfer, flow duality, and population thickening\\
COT & Chapters 6, 13, and 18; rectangularity, coordination tax, architecture base change, and dequantization\\
Singular EG & Chapters 14--15 and Appendix C; radical tax, degree/discriminant mechanisms, and atlas/quotient repair\\
Lift Complexity & Chapter 10; target-calling no-go, slack factorization, quotient reduction, and complexity boundaries\\
Resource-Constrained Architectures & Chapters 10, 16, 19, and 21; target-visible extraction, carrier--decoder rate--distortion, common deployment, and typed realization\\
Exact Data Selection & Chapter 16; exact low-dimensional and budget-two selection frontiers, auxiliary-law elimination, and merge--split recovery\\
Operational Semantics & Chapter 17; task envelopes, context closure, full abstraction, and ordering curvature\\
Marton counterexample & Chapter 18; credit--semantic-tax identity, rectangular-switch frontier, and a certified positive Markov-architecture obstruction; exact verification package shipped with the book\\
Statistical EG & Chapter 19; confidence worlds, identified images, honest frontier brackets, persistent atlases, typed architecture certificates, and the active-set depth-separation candidate family\\
Certificate Statistics & Chapters 20--21; resolution complexity, evidence carriers, recursive state, and typed no-compensation\\
EGML & Chapters 1, 3, 11, and 21--24; four-component risk theory, integrated ML positioning, OALI, and application protocols\\
EGML II & Chapters 20--22; typed carriers, non-compensation, and adaptive learning interfaces\\
OAI packages & Chapters 22--24; transport recovery, cycle audit, atlas construction, and synthetic validation\\
Quantum EG & Chapter 25; Gibbs base-change rigidity, sectors, recoverability, and measurement limits\\
Chapter 24 reproducibility packages & Chapter 24; posterior-family and cytometry proof-of-mechanism checks\\
\bottomrule
\end{longtable}

\section{Principal-result audit ledger}
\index{principal result!provenance audit}
\label{app:principal-result-audit}

This section inventories the 37 formal-result families selected for complete proof in
Version 1.9.  The number is a proof-coverage count, not a count of independent
contributions: the rows range from classical identities and direct workflow
lemmas to book-level interfaces and a small number of narrow priority
candidates.  Calling all 37 ``principal results'' would overstate the density
of original mathematics.  The classification concerns intellectual
provenance, not correctness.

\begin{description}[leftmargin=3.6em,style=nextline]
\item[\textsf{C}] A classical or explicitly imported result supplies the
mathematical core.
\item[\textsf{S}] The statement is a direct specialization, corollary, or
workflow lemma; the application may be useful but is not a priority claim.
\item[\textsf{P}] A possible narrow program-specific increment.  Close
antecedents remain, so this is not a general originality claim.
\item[\textsf{O}] A priority-audit candidate.  This code records an unresolved
literature question, not an established original theorem.
\item[\textsf{B}] A book-level synthesis or interface theorem that organizes
several results and should not be cited as one wholly new mathematical
mechanism.
\end{description}

\begingroup
\small
\setlength{\tabcolsep}{3pt}
\begin{longtable}{@{}p{0.035\textwidth}p{0.27\textwidth}p{0.055\textwidth}p{0.57\textwidth}@{}}
\toprule
No. & Formal result & Class & Antecedent and audited increment \\
\midrule
\endfirsthead
\toprule
No. & Formal result & Class & Antecedent and audited increment \\
\midrule
\endhead
\midrule
\multicolumn{4}{r}{\textit{continued on next page}}\\
\endfoot
\bottomrule
\endlastfoot
1 & \cref{prop:four-component-identity}: exact four-component risk identity
& \textsf{B} & Algebraic telescoping over the classical approximation,
estimation, and optimization ledger.  The book-level increment is the
contract-relative split of approximation into local-model and shared-
architecture accounts; compare amortization-gap work
\citep{CremerLiDuvenaud2018,MargossianBlei2024}.\\
\addlinespace
2 & \cref{thm:four-component-certified}: certified learning bound
& \textsf{S} & Standard uniform convergence plus approximate empirical-risk
minimization.  The specialization preserves the model/architecture floor; it
adds no new concentration inequality.\\
\addlinespace
3 & \cref{thm:conjugate-defect}: conjugate defect identity
& \textsf{C} & Fenchel--Young/Bregman identity
\citep{Bregman1967,Rockafellar1970}.  The book uses it to fix the native
defect and its direction.\\
\addlinespace
4 & \cref{prop:tower-associativity}: tower associativity
& \textsf{C} & Classical infimal-convolution/Bellman associativity
\citep{Bellman1957,BaccelliEtAl1992}; the typed tax interpretation is the
book's organizational layer.\\
\addlinespace
5 & \cref{thm:jet-exactification}: jet exactification
& \textsf{S} & Elementary Taylor subtraction, adjacent to first-order
surrogate touching \citep{Mairal2013FirstOrder}.  The specialization uses the
exact elimination defect rather than an arbitrary correction.\\
\addlinespace
6 & \cref{thm:converse-exactification}: converse exactification normal form
& \textsf{P} & Within the declared frozen-lift class, target-jet preservation
forces defect-jet subtraction modulo a flat term.  The source claims this
normal form, not tangent-surrogate methodology in general.\\
\addlinespace
7 & \cref{prop:cdf-validity-repair}: CDF validity repair
& \textsf{C} & Hilbert projection and coordinatewise nonexpansiveness,
specialized to squared-CDF/CRPS risk and graph roughness.\\
\addlinespace
8 & \cref{thm:flat-elimination}: flat elimination criterion
& \textsf{S} & Classical potential/cycle integrability
\citep{MondererShapley1996,JiangEtAl2011}; nonnegativity and fiberwise
touching specialize the potential to partial-minimum defects.\\
\addlinespace
9 & \cref{thm:target-calling-nogo}: finite-complexity target-calling no-go
& \textsf{S} & An elementary augmentation argument specialized to
target-calling lift geometry.  Its value is diagnostic: unrestricted lifts
make finite-overhead complexity vacuous.  No new extension-complexity
mechanism is claimed.\\
\addlinespace
10 & \cref{thm:lift-factorization}: conic lift--factorization gate
& \textsf{C} & Explicitly imported from \citet{GouveiaParriloThomas2013},
extending \citet{Yannakakis1991}.\\
\addlinespace
11 & \cref{thm:lift-visible-transfer}: visible quotient and lift-to-carrier
transfer & \textsf{P} & Conditional Bregman prediction is classical
\citep{BanerjeeGuoWang2005}.  The claimed increment is the
coarsest dual-signature quotient in the reduced-lift interface and the
factorized resource-to-native-defect transfer.\\
\addlinespace
12 & \cref{thm:oracle-transport}: oracle-variation transport bound
& \textsf{P} & Coupling, Lipschitz, and strong-growth ingredients are
classical.  The companion EOT manuscript supplies the stated sharp
native-objective lower certificate;
this audit does not assert exhaustive priority over equivalent metric forms.\\
\addlinespace
13 & \cref{thm:rectangular-decomposition}: rectangular Bellman and
coordination residual & \textsf{P} & Robust-control rectangularity is
classical \citep{EpsteinSchneider2003,Iyengar2005,NilimElGhaoui2005}.  COT's
increment is to rectangularize the deployable architecture and retain failed
sharing as an exact forward-KL tax.\\
\addlinespace
14 & \cref{thm:radical-tax}: exact radical catastrophe tax
& \textsf{P} & Winding and loop lifting are classical.  The exact uniform
value $r^2$ is a useful quantitative corollary for the declared objective,
not evidence of a new topological mechanism.\\
\addlinespace
15 & \cref{thm:atlas-saturation}: atlas saturation
& \textsf{S} & Direct witness lemma once a covering by exact local sections is
given; sectional category and Schwarz genus are classical
\citep{Schwarz1961}.\\
\addlinespace
16 & \cref{thm:architecture-rd}: architecture rate--distortion decomposition
& \textsf{P} & Conditional Bregman decomposition and rate--distortion are
classical neighbors \citep{BanerjeeEtAl2005,Shannon1959}.  The source claims
the native carrier-erasure/decoder-nonsaturation ledger under a declared
resource and extraction grammar.\\
\addlinespace
17 & \cref{prop:task-envelope}: task-envelope isometry
& \textsf{P} & Classical residuation and conjugacy isometries precede it
\citep{CohenGaubertQuadrat2004,AttouchWets1986}.  The Attouch--Wets result
concerns Legendre--Fenchel/epigraph geometry, not the finite task-envelope
sup-norm proposition.  The source claims the latter's task-restricted,
two-sided min-plus kernel realization, which the book proves directly.\\
\addlinespace
18 & \cref{thm:contextual-full-abstraction}: contextual completion and full
abstraction & \textsf{P} & Contextual equivalence and full abstraction are
classical \citep{Milner1977,Plotkin1977}.  The claimed increment is their
simultaneous realization by one explicit closed min-plus task envelope.\\
\addlinespace
19 & \cref{thm:book-soft-conditional-chain}: soft conditional chain theorem
& \textsf{C} & Gibbs variational identity, telescoping normalization, and the
relative-entropy chain rule \citep{Csiszar1975,CoverThomas2006}, with the
book's common-reference typing.\\
\addlinespace
20 & \cref{thm:three-way}: honesty and maximal decisiveness
& \textsf{C} & Confidence-set projection \citep{Dufour1997}, partial
identification \citep{Manski2003}, and equivalence/noninferiority threshold
logic \citep{Blackwelder1982,BergerHsu1996}.  The $F/I/U/M$ record is book
terminology, not a claim to invent the projection rule.\\
\addlinespace
21 & \cref{thm:honest-elimination-bracket}: honest elimination bracket
& \textsf{S} & Monotone projection through simultaneous objective and
feasible-set uncertainty; closest explicit neighbor is universal confidence
sets for random optimization \citep{Vogel2008}.\\
\addlinespace
22 & \cref{thm:atlas-stability}: atlas interleaving and stable inverse
& \textsf{P} & The canonical companion source is Statistical EG
\citep{StatisticalEG}; EGML reuses the interface in its learning-facing atlas
ledger \citep{EGML}.  Interleaving stability is classical
\citep{deSilvaMunchStefanou2017}.  The claimed increment is the sharp inverse
$K$-chart loss frontier in native objective units.\\
\addlinespace
23 & \cref{thm:book-common-deployment}: common-deployment quantifier theorem
& \textsf{P} & The minimax inequality is elementary.  Statistical antecedents
include Hodges--Le Cam superefficiency, H\'ajek local asymptotic minimaxity,
and Leeb--P\"otscher analyses of oracle-property nonuniformity
\citep{LeCam1953,Hajek1970,Hajek1972,LeebPotscher2005,LeebPotscher2008}.
The source-program increment is only the explicit separation of pointwise
feasibility from one common deployable witness over a confidence world.\\
\addlinespace
24 & \cref{prop:posterior-not-worldwise}: posterior credibility is not
worldwise validity & \textsf{C} & Standard prior-average versus uniform-
frequentist distinction, illustrated by observationally equivalent opposite-
label worlds; retained as an expository typed counterexample.\\
\addlinespace
25 & \cref{cor:certificate-honesty}: truth-map form of three-way honesty
& \textsf{C} & A notational corollary of row 20 and Dufour's projection
principle; it has no independent novelty claim.\\
\addlinespace
26 & \cref{thm:certificate-resolution-lower}: worldwise resolution lower
bound & \textsf{C} & Imported sequential change-of-measure and partition-
identification geometry \citep{KaufmannEtAl2016,GarivierKaufmann2016}; the
certificate truth map selects the alternative set.\\
\addlinespace
27 & \cref{thm:certificate-evidence-slack}: exact certificate preservation by
slack & \textsf{S} & A short specialization of KL data processing and
max--min rearrangement to preservation of one certificate floor.  It is not
presented as a new information-theoretic mechanism.\\
\addlinespace
28 & \cref{thm:certificate-recursive-state}: recursive quotient and minimal
exact state & \textsf{S} & Strong probabilistic bisimulation and partition
refinement are classical
\citep{LarsenSkou1991,GivanEtAl2003,PaigeTarjan1987}.  Certificate color and
the state/chart reading are the specialization.\\
\addlinespace
29 & \cref{thm:book-typed-no-compensation}: typed no-compensation
& \textsf{B} & Synthesizes a zero-information lower bound, a bisimulation
collision, and a decoder collision.  The contribution is their typed logical
independence, not new proofs of each component impossibility.\\
\addlinespace
30 & \cref{thm:book-typed-architecture-realization}: typed architecture
realization interface & \textsf{B} & Collects earlier sample-time, evidence,
state, common-witness, native-loss, and contextual gates.  It is an end-to-end
interface theorem rather than one new underlying bound.\\
\addlinespace
31 & \cref{thm:heldout-improvement}: finite-library held-out improvement
& \textsf{C} & Hoeffding's inequality plus a finite union bound, with the
book's independent-unit and predeclaration contract.\\
\addlinespace
32 & \cref{thm:population-improvement-template}: population improvement
template & \textsf{S} & Direct subtraction of a baseline lower certificate
and repair upper certificate on one loss scale and event.\\
\addlinespace
33 & \cref{thm:partial-transport-recovery}: noisy partial-transport recovery
& \textsf{C} & Finite-class argmin stability under a strict margin, with a
metric perturbation bound.  The partial matched/unmatched semantics are the
application-specific contract.\\
\addlinespace
34 & \cref{thm:cycle-space}: cycle-space characterization
& \textsf{S} & Classical spanning-tree holonomy and graph consistency
\citep{Singer2011,JiangEtAl2011}; the specialization retains domains of
partial maps and requires a common surviving fixed branch.\\
\addlinespace
35 & \cref{thm:quantum-base-change}: universal Gibbs base-change rigidity
& \textsf{S} & Direct finite-dimensional Hamiltonian specialization of the
factorization and quantum-exponential-family sufficiency criteria
\citep[Theorems~5 and~7]{JencovaPetz2006}.  The book supplies an independent
pressure-duality/Petz proof and makes no standalone priority claim.\\
\addlinespace
36 & \cref{thm:book-deployment-conflict}: deployment conflict under
observational overlap, with repair trichotomy
& \textsf{B} & The exact intersection condition expands the existence of a
common measurable witness.  The quantitative bound is Le Cam's two-point
testing/total-variation identity \citep{LeCam1953,Tsybakov2009}; the
information-repair interpretation is adjacent to Blackwell comparison
\citep{Blackwell1953}.  The book-level increment is the typed placement of
solver, information, and architecture repair in one deployment interface.
No new decision-theoretic lower bound is claimed.\\
\addlinespace
37 & \cref{thm:active-set-conflict-modulus,thm:active-set-repair-budget}:
active-set conflict modulus, one-pass native-loss saturation, and sufficient
repair depth
& \textsf{O} & Active-set affine sensitivity, shallow sparse encoders,
LISTA/unrolling, and proximal contraction are established
\citep{GregorLeCun2010,ChenLiuWangYin2018,BeckTeboulle2009,
MuthukumarSulam2023}.  Most importantly,
\citet{ONeillGumranKlindt2025} already prove a global one-layer
linear--nonlinear sparse-autoencoder amortisation gap and study
inference-time optimization.  The narrower candidate increment is the
input-computable two-neighborhood Jacobian conflict modulus along a declared
perturbation subspace, its quantitative native-loss floor, and a sufficient
crossing depth on the same compact set.
The exact combination was not located, but independent priority is not
established; see the dated dedicated audit shipped with the book.\\
\end{longtable}
\endgroup

The ledger should not be summarized as four original and nine
program-specific theorems.  The finite target-calling construction is an
elementary augmentation argument; the radical value is a quantitative
winding corollary; and the evidence-slack criterion is a short KL
data-processing/max--min consequence.  They are retained for their fit with
the audit framework, not presented as established new mechanisms.  The
dedicated result-level audit of universal Gibbs base-change rigidity closes it
as a direct specialization, not an originality candidate.  Several
\textsf{P} rows may contain narrow useful formulations, but their engines are
close to established approximation, control, information-theoretic, or
semantic results.  The main book-level contribution claimed here is the
typed, native-loss, audit-oriented synthesis.

\section{Editorial principle}
\index{source manuscript crosswalk!editorial principle}

The source manuscripts remain the record of model-specific variants and
advanced results not selected for the book.  Every formal result inventoried
here has a complete proof in its chapter-local appendix.  The monograph's
claimed increment is its terminology, typed interfaces, audit order, and
cross-domain interpretation.  Results retained for context without a complete
book proof are identified in place by a \emph{Source status} box and are
excluded from the 37-result proof-coverage inventory.

%% file: thematic_index_crossrefs.tex
\index{architecture gap|see{four-component risk decomposition, architecture obstruction component}}
\index{architecture obstruction|see{obstruction, architecture}}
\index{architecture rate--distortion|see{resource constraints, rate--distortion}}
\index{certified elimination|see{elimination, certified}}
\index{coordination tax|see{rectangularity, coordination tax}}
\index{CRPS|see{continuous ranked probability score}}
\index{holonomy|see{transport, cycle holonomy}}
\index{monodromy|see{topology, monodromy}}
\index{native defect|see{defect, native}}
\index{OAI|see{obstruction-aware inference}}
\index{OALI|see{obstruction-aware learning and inference}}
\index{rate--distortion|see{resource constraints, rate--distortion}}
\index{rectangular hull|see{rectangularity, rectangular hull}}
\index{three-way rule|see{certificate, three-way rule}}
\index{unresolved|see{certificate, unresolved}}